\def\isarxiv{1}
\def\submissionappendix{1}
\def\arxivsubmission{1}
\ifdefined\submissionappendix\else
\def\fullarchive{1}
\def\vTwentyTwoArchive{1}
\fi
\ifdefined\isarxiv
\documentclass[11pt]{article}
\usepackage[letterpaper,margin=1in]{geometry}
\usepackage[section]{placeins}   
\ifdefined\arxivsubmission
  \usepackage[authoryear,round]{natbib}
  \let\cite\citep
\else
  \usepackage{natbib}
\fi
\usepackage{lineno}

\else
\documentclass[11pt]{article}
\usepackage[review]{acl}   
\ifdefined\fullarchive\else
\def\submissionappendix{1}
\fi
\usepackage{times}
\usepackage{microtype}

\AtBeginDocument{%
  \setlength{\abovedisplayskip}{5pt plus 2pt minus 3pt}%
  \setlength{\belowdisplayskip}{5pt plus 2pt minus 3pt}%
  \setlength{\abovedisplayshortskip}{2pt plus 1pt minus 1pt}%
  \setlength{\belowdisplayshortskip}{3pt plus 1pt minus 2pt}%
  \setlength{\abovecaptionskip}{4pt}%
  \setlength{\belowcaptionskip}{2pt}%
  \setlength{\textfloatsep}{9pt plus 2pt minus 3pt}%
  \setlength{\floatsep}{8pt plus 2pt minus 2pt}%
  \setlength{\intextsep}{8pt plus 2pt minus 2pt}%
  \setlist{topsep=3pt, partopsep=0pt}%
}
\makeatletter
\def\thm@space@setup{\thm@preskip=5pt plus 1pt minus 2pt \thm@postskip=\thm@preskip}
\makeatother
\fi

\usepackage{amsmath, amssymb, amsthm, mathtools}
\usepackage{bm}
\usepackage{xcolor}
\usepackage{graphicx}
\usepackage{adjustbox}
\usepackage{booktabs}
\usepackage{tabularx}
\usepackage{multirow}
\usepackage{wrapfig}
\usepackage{float}
\ifdefined\arxivsubmission\else
\usepackage{dblfloatfix}
\fi
\usepackage{algorithm}
\usepackage{algpseudocode}
\usepackage{enumitem}
\usepackage{hyperref}
\hypersetup{
  colorlinks=true,
  linkcolor=blue,
  citecolor=blue,
  urlcolor=blue,
  pdftitle={Beyond Sparse Weights: When Is Attention Compressible?},
  pdfsubject={Attention compressibility and exact-budget KV-cache reduction},
  pdfauthor={Anonymous}
}   
\ifdefined\arxivsubmission
\hypersetup{pdfauthor={Chiwun Yang and Xiaoyu Li}}
\fi
\usepackage{aliascnt}
\usepackage[capitalize]{cleveref}
\usepackage{tikz}
\usetikzlibrary{positioning,fit,arrows.meta,backgrounds,calc,shapes.geometric}

\allowdisplaybreaks

\theoremstyle{plain}
\newtheorem{theorem}{Theorem}[section]

\newaliascnt{lemma}{theorem}
\newtheorem{lemma}[lemma]{Lemma}
\aliascntresetthe{lemma}

\newaliascnt{proposition}{theorem}
\newtheorem{proposition}[proposition]{Proposition}
\aliascntresetthe{proposition}

\newaliascnt{corollary}{theorem}
\newtheorem{corollary}[corollary]{Corollary}
\aliascntresetthe{corollary}

\theoremstyle{definition}
\newaliascnt{definition}{theorem}
\newtheorem{definition}[definition]{Definition}
\aliascntresetthe{definition}

\newaliascnt{assumption}{theorem}

\aliascntresetthe{assumption}

\newaliascnt{fact}{theorem}
\newtheorem{fact}[fact]{Fact}
\aliascntresetthe{fact}

\theoremstyle{remark}
\newaliascnt{remark}{theorem}
\newtheorem{remark}[remark]{Remark}
\aliascntresetthe{remark}

\crefname{theorem}{Theorem}{Theorems}
\crefname{lemma}{Lemma}{Lemmas}
\crefname{proposition}{Proposition}{Propositions}
\crefname{corollary}{Corollary}{Corollaries}
\crefname{definition}{Definition}{Definitions}
\crefname{assumption}{Assumption}{Assumptions}
\crefname{fact}{Fact}{Facts}
\crefname{remark}{Remark}{Remarks}
\crefname{equation}{Eq.}{Eqs.}
\crefname{table}{Table}{Tables}
\crefname{section}{Section}{Sections}
\crefname{appendix}{Appendix}{Appendices}

\DeclareMathOperator{\poly}{poly}

\DeclareMathOperator{\Tr}{Tr}
\DeclareMathOperator{\rank}{rank}

\newcommand{\Otil}{\widetilde{O}}   
\newcommand{\wt}{\widetilde}
\newcommand{\wh}{\widehat}
\newcommand{\R}{\mathbb{R}}
\newcommand{\N}{\mathbb{N}}
\newcommand{\E}{\mathbb{E}}
\renewcommand{\Pr}{\mathbb{P}}
\newcommand{\1}{\mathbf{1}}
\providecommand{\eps}{\varepsilon}                

\newcommand{\arxivonly}[1]{\ifdefined\isarxiv #1\fi}
\newcommand{\iclronly}[1]{\ifdefined\isarxiv\else #1\fi}

\newcommand{\norm}[1]{\left\| #1 \right\|}
\newcommand{\inner}[2]{\left\langle #1, #2 \right\rangle}

\newcommand{\toP}{\xrightarrow{\;\Pr\;}}
\newcommand{\tod}{\xrightarrow{\;d\;}}
\newcommand{\toas}{\xrightarrow{\;\mathrm{a.s.}\;}}

\newcommand{\Bin}{\mathrm{Bin}}
\newcommand{\Poisson}{\mathrm{Poisson}}
\newcommand{\Exp}{\mathrm{Exp}}
\newcommand{\PD}{\mathrm{PD}}

\newcommand{\Ens}{\mathcal{E}}
\newcommand{\Espike}{\Ens_{\mathrm{spike}}}

\newcommand{\Vmax}{V_{\max}}
\newcommand{\barPhi}{\overline{\Phi}}

\newcommand{\cov}{p}              
\newcommand{\bud}{k_\eta}         
\newcommand{\gap}{G}              
\newcommand{\cratio}{\rho_\eta}   

\newcommand{\certkv}{\textsc{CertKV}}

\ifdefined\arxivsubmission
\renewcommand{\certkv}{\textnormal{\textsc{CertKV}}}
\fi

\begin{document}

\ifdefined\isarxiv
\title{Beyond Sparse Weights: When Is Attention Compressible?}
\author{Chiwun Yang\thanks{\texttt{christiannyang37@gmail.com}. City University of Hong Kong.}
\and
Xiaoyu Li\thanks{\texttt{xiaoyu.li2@unsw.edu.au}. University of New South Wales.}}
\date{}
\maketitle
\begin{abstract}
KV-cache compression is often justified by attention maps with a few large
weights.  This is incomplete: large weights may not contain most of the mass,
omitted values can cancel, and preserving the attention output may not preserve
the task.  We separate these questions.  Global score gaps---not threshold
counts---determine how many tokens are needed to retain a target mass.  For a
realized row, the weighted sum of omitted values is the exact missing statistic.
A controlled retrieval--aggregation model explains when truncation helps and
when it hurts.  These results motivate \certkv{}, a training-free compressor
that reserves one tail-summary slot per head and allocates the rest by value
dispersion.  Under matched budgets, \certkv{} is top-two in seven of nine
LongBench-v2 settings, remains in the leading compressed tier on 128K
\textsc{RULER}, and realizes a ten-fold cache budget in a packed Llama
prototype.  Compressibility depends on the mass, values, future queries, and
task---not on a sparse-looking map alone.

\end{abstract}
\else
\title{Beyond Sparse Weights: When Is Attention Compressible?}
\author{Anonymous}   
\date{}
\maketitle
\begin{abstract}
KV-cache compression is often justified by attention maps with a few large
weights.  This is incomplete: large weights may not contain most of the mass,
omitted values can cancel, and preserving the attention output may not preserve
the task.  We separate these questions.  Global score gaps---not threshold
counts---determine how many tokens are needed to retain a target mass.  For a
realized row, the weighted sum of omitted values is the exact missing statistic.
A controlled retrieval--aggregation model explains when truncation helps and
when it hurts.  These results motivate \certkv{}, a training-free compressor
that reserves one tail-summary slot per head and allocates the rest by value
dispersion.  Under matched budgets, \certkv{} is top-two in seven of nine
LongBench-v2 settings, remains in the leading compressed tier on 128K
\textsc{RULER}, and realizes a ten-fold cache budget in a packed Llama
prototype.  Compressibility depends on the mass, values, future queries, and
task---not on a sparse-looking map alone.

\end{abstract}
\fi

\section{Introduction}\label{sec:intro}

Long-context decoding repeatedly reads a key--value (KV) cache that grows with
sequence length.  Cache reduction is therefore both a systems requirement and
a scientific claim about what state later queries need
\cite{dao2022flashattention,kwon2023pagedattention}.  Existing methods retain
tokens deemed important, or impose windows, sinks, per-head budgets, and trained
sparse patterns
\cite{zhang2023h2o,liu2023scissorhands,li2024snapkv,tang2024quest,
xiao2024streamingllm,feng2024adakv,cai2024pyramidkv,jiang2024minference}.
Their justification often starts from an attention map with a few bright
entries.  Yet a cache does not store a heat map: it stores state for future
queries and downstream computation.  The real question is what must be true
before sparse-looking attention licenses compression.

We separate four objects: the number of large weights, the number of tokens
needed for a target mass, the retained state needed to preserve an attention
output, and the state needed by the task.  Without further assumptions,
\[
\begin{aligned}
\text{count sparsity}&\not\Rightarrow\text{coverage sparsity},\\[-1mm]
\text{coverage sparsity}&\not\Rightarrow\text{output fidelity},\\[-1mm]
\text{output fidelity}&\not\Rightarrow\text{task utility}.
\end{aligned}
\]
The missing objects are, respectively, the global score-gap profile, the values
carried by omitted mass, and the task target.  A dim bulk can carry most of the
probability; substantial omitted mass can also cancel in value space; and exact
softmax can preserve noise that a retrieval task would prefer to discard.
Thus ``attention is sparse'' is incomplete until the metric and the required
contract are specified.

The first missing link is \emph{mass coverage}.  A threshold count records
visible entries; a cache budget asks how many retain, say, \(90\%\) of the
probability mass.  Fixed-scale i.i.d.\ sub-Gaussian scores can have a logarithmic
calibrated count, yet Gaussian scores at initialization scale require an
asymptotic fraction near \(0.61\) for \(90\%\) coverage---only \(1.6\times\)
compression.  Complementary all-rank conditions separate the regimes: a
log-rank lower gap envelope with slope above one makes the tail summable,
whereas a sub-unit upper envelope forces a growing budget.  The unit-slope
boundary depends on lower-order structure.  Peaks alone are not enough; the
whole plateau--slope geometry matters.

The second link is \emph{attention-output fidelity}.  Omitted mass is not an
error certificate: its values may cancel or reinforce.  For a fixed realized
row, the output factors into a retained contribution and the omitted weighted
value sum.  That sum is the unique additive correction.  Under the stronger
requirement that one linear sketch recover both the full output and every
retained sub-budget, the kept values plus one tail aggregate attain the matching
\((|S|+1)d\) dimension bound.

The third link is \emph{task utility}.  Exact attention is an estimator, not the
task target.  In a planted-token model, the informative token can enter
top-\(k\) before it dominates softmax; renormalization then suppresses a
diffuse noisy bulk and can win for retrieval-like targets, while exact
attention remains preferable for aggregation.  The controlled test recovers
this transition; the downstream test exposes the proxy's limits.

Compressibility is consequently not a scalar property of a weight histogram.
It is a contract among score geometry, value geometry, future queries, and the
task.  This view suggests a cache design: \certkv{} reserves one explicitly
charged tail-summary slot per KV head, ranks tokens from SnapKV's observation
window, allocates more real-token slots where the omitted values are harder to
summarize, and stores a weighted key--value tail summary.  Synthetic and real
slots obey the same physical budget.

\paragraph{Contributions.}
\begin{enumerate}[leftmargin=1.4em,label=\textbf{\arabic*.},
                  itemsep=2pt,topsep=3pt,parsep=0pt]
\item \textbf{Coverage geometry.}
Complementary all-rank gap conditions give a length-independent fixed-mass
budget above unit log-rank slope and polynomial growth below it.  Fixed-scale
i.i.d.\ Gaussian logits simultaneously exhibit a logarithmic calibrated count
and linear coverage.  A \(4\)K profile's failed \(128\)K prediction isolates
cross-length stability as a separate requirement
(\cref{thm:gap-envelope-dichotomy,fig:headline}).

\item \textbf{Fixed-row output fidelity.}
We identify the omitted weighted value sum as the unique fixed-row correction.
For the stronger linear-sketch requirement above, one tail aggregate also
attains the matching dimension bound; trained-model controls instantiate the
identity (\cref{thm:fixed-row-tail-factorization,tab:bandwidth-main}).

\item \textbf{Task-dependent utility.}
A planted-token analysis characterizes when renormalized top-\(k\) suppresses
a diffuse bulk and overtakes exact attention.  Experiments recover the
controlled transition and show why it is not a calibrated downstream rule
(\cref{thm:denoising-phase-law,fig:alpha}).

\item \textbf{Exact-budget compression.}
\certkv{} is evaluated on all \(503\) LongBench-v2 examples with three
backbones and three ratios, multi-length \textsc{RULER}, one-factor studies,
and a packed-cache prototype.  On the pre-specified four-baseline grid, its
compressed point estimate is top-two in seven of nine settings; at \(128\)K it
joins the leading \textsc{RULER} tier and, at Llama \(64\)K/\(10\times\),
realizes the requested physical cache reduction
(\cref{tab:lbv2-main-v2,tab:ruler-main-v2}).
\end{enumerate}



\ifdefined\arxivsubmission
\section{Related Work}\label{sec:related}

\paragraph{KV-cache compression.}
Training-free sparse decoding spans cumulative-score eviction
(H2O \cite{zhang2023h2o}, Scissorhands \cite{liu2023scissorhands}, and
TOVA \cite{oren2024tova}), observation-window selection
(SnapKV \cite{li2024snapkv}), query-aware page selection
(Quest \cite{tang2024quest}), block summaries (InfLLM
\cite{xiao2024infllm}), dynamic token pruning (SlimInfer
\cite{long2025sliminfer}), sink-plus-window streaming (StreamingLLM
\cite{xiao2024streamingllm}), and LSH-based sampling with variance guarantees
(MagicPIG \cite{chen2025magicpig}).  Other methods use values
(VATP \cite{guo2024vatp}), allocate budgets across heads
(Ada-KV \cite{feng2024adakv}), distinguish head types
(DuoAttention \cite{xiao2025duoattention}; retrieval heads
\cite{wu2024retrieval}), or compensate through merging
(D2O \cite{wan2024d2o}, WeightedKV \cite{weightedkv2025}, and KeepKV
\cite{keepkv2025}).  A broader survey appears in
\ifdefined\arxivsubmission
\citet{zhang2025efficientattnsurvey}.
\else
\cite{zhang2025efficientattnsurvey}.
\fi

Prefill and layer-wise methods select structure before or across decoding
(MInference \cite{jiang2024minference},
\ifdefined\arxivsubmission
ParallelComp \cite{xiong2025parallelcomp},
\fi
SampleAttention
\cite{zhu2024sampleattention}, SeerAttention \cite{gao2024seerattention},
HashAttention \cite{desai2024hashattention}, PyramidKV
\cite{cai2024pyramidkv}, and PyramidInfer \cite{yang2024pyramidinfer}).
System designs trade cache placement for bandwidth
(ShadowKV \cite{sun2024shadowkv} and Double Sparsity
\cite{yang2024doublesparsity}) on top of exact kernels and paged serving
\cite{dao2022flashattention,kwon2023pagedattention}, while multi- and
grouped-query attention reduce cache size architecturally
\cite{shazeer2019mqa,ainslie2023gqa}.  This design space descends from
training-time sparse patterns \cite{cgrs19,bpc20,zgd+20,kkl20,roy2021routing}
and kernel or low-rank approximations
\ifdefined\arxivsubmission
\cite{cld+20,katharopoulos2020linear,xzc21arxiv,qin2022cosformer}.
\else
\cite{cld+20,katharopoulos2020linear,xzc+21,qin2022cosformer}.
\fi
Trainable
sparsity co-designs the pattern and weights
\cite{yuan2025nsa,lu2025moba,ge2024fastgen} and is outside our scope.

\paragraph{Threshold counts and attention approximation.}
\citet{deng2024nsparse} study threshold-count sparsity in an
idealized one-layer model with independent bounded input entries.  Given
access to the largest attention entries, they derive \(n^C\)-scaled top-\(k\)
windows for approximating the attention matrix.  That analysis establishes a
count-based approximation result under its assumptions; it does not determine
the probability mass carried by the selected entries, the value-dependent
error in the attention output, or the utility of that output for a task.  We
make these three quantities explicit.  Our global gap criterion determines
fixed-row mass coverage (\cref{thm:gap-envelope-dichotomy}); our tail
factorization identifies the value statistic required for exact fixed-row
completion (\cref{thm:fixed-row-tail-factorization}); and our task law
characterizes when reproducing exact attention is preferable to renormalized
selection (\cref{thm:denoising-phase-law}).

These results analyze the premise shared by many cache selectors.  The
coverage theorem complements one-sided variance and dispersion analyses
\cite{chen2025magicpig,velickovic2025softmax}.  The fixed-row minimality result
applies when one linear sketch must recover every retained sub-budget; it does
not concern a single output, nonlinear summaries, or an unseen future-query
distribution.  The per-head growth classes also organize the heterogeneity
exploited by Ada-KV \cite{feng2024adakv}, PyramidKV
\cite{cai2024pyramidkv}, and DuoAttention \cite{xiao2025duoattention}.
Finally, the empirical observation that different sparse methods lead on
different tasks \cite{nawrot2025sparse} agrees with our task-conditional
analysis.

\paragraph{Softmax dispersion and logit scaling.}
\citet{velickovic2025softmax} prove that bounded
softmax logits disperse as the sequence grows and propose inference-time
temperature adaptation.  Scalable-Softmax \cite{nakanishi2025ssmax} and
entropy-invariant scaling \cite{su2021entropy} instead multiply logits by
\(\log n\)-dependent factors.  These interventions change the score scale and
therefore the sorted-logit gap profile that governs coverage.  Our proved
submission claim is the distribution-free row-wise statement:
a global lower envelope steeper than \(\log j\) gives a length-independent
coverage budget, while a global upper envelope below unit slope forces the
budget to grow with context length
(\cref{thm:gap-envelope-dichotomy}).  This criterion deliberately separates
coverage from concentration: a bounded normalized count or a large top weight
does not by itself determine the mass budget
(\cref{sub:count-versus-coverage,fig:headline}).

Related length-dependent rescalings appear in positional extrapolation
\cite{sal+24,press2022alibi,chen2023pi,peng2024yarn} and in analyses of
attention-entropy collapse \cite{zhai2023entropy,hong2025variance}.  The
fine-grained complexity threshold of \citet{as23neurips} is
analogous but distinct: it prices \(O(\sqrt{\log n})\) logit magnitude for
almost-linear-time approximation, whereas our criterion prices the global
log-rank gap slope needed for fixed-row mass coverage.

\paragraph{Theory of attention concentration.}
Modern Hopfield networks \cite{ramsauer2021hopfield} study when an attention
update retrieves a stored pattern rather than a metastable average.  Our
planted-token analysis gives a statistical complement: once the signal enters
the retained set but before it dominates the full softmax, renormalized
selection can suppress a noisy bulk
(\cref{thm:denoising-phase-law}).  Random Energy Model and
Poisson--Dirichlet laws \cite{derrida1981random,bovier2006statistical} provide
classical machinery for analyzing condensed weights; the three submission
proofs do not rely on that additional asymptotic route.  Rank collapse across
depth \cite{dong2021attention} concerns compositions of layers and is
orthogonal to our fixed-row summary question.  Likewise, recent empirical
``lossless'' top-\(k\) claims \cite{ruizwilliams2026condensate} concern
trained values and future queries; our exact-completion statement is only for
the specified fixed-row additive statistic
(\cref{thm:fixed-row-tail-factorization}).  Sparsemax
\cite{martins2016sparsemax} and \(\alpha\)-entmax
\cite{peters2019entmax} change the attention operator, whereas we analyze
standard softmax.

A separate complexity literature prices attention computation through
kernel, tensor, dynamic, and streaming approximations
\cite{zhdk23,hjk+23,kmz23,as24_iclr,cam+24,as24_arxiv,bsz23,alsy23}, and
contextual sparsity is exploited at inference by
\ifdefined\arxivsubmission
\citet{lwd+23}.
\else
\cite{lwd+23}.
\fi
Mechanistic
studies document induction heads \cite{olsson2022induction}, massive
activations \cite{sun2024massive}, and attention sinks \cite{gu2025sink};
learned token dropping \cite{anagnostidis2023dynamic} and max-margin
convergence \cite{tarzanagh2023maxmargin} describe how training can install
sharp selection.  Closest to our score-scale discussion,
\citet{chen2026focusdilution} show in a one-layer Markov model that
attention scale grows only after embeddings condense to rank one.  Their
parameter-level condensation differs from our realized-row criterion, but
provides a training-dynamics counterpart to the length-growing score geometry
excluded by a fixed-weight, bounded-input model
(\cref{fac:score-scale}).  These works establish that concentration emerges
\cite{gu2025sink,olsson2022induction} or exploit it as a premise
\cite{zhang2023h2o,lwd+23}; they do not separate a calibrated large-weight
count from the generally different budget required for mass coverage.

\fi
\section{From Sparse Weights to Attention Compressibility: Three Separations}
\label{sec:rethinking}
\label{sec:prelim}

``Sparse attention'' can refer to four different properties: a small threshold
count, a small fixed-mass support, a compact representation of the attention
output, or preservation of the downstream task.  Without additional
structure, none determines the next:
\[
\begin{aligned}
\text{count}&\not\Rightarrow\text{coverage}
\not\Rightarrow\text{output fidelity},\\[-1mm]
\text{output fidelity}&\not\Rightarrow\text{task utility}.
\end{aligned}
\]
This section identifies what each arrow is missing.  Global score gaps govern
mass coverage, omitted values govern output fidelity, and the task target
governs whether fidelity is useful.  The result is a set of quantities that a
cache-compression argument can state, measure, or model.

\paragraph{One-row setup.}
Fix one realized query attending over logits
\(\ell_1,\ldots,\ell_n\) and values \(V_1,\ldots,V_n\in\mathbb R^d\), with
\(\max_j\lVert V_j\rVert_2\le V_{\max}\).  Write
\[
A_j=\frac{e^{\ell_j}}{\sum_t e^{\ell_t}},
\qquad
O=\sum_j A_jV_j.
\]
In a transformer, conditioning on query row \(i\) gives the input-to-score
bridge
\begin{equation}
\ell_j=\langle u_i,X_j\rangle,
\qquad
u_i=\frac{W_KW_Q^\top X_i^\top}{\sqrt d}.
\label{eq:score-bridge}
\end{equation}
\begin{fact}[Fixed weights do not create length-growing score scale]
\label{fac:score-scale}
If \(\lVert W_Q\rVert_{\mathrm{op}},\lVert W_K\rVert_{\mathrm{op}}\le W\)
and \(\lVert X_i\rVert_2\le B\), then
\(\lVert u_i\rVert_2\le W^2B/\sqrt d\).  If, conditional on \(X_i\), each key
\(X_j\) is centered and \(\kappa^2\)-sub-Gaussian in every direction, then
\(\ell_j\mid X_i\) is centered
\(\kappa^2\lVert u_i\rVert_2^2\)-sub-Gaussian.  Thus fixed
\(B,W,\kappa,d\) prevent the logit scale from growing with context length;
they do not imply a favorable gap profile.
\end{fact}

Let \(\pi\) order the logits nonincreasingly, breaking ties by index, and write
\(\ell_{(j)}=\ell_{\pi(j)}\) and \(A_{(j)}=A_{\pi(j)}\).  Define the gap profile
\begin{equation}
G(j)=\ell_{(1)}-\ell_{(j)},
\label{eq:gap-def}
\end{equation}
and let
\begin{equation}
K_\eta=\min\Bigl\{k\in[n]:\sum_{j=1}^kA_{(j)}\ge1-\eta\Bigr\}.
\label{eq:coverage-def}
\end{equation}
The count of large weights is a different quantity.
\begin{definition}[Calibrated count sparsity]
\label{def:abs-sparse}
\label{def:main-relative-sparse}
For \(\epsilon>0\), define
\(k(\epsilon)=\#\{j\in[n]:A_j>\epsilon\}\).  A row is
\((\epsilon,k)\)-count-sparse when \(k(\epsilon)\le k\).
\end{definition}
A realized count is informative only between the uniform floor and
\(A_{(1)}\).  The first result below is a conservative simultaneous upper
certificate and may be empty; its Gaussian clause and the subsequent
relative-threshold law supply non-vacuous counts.
\begin{remark}[Threshold scope]
\label{rem:eps-window}
Writing a threshold merely as \(n^{-1+o(1)}\) is insufficient.  The formal
i.i.d.\ result uses
\(\epsilon\ge n^{-1}\exp\{O(\sqrt{\log(n/\delta)})\}\); below that calibrated
window even a uniform row can have a linear count.  Its conservative threshold
need not lie below the realized maximum weight, so non-vacuity must be proved
separately rather than inferred from the \(n^{-1+o(1)}\) notation.
\end{remark}

\begin{proposition}[Calibrated logarithmic count and a non-vacuous Gaussian level]
\label{prop:calibrated-log-count}
Fix \(\sigma>0\).  For each \(n\ge2\), let
\(\ell_1,\ldots,\ell_n\) be independent, centered,
\(\sigma^2\)-sub-Gaussian logits, and fix \(\delta\in(0,1)\).  Set
\begin{align*}
S_\delta&=\sigma\sqrt{2\log(2n/\delta)},\\[-2pt]
\epsilon_\delta(n)&=\frac1n
\exp\!\left\{\sigma\sqrt{2\log(n/\delta)}+S_\delta\right\}.
\end{align*}
With probability at least \(1-2\delta\), simultaneously for every
\(\epsilon\ge\epsilon_\delta(n)\),
\[
k(\epsilon)\le
\frac23\log\frac{2n}{\delta}
+\sqrt{2\log\frac{2n}{\delta}}+2.
\]
Thus, at fixed \((\sigma,\delta)\), the calibrated threshold is
\(n^{-1+o(1)}\) and the large-weight count is \(O(\log n)\).  This is a
possibly empty upper certificate; it neither controls retained mass nor
licenses a cache budget.

If additionally
\(\ell_j\stackrel{\mathrm{iid}}{\sim}\mathcal N(0,\sigma^2)\), define
\(m_n=\log n\),
\(\tau_n=\sigma\Phi^{-1}(1-m_n/n)\), and
\(\epsilon_n=n^{-1}\exp(\tau_n-\sigma^2/2)\).  Then
\(\epsilon_n=n^{-1+o(1)}\) and
\(k(\epsilon_n)/\log n\xrightarrow{\Pr}1\).
\end{proposition}

\begin{proposition}[Relative-threshold count under an additive near-top law]
\label{prop:relative-count-law}
Let \(b_n=\sqrt{2\log n}\).  Suppose that for some \(\sigma_s>0\) and every
fixed \(C>0\),
\[
\sup_{1\le j\le\lfloor\exp(Cb_n)\rfloor}
\left|
\ell_{(j)}-\sigma_s\sqrt{2\log(n/j)}
\right|\xrightarrow{\Pr}0.
\]
Then, for every fixed \(\epsilon\in(0,1)\) and \(t=\log(1/\epsilon)\),
\[
\frac{\log k(\epsilon A_{(1)})}{\sqrt{2\log n}}
\xrightarrow{\Pr}\frac{t}{\sigma_s},
\qquad
\frac{\log k(\epsilon A_{(1)})}{\log n}\xrightarrow{\Pr}0.
\]
The additive precision is load-bearing: first-order tails alone do not imply
this non-vacuous conditional law.
\end{proposition}

\subsection{Count sparsity does not imply coverage sparsity}
\label{sub:count-versus-coverage}

A threshold count looks only at entries that cross one horizontal level.
Coverage instead adds the whole ranked tail until it reaches a prescribed
mass.  Softmax monotonicity makes the top-\(K_\eta\) set coverage-optimal, and
\(A_{(j)}/A_{(1)}=e^{-G(j)}\) turns this cumulative question into a
summability test for the ranked gaps.

\begin{theorem}[Coverage regimes from global gap envelopes]
\label{thm:gap-envelope-dichotomy}
Fix \(\eta\in(0,1)\), with the envelope parameters below independent of
\(n\).  For one realized row, suppose the \emph{global pointwise} envelope
\(G(j)\ge\gamma\log j-b\) holds at every rank for \(b\ge0\).  If
\(\gamma>1\), then
\[
K_\eta\le
\left\lceil
\left(\frac{e^b}{(\gamma-1)\eta}\right)^{1/(\gamma-1)}
\right\rceil ,
\]
independently of \(n\).  On the other hand, if
\(G(j)\le\gamma'\log j+b'\) for every rank, with \(b'\ge0\), and
\(0\le\gamma'<1\), then
\[
K_\eta\ge
\frac{(1-\eta)e^{-b'}}{2}\,n^{1-\gamma'}.
\]
At \(\gamma'=0\), \(K_\eta=\Theta(n)\).
\end{theorem}

The theorem has a direct interpretation.  If the ranked logits fall faster
than \(\log j\), their exponentiated tail is summable and a fixed cache budget
can retain the target mass.  If they fall more slowly, the required budget
must grow with context length.  The proof is the \(p\)-series test applied to
\(e^{-G(j)}\), with matching tail lower bounds on the flat side.  The
all-rank quantifier is essential: a slope fitted on a finite range describes
that range only.  A leading plateau of size \(J\) multiplies the sparse-side
budget by \(J\), while lower-order terms determine the boundary
\(\gamma=1\).  Self-contained proofs of the three headline results are in
\cref{sec:submission-theory-proofs}.

Count calibration supplies the contrast.  The conservative sub-Gaussian
certificate in \cref{prop:calibrated-log-count} gives a simultaneous
logarithmic upper bound, while its Gaussian-specific deterministic threshold
has \((1+o_\Pr(1))\log n\) exceedances and is non-vacuous.  Geometric mixing
weakens the available generic upper bound to a larger sublinear rate.  Yet for
i.i.d.\ \(\mathcal N(0,\sigma^2)\) logits the asymptotic kept fraction is
\[
  \frac{K_\eta}{n}
  \xrightarrow{\Pr}
  \Phi\!\left(\Phi^{-1}(1-\eta)-\sigma\right).
\]
At \(\sigma=1\) and \(\eta=0.1\), it is about \(0.61\): logarithmic
large-weight count coexists with only \(1.6\times\) coverage compression.
Length-growing score scale can improve coverage under additional tail-shape
assumptions, but scale alone does not imply a universal coverage budget and the
critical boundary remains sensitive to lower-order structure.

\paragraph{Connection to measured attention profiles.}
The relative-weight count
\(\hat k_{\rm rel}=\#\{j:A_j>0.1A_{(1)}\}\) has estimated sublinear growth over
the measured lengths of the multi-length language-model anchors.  This is the
observable described by
\cref{prop:relative-count-law} when its Gaussian-type order-statistic
assumption holds; it differs from the calibrated absolute threshold in
\cref{prop:calibrated-log-count}.  To test whether a short-context profile
provides a reusable cache budget, we fit the profile at \(4\)K and predict the
\(128\)K coverage requirement without refitting.  The predicted and measured
budgets differ by more than a factor of \(1.5\) (\cref{fig:headline}).  Thus a
short-context profile must demonstrate stability before it is reused as a
long-context cache budget.  This does not contradict the all-rank theorem: a
\(4\)K fit does not establish its global envelope.

\subsection{Coverage does not determine output fidelity}
\label{sub:learned-approximability}

For a nonempty proper retained set \(S\subset[n]\), let
\(T=[n]\setminus S\) and \(p_S=\sum_{j\in S}A_j\).  The truncated output, its
renormalized counterpart, and the conditional tail mean are
\begin{equation}
\begin{aligned}
O_S&=\sum_{j\in S}A_jV_j,
&\widetilde O_S&=\sum_{j\in S}\frac{A_j}{p_S}V_j,\\
\bar V_T&=\sum_{j\in T}\frac{A_j}{1-p_S}V_j.
\end{aligned}
\label{eq:schemes}
\end{equation}
Coverage controls only the scalar \(p_S\).  It does not say whether the
omitted values reinforce one another, cancel, or point in a direction already
represented by the retained set.  The next identity exposes this missing
quantity exactly.

\begin{theorem}[Fixed-row tail factorization]
\label{thm:fixed-row-tail-factorization}
For every fixed realized row and nonempty proper retained set,
\begin{align}
O&=p_S\widetilde O_S+(1-p_S)\bar V_T,
\label{eq:rethinking-factor}\\
\widetilde O_S-O
&=(1-p_S)(\widetilde O_S-\bar V_T),
\label{eq:rethinking-bias}\\
\lVert\widetilde O_S-O\rVert_2
&\le2V_{\max}(1-p_S).
\label{eq:rethinking-mass-bound}
\end{align}
The unique \(u\in\mathbb R^d\) satisfying \(O_S+u=O\) is
\[
u^\star=\sum_{j\in T}A_jV_j=(1-p_S)\bar V_T.
\]
\end{theorem}

The identities follow by splitting kept and omitted terms.  They are tight:
the same omitted mass can either cancel or create constant error, depending on
the values.  Under the explicitly stronger requirement that a \emph{linear}
sketch recover the full output and every retained sub-budget, the kept values
plus one tail aggregate also meet a matching \((|S|+1)d\) rank floor
(\cref{thm:factor-map}).  That scoped statement is not a lower bound for one
output, nonlinear compression, or a future-query KV cache.

\paragraph{From a realized row to later queries.}
The theorem is exact only after the attention weights of one row are known.
A cache evictor acts before later queries reveal their weights.  Our
future-query result gives an a posteriori bound separating two errors:
synthetic tail-mass mismatch and conditional tail-distribution mismatch.
Computing these quantities still requires the evicted keys
(\cref{thm:future-query-lift}).  \Cref{sec:certkv} therefore uses the
fixed-row result as a design principle and approximates the unavailable
future-query quantities at eviction time.

\subsection{Output fidelity does not determine task utility}
\label{sub:task-ordering}

Approximating \(O\) is appropriate when the task target is the full aggregate.
It can be wrong when the bulk acts as noise around a planted or retrieval-like
value.  For target \(y\), use
\(L(\widehat O)=\lVert\widehat O-y\rVert_2\), and compare exact attention,
renormalized top-\(k\), and the value-aware oracle set minimizing this loss.

\paragraph{Planted-token setup.}
For each \(n\), one token \(j^\star\) has logit \(\mu_n\) and fixed value
\(V_{j^\star}\).  The other logits are i.i.d.\
\(\mathcal N(0,\sigma^2)\) for fixed \(\sigma>0\); their values are i.i.d.,
independent of the logits, with mean \(\bar V_b\) and second moment
\(\mathbb E\lVert V_j-\bar V_b\rVert_2^2\le\sigma_V^2\) for fixed
\(\sigma_V>0\).  Write
\(D=\lVert V_{j^\star}-\bar V_b\rVert_2>0\).  The retained budget and planted
logit satisfy
\[
k_n\to\infty,\qquad
\log k_n=o(\sqrt{\log n}),
\]
\[
\sigma\sqrt{2\log n}+\log k_n+C
\le\mu_n\le(1-\varepsilon)\log n.
\]
Here \(C>0\) and \(\varepsilon\in(0,1)\) are fixed.  Let \(S_{k_n}\) be the
top-\(k_n\) set, and define
\[
\begin{aligned}
\widetilde A_j&=
\frac{A_j\mathbf 1\{j\in S_{k_n}\}}
{\sum_{t\in S_{k_n}}A_t},
&\widetilde O_{S_{k_n}}&=\sum_j\widetilde A_jV_j,\\
w_e&=A_{j^\star},
&w_s&=\widetilde A_{j^\star}.
\end{aligned}
\]
The retained bulk is \emph{diffuse} when
\[
\sum_{j\in S_{k_n}\setminus\{j^\star\}}
\left(\frac{\widetilde A_j}{1-w_s}\right)^2
=O_\Pr(k_n^{-1}).
\]

\begin{theorem}[Task ordering in the planted-token regime]
\label{thm:denoising-phase-law}
Under the setup above, assume the retained bulk is diffuse.  Then:
\begin{enumerate}[label=(\roman*),leftmargin=1.45em,itemsep=2pt,topsep=2pt]
\item With probability tending to one, \(j^\star\in S_{k_n}\),
\(w_s=\Theta_\Pr(1)\), and
\(w_e\le n^{-\varepsilon+o_\Pr(1)}\).
\item For \(y_\alpha=\alpha V_{j^\star}+(1-\alpha)\bar V_b\),
\begin{align*}
L(O)&=|\alpha-w_e|D+O_\Pr(\sigma_Vn^{-1/2}),\\
L(\widetilde O_{S_{k_n}})
&=|\alpha-w_s|D+O_\Pr(\sigma_Vk_n^{-1/2}).
\end{align*}
\item Let \(\alpha_n^\star=(w_e+w_s)/2\) and
\(\Delta_n(\alpha)=\bigl||\alpha-w_e|-|\alpha-w_s|\bigr|\).  For any
deterministic or logit-measurable \(\alpha_n\in[0,1]\) satisfying
\[
\frac{D\Delta_n(\alpha_n)}
{\sigma_V(n^{-1/2}+k_n^{-1/2})}
\xrightarrow{\Pr}\infty,
\]
exact attention has smaller loss below \(\alpha_n^\star\), and renormalized
top-\(k_n\) has smaller loss above it, with probability tending to one.
\end{enumerate}
\end{theorem}

The value-aware oracle is defined by minimizing over all \(k_n\)-subsets, so,
deterministically,
\[
\min_{S:\,|S|=k_n}L(\widetilde O_S)
\le L(\widetilde O_{S_{k_n}}).
\]

Gaussian maxima establish that the planted token is detected, while the
lognormal law of large numbers and retained-bulk diffuseness control the two
value averages.  The result isolates a precise denoising mechanism: truncation
can help after a signal becomes rank-visible but before full softmax gives it
enough mass.  Rank inclusion alone does not imply the diffuseness condition,
so applying the law to a deployed head requires checking that additional
property.

\paragraph{Connection to downstream tasks.}
The controlled mixture exhibits the predicted single flip.  We then asked
whether an observable estimate of the same mixture parameter predicts which
method is more accurate on downstream retrieval and aggregation tasks.  The
estimated parameter varies, but the exact--top-\(k\) accuracy difference does
not change sign as the model predicts; on the frozen dense model, the two
methods remain statistically indistinguishable.  Thus the controlled result
isolates a task-dependent mechanism, while the downstream measurement shows
that one mixture coordinate does not describe real tasks adequately.
Separately, trained sparse attention shows a finite
extrapolation advantage at \(4\times\) and \(8\times\), not at \(16\times\)
or \(32\times\) (\cref{fig:alpha}).  This is an independent observation, not
a validation of the planted-token model.

\paragraph{Design implications.}
The three results suggest a concrete workflow for cache compression.  Estimate
the current row's coverage from its full score profile; retain information
about the values carried by the omitted tail; model how later queries will
reweight that tail; and evaluate the resulting state on the target task.  The
next section instantiates these principles under an exact physical cache
budget.

\section{\certkv{}: Value-Aware Compression under Exact Budgets}
\label{sec:certkv}

The preceding results yield a design rule, not an end-to-end certificate:
retain score-relevant tokens, preserve information about omitted values, and
charge every stored summary to the physical budget.  \certkv{} implements this
rule by combining SnapKV's observation-window score \cite{li2024snapkv} with
value-dispersion allocation and one synthetic KV pair per head.  These three
choices determine which tokens remain, how the shared budget is split, and how
the tail is represented.  Here ``Cert'' refers to exact slot accounting; it
does not denote an unconditional future-query guarantee.

\paragraph{From the fixed-row decomposition to a cache design.}
The fixed-row factorization identifies the weighted tail value discarded by
subset-only eviction (\cref{thm:fixed-row-tail-factorization}).  Later-query
weights are unavailable at eviction time, so \certkv{} estimates this
statistic from the observation window and stores it as one synthetic KV pair.
The slot budget is exact; the scorer, estimator, and cross-head allocation
remain training-free approximations evaluated below.

\subsection{Dispersion-guided allocation with exact slot accounting}

For head \(h\), sort the \(L\) real tokens by the scorer and normalize its
scores to positive importance weights \(\pi_{h,j}\) that sum to one.  For a
candidate \(0\le b\le L-1\), let \(E_h(b)\) be the evicted suffix.  Its total
importance and normalized tail weights are
\[
m_h(b)=\sum_{j\in E_h(b)}\pi_{h,j},
\qquad
r_{h,j}(b)=\frac{\pi_{h,j}}{m_h(b)}.
\]
Thus \(r_h(b)\) is the conditional importance distribution within the evicted
tail.  Its key and value centroids form the synthetic pair:
\[
\bigl(k_h^\star(b),v_h^\star(b)\bigr)
=\mathbb E_{j\sim r_h(b)}
\bigl[(k_{h,j},v_{h,j})\bigr].
\]
We allocate by the mass-weighted spread of the omitted values,
\begin{equation}
c_h(b)=m_h(b)
\sqrt{\mathbb E_{j\sim r_h(b)}
\bigl[\lVert v_{h,j}-v_h^\star(b)\rVert_2^2\bigr]}.
\label{eq:certkv-curve}
\end{equation}
This is the tight envelope within the declared
mass--centroid--second-moment summary class
(\cref{thm:cert-stat-lb}).  Intuitively, a head receives more slots when the
tail carries more mass or is harder to summarize by one value.

With \(H\) KV heads and requested evicted fraction \(\rho\), every method
receives \(B_{\mathrm{live}}=H\lfloor L(1-\rho)\rfloor\) live slots.  Because
\certkv{}
charges one aggregate per head, its real-token allocation satisfies
\begin{align*}
B_{\mathrm{real}}&=B_{\mathrm{live}}-H,\\[-2pt]
\sum_h b_h&=B_{\mathrm{real}},\\[-2pt]
b_h&\ge
\max\!\left\{1,\left\lfloor0.2\,B_{\mathrm{real}}/H\right\rfloor\right\}.
\end{align*}
It selects per-head minimizers of \(c_h(b)+\lambda b\), then applies
deterministic best-marginal repair until the equality budget holds.  This
construction enforces the global slot budget exactly while adapting the
per-head allocation to the measured tail dispersion.

The method keeps \(b_h\le L-1\), so \(m_h(b_h)>0\), and writes
\((k_h^\star(b_h),v_h^\star(b_h))\) into the charged slot.  This synthetic pair
is the eviction-time estimate of the future conditional tail.
For realized weights, the exact \(d\)-dimensional tail aggregate is sufficient
and minimal among linear representations supporting every sub-budget
(\cref{thm:fixed-row-tail-factorization,thm:factor-map}).  The synthetic pair
uses this structure as a later-query approximation; implementation details
are in the appendix.

\subsection{Matched-budget evaluation and one-factor ablations}
\label{sub:certkv-results}

The pre-specified \(3\times3\) grid uses all \(503\) aligned LongBench-v2
examples per setting (inference seed \(42\); bootstrap seed \(0\), \(B=5000\)).
Among the four matched-budget baselines in this grid, \certkv{} has a top-two
compressed point estimate in seven of nine settings
(\cref{tab:lbv2-main-v2}).  After Holm correction across \(36\) paired
comparisons, the only decisive difference is Llama at \(2\times\) versus
PyramidKV: \(+8.75\) points, 95\% paired CI \([3.58,13.92]\), exact McNemar
\(p_{\mathrm{Holm}}=0.0407\).  We therefore read the grid as evidence for a
competitive operating point, not a universal method ranking.

\begin{table*}[t]
\vspace{-1mm}
\centering
\begingroup
\small
\setlength{\tabcolsep}{4.5pt}
\renewcommand{\arraystretch}{0.86}
\begin{adjustbox}{max width=.96\textwidth}
\begin{tabular}{@{}llcccc@{}}
\toprule
Model & Method & \(2\times\) & \(4\times\) & \(10\times\) & Avg. \\
\midrule
Llama-3.1-8B & Full cache & \(26.84{\pm}3.88\) & \(26.84{\pm}3.88\) & \(26.84{\pm}3.88\) & \(26.84{\pm}3.88\) \\
 & SnapKV & \(\mathbf{27.04{\pm}3.88}\) & \(25.65{\pm}3.78\) & \(23.46{\pm}3.68\) & \(25.38{\pm}3.51\) \\
 & StreamingLLM & \(\mathbf{27.04{\pm}3.88}\) & \(\mathbf{27.24{\pm}3.88}\) & \(\mathbf{26.04{\pm}3.78}\) & \(\mathbf{26.77{\pm}3.55}\) \\
 & PyramidKV & \(18.09{\pm}3.28\) & \(18.89{\pm}3.48\) & \(18.29{\pm}3.38\) & \(18.42{\pm}3.02\) \\
 & Ada-KV & \(26.64{\pm}3.88\) & \(25.25{\pm}3.88\) & \(\underline{24.06{\pm}3.68}\) & \(25.31{\pm}3.45\) \\
 & CertKV & \(\underline{26.84{\pm}3.88}\) & \(\underline{26.04{\pm}3.78}\) & \(\underline{24.06{\pm}3.68}\) & \(\underline{25.65{\pm}3.48}\) \\
\midrule
Qwen3-8B & Full cache & \(33.20{\pm}4.08\) & \(33.20{\pm}4.08\) & \(33.20{\pm}4.08\) & \(33.20{\pm}4.08\) \\
 & SnapKV & \(\mathbf{32.41{\pm}4.08}\) & \(29.62{\pm}3.98\) & \(\mathbf{30.22{\pm}3.98}\) & \(\mathbf{30.75{\pm}3.58}\) \\
 & StreamingLLM & \(29.62{\pm}3.98\) & \(29.22{\pm}3.98\) & \(\underline{29.62{\pm}3.98}\) & \(29.49{\pm}3.58\) \\
 & PyramidKV & \(28.03{\pm}3.88\) & \(28.03{\pm}3.88\) & \(28.03{\pm}3.88\) & \(28.03{\pm}3.64\) \\
 & Ada-KV & \(\underline{31.61{\pm}4.08}\) & \(\mathbf{30.82{\pm}4.08}\) & \(28.43{\pm}3.88\) & \(\underline{30.28{\pm}3.68}\) \\
 & CertKV & \(31.01{\pm}4.08\) & \(\underline{30.02{\pm}3.98}\) & \(29.03{\pm}3.98\) & \(30.02{\pm}3.68\) \\
\midrule
Mistral-7B & Full cache & \(28.03{\pm}3.88\) & \(28.03{\pm}3.88\) & \(28.03{\pm}3.88\) & \(28.03{\pm}3.88\) \\
 & SnapKV & \(27.24{\pm}3.88\) & \(\mathbf{28.43{\pm}3.88}\) & \(26.44{\pm}3.78\) & \(27.37{\pm}3.55\) \\
 & StreamingLLM & \(\mathbf{28.83{\pm}3.88}\) & \(28.03{\pm}3.88\) & \(25.05{\pm}3.78\) & \(27.30{\pm}3.64\) \\
 & PyramidKV & \(22.66{\pm}3.68\) & \(21.87{\pm}3.58\) & \(22.27{\pm}3.58\) & \(22.27{\pm}3.25\) \\
 & Ada-KV & \(27.83{\pm}3.88\) & \(27.44{\pm}3.88\) & \(\underline{27.04{\pm}3.88}\) & \(\underline{27.44{\pm}3.55}\) \\
 & CertKV & \(\underline{28.03{\pm}3.88}\) & \(\underline{28.23{\pm}3.88}\) & \(\mathbf{28.23{\pm}3.88}\) & \(\mathbf{28.16{\pm}3.58}\) \\
\bottomrule
\end{tabular}

\end{adjustbox}
\endgroup
\caption{\textbf{LongBench-v2 under matched KV budgets.}
Accuracy (\%) on 503 aligned examples per setting; entries are point estimate
\({\pm}\) 95\% example-bootstrap half-width
(\(B=5000\); fixed checkpoints/configurations).  Avg.\ averages ratios; Full
is unranked; bold/underline mark the best/second distinct compressed estimates.}
\label{tab:lbv2-main-v2}
\label{tab:lbv2}
\vspace{-1mm}
\end{table*}

At \(128\)K \textsc{RULER}, \certkv{} and Ada-KV form the leading compressed
tier: \(73.6{\pm}17.9\) versus \(73.2{\pm}18.0\) on Llama, and
\(61.5{\pm}21.0\) versus \(60.2{\pm}21.0\) on Qwen
(\cref{tab:ruler-main-v2}).  The \certkv{} entries are the strongest
compressed point estimates, but the task-bootstrap uncertainty is wide.

\begin{table*}[t]
\vspace{-1mm}
\centering
\begingroup
\small
\setlength{\tabcolsep}{3.5pt}
\renewcommand{\arraystretch}{0.86}
\begin{adjustbox}{max width=.96\textwidth}
\begin{tabular}{@{}llccccc@{}}
\toprule
Model & Method & 16K & 32K & 64K & 128K & Avg. \\
\midrule
\textsc{Llama-3.1-8B} & Full cache & $90.8{\pm}8.3$ & $83.3{\pm}14.1$ & $79.9{\pm}16.5$ & $72.2{\pm}18.4$ & $81.6{\pm}7.4$ \\
 & SnapKV & $\underline{90.7{\pm}8.6}$ & $83.3{\pm}13.8$ & $79.2{\pm}16.4$ & $71.7{\pm}18.2$ & $\underline{81.2{\pm}7.4}$ \\
 & StreamingLLM & $48.4{\pm}10.4$ & $45.7{\pm}10.3$ & $49.7{\pm}11.4$ & $42.0{\pm}10.4$ & $46.4{\pm}5.3$ \\
 & PyramidKV & $87.1{\pm}9.5$ & $\mathbf{84.1{\pm}12.7}$ & $\underline{79.6{\pm}16.2}$ & $68.9{\pm}17.4$ & $79.9{\pm}7.2$ \\
 & Ada-KV & $\mathbf{90.8{\pm}8.5}$ & $\underline{83.5{\pm}13.9}$ & $\underline{79.6{\pm}16.4}$ & $\underline{73.2{\pm}18.0}$ & $\mathbf{81.8{\pm}7.4}$ \\
 & \certkv{} (ours) & $\mathbf{90.8{\pm}8.5}$ & $83.0{\pm}14.1$ & $\mathbf{79.9{\pm}16.2}$ & $\mathbf{73.6{\pm}17.9}$ & $\mathbf{81.8{\pm}7.4}$ \\
\midrule
\textsc{Qwen3-8B} & Full cache & $80.9{\pm}17.0$ & $79.6{\pm}17.1$ & $73.4{\pm}17.1$ & $61.0{\pm}22.0$ & $73.7{\pm}9.2$ \\
 & SnapKV & $\mathbf{81.7{\pm}15.8}$ & $78.6{\pm}17.0$ & $71.6{\pm}17.5$ & $55.9{\pm}21.7$ & $72.0{\pm}9.1$ \\
 & StreamingLLM & $46.2{\pm}11.3$ & $46.8{\pm}10.4$ & $47.2{\pm}12.9$ & $39.2{\pm}18.2$ & $44.9{\pm}6.8$ \\
 & PyramidKV & $77.2{\pm}16.4$ & $78.0{\pm}15.6$ & $69.2{\pm}16.5$ & $52.9{\pm}21.1$ & $69.3{\pm}9.0$ \\
 & Ada-KV & $\underline{81.5{\pm}16.4}$ & $\mathbf{79.9{\pm}16.6}$ & $\underline{72.8{\pm}17.2}$ & $\underline{60.2{\pm}21.0}$ & $\underline{73.6{\pm}9.0}$ \\
 & \certkv{} (ours) & $\underline{81.5{\pm}16.4}$ & $\underline{79.7{\pm}16.8}$ & $\mathbf{73.0{\pm}17.2}$ & $\mathbf{61.5{\pm}21.0}$ & $\mathbf{73.9{\pm}9.0}$ \\
\midrule
\textsc{Mistral-7B} & Full cache & $87.0{\pm}9.3$ & -- & -- & -- & $87.0{\pm}9.3$ \\
 & SnapKV & $71.9{\pm}13.6$ & -- & -- & -- & $71.9{\pm}13.6$ \\
 & StreamingLLM & $50.5{\pm}9.6$ & -- & -- & -- & $50.5{\pm}9.6$ \\
 & PyramidKV & $70.4{\pm}14.4$ & -- & -- & -- & $70.4{\pm}14.4$ \\
 & Ada-KV & $\underline{77.4{\pm}11.6}$ & -- & -- & -- & $\underline{77.4{\pm}11.6}$ \\
 & \certkv{} (ours) & $\mathbf{79.5{\pm}11.2}$ & -- & -- & -- & $\mathbf{79.5{\pm}11.2}$ \\
\bottomrule
\end{tabular}

\end{adjustbox}
\endgroup
\caption{\textbf{\textnormal{\textsc{RULER}} across context lengths at \(2\times\).}
Accuracy (\%) for shared denominators; entries report the point estimate
\({\pm}\) 95\% task-bootstrap half-width
(\(B=5000\), fixed checkpoints and configurations).  At 128K, all methods use
3\% subsets and Qwen uses YaRN.  Avg.\ averages available lengths
(Mistral: 16K only).  Full is unranked; styling marks compressed point ranks,
not significance.}
\label{tab:ruler-main-v2}
\label{tab:ruler-128k}
\label{tab:ruler}
\vspace{-1mm}
\end{table*}

The one-factor study in \cref{tab:certkv-ablation-v2} varies the tail statistic,
allocator, summary, and safeguard.  Block-average point estimates span
\(24.3\%\)--\(25.1\%\); none of the nine adjusted paired contrasts is decisive.
This is local descriptive insensitivity, not statistical equivalence.  We
retain centered dispersion with deterministic repair because it directly
implements the value-heterogeneity objective while preserving slot equality.

The packed Llama prototype realizes the nominal \(10\times\) persistent-KV
budget.  At \(64\)K, it reduces storage from \(8.59\) to \(0.86\) GB and
decode peak memory from \(23.11\) to \(16.34\) GiB relative to the full cache
(\cref{tab:physical-cache}).  This validates physical accounting; it
does not establish latency gains or a memory advantage over other methods at
the same budget.

The \textsc{RULER} result uses all nine model--length settings with fixed task
denominators; historical no-slot and \(20\times\) measurements remain
descriptive appendix material.

\section{Evidence for the Three Separations}
\label{sec:experiments}

\subsection{Design}
\label{sub:results}

The experiments follow the three failed arrows.  Multi-length profiles compare
threshold count with mass coverage; matched tail summaries instantiate the
fixed-row factorization; and controlled plus trained-sparse tasks probe utility.
They test the scope of each mechanism, not an end-to-end guarantee.  Five
language-model anchors span \(4\)K--\(128\)K, with a \(4\)K Gemma control.
Heads follow fixed MInference groups \cite{jiang2024minference}; captions give
the resampling unit, and the appendix gives the roster and controls.

\begin{figure*}[t]
\vspace{-1mm}
\centering
\begin{minipage}[t]{0.32\textwidth}
\centering
\includegraphics[width=\linewidth]{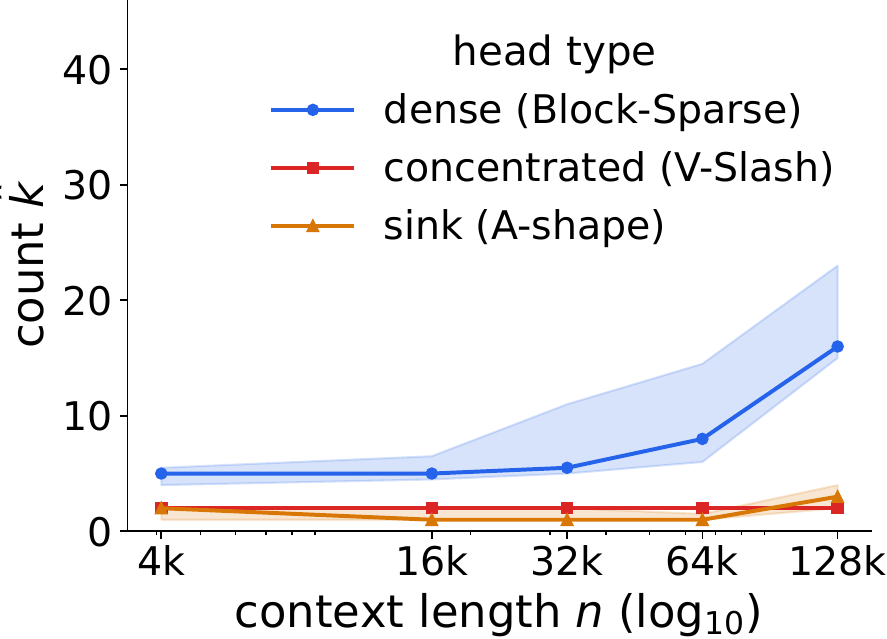}
\\[-1mm]{\small (a) relative-weight count}
\end{minipage}\hfill
\begin{minipage}[t]{0.32\textwidth}
\centering
\includegraphics[width=\linewidth]{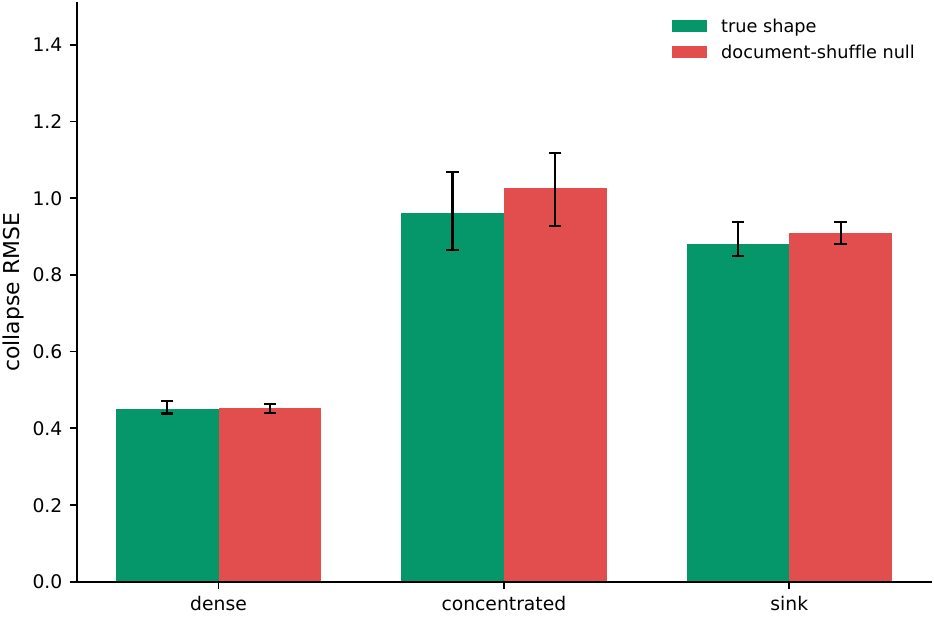}
\\[-1mm]{\small (b) coverage-shape prediction}
\end{minipage}\hfill
\begin{minipage}[t]{0.32\textwidth}
\centering
\includegraphics[width=\linewidth]{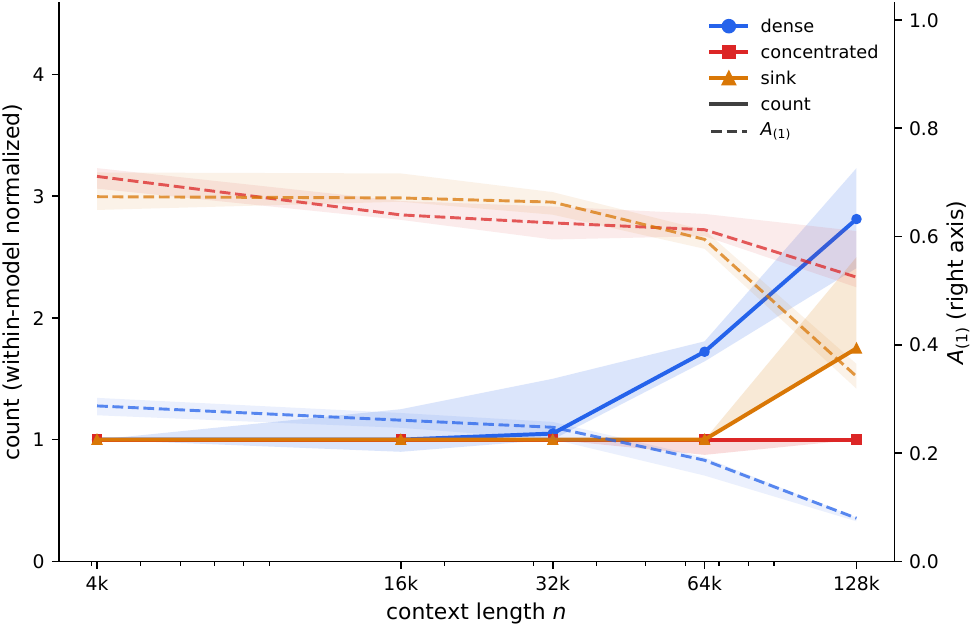}
\\[-1mm]{\small (c) count and concentration}
\end{minipage}
\caption{\textbf{Finite-range count sparsity does not yield a transferable mass
budget.}
\textbf{(a)} Within fixed head types, \(\hat k(0.1A_{(1)})\) grows sublinearly
over the measured lengths. \textbf{(b)} A profile that beats a shuffled control
on five of six anchors at fixed length still misses the measured \(128\)K
budget by \(>1.5\times\) when fitted at \(4\)K. \textbf{(c)} \(A_{(1)}\) can
fall while normalized count stays bounded.  Intervals are 95\% document
bootstraps (\(B=5000\)); (b,c) use an equal-model median.  Gemma is a \(4\)K
descriptive point.}
\label{fig:headline}
\vspace{-2mm}
\end{figure*}

\subsection{Count, Coverage, Output, and Task}

\paragraph{Count and coverage follow different scaling laws.}
Among eligible heads of each fixed checkpoint observed at three or more
lengths, after excluding non-finite document slopes (seed \(0\), \(B=5000\),
BH-FDR \(q=0.05\)), the measured sublinear-growth fractions (estimate
\({\pm}\) Wilson 95\% half-width) are
\(99.87{\pm}0.22\%\) (\textsc{gpt-oss}), \(98.93{\pm}0.66\%\) (Llama-8B),
\(96.05{\pm}0.53\%\) (Llama-70B), \(98.63{\pm}0.73\%\) (Mistral), and
\(98.96{\pm}0.61\%\) (Qwen).  Sink/concentrated heads typically use
\(1\)--\(7\) large weights; dense-heavy Qwen3/\textsc{gpt-oss} use
\(12\)--\(19\) at the longest contexts.  Gemma is a \(4\)K descriptive
control.  The result establishes finite-range count sparsity, not cheap
coverage.

\paragraph{Short-context coverage does not transfer without profile stability.}
At fixed length, the two-parameter gap-profile fit beats its within-head
shuffle on five of six anchors (fit/null RMSE:
\(0.56/0.68\), \(0.58/0.77\), \(0.57/1.19\), \(0.53/0.75\), and
\(0.37/0.58\)); \textsc{gpt-oss} is the exception (\(0.49/0.46\)).
The \(4\)K fit nevertheless misses the measured \(128\)K coverage budget by
more than \(1.5\times\): \(A_{(1)}\), hence concentration, changes with length
even when normalized count stays small (\cref{fig:headline}c).  Reusing a
short-context cache budget therefore requires profile stability.

\paragraph{The fixed-row tail aggregate is the operative oracle statistic.}
At \(4\)K context and \(k=64\), the true weighted tail aggregate reconstructs
the realized attention output exactly on every anchor, while the norm-matched
random, covariance-matched, and cross-document shuffled controls are all worse
than keeping no extra statistic (\cref{tab:bandwidth-main}).  This experiment
instantiates the algebraic identity on trained-model activations; it is a
fixed-row diagnostic, not evidence for unseen future queries.

\begin{table}[t]
\vspace{-2mm}
\centering
\begin{adjustbox}{max width=\columnwidth}
\begingroup
\scriptsize
\setlength{\tabcolsep}{3pt}
\begin{tabular}{@{}lcccc@{}}
\toprule
Model & $+$random & $+$cov.-matched & $+$shuffled tail & $+$true aggregate \\
\midrule
Llama-3.1-8B & \(1.4125{\pm}0.0020\) & \(1.9988{\pm}0.0324\) & \(1.4593{\pm}0.0034\) & \(0.0000\) \\
Qwen3-8B & \(1.4135{\pm}0.0026\) & \(1.6893{\pm}0.1018\) & \(1.3906{\pm}0.0117\) & \(0.0000\) \\
Llama-3.3-70B & \(1.4131{\pm}0.0005\) & \(1.7958{\pm}0.2033\) & \(1.4533{\pm}0.0050\) & \(0.0000\) \\
Mistral-7B & \(1.4122{\pm}0.0020\) & \(2.9515{\pm}0.0766\) & \(1.4679{\pm}0.0055\) & \(0.0000\) \\
Gemma-2-9B & \(1.4128{\pm}0.0012\) & \(1.7967{\pm}0.1773\) & \(1.4321{\pm}0.0193\) & \(0.0000\) \\
\midrule
Avg. (5 anchors) & \(1.4128{\pm}0.0004\) & \(2.0464{\pm}0.3776\) & \(1.4406{\pm}0.0236\) & \(0.0000\) \\
\bottomrule
\end{tabular}
\endgroup

\end{adjustbox}
\vspace{-1mm}
\caption{\textbf{Fixed-row tail reconstruction at \(4\)K, \(k=64\), on five
anchors.}
Keep-only-normalized error; \(+\)random is norm-matched.  Intervals are 95\%
document-bootstrap half-widths.  Among the tested summaries,
the oracle aggregate is exact for the realized row.}
\label{tab:bandwidth-main}
\label{tab:bandwidth}
\vspace{-2mm}
\end{table}

\begin{figure}[t]
\vspace{-2.5mm}
\centering
\begin{minipage}[t]{0.49\linewidth}
\centering
\includegraphics[width=\linewidth]{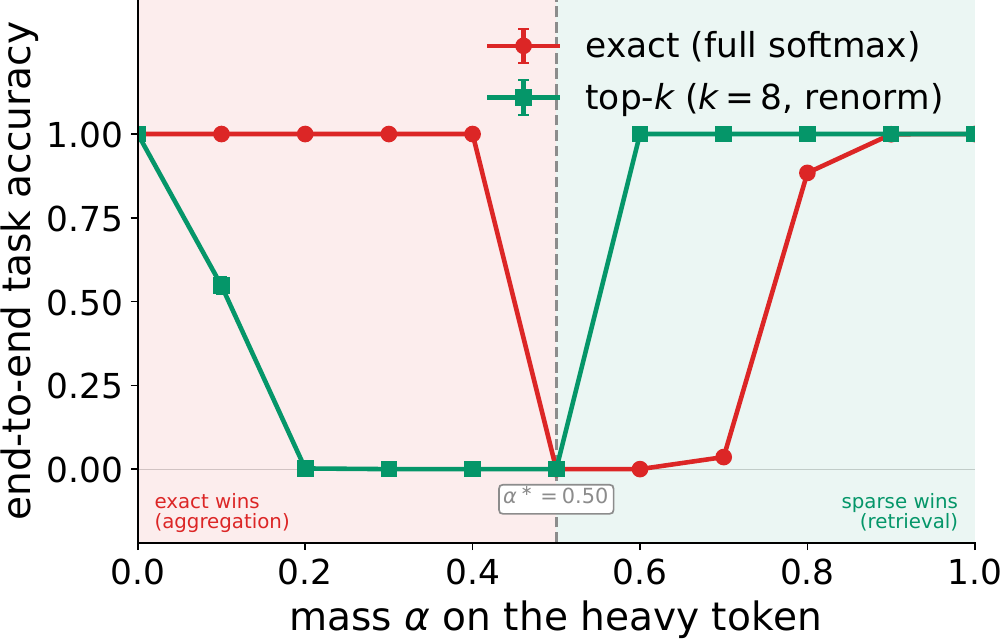}
\\[-1mm]{\small (a) controlled task phase law}
\end{minipage}\hfill
\begin{minipage}[t]{0.49\linewidth}
\centering
\includegraphics[width=\linewidth]{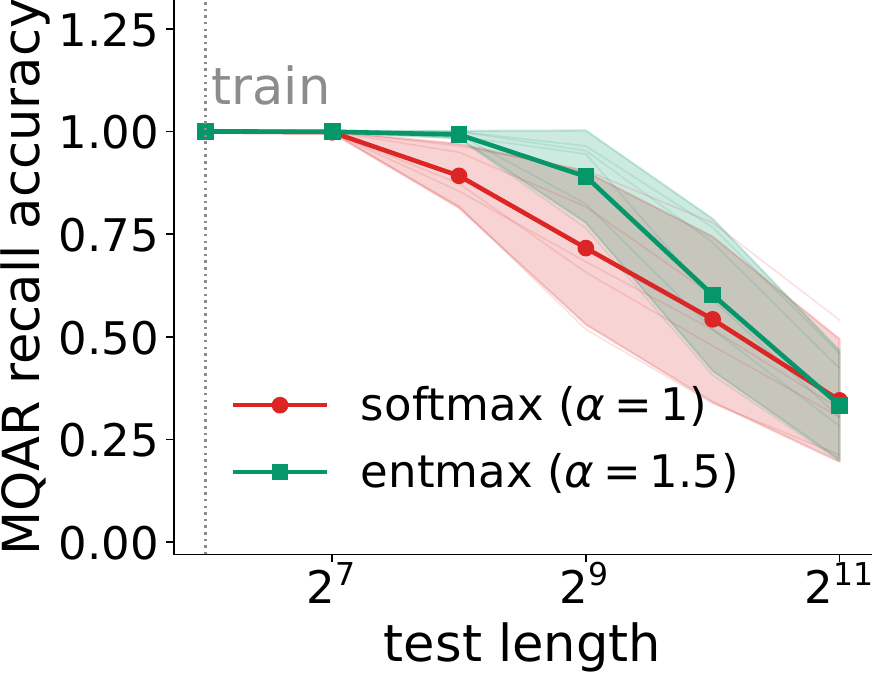}
\\[-1mm]{\small (b) trained-sparse length extrapolation}
\end{minipage}
\vspace{-1mm}
\caption{\textbf{Task utility depends on regime.}
\textbf{(a)} Exact and top-\(k\) exchange order at \(\alpha^\star=0.50\) in the
controlled mixture; the real-task observable does not predict a change (1500
trials/\(\alpha\), \(B=5000\), 95\% trial bootstrap). \textbf{(b)} Five-seed
paired-\(t\) differences are \(0.10{\pm}0.08\) at \(4\times\) and
\(0.17{\pm}0.13\) at \(8\times\); \(16/32\times\) include zero.  This test is
separate from the controlled theorem.}
\label{fig:alpha}
\label{fig:extrapolate-band}
\vspace{-2mm}
\end{figure}

\paragraph{A controlled task law does not reduce real tasks to one coordinate.}
The controlled task recovers the predicted transition: top-\(k\) helps
retrieval after retaining the planted signal, whereas exact attention remains
preferable for aggregation (\cref{fig:alpha}a).  The analogous real-task
coordinate varies but does not predict method ordering, so the mechanism does
not transfer without a richer signal model.  Trained entmax separately
improves at \(4/8\times\), while the \(16/32\times\) intervals include zero
(\cref{fig:alpha}b).

\paragraph{Evidence synthesis.}
The tests rule out three shortcuts: treating count as a transferable coverage
budget, generic tail moments as fixed-row output summaries, and one controlled
coordinate as a real-task predictor.  Together they show where the formal
mechanisms instantiate cleanly and where additional structure is required.

\ifdefined\ShowEditorialBoundaryNote
\paragraph{Evidence boundary.}
The complete \textsc{RULER} evaluation is reproducible from its recorded
inputs and outputs; the appendix records their data sources and checksums, and
legacy results that cannot be regenerated remain excluded.
\fi

\newcommand{\bandwidthtabular}{%
\begin{tabular}{@{}lcccc@{}}
\toprule
Model & \(+\)random & \(+\)cov.-matched & \(+\)shuffled tail & \(+\)\textbf{true} aggregate \\
\midrule
Llama-3.1-8B  & \(1.41{\pm}0.00\) & \(2.00{\pm}0.03\) & \(1.46{\pm}0.00\) & \(\mathbf{0.00{\pm}0.00}\) \\
Qwen3-8B      & \(1.41{\pm}0.00\) & \(1.69{\pm}0.10\) & \(1.39{\pm}0.01\) & \(\mathbf{0.00{\pm}0.00}\) \\
Llama-3.3-70B & \(1.41{\pm}0.00\) & \(1.80{\pm}0.20\) & \(1.45{\pm}0.01\) & \(\mathbf{0.00{\pm}0.00}\) \\
Mistral-7B    & \(1.41{\pm}0.00\) & \(2.95{\pm}0.08\) & \(1.47{\pm}0.01\) & \(\mathbf{0.00{\pm}0.00}\) \\
Gemma-2-9B    & \(1.41{\pm}0.00\) & \(1.80{\pm}0.18\) & \(1.43{\pm}0.02\) & \(\mathbf{0.00{\pm}0.00}\) \\
\bottomrule
\end{tabular}}

\newcommand{\bandwidthcaption}{\textbf{Bandwidth floor}, five anchors:
output-reconstruction error versus tail-sketch composition, normalized to the
kept-only baseline (top-\(k\), \(k=64\), \(4\)K context; estimate
\({\pm}\) 95\% document-bootstrap half-width over 12 documents).  Only the
true \(d\)-dimensional tail aggregate reconstructs the fixed-weight output
exactly; the other controls are worse than keeping nothing extra.}

\newcommand{\bandwidthcaptionshort}{\textbf{Bandwidth floor}: normalized
output-reconstruction error versus tail sketch on five anchors.}

\ifdefined\arxivsubmission
\section{Discussion}\label{sec:discussion}

\subsection{What the three separations change}

The paper separates four objects that are often collapsed into the phrase
``attention sparsity.''  A small threshold count describes how many unusually
large weights a row has; a coverage budget asks how many entries retain a
specified mass; output recovery additionally depends on the omitted values;
and task utility depends on which value directions the task reads.  None of
the three forward transitions follows without additional assumptions.
\Cref{thm:gap-envelope-dichotomy} makes the count--coverage break explicit,
\cref{thm:fixed-row-tail-factorization} identifies the exact fixed-row
statistic lost by subset-only retention, and
\cref{thm:denoising-phase-law} gives one controlled task family in which exact
and top-$k$ attention exchange order.

This decomposition also changes how an empirical compression claim should be
read.  Count curves describe observed weights; they are not memory budgets.  A
coverage predictor must be checked under length transfer rather than only
within a fixed length.  A fixed-row aggregate is not automatically a
future-query cache.  Finally, a downstream average cannot establish that
sparsification is uniformly beneficial when aggregation- and retrieval-like
targets price the omitted values differently.

\subsection{Open problems}

Three directions would materially strengthen the result.  First, a
multi-layer stability theorem should propagate row-level omitted-mass and
value-dispersion errors through a transformer block.  Second, a useful
future-query result must bound tail log-partition and conditional-value
mismatch using statistics available before eviction.  Third, training theory
should explain when a head moves across the critical coverage regime and what
happens at the \(\gamma=1\) boundary.  These are mathematical gaps, not
promises that the current experiments have already resolved.

\fi
\vspace{-2mm}
\section{Conclusion}\label{sec:conclusion}
Attention compressibility is a contract---not a heat-map property---among
retained mass, omitted values, future queries, and task.  Score gaps govern
coverage; the weighted tail sum completes a realized row; task structure can
reverse exact and top-\(k\) attention.  \certkv{} converts these lessons into a
competitive exact-budget compressor with measured cache savings, while claiming
no unconditional future-query or multilayer guarantee.  Sparse-looking weights
are clues, not certificates.

\ifdefined\arxivsubmission
\section*{Acknowledgement}

Chiwun is supported by Hong Kong Future Talents Scholarship Scheme for Advanced Studies (FTSS).
\fi

\ifdefined\arxivsubmission\else
\clearpage
\iclronly{%
\section*{Limitations}
The theory analyzes one attention row in one layer and assumes oracle top-\(k\)
selection; extending its error control through a practical selector and
multiple layers remains open.  Empirically, a profile fitted at \(4\)K does not
predict the \(128\)K coverage budget: predicted and measured budgets differ by
more than \(1.5\times\).  Cross-length allocation therefore additionally
requires profile stability.
The exact tail identity concerns realized attention weights; predicting later
queries requires a model of their distribution.  Our controlled ordering law
is restricted to the stated planted-token model, and the current vision/audio
measurements are insufficient to claim cross-modal universality.  Empirical
uncertainty is conditional on fixed checkpoints and configurations:
LongBench-v2 uses \(32\)K truncation, the \(128\)K \textsc{RULER} settings use
pre-specified \(3\%\) subsets, and the profile study uses one natural-text
surface.  The systems prototype measures cache capacity rather than latency.
Training dynamics near the critical \(\gamma=1\) regime and the boundary case
\(\gamma=1\) remain open.
\Cref{sec:app-discussion} gives the extended scope, future work, and broader
impact discussion.

\section*{Ethical Considerations}

This work studies the mathematical and systems properties of attention and
KV-cache compression.  We conducted no human-subject study, user study, new
human annotation, or collection of personal data.  The empirical evaluation
uses previously released pretrained checkpoints and established text
benchmarks.  These artifacts may inherit biases, privacy concerns, and usage
constraints from their original data and model releases; reproduction should
therefore obtain them from their original providers and follow the
corresponding terms of use.

The intended benefit is more auditable and memory-efficient long-context
inference.  Lower inference costs may, however, also reduce the cost of
deploying language models for harmful purposes or operating them at greater
scale.  In addition, the proposed certificate concerns attention-output
approximation only under its stated assumptions.  It does not certify
factuality, fairness, privacy, robustness, or downstream safety.  Because
cache compression can alter model behavior, deployments in consequential
settings should validate both task performance and safety for the specific
model, task, context length, and cache budget being used, and should fall back
to a less compressed or full-cache configuration when these checks are
inadequate.}
\fi

\ifdefined\isarxiv
  \ifdefined\arxivsubmission
    \bibliographystyle{abbrvnat}
  \else
    \bibliographystyle{alpha}
  \fi
\fi   
\ifdefined\arxivsubmission
  \bibliography{refs,refs_prior,refs_arxiv}
\else
  \bibliography{refs,refs_prior}
\fi

\clearpage
\appendix
\ifdefined\submissionappendix
\ifdefined\arxivsubmission
\begin{center}
\textbf{\LARGE Appendix}
\end{center}
{\hypersetup{linkcolor=black}
\tableofcontents
}
\newpage
\section{Appendix Roadmap}\label{sec:appendix-roadmap}

The appendix is organized as follows.

\begin{itemize}
  \item \Cref{sec:submission-theory-proofs} proves the three main separations:
  the distinction between calibrated large-weight counts and mass coverage,
  the fixed-row tail factorization, and the stylized task-ordering law.

  \item \Cref{sec:submission-certkv-proofs} records the scoped guarantees behind
  \certkv{}, including the exact-budget construction, fixed-row completion,
  allocation rule, and the boundary between realized-row guarantees and
  future-query claims.

  \item \Cref{sec:app-experiments} gives the experimental protocol and
  supplementary results, including ablations, physical cache measurements,
  domain-level breakdowns, uncertainty procedures, and reproducibility
  details.
\end{itemize}

\else

\fi
\section{Proofs for the Three Main Separations}
\label{sec:submission-theory-proofs}

This appendix proves the three results stated in
\cref{sec:rethinking}.  The proofs use the same scope as the main text: one
realized row for coverage and output, and an explicit planted model for the
task ordering.  They do not assert an unconditional guarantee for future
queries.

\subsection{Calibrated large-weight count}

\begin{proof}[Proof of \cref{prop:calibrated-log-count}]
The sub-Gaussian tail and a union bound give
\[
\Pr\!\left\{\max_{j\le n}|\ell_j|>S_\delta\right\}\le\delta.
\]
On the complementary event,
\(Z=\sum_j e^{\ell_j}\ge ne^{-S_\delta}\).  Therefore
\(A_j>\epsilon\) implies
\(\ell_j>\log(\epsilon n)-S_\delta\).  For every
\(\epsilon\ge\epsilon_\delta(n)\), the latter threshold is at least
\(\tau_\delta=\sigma\sqrt{2\log(n/\delta)}\).  Hence, simultaneously over
that entire threshold range,
\[
k(\epsilon)\le
N_\delta\coloneqq\sum_{j=1}^n
\mathbf 1\{\ell_j>\tau_\delta\}.
\]
Writing \(p_j=\Pr\{\ell_j>\tau_\delta\}\), sub-Gaussianity gives
\(p_j\le\delta/n\).  Hence
\[
\begin{aligned}
\mu_\delta\coloneqq\mathbb E N_\delta
  &=\sum_{j=1}^n p_j\le\delta,\\
\operatorname{Var}(N_\delta)&\le\mu_\delta\le1.
\end{aligned}
\]
Bernstein's inequality, with \(x=\log(2n/\delta)\), gives with failure
probability at most \(e^{-x}=\delta/(2n)\),
\[
\begin{aligned}
N_\delta\le
\mu_\delta+\sqrt{2\mu_\delta x}+\frac23x
&\le \frac23x+\sqrt{2x}+2 .
\end{aligned}
\]
Combining the two events proves the stated probability \(1-2\delta\).
Finally,
\(\exp\{O(\sqrt{\log n})\}=n^{o(1)}\) for fixed
\((\sigma,\delta)\), so \(\epsilon_\delta(n)=n^{-1+o(1)}\).

For the Gaussian clause,
\(\mathbb E e^{\ell_j}=e^{\sigma^2/2}\) and
\(\operatorname{Var}(e^{\ell_j})<\infty\), so
\[
  \xi_n\coloneqq
  \log\frac{Z}{ne^{\sigma^2/2}}=O_\Pr(n^{-1/2}).
\]
The event \(A_j>\epsilon_n\) is exactly
\(\ell_j>\tau_n+\xi_n\).  Put \(a_n=n^{-1/4}\) and let
\(N(u)=\sum_j\mathbf 1\{\ell_j>u\}\).  On
\(\{|\xi_n|\le a_n\}\), whose probability tends to one,
\[
N(\tau_n+a_n)\le k(\epsilon_n)\le N(\tau_n-a_n).
\]
By the definition of \(\tau_n\) and the Gaussian Mills ratio,
\[
n\Pr\{\ell_j>\tau_n\pm a_n\}=(1+o(1))\log n,
\]
because \(\tau_na_n\to0\).  Each bounding count is binomial with variance no
larger than its mean, so Chebyshev's inequality makes both
\((1+o_\Pr(1))\log n\).  The sandwich proves the claimed count, while
\(\tau_n=O(\sqrt{\log n})\) gives
\(\epsilon_n=n^{-1+o(1)}\).
\end{proof}

\begin{proof}[Proof of \cref{prop:relative-count-law}]
Let \(t=\log(1/\epsilon)\) and \(b_n=\sqrt{2\log n}\).  The assumed additive
near-top law gives, uniformly for \(\log j=O(b_n)\),
\begin{align*}
G(j)
&=\sigma_s\!\left(b_n-\sqrt{b_n^2-2\log j}\right)+o_\Pr(1)\\
&=\sigma_s\frac{\log j}{b_n}+o_\Pr(1).
\end{align*}
For fixed \(\xi\in(0,t)\), set
\[
j_\pm=
\left\lfloor
\exp\!\left\{\frac{t\pm\xi}{\sigma_s}b_n\right\}
\right\rfloor .
\]
With probability tending to one, \(G(j_-)<t<G(j_+)\).  Since \(G\) is
nondecreasing, \(j_-\le k(\epsilon A_{(1)})<j_+\).  Dividing logarithms by
\(b_n\), then letting \(\xi\downarrow0\), gives the first convergence.  Since
\(b_n/\log n\to0\), the second follows.
\end{proof}

\subsection{Global gap envelopes and coverage}

\begin{proof}[Proof of \cref{thm:gap-envelope-dichotomy}]
Normalize by the largest logit and write
\[
q_j=e^{-G(j)},\qquad Z=\sum_{j=1}^n q_j.
\]
The top-\(k\) mass is \(\sum_{j\le k}q_j/Z\).
On the sparse branch, \(q_j\le e^b j^{-\gamma}\), while \(Z\ge q_1=1\).
Hence
\begin{align*}
1-\sum_{j\le k}A_{(j)}
&\le \sum_{j>k}q_j\\
&\le e^b\int_k^\infty x^{-\gamma}\,dx
=\frac{e^b}{\gamma-1}k^{1-\gamma}.
\end{align*}
The displayed upper bound on \(K_\eta\) follows by making the last quantity
at most \(\eta\).

On the dense branch, \(q_j\ge e^{-b'}j^{-\gamma'}\).  In particular,
\[
Z\ge\sum_{j=\lceil n/2\rceil}^{n}q_j
\ge\frac12e^{-b'}n^{1-\gamma'}.
\]
Since every \(q_j\le1\), any \(k\)-set has mass at most \(k/Z\).
Reaching mass \(1-\eta\) therefore requires
\[
k\ge(1-\eta)Z
\ge\frac{(1-\eta)e^{-b'}}2n^{1-\gamma'}.
\]
At \(\gamma'=0\), this lower bound is linear and the trivial upper bound
\(K_\eta\le n\) gives \(K_\eta=\Theta(n)\).
\end{proof}

\paragraph{Gaussian calibration.}
For i.i.d.\ \(X_j\sim\mathcal N(0,\sigma^2)\), selecting the largest
\(\rho n\) logits is asymptotically equivalent to thresholding at a quantile
\(\tau\).  Exponential tilting by \(e^X\) sends
\(\mathcal N(0,\sigma^2)\) to \(\mathcal N(\sigma^2,\sigma^2)\).  Thus the
threshold capturing weighted mass \(1-\eta\) obeys
\[
\Pr\{\mathcal N(\sigma^2,\sigma^2)\ge\tau\}=1-\eta,
\]
and the unweighted fraction above it is
\[
\rho
=\Pr\{\mathcal N(0,\sigma^2)\ge\tau\}
=\Phi\!\left(\Phi^{-1}(1-\eta)-\sigma\right).
\]
To justify the random minimal rank, define
\[
h(t)=\frac{\mathbb E[e^X\mathbf 1\{X\ge t\}]}
           {\mathbb E e^X}.
\]
This function is continuous and strictly decreasing, and \(\tau\) is its
unique solution to \(h(\tau)=1-\eta\).  For every \(a>0\), the law of large
numbers at \(\tau-a\) and \(\tau+a\) implies, with probability tending to one,
\[
\begin{aligned}
\frac1n\sum_j\mathbf 1\{X_j>\tau+a\}
&\le\frac{K_\eta}{n}\\
&\le\frac1n\sum_j\mathbf 1\{X_j>\tau-a\}.
\end{aligned}
\]
Letting \(n\to\infty\) and then \(a\downarrow0\) proves the stated limit.  The
formula is a distribution-specific calibration, not a consequence of a small
threshold count.

\subsection{Fixed-row tail factorization}

\begin{proof}[Proof of \cref{thm:fixed-row-tail-factorization}]
Split the exact output into retained and omitted indices:
\[
O=\sum_{j\in S}A_jV_j+\sum_{j\notin S}A_jV_j
=p_S\widetilde O_S+(1-p_S)\bar V_T.
\]
Subtracting this identity from \(\widetilde O_S\) gives
\[
\widetilde O_S-O
=(1-p_S)(\widetilde O_S-\bar V_T).
\]
Both \(\widetilde O_S\) and \(\bar V_T\) are convex combinations of values
with norm at most \(V_{\max}\), proving the \(2V_{\max}(1-p_S)\) bound.
For an additive correction \(u\),
\[
\left\|\sum_{j\in S}A_jV_j+u-O\right\|_2
=\left\|u-\sum_{j\notin S}A_jV_j\right\|_2,
\]
which vanishes uniquely at
\(u^\star=\sum_{j\notin S}A_jV_j=(1-p_S)\bar V_T\).
\end{proof}

\subsection{Task-conditional denoising}

\begin{proof}[Proof of \cref{thm:denoising-phase-law}]
Let \(M_n\) be the maximum bulk logit.  We only need a sharp upper bound.
Set
\begin{align*}
t_n&=\frac{\sigma\log\log n}{\sqrt{2\log n}}=o(1),\\
u_n&=\sigma\sqrt{2\log n}+t_n.
\end{align*}
A Gaussian tail bound and a union bound give
\begin{align*}
\Pr\!\left\{M_n>u_n\right\}
&\le n\exp\!\left(-\frac{u_n^2}{2\sigma^2}\right)\\
&=\exp\!\left(
-\frac{\sqrt{2\log n}}{\sigma}t_n
-\frac{t_n^2}{2\sigma^2}
\right)\\
&=o(1).
\end{align*}
Thus \(M_n\le\sigma\sqrt{2\log n}+o_\Pr(1)\).
The lower bound on \(\mu_n\) therefore puts the planted token above the bulk
with probability tending to one.  Moreover, the retained bulk contributes at
most \(k_ne^{M_n}\), so
\[
\frac{\sum_{j\in S_{k_n}\setminus\{j^\star\}}e^{\ell_j}}
{e^{\mu_n}}
\le e^{-C+o_\Pr(1)}.
\]
Consequently \(w_s\ge(1+e^{-C+o_\Pr(1)})^{-1}=\Theta_\Pr(1)\).
For full softmax, the lognormal law of large numbers gives
\[
\sum_{\mathrm{bulk}}e^{\ell_j}
=n\,e^{\sigma^2/2+o_\Pr(1)}.
\]
The upper bound on \(\mu_n\) then yields
\(w_e\le n^{-\varepsilon+o_\Pr(1)}\).

Condition on the logits.  Because the bulk values are i.i.d.\ and independent
of the logits, their exact attention-weighted average has mean
\(\bar V_b\); fixed logit
scale gives squared normalized weights of order \(O_\Pr(n^{-1})\), hence
deviation \(O_\Pr(\sigma_Vn^{-1/2})\).  The explicitly assumed retained-bulk
diffuseness gives the corresponding
\(O_\Pr(\sigma_Vk_n^{-1/2})\) deviation for top-\(k_n\).
An estimator placing planted-token weight \(w\) therefore satisfies
\[
L=|\alpha-w|D+\text{the stated stochastic error}.
\]
Since \(w_e<w_s\) with probability tending to one,
\(|\alpha-w_e|=|\alpha-w_s|\) has the unique solution
\(\alpha_n^\star=(w_e+w_s)/2\).  On that event, the exact noiseless loss gap is
\begin{align*}
\Delta_n(\alpha)
&=\bigl||\alpha-w_e|-|\alpha-w_s|\bigr|\\
&=\min\!\left\{w_s-w_e,\,2|\alpha-\alpha_n^\star|\right\}.
\end{align*}
The stated margin absorbs the sum of the two stochastic remainders and
therefore determines the ordering.  Finally, the value-aware oracle minimizes
the loss over all \(k_n\)-subsets, so it cannot be worse than the
score-selected set.
\end{proof}

\paragraph{Scope.}
The retained-bulk diffuseness assumption is load bearing and is not implied by
rank inclusion.  The controlled mixture exhibits the predicted ordering
change.  The downstream test is different: it asks whether a pre-specified
estimate of \(\alpha\) predicts the sign of the exact--top-\(k\) accuracy
difference on retrieval and aggregation tasks.  It does not.  We therefore
treat the controlled experiment as a validation within the planted model and
the downstream result as evidence that this one-parameter task summary is not
a calibrated decision rule.

\section{Scoped Guarantees Behind \certkv{}}
\label{sec:submission-certkv-proofs}

The fixed-row identity in
\cref{thm:fixed-row-tail-factorization} motivates a charged tail-summary
slot.  The results below make precise what else is, and is not, guaranteed.

\subsection{A posteriori future-query error decomposition}

\begin{theorem}[Future-query lift and summary mismatch]
\label{thm:future-query-lift}
Let nonempty sets \(S,T\) partition the original cache.  For one fixed future
query, let
\(A\) be exact attention on \(S\cup T\), and let \(\widehat A\) be compressed
attention on \(S\cup\{\star\}\).  For \(r\in\Delta(T)\), store
\(V^\star=\sum_{j\in T}r_jV_j\).  Put
\[
p=A(T),\qquad q=\widehat A_\star,\qquad a=A_T/p
\]
when \(p>0\).  If all retained logits are unchanged and
\(\lVert V_j\rVert_2\le V_{\max}\), then
\begin{align*}
\lVert O-\widehat O\rVert_2
&\le V_{\max}\!\left[2|p-q|\right.\\[-1mm]
&\hspace{10mm}\left.
{}+\min\{p,q\}\lVert a-r\rVert_1\right].
\end{align*}
If \(\Lambda_T=\log\sum_{j\in T}e^{z_j}\) is the exact tail log-partition and
\(z_\star\) is the synthetic-slot logit, then
\[
|p-q|
\le\tanh\!\left(\frac{|\Lambda_T-z_\star|}{4}\right)
\le\frac14|\Lambda_T-z_\star|.
\]
\end{theorem}

\begin{proof}
Lift the compressed distribution back to \(S\cup T\) by setting
\(\bar A_i=\widehat A_i\) on \(S\) and
\(\bar A_j=qr_j\) on \(T\).  Expanding \(V^\star\) gives
\(\widehat O=\sum_j\bar A_jV_j\), so
\[
\lVert O-\widehat O\rVert_2
\le V_{\max}\lVert A-\bar A\rVert_1.
\]
Unchanged retained logits preserve all within-\(S\) softmax ratios.  Splitting
the \(L_1\) difference into its retained and tail parts, and matching the
smaller common mass on each side, gives
\[
\lVert A-\bar A\rVert_1
\le2|p-q|+\min\{p,q\}\lVert a-r\rVert_1.
\]
Finally, with \(Z_S=\sum_{i\in S}e^{z_i}\),
\begin{align*}
p&=\operatorname{sigmoid}(\Lambda_T-\log Z_S),\\
q&=\operatorname{sigmoid}(z_\star-\log Z_S).
\end{align*}
The largest displacement of a sigmoid under an input shift of magnitude
\(d\) is \(\tanh(d/4)\le d/4\).
\end{proof}

Evaluating \(p\), \(a\), or \(\Lambda_T\) after eviction requires the omitted
keys.  The theorem is therefore an a posteriori bound separating two
mismatches, not a pre-eviction guarantee for \certkv{}.

\subsection{Linear-sketch scope of the tail aggregate}

\begin{theorem}[Fixed-row factor map and scoped minimality]
\label{thm:factor-map}
Fix \(A_1,\ldots,A_n>0\) with \(\sum_jA_j=1\), and let
\(S\subset[n]\) satisfy \(|S|=k<n\).  Write \(T=[n]\setminus S\),
\(p_S=\sum_{j\in S}A_j\), and
\[
\bar V_T=\frac{1}{1-p_S}\sum_{j\in T}A_jV_j.
\]
Treating \(V\in\mathbb R^{nd}\) as variable, the linear map
\[
\Phi(V)=\bigl((V_j)_{j\in S},\bar V_T\bigr)
\in\mathbb R^{(k+1)d}
\]
is surjective and has kernel dimension \((n-k-1)d\).  It recovers the full
output and every truncated sub-budget output
\(\sum_{j\in S'}A_jV_j\), \(S'\subseteq S\).  Conversely, any linear sketch
\(\Psi:\mathbb R^{nd}\to\mathbb R^m\) from which all these quantities can be
recovered, even by nonlinear decoders, has \(m\ge(k+1)d\).
\end{theorem}

\begin{proof}
Surjectivity follows by assigning the requested retained values and setting
all tail values equal to the requested \(\bar V_T\); rank--nullity gives the
kernel dimension.  The output factorization is the split in
\cref{thm:fixed-row-tail-factorization}, and sub-budget outputs read retained
coordinates directly.

For minimality, singleton sub-budgets recover each of the \(kd\) retained
value coordinates up to its positive scalar weight.  Subtracting the retained
output from the full output recovers the \(d\) coordinates of the weighted
tail sum.  These \((k+1)d\) coordinate functionals are linearly independent.
If any factored functional varied on \(\ker\Psi\), two inputs with the same
sketch would require different decoder outputs.  Thus \(\ker\Psi\) lies in
the joint kernel of all \((k+1)d\) functionals, forcing
\(\operatorname{rank}\Psi\ge(k+1)d\) and hence \(m\ge(k+1)d\).
\end{proof}

The lower bound concerns a linear sketch that supports every retained
sub-budget.  It is not a lower bound for one fixed output, a nonlinear
summary, or a future-query cache.

\subsection{Why centered dispersion is a principled feature}

For \(m>0\) and \(\lVert v^\star\rVert_2\le V_{\max}\), let
\(\mathcal T(m,v^\star)\) be the class of all finite weighted tails
\(\{(a_j,v_j)\}_{j=1}^N\), with arbitrary \(N\ge1\), satisfying
\[
\begin{aligned}
a_j&\ge0,
&\sum_j a_j&=m,\\
\sum_j a_jv_j&=mv^\star,
&\lVert v_j\rVert_2&\le V_{\max}.
\end{aligned}
\]
For a member of this class, define
\[
\begin{aligned}
M_2&=\sum_j a_j\lVert v_j-v^\star\rVert_2^2,\\
\operatorname{Err}_1&=\sum_j a_j\lVert v_j-v^\star\rVert_2.
\end{aligned}
\]
Let \(\mathcal T(m,v^\star,M_2)\) denote the subclass with the displayed
centered moment fixed to \(M_2\).

\begin{theorem}[Exact envelopes for two declared summary classes]
\label{thm:cert-stat-lb}
For \(d\ge2\),
\[
\sup_{\mathcal T(m,v^\star)}\operatorname{Err}_1
=m\sqrt{V_{\max}^2-\lVert v^\star\rVert_2^2},
\]
although the same \((m,v^\star)\) also permits zero deviation.  If the
centered moment \(M_2\) is additionally fixed, the subclass is nonempty
exactly when
\[
0\le M_2\le
m\bigl(V_{\max}^2-\lVert v^\star\rVert_2^2\bigr),
\]
and in that case
\[
\sup_{\mathcal T(m,v^\star,M_2)}\operatorname{Err}_1=\sqrt{mM_2}.
\]
\end{theorem}

\begin{proof}
Normalize the weights by \(p_j=a_j/m\).  Jensen followed by the variance
identity gives
\[
\frac{\operatorname{Err}_1}{m}
\le\sqrt{\sum_jp_j\lVert v_j-v^\star\rVert_2^2}
\le\sqrt{V_{\max}^2-\lVert v^\star\rVert_2^2}.
\]
Choose a unit vector \(u\perp v^\star\) and two equally weighted points
\(v^\star\pm\delta u\).  With
\(\delta=\sqrt{V_{\max}^2-\lVert v^\star\rVert_2^2}\), this attains the first
envelope; a one-point tail at \(v^\star\) has zero deviation.

For fixed \(M_2\), weighted Cauchy--Schwarz yields
\[
\operatorname{Err}_1
\le\sqrt{\sum_ja_j}\,
\sqrt{\sum_ja_j\lVert v_j-v^\star\rVert_2^2}
=\sqrt{mM_2}.
\]
The same two-point construction with \(\delta=\sqrt{M_2/m}\) is feasible
exactly on the displayed interval and attains equality.
\end{proof}

This theorem compares two declared summary classes.  It neither proves
optimality over arbitrary constant-size summaries nor bounds future-query
output error at fixed summaries.  It justifies \(\sqrt{mM_2}\) as a tight
observation-weighted dispersion envelope for allocation.

\ifdefined\arxivsubmission\else
\section{Discussion, Limitations, and Broader Impact}
\label{sec:app-discussion}

\subsection{What the three separations change}

The paper separates four objects that are often collapsed into the phrase
``attention sparsity.''  A small threshold count describes how many unusually
large weights a row has; a coverage budget asks how many entries retain a
specified mass; output recovery additionally depends on the omitted values;
and task utility depends on which value directions the task reads.  None of
the three forward transitions follows without additional assumptions.
\Cref{thm:gap-envelope-dichotomy} makes the count--coverage break explicit,
\cref{thm:fixed-row-tail-factorization} identifies the exact fixed-row
statistic lost by subset-only retention, and
\cref{thm:denoising-phase-law} gives one controlled task family in which exact
and top-\(k\) attention exchange order.

This decomposition also changes how an empirical compression claim should be
read.  Count curves describe observed weights; they are not memory budgets.  A
coverage predictor must be checked under length transfer rather than only
within a fixed length.  A fixed-row aggregate is not automatically a
future-query cache.  Finally, a downstream average cannot establish that
sparsification is uniformly beneficial when aggregation- and retrieval-like
targets price the omitted values differently.

\subsection{Scope of the claims}

\paragraph{Per-row rather than end-to-end guarantees.}
The main results price one realized attention row and do not propagate error
through residual streams, normalization, nonlinear blocks, or depth.  The
coverage theorem assumes a global rank envelope; a fitted log--log slope is
only a local summary unless the envelope is verified over the declared rank
range.  The task theorem additionally assumes a planted logit, independent
bulk values, and retained-bulk diffuseness.  These assumptions are visible in
the statement and proof rather than inferred from a trained model.

\paragraph{Fixed-row reconstruction versus future queries.}
The exact tail aggregate reconstructs the output only for the weights that
defined it.  After eviction, a new query changes both the tail log-partition
and the conditional distribution over omitted values.  The a posteriori
decomposition in \cref{thm:future-query-lift} measures both mismatches, but
evaluating it requires the evicted keys.  Consequently, \certkv{} is a
theory-motivated, exactly budgeted compressor; the fixed-row identity alone
does not give it an unconditional guarantee for later queries.

\paragraph{What transfers across length and modality.}
The large-weight count pattern recurs across all five language models measured
at multiple context lengths.  Mass coverage is more length-sensitive: a
profile fitted at \(4\)K predicts the \(128\)K coverage budget with more than
the \(1.5\times\) multiplicative error allowed by our test.  A short-context
coverage profile therefore needs recalibration before it is used as a
long-context cache budget.  The vision and audio measurements also do not
support one shared two-regime phase description.  We use these experiments to
locate the scope of the observed language-model geometry rather than
extrapolating it to new architectures, learned sinks, or sliding-window
attention.  Gemma has only one measured length and is excluded from growth
estimates.

\paragraph{Breadth on the pre-specified main evaluation grid.}
Across the four-baseline LongBench-v2 grid, \certkv{} has a top-two compressed
point estimate in seven of nine settings.  This cross-backbone,
cross-budget breadth supports a competitive operating point within the
declared grid, not a universal method ranking.  The one-factor block averages
span \(24.3\%\)--\(25.1\%\), and none of the nine adjusted paired contrasts is
decisive at the present sample size; this is descriptive local insensitivity,
not statistical equivalence.
\subsection{Open problems}

Three directions would materially strengthen the result.  First, a
multi-layer stability theorem should propagate row-level omitted-mass and
value-dispersion errors through a transformer block.  Second, a useful
future-query result must bound tail log-partition and conditional-value
mismatch using statistics available before eviction.  Third, training theory
should explain when a head moves across the critical coverage regime and what
happens at the \(\gamma=1\) boundary.  These are mathematical gaps, not
promises that the current experiments have already resolved.

\subsection{Broader impact and responsible use}

KV compression can reduce persistent memory and make long-context inference
available on smaller hardware, but lower per-request cost may also increase
total deployment volume.  Approximation error is input- and task-dependent:
aggregation, rare-token retrieval, code, and structured reasoning can fail in
different ways even at the same nominal cache ratio.  A deployment should
therefore report the physical cache budget, task-stratified accuracy, worst
observed degradation, and fallback behavior rather than relying on a single
average.  Our experiments use public model checkpoints and benchmark
datasets under their respective licenses.  The anonymous release will include
the exact environment, configurations, and compute disclosure.

\fi
\section{Experimental Protocol and Supplementary Results}
\label{sec:app-experiments}

\subsection{Supplementary diagnostics and scope checks}
\label{sub:submission-supp-diagnostics}

The following measurements test different parts of the argument and therefore
use different sampling units and decision rules.  We present them separately
so that each diagnostic, uncertainty interval, and negative result can be read
at its native scale.

\paragraph{Cross-modal scope of the fitted phase structure.}
We first ask whether the clean two-regime structure fitted on language-model
attention also appears in bidirectional vision and audio encoders.  The
pre-specified gate requires a sufficiently deep trough between two fitted
modes, not merely a non-unimodal histogram.  Neither modality passes this
gate.  The result limits the empirical phase-structure claim; it does not
limit the deterministic coverage criterion, which can be evaluated on any
realized attention row.  The plotted audit uses per-input/layer/head slopes
from \textsc{SigLIP2} and \textsc{Whisper}; the broader modality check also
includes \textsc{DINOv2} and a causal \textsc{ImageGPT} control.  The fitted
mode separation is only moderate (Cohen's \(d=0.66\) for vision and \(1.23\)
for audio), whereas a synthetic positive control recovers a clearly separated
effect (\(d=7.1\)).  The negative gate is therefore not explained by an
instrument that is intrinsically unable to detect strong bimodality.

\begin{figure*}[!t]
\centering
\includegraphics[width=\textwidth]{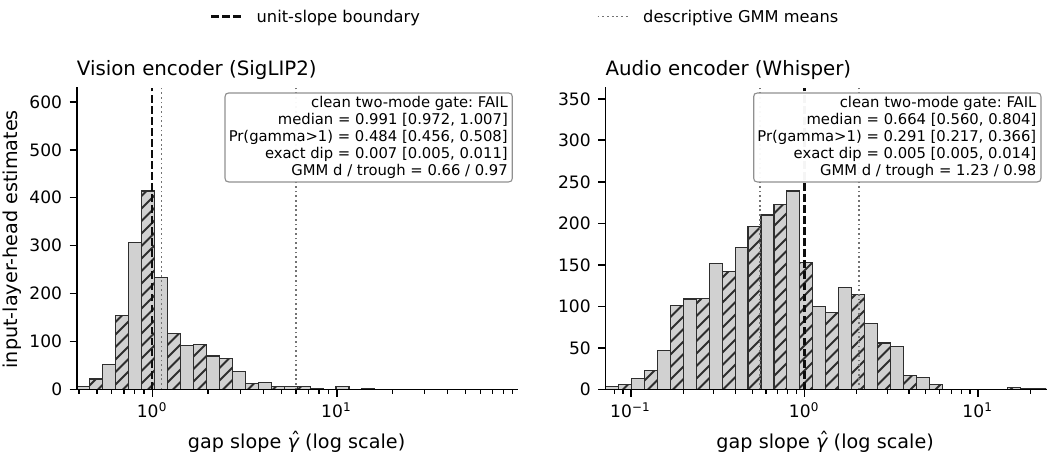}
\caption{\textbf{Cross-modal clean two-phase universality is not established.}
Per-input/\allowbreak layer/\allowbreak head gap-slope estimates are shown on a log axis; the dashed
line is the unit-slope boundary and dotted lines are descriptive two-Gaussian
means.  Neither modality meets the clean two-mode criterion because the fitted
mixtures have no sufficiently deep between-mean trough.  Brackets are \(95\%\)
percentile sensitivity intervals from resampling the four structured inputs
(seed \(0\), \(B=5000\)).  Exact Hartigan dip statistics are reported without
iid head-level \(p\)-values.}
\label{fig:supp-diagnostics}
\label{fig:phase4uni}
\end{figure*}

\paragraph{Stability of the count trend on real books.}
This check resamples documents and positions from a fixed
\textsc{Project Gutenberg} capture.  Across six such resampling replicates,
the fraction of heads whose large-weight count grows sublinearly is \(98.9\%\)
for \textsc{Llama-3.1-8B}, \(99.1\%\) for \textsc{Qwen3-8B}, and \(99.4\%\)
for \textsc{Mistral-7B}.  The intervals quantify stability to that fixed
resampling procedure; they are not evidence from six independent training
runs or six independently collected corpora.  Each replicate repeats the
same head-level slope decision after resampling positions from eight books,
and the displayed uncertainty is a two-sided Student-\(t\) interval across
those six estimates.  This check addresses data-resampling sensitivity of the
finite-range count trend; it does not validate short-to-long mass-coverage
transfer.

\begin{figure*}[!t]
\centering
\includegraphics[width=0.42\textwidth]{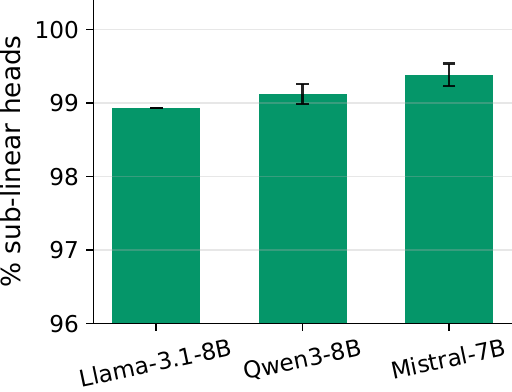}
\caption{\textbf{Count-sublinearity on real books, with resampling intervals}
(real \textsc{Project Gutenberg}; fraction of heads with sublinear
large-weight count).  Error bars are two-sided \(95\%\) Student-\(t\)
intervals over six document/position resampling replicates (seed base \(0\),
eight documents), not six independent training runs.}
\label{fig:e5count}
\end{figure*}

\paragraph{Distributional scope of the probabilistic count model.}
The count theorem's i.i.d.\ sub-Gaussian model is a sufficient-condition
analysis, whereas the operative coverage criterion is deterministic.  To
separate the two layers, we measure effective rank, per-coordinate excess
kurtosis, and the fraction of condensed heads across relative depth on three
trained architectures.  The trained hidden states are anisotropic and
heavier-tailed, and the prevalence of condensed heads is architecture- and
depth-dependent.  These measurements motivate applying the deterministic
row-wise criterion directly rather than treating the clean probabilistic
regimes as a literal model of every deployed hidden state.  Across the
recorded trained models, effective rank is roughly \(43\)--\(182\) out of
hidden dimensions \(2880\) or \(4096\), median per-coordinate excess kurtosis
reaches \(0.55\), and the condensed-head fraction ranges from low single
digits on \textsc{Llama} to about \(60\%\) on some \textsc{gpt-oss} layers.
The intervals resample corpora first and heads second; identical PG19/code
inputs are not counted as independent corpora.  These are scope diagnostics,
not new assumptions inserted into the deterministic theorem.

\begin{figure*}[!t]
\centering
\includegraphics[width=0.94\textwidth]{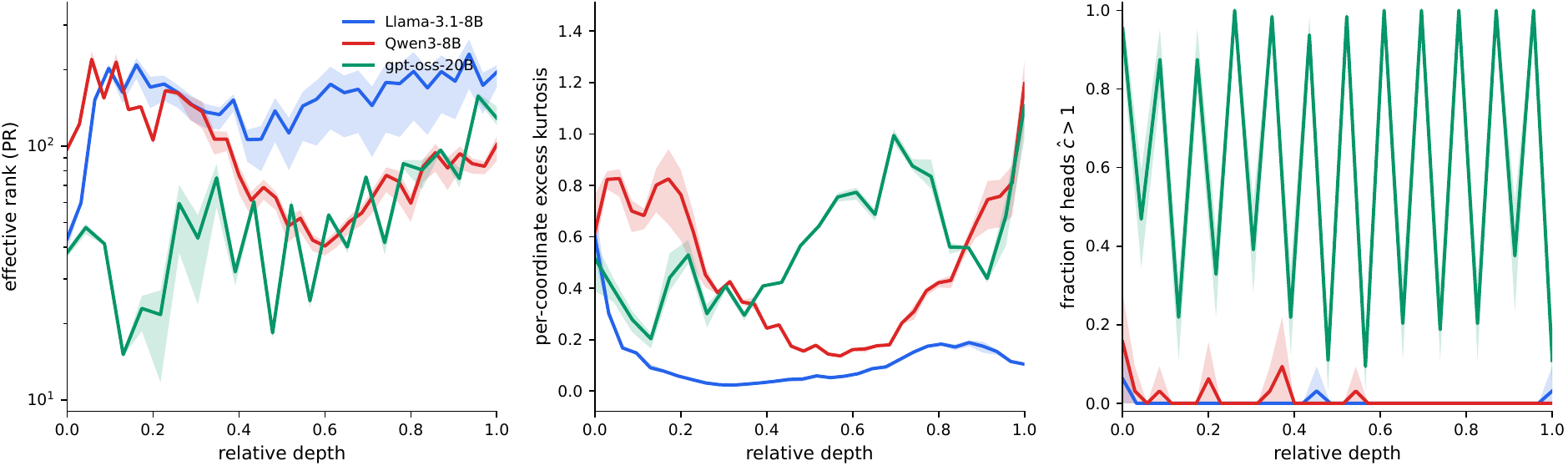}
\caption{\textbf{Real hidden states sit outside the clean distributional
regimes} (\cref{fac:score-scale,thm:gap-envelope-dichotomy}; three
architectures).  \textbf{(a,b)} Effective rank and excess kurtosis, with
seed-\(0\), \(B=5000\), \(95\%\) corpus-bootstrap bands over three trained
corpora.  \textbf{(c)} Condensed-head fraction with corpus-then-head
bootstrap; the single-corpus untrained condition is omitted, and identical
PG19/code inputs are not treated as independent corpora.}
\label{fig:xdist}
\end{figure*}

\paragraph{Controlled emergence on real versus shuffled text.}
The controlled language-model run asks whether decreasing kept fraction is
accompanied by learning.  With real books, next-token accuracy rises while the
median kept fraction falls; with shuffled tokens, accuracy remains near
chance even though its kept-fraction trajectory also changes.  The supported
observation is therefore the conjunction of learned task structure and
attention concentration, not the claim that a falling kept fraction alone
certifies useful learning.  Over the plotted horizon, the real-book median
kept fraction falls from \(0.67\) to \(0.30\) while next-token accuracy rises
to \(0.41\); shuffled-token accuracy remains near its \(0.05\) chance level.
Three training seeds supply the mean and sample-standard-deviation bands.
Because the shuffled control's kept fraction also falls, it is the paired
accuracy trajectory---not sparsity in isolation---that makes the real-text
observation informative.

\begin{figure*}[!t]
\centering
\includegraphics[width=0.72\textwidth]{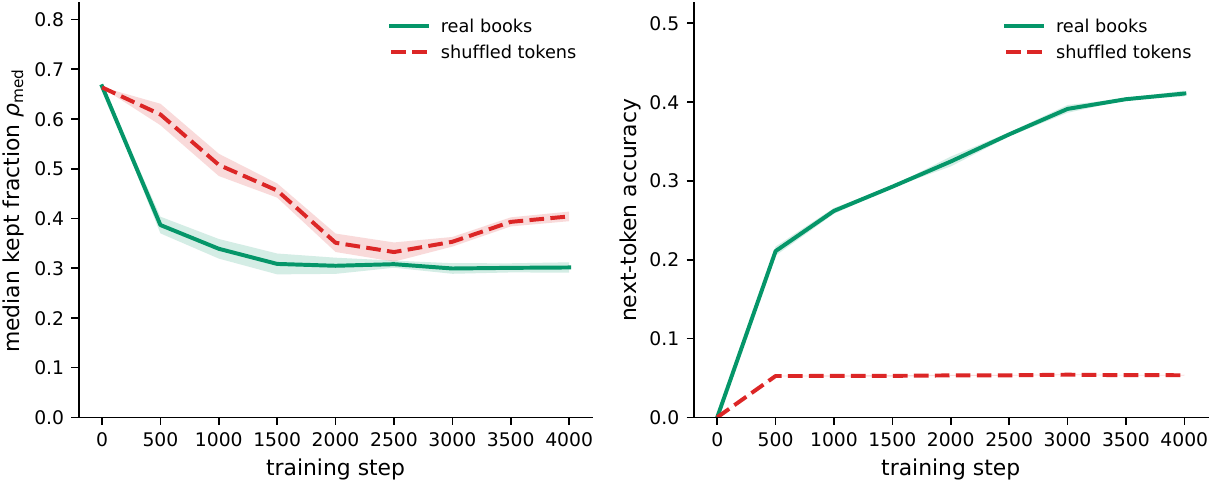}
\caption{\textbf{Sparsity accompanies learned alignment in a controlled
from-scratch replication} (\textsc{Qwen3}-architecture byte-LM).
\textbf{(a)} Median kept fraction, real versus shuffled text;
\textbf{(b)} next-token accuracy.  Curves show mean \({\pm}\) sample standard
deviation over seeds \(0,1,2\); the single-checkpoint legacy control is
omitted.}
\label{fig:fromscratch}
\end{figure*}

\paragraph{Heterogeneity across fixed head types.}
Pooling all heads would mix different structural profiles.  We therefore
assign the MInference-inspired static types before aggregation and display
their count, concentration, and model-composition distributions separately.
The violin plots are descriptive finite distributions; the black summaries
use model-balanced centers and document-bootstrap intervals.  These static
profile types should not be read as functional retrieval or induction labels.
The three types are \emph{sink} (A-shape, with high initial-token mass),
\emph{concentrated} (Vertical-Slash, with a few dominant non-sink columns),
and \emph{dense} (Block-Sparse, with broad mass).  Their composition differs
substantially by architecture: the Llama family is sink-heavy, while
\textsc{Qwen3} and \textsc{gpt-oss} are dense-heavy under this static
classifier.  We therefore pool within type only after assignment and use an
equal-model center, preventing a large architecture or a common head type
from silently dominating the descriptive summary.

\begin{figure*}[!t]
\centering
\includegraphics[width=\textwidth]{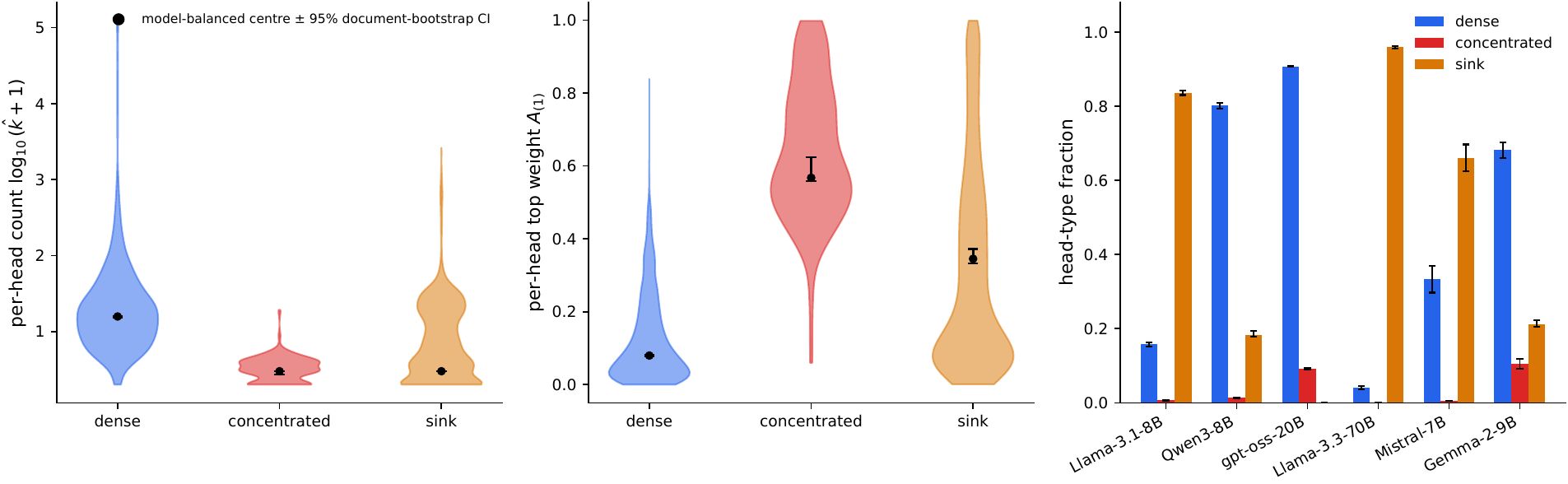}
\caption{\textbf{Per-head distributions by static type.}  Left/middle:
violins show the finite observed distributions of per-head count \(\hat k\)
and top weight \(A_{(1)}\) at the longest context; black points and bars are
the model-balanced center and seed-\(0\), \(B=5000\), \(95\%\) percentile
document-bootstrap interval.  Right: head-type composition per anchor with
the same document-bootstrap intervals.  The types summarize static profiles
and should not be read as functional head labels.}
\label{fig:perhead}
\end{figure*}

\paragraph{Replication across two controlled architectures.}
A second from-scratch check measures the sparse-head fraction throughout
training in two small byte-level language-model architectures, with matched
random-byte controls.  Both real-text trajectories cross the pre-specified
\(15\%\) guide and remain above their controls.  This supports recurrence
across the two controlled architectures; it does not establish an
in-the-wild model-scale trend or task utility from sparsity alone.  The
\textsc{GPT-NeoX}-like and \textsc{Qwen3}-like models have \(3.22\)M and
\(4.26\)M parameters, respectively, and each real/random trajectory is
averaged over three seeds.  The real-text endpoints reach approximately
\(93\%\) and \(75\%\) sparse heads, respectively; the random-byte controls
remain nonzero but lower, so the comparison supports a training-data effect
rather than an architectural zero baseline.

\begin{figure}[t]
\centering
\includegraphics[width=0.8\columnwidth]{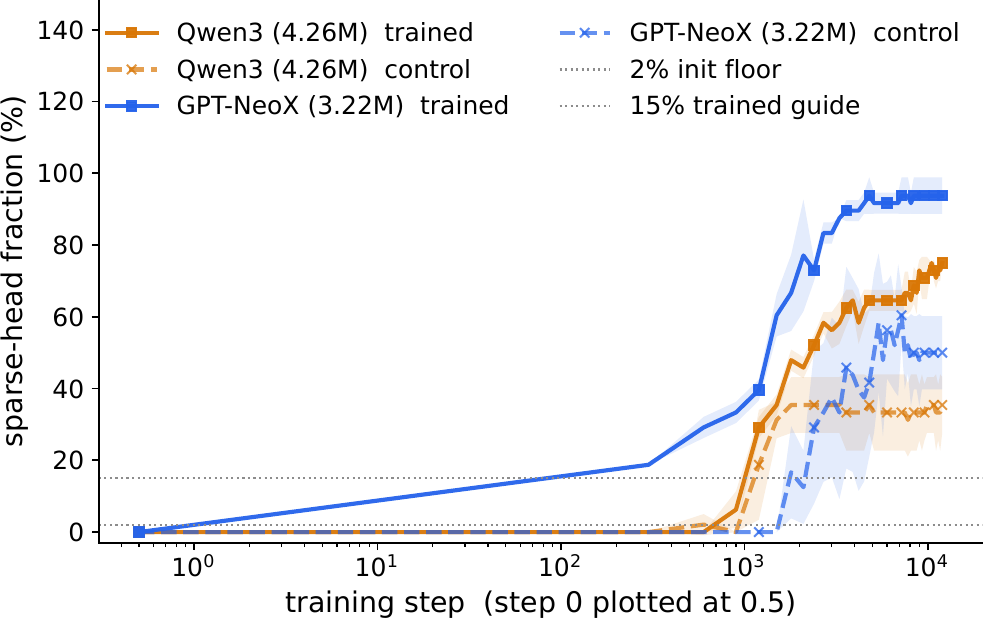}
\caption{\textbf{Sparse-head growth recurs in two controlled architectures.}
Sparse-head fraction (kept-fraction
\(\rho_{\eta=0.1}\leq0.1\)) versus training step for two from-scratch
byte-level language models (\textsc{GPT-NeoX}-architecture, \(3.22\)M
parameters; \textsc{Qwen3}-architecture, \(4.26\)M parameters; three seeds,
solid \(=\) mean, band \(=\pm\) sample standard deviation).  On real text
both trajectories cross the \(15\%\) guide (\textsc{GPT-NeoX} to
\({\sim}93\%\), \textsc{Qwen3} to \({\sim}75\%\) of heads sparse); the
i.i.d.-random-byte controls (dashed) are nonzero but remain lower.}
\label{fig:emergence}
\end{figure}

\subsection{Evaluation design}
\label{sub:submission-roster}

\paragraph{Profile anchors.}
Five models---\textsc{Llama-3.1-8B} and \textsc{Llama-3.3-70B}
\cite{grattafiori2024llama3}, \textsc{Qwen3-8B} \cite{yang2025qwen3},
\textsc{gpt-oss-20B} \cite{openai2025gptoss}, and
\textsc{Mistral-7B}---provide profiles between \(4\)K and \(128\)K.
The recovered \textsc{Gemma-2-9B} surface contains only \(4\)K and is used as
a descriptive control, never for growth inference.  The available profile
records contain one natural-text surface: Wikipedia fallback documents
stored under a legacy \texttt{pg19} key.  We preserve this data-source label
and do not describe the surface as a multi-domain atlas.

\paragraph{Theory-validation protocol.}
The relative-count thresholds are
\(\epsilon\in\{0.5,0.1,0.05\}\) times the row maximum, and coverage targets are
\(\eta\in\{0.2,0.1,0.05,0.02,0.01\}\).  Growth inference requires at least
three distinct lengths and at least two finite document slopes per head.
The hardened report uses document bootstrap seed \(0\), \(B=5000\), and
Benjamini--Hochberg correction at \(q=0.05\)
\cite{benjamini1995fdr}; underidentified and non-finite fits never enter the
denominator.  The short-to-long-context test fits the profile at \(4\)K and,
without refitting, predicts the \(128\)K coverage budget.  Its pre-specified
success criterion is multiplicative error at most \(1.5\times\).  Controlled
task curves use shared evaluation examples and state their resampling unit in
the caption.  The separately trained sparse-model experiment is an
exploratory observation, not an instantiation of
\cref{thm:denoising-phase-law}.

\paragraph{Matched-budget task protocol.}
LongBench-v2 \cite{bai2024longbench2} uses all \(503\) unique, aligned
multiple-choice examples with \(32\)K truncation in every
\(2/4/10\times\) setting.  Inference uses seed \(42\); example-bootstrap
summaries use seed \(0\), \(B=5000\), and 95\% percentile intervals.  The
\(36\) planned method contrasts form one Holm family.  The complete
\textsc{RULER} evaluation \cite{hsieh2024ruler} uses nine fixed model--length
settings.  The example fractions are \((.10,.10,.10,.03)\) for Llama at
\(16/32/64/128\)K, \((.10,.10,.05,.03)\) for Qwen, and \(.10\) for Mistral
at \(16\)K; all methods within a setting share the same sampled examples.
Task-level resampling uses the same statistics seed and bootstrap count.  For
every setting we retain the configuration, raw predictions, scored outputs,
item-level outcomes, and immutable log, and verify their file checksums before
computing a table.

\paragraph{Software and compute.}
Production evaluation uses \textsc{PyTorch} 2.8.0+cu128 and
\textsc{Transformers} 4.57.6 in \texttt{bfloat16}; attention-row logits are
recomputed in \texttt{fp32}.  Experiments run on one 96\,GB NVIDIA RTX PRO
6000.  Table generators and configurations are identified by SHA-256; the
anonymous code release will also include the complete environment export.

Across all experiments conducted on the audited instance, we used 205.70 GPU-hours, corresponding to 160.13 powered instance-hours across one- and two-GPU configurations on 96 GB NVIDIA RTX PRO 6000 GPUs.

\subsection{One-factor ablations}

\begin{table}[H]
\centering
\scriptsize
\setlength{\tabcolsep}{1.6pt}
\renewcommand{\arraystretch}{0.82}
\begin{adjustbox}{max width=\columnwidth}
\begin{tabular}{@{}llccc@{}}
\toprule
Component & Variant & $4\times$ & $10\times$ & Avg. \\
\midrule
Statistic & Tail mass & $24.7{\pm}3.7$ & $23.9{\pm}3.7$ & $24.3{\pm}3.5$ \\
 & Uncentered dispersion & $25.8{\pm}3.8$ & $23.5{\pm}3.7$ & $24.7{\pm}3.5$ \\
 & Centered dispersion & $26.0{\pm}3.8$ & $24.1{\pm}3.7$ & $25.0{\pm}3.6$ \\
\midrule
Allocator & True uniform & $26.2{\pm}3.8$ & $23.1{\pm}3.7$ & $24.7{\pm}3.5$ \\
 & Greedy marginal & $26.4{\pm}3.8$ & $23.1{\pm}3.7$ & $24.8{\pm}3.5$ \\
 & Lagrangian + repair & $26.0{\pm}3.8$ & $24.1{\pm}3.7$ & $25.0{\pm}3.6$ \\
\midrule
Tail summary & No slot & $26.0{\pm}3.8$ & $23.7{\pm}3.7$ & $24.9{\pm}3.5$ \\
 & Plain slot & $26.0{\pm}3.8$ & $24.1{\pm}3.7$ & $25.0{\pm}3.6$ \\
 & Calibrated $\bar q$ slot & $25.4{\pm}3.8$ & $24.5{\pm}3.7$ & $25.0{\pm}3.5$ \\
\midrule
Safeguard & $\alpha=0$ & $25.8{\pm}3.8$ & $24.3{\pm}3.7$ & $25.0{\pm}3.6$ \\
 & $\alpha=0.1$ & $26.0{\pm}3.8$ & $24.3{\pm}3.7$ & $25.1{\pm}3.6$ \\
 & $\alpha=0.2$ & $26.0{\pm}3.8$ & $24.1{\pm}3.7$ & $25.0{\pm}3.6$ \\
 & $\alpha=0.4$ & $25.8{\pm}3.8$ & $23.9{\pm}3.7$ & $24.9{\pm}3.5$ \\
\bottomrule
\end{tabular}

\end{adjustbox}
\caption{\textbf{\certkv{} ablations on Llama LongBench-v2.}
Accuracy is estimate \({\pm}\) 95\% example-bootstrap half-width (seed \(0\),
\(B=5000\)), conditional on one fixed checkpoint and configuration per row.
Rows are unranked one-factor changes; nine paired \(p\)-values form one family.}
\label{tab:certkv-ablation-v2}
\end{table}

\paragraph{Component attribution under the same physical budget.}
We further isolate which \certkv{} components improve attention-output
fidelity under exact live-slot accounting.  Starting from one shared
full-cache \textsc{Llama-3.1-8B-Instruct} capture, we evaluate \(40\)
preselected held-out GovReport documents (indices \(6\)--\(45\)), each
truncated to \(8{,}192\) tokens.  The evaluation crosses \(4\times\) and
\(10\times\) compression, layers \(0,8,16,24,31\), all eight KV heads, and
\(32\) realized future-query positions.  Each contrast changes exactly one
design choice: centering, cross-head allocation, the charged tail-summary
slot, or the per-head safeguard.  When the tail summary is removed, its
physical slot is reassigned to one additional real token per head, preserving
the same global live-slot budget.

Within each document, errors are averaged over the
ratio--layer--head grid before statistical inference, making the document the
independent unit.  Inference uses seed \(42\).  We report paired percentile
bootstrap intervals with seed \(0\), \(B=5000\), and 95\% coverage.  The four
two-sided paired sign-flip tests form one Holm-corrected family.  A component
passes the pre-specified practical-effect gate only when the upper confidence
endpoint is negative, \(p_{\mathrm{Holm}}<0.05\), and the relative error
reduction is at least \(10\%\).  The safeguard must additionally keep the
upper confidence bound on its relative mean-error increase below \(2\%\).

\begin{table}[H]
\centering
\scriptsize
\setlength{\tabcolsep}{1.2pt}
\renewcommand{\arraystretch}{0.86}
\begin{adjustbox}{max width=\columnwidth}
\begin{tabular}{@{}lrrrc@{}}
\toprule
Intervention
  & Effect \(\Delta\downarrow\)
  & Rel.
  & Holm \(p\)
  & \(10\%\) gate \\
\midrule
Centering vs.\ uncentered statistic
  & \((-8.044 \pm 3.907)\times 10^{-4}\)
  & \(0.203\%\)
  & \(1.320\times 10^{-3}\)
  & No \\
Adaptive vs.\ uniform allocation
  & \((-9.995 \pm 2.519)\times 10^{-3}\)
  & \(2.466\%\)
  & \(4.000\times 10^{-5}\)
  & No \\
Tail summary vs.\ \(+1\) real token
  & \((1.494 \pm 0.994)\times 10^{-4}\)
  & \(-0.038\%\)
  & \(9.180\times 10^{-3}\)
  & No \\
Safeguard (p95) vs.\ \(\alpha=0\)
  & \((-2.194 \pm 4.226)\times 10^{-5}\)
  & \(0.002\%\)
  & \(1.000\)
  & No \\
\bottomrule
\end{tabular}
\end{adjustbox}
\caption{\textbf{Exact-budget component interventions.}
Relative attention-output error is lower-is-better.  The first three rows
evaluate document-level mean error; the safeguard row evaluates document-level
p95 error.  We define
\(\Delta=\text{design}-\text{control}\), so negative values favor the design.
Entries report the point estimate \({\pm}\) half-width of the registered
95\% percentile interval.  Rel.\ denotes relative error reduction.  The final
column reports the pre-specified practical-effect gate rather than statistical
significance alone.}
\label{tab:certkv-component-ablation}
\end{table}

Adaptive allocation gives the largest resolved improvement: its mean relative
output error is \(0.395369\), compared with \(0.405364\) under true-uniform
allocation
(\(\Delta=-9.995\times10^{-3}\), 95\% CI
\([-1.246\times10^{-2},-7.420\times10^{-3}]\),
\(2.466\%\) relative reduction,
\(p_{\mathrm{Holm}}=4.000\times10^{-5}\)).
Centering provides a smaller but consistently directed improvement over the
uncentered statistic
(\(\Delta=-8.044\times10^{-4}\), 95\% CI
\([-1.186\times10^{-3},-4.047\times10^{-4}]\),
\(0.203\%\) relative reduction,
\(p_{\mathrm{Holm}}=1.320\times10^{-3}\)).
Thus, under this controlled fidelity metric, cross-head allocation accounts
for the clearest component-level gain, with centering providing an additional
smaller improvement.  Both effects are statistically resolved but remain
below the pre-specified \(10\%\) practical-effect threshold.

The charged tail summary is slightly worse than assigning the same physical
slot to one additional real token
(\(\Delta=1.494\times10^{-4}\), 95\% CI
\([4.664\times10^{-5},2.454\times10^{-4}]\),
\(p_{\mathrm{Holm}}=9.180\times10^{-3}\)).
The safeguard does not resolve a p95-error improvement
(\(\Delta=-2.194\times10^{-5}\), 95\% CI
\([-7.517\times10^{-5},9.347\times10^{-6}]\),
\(p_{\mathrm{Holm}}=1.000\)), although its registered mean-error
non-inferiority check passes.  These results attribute the measured
shared-capture fidelity gains primarily to allocation and, secondarily, to
centering.  They characterize attention-output fidelity under realized
queries and should not be interpreted as downstream-accuracy evidence or as
an unconditional guarantee for arbitrary future queries.

\subsection{Physical cache capacity}

The physical prototype materializes a packed \texttt{bfloat16} cache for
\textsc{Llama-3.1-8B}.  At \(64\)K and \(10\times\), the packed \certkv{}
cache stores \(0.86\) GB of persistent K/V state, compared with \(8.59\) GB
for the full cache, and reduces decode peak memory from \(23.11\) to
\(16.34\) GiB.  Persistent bytes are exact; the displayed peak is unchanged
across five repetitions at the reported precision.

\begin{table}[H]
\centering
\begin{tabular}{@{}lcc@{}}
\toprule
Method & Persistent KV & Decode peak \\
\midrule
Full cache (\(1\times\)) & \(8.59\) GB & \(23.11\) GiB \\
\certkv{} (\(10\times\)) & \(0.86\) GB & \(16.34\) GiB \\
\bottomrule
\end{tabular}

\caption{\textbf{Physical memory relative to the full cache.}
The \(64\)K \certkv{} row uses the exact \(10\times\) live-slot budget.}
\label{tab:physical-cache}
\label{tab:physical-memory}
\label{tab:efficiency}
\end{table}

\subsection{LongBench-v2 domain breakdown}

The domain tables retain the exact 503-example denominator.  Intervals
resample examples within the displayed domain; Avg.\ is weighted by the
actual domain counts rather than averaging six domain percentages.  Full
cache is shown as an unranked reference.  Bold/underline mark the best/second
distinct compressed point estimates; statistical decisions remain governed
by the paired Holm family defined above.

\subsection{Reproducibility and data traceability}

For every reported number, the release links the configuration and code hash
to raw predictions, item-level outcomes, the resampling report, and the final
table or figure.  Exploratory sweeps, incomplete settings, and results without
the stated denominator are excluded.  The release records the exact
generators, configurations, environment, and a one-command validation entry
point.

\subsection{Reading the domain tables}

\paragraph{The aggregate pattern is architecture-dependent.}
At \(2/\allowbreak4/\allowbreak10\times\), \certkv{} attains
\(26.8/\allowbreak26.0/\allowbreak24.1\) on \textsc{Llama-3.1-8B},
\(31.0/\allowbreak30.0/\allowbreak29.0\) on \textsc{Qwen3-8B}, and
\(28.0/\allowbreak28.2/\allowbreak28.2\) on \textsc{Mistral-7B}.  The strongest
compressed point estimates are, respectively,
\(27.0/\allowbreak27.2/\allowbreak26.0\),
\(32.4/\allowbreak30.8/\allowbreak30.2\), and
\(28.8/\allowbreak28.4/\allowbreak28.2\).  Thus the exact-budget method is
close to the leading point across the grid and reaches the leading
\textsc{Mistral} average at \(10\times\), but it is not uniformly the largest
entry.  Formal comparative claims use the paired Holm family rather than these
unpaired domain intervals.

\paragraph{Domain estimates are deliberately descriptive.}
Half-widths are often \(6\)--\(18\) points because each domain contains only
a subset of the \(503\) examples.  They localize aggregate differences but do
not license separate domain significance claims or a domain-wise SOTA label.

\begin{table*}[t]
\centering
\resizebox{\textwidth}{!}{
\begingroup
\scriptsize
\setlength{\tabcolsep}{1.5pt}
\begin{tabular}{@{}llccccccc@{}}
\toprule
Budget & Method & Single-Doc & Multi-Doc & ICL & Dialogue & Code & Structured & Avg. (503-wtd.) \\
\midrule
$2\times$ & Full cache & \(30.3{\pm}6.9\) & \(32.0{\pm}8.0\) & \(23.5{\pm}9.3\) & \(15.4{\pm}11.5\) & \(30.0{\pm}12.0\) & \(6.1{\pm}7.6\) & \(26.8{\pm}3.9\) \\
$2\times$ & SnapKV & \(\mathbf{34.3{\pm}6.9}\) & \(28.0{\pm}8.0\) & \(\mathbf{22.2{\pm}9.3}\) & \(\underline{15.4{\pm}11.5}\) & \(\underline{28.0{\pm}12.0}\) & \(\underline{9.1{\pm}10.6}\) & \(\mathbf{27.0{\pm}3.9}\) \\
$2\times$ & StreamingLLM & \(32.6{\pm}6.9\) & \(\underline{28.8{\pm}8.0}\) & \(19.8{\pm}8.6\) & \(\mathbf{17.9{\pm}11.5}\) & \(\underline{28.0{\pm}12.0}\) & \(\mathbf{18.2{\pm}13.6}\) & \(\mathbf{27.0{\pm}3.9}\) \\
$2\times$ & PyramidKV & \(20.6{\pm}6.0\) & \(19.2{\pm}6.8\) & \(19.8{\pm}8.6\) & \(\mathbf{17.9{\pm}11.5}\) & \(4.0{\pm}5.0\) & \(\mathbf{18.2{\pm}13.6}\) & \(18.1{\pm}3.3\) \\
$2\times$ & Ada-KV & \(\underline{33.1{\pm}6.9}\) & \(\mathbf{29.6{\pm}8.0}\) & \(\underline{21.0{\pm}8.6}\) & \(\underline{15.4{\pm}11.5}\) & \(\mathbf{30.0{\pm}12.0}\) & \(3.0{\pm}4.5\) & \(26.6{\pm}3.9\) \\
$2\times$ & CertKV & \(\underline{33.1{\pm}6.9}\) & \(\mathbf{29.6{\pm}8.0}\) & \(\mathbf{22.2{\pm}9.3}\) & \(\underline{15.4{\pm}11.5}\) & \(\mathbf{30.0{\pm}12.0}\) & \(3.0{\pm}4.5\) & \(\underline{26.8{\pm}3.9}\) \\
\midrule
$4\times$ & Full cache & \(30.3{\pm}6.9\) & \(32.0{\pm}8.0\) & \(23.5{\pm}9.3\) & \(15.4{\pm}11.5\) & \(30.0{\pm}12.0\) & \(6.1{\pm}7.6\) & \(26.8{\pm}3.9\) \\
$4\times$ & SnapKV & \(32.0{\pm}6.9\) & \(\mathbf{26.4{\pm}8.0}\) & \(\mathbf{23.5{\pm}9.3}\) & \(\underline{17.9{\pm}11.5}\) & \(24.0{\pm}12.0\) & \(6.1{\pm}7.6\) & \(25.6{\pm}3.8\) \\
$4\times$ & StreamingLLM & \(\underline{32.6{\pm}6.9}\) & \(\underline{25.6{\pm}8.0}\) & \(19.8{\pm}8.6\) & \(\mathbf{23.1{\pm}12.8}\) & \(\mathbf{30.0{\pm}12.0}\) & \(\mathbf{24.2{\pm}15.2}\) & \(\mathbf{27.2{\pm}3.9}\) \\
$4\times$ & PyramidKV & \(18.9{\pm}5.7\) & \(20.0{\pm}7.2\) & \(\underline{22.2{\pm}9.3}\) & \(\mathbf{23.1{\pm}12.8}\) & \(6.0{\pm}7.0\) & \(\underline{21.2{\pm}13.6}\) & \(18.9{\pm}3.5\) \\
$4\times$ & Ada-KV & \(\mathbf{33.7{\pm}6.9}\) & \(\underline{25.6{\pm}8.0}\) & \(19.8{\pm}8.6\) & \(15.4{\pm}11.5\) & \(24.0{\pm}12.0\) & \(6.1{\pm}7.6\) & \(25.2{\pm}3.9\) \\
$4\times$ & CertKV & \(\mathbf{33.7{\pm}6.9}\) & \(\mathbf{26.4{\pm}8.0}\) & \(21.0{\pm}8.6\) & \(15.4{\pm}11.5\) & \(\underline{26.0{\pm}12.0}\) & \(9.1{\pm}10.6\) & \(\underline{26.0{\pm}3.8}\) \\
\midrule
$10\times$ & Full cache & \(30.3{\pm}6.9\) & \(32.0{\pm}8.0\) & \(23.5{\pm}9.3\) & \(15.4{\pm}11.5\) & \(30.0{\pm}12.0\) & \(6.1{\pm}7.6\) & \(26.8{\pm}3.9\) \\
$10\times$ & SnapKV & \(\underline{29.7{\pm}6.6}\) & \(\underline{20.8{\pm}7.2}\) & \(23.5{\pm}9.3\) & \(\mathbf{20.5{\pm}12.8}\) & \(22.0{\pm}11.0\) & \(6.1{\pm}7.6\) & \(23.5{\pm}3.7\) \\
$10\times$ & StreamingLLM & \(\mathbf{34.3{\pm}6.9}\) & \(\mathbf{23.2{\pm}7.2}\) & \(18.5{\pm}8.6\) & \(\mathbf{20.5{\pm}12.8}\) & \(\mathbf{30.0{\pm}12.0}\) & \(\underline{12.1{\pm}10.6}\) & \(\mathbf{26.0{\pm}3.8}\) \\
$10\times$ & PyramidKV & \(21.7{\pm}6.3\) & \(16.8{\pm}6.4\) & \(18.5{\pm}8.6\) & \(\mathbf{20.5{\pm}12.8}\) & \(10.0{\pm}8.0\) & \(\mathbf{15.2{\pm}12.1}\) & \(18.3{\pm}3.4\) \\
$10\times$ & Ada-KV & \(28.0{\pm}6.9\) & \(\mathbf{23.2{\pm}7.2}\) & \(\underline{25.9{\pm}9.9}\) & \(\mathbf{20.5{\pm}12.8}\) & \(\underline{24.0{\pm}12.0}\) & \(6.1{\pm}7.6\) & \(\underline{24.1{\pm}3.7}\) \\
$10\times$ & CertKV & \(27.4{\pm}6.9\) & \(\mathbf{23.2{\pm}7.2}\) & \(\mathbf{27.2{\pm}9.9}\) & \(\mathbf{20.5{\pm}12.8}\) & \(\underline{24.0{\pm}12.0}\) & \(6.1{\pm}7.6\) & \(\underline{24.1{\pm}3.7}\) \\
\bottomrule
\end{tabular}
\endgroup
}
\caption{\textbf{Llama-3.1-8B LongBench-v2 by domain.}
Accuracy (\%) is estimate \({\pm}\) 95\% within-domain bootstrap half-width;
Avg.\ is 503-weighted.}
\label{tab:lbv2-domain}
\end{table*}

\paragraph{No single domain explains the averages.}
On \textsc{Llama} at \(4\times\), \certkv{} is strongest or tied on
single- and multi-document questions, whereas StreamingLLM gains its larger
average from dialogue, code, and structured-data items.  The leader changes
with ratio and domain on \textsc{Qwen}.  At \textsc{Mistral} \(10\times\),
\certkv{} leads single-document accuracy and is leading or tied on ICL,
dialogue, and structured data.  This heterogeneity supports an
architecture-dependent operating region, not a universal method ordering.

\paragraph{Compression-ratio sensitivity is likewise model-specific.}
The \textsc{Llama} average decreases from \(26.8\) at \(2\times\) to \(24.1\)
at \(10\times\), while \textsc{Qwen} changes from \(31.0\) to \(29.0\).
\textsc{Mistral} remains at \(28.0\)--\(28.2\) across the same budgets.
These trajectories should not be read as three estimates of one common
compression curve: each model has a different full-cache reference, and the
same nominal ratio acts on different attention and value geometries.  In
particular, the small non-monotonic changes on \textsc{Mistral} are well inside
the reported uncertainty and are not evidence that stronger compression
improves accuracy.

\paragraph{Point-estimate ranks are not significance decisions.}
Bold and underlined entries make the large table navigable within a
model--ratio block; they do not create a separate hypothesis test for each
domain.  The domain subsets overlap neither in size nor in difficulty, and
their intervals are marginal rather than paired method-difference intervals.
Accordingly, the paper bases comparative claims on the pre-specified
example-paired Holm family over the complete \(503\)-example evaluation.
The full cache is retained as a calibration reference but is deliberately
excluded from compressed-method ranking.

\paragraph{What the breakdown rules out.}
The tables show that the aggregate result is not produced by a single
favorable task family repeated across all architectures.  \textsc{Llama},
\textsc{Qwen}, and \textsc{Mistral} obtain their strongest cells from different
domains, yet \certkv{} remains in the competitive band in all three aggregate
rows.  This supports the narrower conclusion used in the main text:
exact-budget, value-aware selection is competitive across the declared
four-baseline grid.  It does not support a domain-wise SOTA claim, nor does it
identify a universal semantic category in which one eviction rule must win.

\begin{table*}[t]
\centering
\resizebox{\textwidth}{!}{
\begingroup
\scriptsize
\setlength{\tabcolsep}{1.5pt}
\begin{tabular}{@{}llccccccc@{}}
\toprule
Budget & Method & Single-Doc & Multi-Doc & ICL & Dialogue & Code & Structured & Avg. (503-wtd.) \\
\midrule
$2\times$ & Full cache & \(32.6{\pm}6.9\) & \(34.4{\pm}8.8\) & \(32.1{\pm}9.9\) & \(33.3{\pm}15.4\) & \(34.0{\pm}13.0\) & \(33.3{\pm}15.2\) & \(33.2{\pm}4.1\) \\
$2\times$ & SnapKV & \(\mathbf{34.3{\pm}6.9}\) & \(\mathbf{33.6{\pm}8.8}\) & \(27.2{\pm}9.9\) & \(\underline{28.2{\pm}14.1}\) & \(\mathbf{34.0{\pm}13.0}\) & \(\underline{33.3{\pm}15.2}\) & \(\mathbf{32.4{\pm}4.1}\) \\
$2\times$ & StreamingLLM & \(\underline{33.7{\pm}6.9}\) & \(28.0{\pm}8.0\) & \(\mathbf{29.6{\pm}9.9}\) & \(12.8{\pm}10.3\) & \(\underline{30.0{\pm}12.0}\) & \(\underline{33.3{\pm}15.2}\) & \(29.6{\pm}4.0\) \\
$2\times$ & PyramidKV & \(27.4{\pm}6.9\) & \(26.4{\pm}8.0\) & \(\underline{28.4{\pm}9.9}\) & \(23.1{\pm}12.8\) & \(26.0{\pm}12.0\) & \(\mathbf{45.5{\pm}18.2}\) & \(28.0{\pm}3.9\) \\
$2\times$ & Ada-KV & \(32.0{\pm}6.9\) & \(\underline{30.4{\pm}8.0}\) & \(\mathbf{29.6{\pm}9.9}\) & \(\mathbf{33.3{\pm}15.4}\) & \(\mathbf{34.0{\pm}13.0}\) & \(\underline{33.3{\pm}15.2}\) & \(\underline{31.6{\pm}4.1}\) \\
$2\times$ & CertKV & \(33.1{\pm}6.9\) & \(28.8{\pm}8.0\) & \(\underline{28.4{\pm}9.9}\) & \(\underline{28.2{\pm}14.1}\) & \(\mathbf{34.0{\pm}13.0}\) & \(\underline{33.3{\pm}15.2}\) & \(31.0{\pm}4.1\) \\
\midrule
$4\times$ & Full cache & \(32.6{\pm}6.9\) & \(34.4{\pm}8.8\) & \(32.1{\pm}9.9\) & \(33.3{\pm}15.4\) & \(34.0{\pm}13.0\) & \(33.3{\pm}15.2\) & \(33.2{\pm}4.1\) \\
$4\times$ & SnapKV & \(32.0{\pm}6.9\) & \(27.2{\pm}8.0\) & \(\underline{28.4{\pm}9.9}\) & \(\underline{23.1{\pm}12.8}\) & \(\mathbf{34.0{\pm}13.0}\) & \(30.3{\pm}15.2\) & \(29.6{\pm}4.0\) \\
$4\times$ & StreamingLLM & \(\underline{32.6{\pm}6.9}\) & \(27.2{\pm}8.0\) & \(\mathbf{30.9{\pm}9.9}\) & \(17.9{\pm}11.5\) & \(30.0{\pm}12.0\) & \(27.3{\pm}15.2\) & \(29.2{\pm}4.0\) \\
$4\times$ & PyramidKV & \(26.9{\pm}6.9\) & \(\underline{28.8{\pm}8.0}\) & \(27.2{\pm}9.9\) & \(\mathbf{28.2{\pm}14.1}\) & \(22.0{\pm}11.0\) & \(\mathbf{42.4{\pm}16.7}\) & \(28.0{\pm}3.9\) \\
$4\times$ & Ada-KV & \(\mathbf{33.1{\pm}6.9}\) & \(\mathbf{30.4{\pm}8.0}\) & \(\underline{28.4{\pm}9.9}\) & \(\underline{23.1{\pm}12.8}\) & \(\underline{32.0{\pm}13.0}\) & \(\underline{33.3{\pm}15.2}\) & \(\mathbf{30.8{\pm}4.1}\) \\
$4\times$ & CertKV & \(30.9{\pm}6.9\) & \(\underline{28.8{\pm}8.0}\) & \(\underline{28.4{\pm}9.9}\) & \(\mathbf{28.2{\pm}14.1}\) & \(\underline{32.0{\pm}13.0}\) & \(\underline{33.3{\pm}15.2}\) & \(\underline{30.0{\pm}4.0}\) \\
\midrule
$10\times$ & Full cache & \(32.6{\pm}6.9\) & \(34.4{\pm}8.8\) & \(32.1{\pm}9.9\) & \(33.3{\pm}15.4\) & \(34.0{\pm}13.0\) & \(33.3{\pm}15.2\) & \(33.2{\pm}4.1\) \\
$10\times$ & SnapKV & \(\underline{29.7{\pm}6.6}\) & \(\underline{24.8{\pm}7.2}\) & \(\mathbf{35.8{\pm}10.5}\) & \(\mathbf{30.8{\pm}14.1}\) & \(\underline{32.0{\pm}13.0}\) & \(\underline{36.4{\pm}15.2}\) & \(\mathbf{30.2{\pm}4.0}\) \\
$10\times$ & StreamingLLM & \(\mathbf{31.4{\pm}6.9}\) & \(\mathbf{28.8{\pm}8.0}\) & \(30.9{\pm}9.9\) & \(17.9{\pm}11.5\) & \(\mathbf{34.0{\pm}13.0}\) & \(27.3{\pm}15.2\) & \(\underline{29.6{\pm}4.0}\) \\
$10\times$ & PyramidKV & \(25.1{\pm}6.3\) & \(\mathbf{28.8{\pm}8.0}\) & \(30.9{\pm}9.9\) & \(\underline{23.1{\pm}12.8}\) & \(24.0{\pm}12.0\) & \(\mathbf{45.5{\pm}18.2}\) & \(28.0{\pm}3.9\) \\
$10\times$ & Ada-KV & \(28.6{\pm}6.9\) & \(24.0{\pm}7.6\) & \(33.3{\pm}9.9\) & \(\mathbf{30.8{\pm}14.1}\) & \(26.0{\pm}12.0\) & \(33.3{\pm}15.2\) & \(28.4{\pm}3.9\) \\
$10\times$ & CertKV & \(29.1{\pm}6.9\) & \(\underline{24.8{\pm}7.2}\) & \(\underline{34.6{\pm}9.9}\) & \(\mathbf{30.8{\pm}14.1}\) & \(28.0{\pm}12.0\) & \(30.3{\pm}15.2\) & \(29.0{\pm}4.0\) \\
\bottomrule
\end{tabular}
\endgroup
}
\caption{\textbf{Qwen3-8B LongBench-v2 by domain.}
Accuracy (\%) is estimate \({\pm}\) 95\% within-domain bootstrap half-width;
Avg.\ is 503-weighted.}
\label{tab:lbv2-domain-qwen}
\end{table*}

\begin{table*}[t]
\centering
\resizebox{\textwidth}{!}{
\begingroup
\scriptsize
\setlength{\tabcolsep}{1.5pt}
\begin{tabular}{@{}llccccccc@{}}
\toprule
Budget & Method & Single-Doc & Multi-Doc & ICL & Dialogue & Code & Structured & Avg. (503-wtd.) \\
\midrule
$2\times$ & Full cache & \(29.1{\pm}6.9\) & \(30.4{\pm}8.0\) & \(29.6{\pm}9.9\) & \(28.2{\pm}14.1\) & \(18.0{\pm}11.0\) & \(24.2{\pm}15.2\) & \(28.0{\pm}3.9\) \\
$2\times$ & SnapKV & \(30.9{\pm}6.9\) & \(\underline{29.6{\pm}8.0}\) & \(25.9{\pm}9.9\) & \(\underline{23.1{\pm}12.8}\) & \(\underline{20.0{\pm}11.0}\) & \(18.2{\pm}13.6\) & \(27.2{\pm}3.9\) \\
$2\times$ & StreamingLLM & \(\underline{32.6{\pm}6.9}\) & \(\mathbf{30.4{\pm}8.0}\) & \(\mathbf{29.6{\pm}9.9}\) & \(\mathbf{25.6{\pm}14.1}\) & \(18.0{\pm}11.0\) & \(\underline{21.2{\pm}13.6}\) & \(\mathbf{28.8{\pm}3.9}\) \\
$2\times$ & PyramidKV & \(21.7{\pm}6.3\) & \(26.4{\pm}8.0\) & \(24.7{\pm}9.3\) & \(12.8{\pm}10.3\) & \(\mathbf{24.0{\pm}12.0}\) & \(18.2{\pm}13.6\) & \(22.7{\pm}3.7\) \\
$2\times$ & Ada-KV & \(31.4{\pm}6.9\) & \(\mathbf{30.4{\pm}8.0}\) & \(\underline{27.2{\pm}9.9}\) & \(20.5{\pm}12.8\) & \(\underline{20.0{\pm}11.0}\) & \(\underline{21.2{\pm}13.6}\) & \(27.8{\pm}3.9\) \\
$2\times$ & CertKV & \(\mathbf{33.1{\pm}6.9}\) & \(\underline{29.6{\pm}8.0}\) & \(\underline{27.2{\pm}9.9}\) & \(17.9{\pm}11.5\) & \(18.0{\pm}11.0\) & \(\mathbf{24.2{\pm}15.2}\) & \(\underline{28.0{\pm}3.9}\) \\
\midrule
$4\times$ & Full cache & \(29.1{\pm}6.9\) & \(30.4{\pm}8.0\) & \(29.6{\pm}9.9\) & \(28.2{\pm}14.1\) & \(18.0{\pm}11.0\) & \(24.2{\pm}15.2\) & \(28.0{\pm}3.9\) \\
$4\times$ & SnapKV & \(\mathbf{37.1{\pm}7.1}\) & \(\underline{28.8{\pm}8.0}\) & \(\underline{27.2{\pm}9.9}\) & \(\underline{20.5{\pm}12.8}\) & \(\underline{18.0{\pm}11.0}\) & \(9.1{\pm}10.6\) & \(\mathbf{28.4{\pm}3.9}\) \\
$4\times$ & StreamingLLM & \(32.0{\pm}6.9\) & \(\underline{28.8{\pm}8.0}\) & \(\mathbf{28.4{\pm}9.9}\) & \(\mathbf{28.2{\pm}14.1}\) & \(16.0{\pm}10.0\) & \(\mathbf{21.2{\pm}13.6}\) & \(28.0{\pm}3.9\) \\
$4\times$ & PyramidKV & \(20.6{\pm}6.0\) & \(\underline{28.8{\pm}8.0}\) & \(21.0{\pm}8.6\) & \(15.4{\pm}11.5\) & \(\mathbf{24.0{\pm}12.0}\) & \(9.1{\pm}10.6\) & \(21.9{\pm}3.6\) \\
$4\times$ & Ada-KV & \(33.7{\pm}6.9\) & \(\underline{28.8{\pm}8.0}\) & \(\mathbf{28.4{\pm}9.9}\) & \(\underline{20.5{\pm}12.8}\) & \(16.0{\pm}10.0\) & \(\underline{12.1{\pm}10.6}\) & \(27.4{\pm}3.9\) \\
$4\times$ & CertKV & \(\underline{35.4{\pm}7.1}\) & \(\mathbf{30.4{\pm}8.0}\) & \(\mathbf{28.4{\pm}9.9}\) & \(\underline{20.5{\pm}12.8}\) & \(16.0{\pm}10.0\) & \(9.1{\pm}10.6\) & \(\underline{28.2{\pm}3.9}\) \\
\midrule
$10\times$ & Full cache & \(29.1{\pm}6.9\) & \(30.4{\pm}8.0\) & \(29.6{\pm}9.9\) & \(28.2{\pm}14.1\) & \(18.0{\pm}11.0\) & \(24.2{\pm}15.2\) & \(28.0{\pm}3.9\) \\
$10\times$ & SnapKV & \(30.3{\pm}6.9\) & \(28.0{\pm}8.0\) & \(\mathbf{28.4{\pm}9.9}\) & \(\underline{23.1{\pm}12.8}\) & \(14.0{\pm}9.0\) & \(\underline{18.2{\pm}13.6}\) & \(26.4{\pm}3.8\) \\
$10\times$ & StreamingLLM & \(27.4{\pm}6.9\) & \(\mathbf{31.2{\pm}8.0}\) & \(\underline{23.5{\pm}9.3}\) & \(20.5{\pm}12.8\) & \(14.0{\pm}9.0\) & \(15.2{\pm}12.1\) & \(25.0{\pm}3.8\) \\
$10\times$ & PyramidKV & \(24.0{\pm}6.3\) & \(\underline{30.4{\pm}8.0}\) & \(16.0{\pm}8.0\) & \(15.4{\pm}11.5\) & \(\mathbf{22.0{\pm}11.0}\) & \(6.1{\pm}7.6\) & \(22.3{\pm}3.6\) \\
$10\times$ & Ada-KV & \(\underline{32.6{\pm}6.9}\) & \(28.0{\pm}8.0\) & \(\mathbf{28.4{\pm}9.9}\) & \(20.5{\pm}12.8\) & \(\underline{16.0{\pm}10.0}\) & \(15.2{\pm}12.1\) & \(\underline{27.0{\pm}3.9}\) \\
$10\times$ & CertKV & \(\mathbf{33.7{\pm}6.9}\) & \(28.0{\pm}8.0\) & \(\mathbf{28.4{\pm}9.9}\) & \(\mathbf{28.2{\pm}14.1}\) & \(14.0{\pm}9.0\) & \(\mathbf{21.2{\pm}13.6}\) & \(\mathbf{28.2{\pm}3.9}\) \\
\bottomrule
\end{tabular}
\endgroup
}
\caption{\textbf{Mistral-7B LongBench-v2 by domain.}
Accuracy (\%) is estimate \({\pm}\) 95\% within-domain bootstrap half-width;
Avg.\ is 503-weighted.}
\label{tab:lbv2-domain-mistral}
\end{table*}

\else
\begin{center}
\textbf{\LARGE Appendix}
\end{center}
{\hypersetup{linkcolor=black}
\tableofcontents
}
\newpage

\ifdefined\submissionappendix
\section{Detailed Theory Statements}
\else
\section{Extended Theory Development}
\fi
\label{sec:app-main-theory-record}

\ifdefined\submissionappendix
For completeness, this appendix gives the detailed statements underlying the
count--coverage--output--task chain in the main text.  It also fixes the
cross-reference surface used by the subsequent proofs.
\else
This appendix develops the broader theory beyond the three main separations,
including distributional regimes, coverage laws, value-aware approximation,
and task-dependent ordering.  The main text isolates the shortest
count--coverage--output--task chain; the chapters below provide its larger
technical context and cross-reference surface.
\fi

\section{Large-Weight Counts Under Explicit Input Regimes}\label{sec:sparsity}
\iclronly{\vspace{-2mm}}

\iclronly{How many weights in a row are \emph{large}?  Under i.i.d.\
sub-Gaussian logits, \(k(\epsilon)=\#\{j:A_j>\epsilon\}\) admits an
\(O(\log n)\) upper certificate at a calibrated threshold; for Gaussian
logits a separate deterministic threshold makes the count non-vacuously
\((1+o_\Pr(1))\log n\) (\cref{prop:calibrated-log-count}).  Under an additive
near-top law the relative-threshold count is sub-polynomial
(\cref{cor:subpoly-count}).  Both statements are strictly separate from the
coverage budget \(\bud\), which stays \(\Theta(n)\) in the Gaussian model.}%
\arxivonly{This section asks how many weights in a row are \emph{large}.  The
sub-Gaussian rung gives an \(O(\log n)\) upper certificate at its calibrated
threshold; that certificate can be empty and is not itself a coverage
statement.  The Gaussian specialization supplies a deterministic
non-vacuous threshold with \((1+o_\Pr(1))\log n\) exceedances
(\cref{prop:calibrated-log-count}), while an explicit additive near-top law
implies a sub-polynomial relative-threshold count
(\cref{cor:subpoly-count}).  We keep these count statements separate from the
coverage budget \(\bud\), which remains \(\Theta(n)\) in the Gaussian model.}

\iclronly{\vspace{-2mm}}
\subsection{The count of large weights}\label{sub:ladder}

\iclronly{Fix a threshold $\epsilon$ in the non-trivial window of \cref{rem:eps-window} and read off the count per regime, with $\delta \in (0,1)$ the failure probability and $\epsilon_*(n,\delta) = n^{-1+o(1)}$ the window floor.}%
\arxivonly{How many weights are large is where ``naturally sparse'' is literally true. Fix a threshold $\epsilon$ in the non-trivial window of \cref{rem:eps-window} and read off the count under each input regime. \textsc{Iid sub-Gaussian} (rung (iii)) is the cleanest generic-input model --- independent scores, abstracting away cross-token dependence; \textsc{Mixing} sub-Gaussian (rung (iv)) drops only that independence, keeping a sub-Gaussian marginal tail $\bar F(t) \le e^{-t^2/(2\sigma_s^2)}$. Throughout, $\delta \in (0,1)$ is the failure probability and $\epsilon_*(n,\delta) = n^{-1+o(1)}$ the window floor of \cref{rem:eps-window}.}

\begin{theorem}[Per-regime large-weight count bounds; informal version of \cref{thm:app-count-ladder}]\label{thm:naturally-sparse}
Write $k(\epsilon) = \#\{j : A_j > \epsilon\}$. Each input regime gives a count bound.
\begin{enumerate}[label=(\roman*),leftmargin=2em,itemsep=0.2em]
\item \textsc{Free}/\textsc{Bounded}: $k(\epsilon) \le \lceil 1/\epsilon \rceil$ deterministically, and $k(\epsilon) = 0$ once $\epsilon > e^{2S}/n$ under $|\ell_j| \le S$.
\item \textsc{Mixing}: $k(\epsilon) \le n\,\bar F\big(\log\tfrac1\epsilon + \log Z\big) + \wt O(\sqrt n)$.
\item \textsc{Iid sub-Gaussian}: with probability $1-2\delta$, for every $\epsilon \ge \epsilon_*(n,\delta)$,
\begin{align}\label{eq:naturally-sparse}
k(\epsilon) \;&\le\; \tfrac23\log(2n/\delta) + \sqrt{2\log(2n/\delta)} + 2 \notag \\
&=\; O\big(\log(n/\delta)\big),
\end{align}
\item \textsc{Mixing sub-Gaussian}: with probability $1-\delta$, simultaneously for every $\epsilon \ge \epsilon_*(n,\delta)$, $k(\epsilon) \le C_{\mathrm{mix}}\big(\sqrt{n\log(n/\delta)} + (\log n)^2\log(n/\delta)\big) = \wt O(\sqrt n)$.
\end{enumerate}
\end{theorem}

\iclronly{The two probabilistic rungs are upper certificates: i.i.d.\
sub-Gaussian scores give \(O(\log(n/\delta))\) at the calibrated threshold,
while mixing leaves a \(\widetilde O(\sqrt n)\) fluctuation.  Neither bound
alone proves that the chosen threshold lies below the realized maximum; the
Gaussian clause of \cref{prop:calibrated-log-count} supplies that separate
non-vacuity statement.}%
\arxivonly{Rung (i) is pigeonhole plus the deterministic weight cap of
\textsc{Bounded}; rung (ii) is the marginal exceedance count.  Rungs (iii)
and (iv) are upper certificates at calibrated thresholds: independence gives
\(O(\log(n/\delta))\), whereas mixing leaves a
\(\widetilde O(\sqrt n)\) fluctuation.  These bounds do not by themselves
show that the threshold is below the realized maximum.  The Gaussian clause
of \cref{prop:calibrated-log-count} establishes that missing non-vacuity
separately.}

\arxivonly{%
\begin{proof}[Proof sketch]
For rung (iii), conditioning on the score range $\max_j|\ell_j| \le S_\delta$ reduces a weight exceedance $A_j > \epsilon$ to a score exceedance $\ell_j > \tau$ at a calibrated level $\tau = \sigma_s\sqrt{2\log(n/\delta)}$, whose i.i.d.\ exceedance count is controlled by Bennett's inequality in its small-variance regime ($n\bar F(\tau) \le \delta$), giving the $\tfrac23$ Bernstein constant. Rung (iv) reruns the same calibration but replaces the i.i.d.\ Bennett bound by the Merlev\`ede--Peligrad--Rio Bernstein inequality for geometrically $\alpha$-mixing exceedance indicators, which leaves the $\wt O(\sqrt n)$ mixing-noise floor. Full proof in \cref{thm:app-count-ladder}.
\end{proof}}

\iclronly{Re-anchoring the threshold \emph{relative} to the top weight yields a separate two-sided law at \emph{fixed} confidence, measuring the count's exponent rather than its constant.}%
\arxivonly{Re-anchoring the threshold \emph{relative} to the top weight yields a separate two-sided law at \emph{fixed} confidence --- at the cost of measuring the count's exponent rather than its constant.}

\iclronly{\noindent\emph{Setup.}  The top order statistics obey the additive
near-top law in \cref{prop:relative-count-law}, and the count is taken at
\(\epsilon A_{(1)}\) for fixed \(\epsilon\in(0,1)\).}%
\arxivonly{\noindent\emph{Setup.}  The top order statistics obey the additive
near-top law in \cref{prop:relative-count-law}; this load-bearing condition is
stronger than a first-order marginal-tail approximation.  The count is taken
at the relative threshold \(\epsilon A_{(1)}\), equivalently within a fixed
logit gap \(\log(1/\epsilon)\) of the maximum.}

\begin{corollary}[Sub-polynomial count under an additive near-top law; informal version of \cref{cor:app-subpoly-count}]\label{cor:subpoly-count}
For every fixed \(\epsilon\in(0,1)\),
\begin{align}\label{eq:subpoly-count}
\frac{\log k(\epsilon A_{(1)})}{\sqrt{2\log n}}
\xrightarrow{\Pr}
\frac{\log(1/\epsilon)}{\sigma_s},
\end{align}
and hence \(k(\epsilon A_{(1)})=n^{o_\Pr(1)}\).
\end{corollary}

\iclronly{The exponent is identified two-sidedly in probability.  No
finite-\(n\) deterministic second-order constant is claimed
(\cref{rem:app-subpoly-rate}).}%
\arxivonly{The exponent is identified two-sidedly in probability.  This is a
conditional conversion from near-top score geometry to a relative-weight
count; it is not asserted from a first-order Gumbel-domain tail alone, and no
finite-\(n\) deterministic second-order constant is claimed
(\cref{rem:app-subpoly-rate}).}

\arxivonly{%
\begin{proof}[Proof sketch]
The additive near-top law yields, uniformly for
\(\log j=O(\sqrt{\log n})\),
\(\ell_{(1)}-\ell_{(j)}
=\sigma_s\log j/\sqrt{2\log n}+o_\Pr(1)\).
Bracketing the crossing of \(\log(1/\epsilon)\) at ranks
\(\exp\{[\log(1/\epsilon)\pm\xi]\sqrt{2\log n}/\sigma_s\}\) proves the
claim.  Full proof in \cref{cor:app-subpoly-count}.
\end{proof}}

\iclronly{The count is training-free, but counting is not covering: the coverage budget $\bud$ --- the keys needed to capture a $1-\eta$ mass fraction, hence to \emph{approximate} exact attention --- stays $\Theta(n)$ under the same regimes, the subject of \cref{sec:error}.}%
\arxivonly{The count is the precise sense in which ``your attention is naturally sublinear-sparse'' is a theorem --- a per-row, training-free property. The same generic regimes, however, leave the \emph{coverage} budget $\bud$ --- the number of keys needed to capture a $1-\eta$ fraction of the softmax \emph{mass}, hence the cost of \emph{approximating} exact attention --- at $\Theta(n)$. Counting is not covering; that gap, and exactly what it takes to close it, is the subject of \cref{sec:error}.}
\iclronly{\vspace{-2mm}}

\section{The Coverage Budget: Why Sparse Cannot Naturally Approximate Exact}\label{sec:error}

\iclronly{The $O(\log n)$ large weights of \cref{sec:sparsity} carry a \emph{vanishing} mass fraction, so covering $1-\eta$ needs $\Theta(n)$ keys and whether a sparse output approximates exact is value-dependent. A row is approximable by few keys only once the score scale grows: the network does not learn to be sparse (it already is), it learns an attention \emph{easily approximated by sparse attention}.}%
\arxivonly{\Cref{sec:sparsity} established the count: under generic inputs only $O(\log n)$ weights are large. We now price the \emph{approximation} of exact attention, and the verdict is: those few large weights carry only a \emph{vanishing} fraction of the mass, so covering a $1-\eta$ fraction needs a coverage budget of $\Theta(n)$ keys, and whether a sparse \emph{output} then approximates exact is value-dependent. A row becomes \emph{approximable} by a few keys only when the score scale grows --- a per-head property installed by training. In one line: the network does not learn to be sparse (it already is), it learns an attention \emph{easily approximated by sparse attention}.}

\subsection{From mass to output error}
\iclronly{\vspace{-2mm}}
\begin{lemma}[Mass-to-output reduction and bias identity; informal version of \cref{prop:bias-identity}]\label{lem:main-mass}
For every kept set $S$ with $p_S < 1$:
\begin{align}
\begin{split}\label{eq:bias-identity}
O \;&=\; p_S\, \wt O_S + (1 - p_S)\, \bar V_T, \\
\wt O_S - O \;&=\; (1 - p_S)\big( \wt O_S - \bar V_T \big),
\end{split}\\
\begin{split}\label{eq:mass-bound}
\norm{ \wt O_S - O }_2 \;&\le\; 2 \Vmax\, (1 - p_S), \\
\norm{ O_S - O }_2 \;&=\; (1 - p_S)\, \norm{ \bar V_T }_2 .
\end{split}
\end{align}
\end{lemma}

\iclronly{All output error thus routes through the uncovered mass $1-p_S$; an imperfect selector is priced by its missed mass and adversarial flat rows force $k=\Theta(n)$ (\cref{lem:main-selector,thm:main-tight}, \cref{sec:app-deferred}), with pricing below for the realized kept set.}%
\arxivonly{All output error thus routes through the uncovered mass $1-p_S$. An imperfect selector is priced one-sidedly by its missed mass (\cref{lem:main-selector}), and on adversarial flat rows every uncompensated $k$-subset needs $k = \Theta(n)$ (\cref{thm:main-tight}); both are deferred to \cref{sec:app-deferred}, and all pricing below is for the realized kept set.}
\iclronly{\vspace{-2mm}}
\subsection{Count is not coverage; the coverage law}

\begin{definition}[Coverage budget and kept fraction]\label{def:compression}
The \emph{coverage budget} $\bud$ (\cref{eq:coverage-def}) is the number of top weights carrying a $1-\eta$ fraction of the mass; the \emph{kept fraction} is $\cratio \coloneqq \bud/n$. Both price \emph{approximation} and are distinct from the count: a row can be $(\epsilon, O(\log n))$-sparse (\cref{thm:naturally-sparse}) yet require $\bud = \Theta(n)$ to approximate exact attention.
\end{definition}

\iclronly{The deterministic criterion (\cref{thm:main-dichotomy}, \cref{sec:app-profile}) classifies the growth of $\bud$ from the gap profile's log-slope, sharpening in a Gaussian null model to a closed form for $\cratio$.}%
\arxivonly{The deterministic criterion (\cref{thm:main-dichotomy}, \cref{sec:app-profile}) classifies the \emph{growth} of $\bud$ from the log-slope of the gap profile, and in a Gaussian null model it sharpens to a closed form for $\cratio$.}

\emph{Setup.} Let $\Phi$ be the standard normal CDF, and let $\sigma$ denote the logit scale, which in the transformer is the query's $G$-energy $\sigma = \sqrt{X_i^\top G X_i}$ (\cref{prop:scaled-inverse}).

\begin{theorem}[Coverage law: the kept fraction; informal version of \cref{thm:t2-coverage}]\label{thm:compression}
\emph{(i)~Distribution-free:} for any row, $\cratio$ is the $(1-\eta)$-quantile of the coverage--Lorenz curve of the normalized weights. \emph{(ii)~Gaussian null model:} if the logits are i.i.d.\ $N(0,\sigma^2)$ then $\cratio \toP \Phi\big(\Phi^{-1}(1-\eta) - \sigma\big)$.
\emph{(iii)~Input reading:} at fixed weights $\sigma = O(1)$, so $\cratio$ is bounded away from $0$ (e.g.\ $\cratio \approx 0.61$ at $\sigma=1$, $\eta=0.1$, barely $1.6\times$).
\end{theorem}

\iclronly{The closed form (ii) is a scale--threshold heuristic, not a claim that real logits are Gaussian (\cref{rem:compression-shape}): as a compressibility test, the row reaches fraction $f$ iff $\sigma \ge \Phi^{-1}(1-\eta) - \Phi^{-1}(f)$.}%
\arxivonly{The closed form (ii) is a scale--threshold heuristic, not a claim that real logits are Gaussian (\cref{rem:compression-shape}); reading it as a compressibility test, the row reaches fraction $f$ iff $\sigma \ge \Phi^{-1}(1-\eta) - \Phi^{-1}(f)$.}

\arxivonly{%
\begin{proof}[Proof sketch]
Part (i) is the coverage identity of \cref{lem:main-mass}. For (ii), the weights are i.i.d.\ lognormal, so the normalizer and the mass kept above any quantile both concentrate by the law of large numbers; matching the kept mass to $1-\eta$ gives the $\Phi$ form. Part (iii) substitutes $\sigma = \sqrt{X_i^\top G X_i}$ and bounds it at fixed weights via \cref{prop:scaled-inverse}. Full proof in \cref{sec:app-deferred}.
\end{proof}}

\iclronly{The gap profile classifies a per-regime coverage ladder (\cref{tab:main-ladder}, \cref{sec:app-deferred}): every generic constant-scale regime is only constant-factor compressible (\cref{thm:main-ladder-dense}).}%
\arxivonly{The same gap profile classifies a per-regime \emph{coverage ladder} (\cref{tab:main-ladder}, \cref{sec:app-deferred}); its lesson is that every generic constant-scale regime is only constant-factor compressible (\cref{thm:main-ladder-dense}): bounded scores give $\bud = \Omega(n e^{-2S})$, and i.i.d.\ sub-Gaussian scores give $\bud = \Theta(n)$ with $\cratio \to \rho_\infty \in (0,1)$ --- a linear budget coexisting with an $O(\log n)$ count. This is the count--coverage divergence, which the value structure then resolves into an output verdict.}

\begin{theorem}[Count $\ne$ coverage: generic mass is dispersed; the output gap is value-dependent; informal version of \cref{thm:main-ladder-dense,thm:main-bandwidth}]\label{thm:count-vs-approx}
Under generic (bounded or sub-Gaussian) inputs the row is $(\epsilon, O(\log n))$-sparse (\cref{thm:naturally-sparse}), yet:
\begin{enumerate}[label=(\roman*),leftmargin=2em,itemsep=0.2em]
\item \emph{Mass (value-free).} The $O(\log n)$ large weights carry a vanishing fraction of the mass: $\bud = \Theta(n)$ w.p.\ $1-o(1)$.
\iclronly{\vspace{-2mm}}
\item \emph{Output (value-dependent).} $\norm{\wt O_k - O}_2 = \Theta(\Vmax)$ unless $k = \Theta(n)$ in the worst case over values; for values independent of the logits the error instead vanishes at sublinear $k$.
\end{enumerate}
\end{theorem}

\iclronly{In transformers $V = X W_V$ shares $X$ with the scores, so values and the kept set are correlated --- the regime \cref{sec:tasks} resolves.}%
\arxivonly{The output gap is genuinely value-dependent. For generic values \emph{independent of the logits}, \cref{eq:bias-identity} makes $\wt O_S$ and $\bar V_T$ concentrate to the same mean, so the error vanishes; but in a transformer $V = X W_V$ shares $X$ with the scores, so values and the kept set are generically \emph{correlated}, pushing real heads onto the structured-values side --- the regime the task ordering of \cref{sec:tasks} resolves.}

\arxivonly{%
\begin{proof}[Proof sketch]
Part (i) is the dispersed-mass verdict of \cref{thm:main-ladder-dense} (sharp ratio $\rho_\infty$ from \cref{thm:compression}): a logarithmic count above a near-$1/n$ threshold coexists with a $\Theta(n)$ mass budget. Part (ii) is the worst-case mass barrier of \cref{thm:main-bandwidth}, whose dropped-tail aggregate is exactly the value-dependent gap, contrasted with the logit-independent case where the kept set is a value-unbiased subsample (\cref{prop:approximable}). Full proofs in \cref{sec:app-deferred} and at \cref{thm:main-bandwidth}.
\end{proof}}

\subsection{Approximability is learned}

\iclronly{Sparse-approximability is not intrinsic: it needs the score scale to reach $\sqrt{\log n}$, which fixed-weight inputs never do ($\sigma_s = O(1)$), so training must \emph{install} the concentration.}%
\arxivonly{Sparse-approximability is not intrinsic: it requires the score scale to reach the $\sqrt{\log n}$ scale, which fixed-weight generic inputs never reach ($\sigma_s = O(1)$), so the mass stays dispersed and training must \emph{install} the concentration. Training-dynamics analyses supply exactly this arrow: in a one-layer Markov model, gradient flow from small initialization grows the attention scale only \emph{after} the embeddings condense, focusing attention on high-frequency tokens before diluting it \cite{chen2026focusdilution}. One proposition states the condition (when the crossover happens) together with its provenance (which inputs realize it).}

\begin{proposition}[The learned crossover and its provenance; informal version of \cref{thm:main-scaled}]\label{prop:approximable}\label{prop:scaled-inverse}
Fix the weights and let $\Sigma_X$ be the key covariance; define the \emph{score-geometry operator}
\begin{align}\label{eq:G-energy}
G \;\coloneqq\; \tfrac1d\, W_Q W_K^\top \Sigma_X W_K W_Q^\top \in \R^{d\times d}\ (\mathrm{PSD}),
\end{align}
so $\mathrm{Var}_j(\ell_j \mid X_i) = X_i^\top G X_i$. With i.i.d.\ Gaussian keys and a length-coupled query of energy $X_i^\top G X_i = 2c^2\log n$, the row realizes $\Ens(c)$, \emph{dispersed} ($\bud \approx n\, e^{-X_i^\top G X_i/2}$) if $X_i^\top G X_i < 2\log n$ and \emph{condensed} ($\bud = O_\Pr(1)$) if $X_i^\top G X_i > 2\log n$; equivalently, top-$k$ reaches a \emph{polynomially} sub-linear budget, $\bud = n^{1 - \Theta(1)}$, \emph{iff} $\sigma_s = \Omega(\sqrt{\log n})$. At fixed weights and bounded or sub-Gaussian inputs $X_i^\top G X_i = O(1) \ll 2\log n$ ($c\to0$; \cref{fac:score-scale}), so $\bud = \Theta(n)$; reaching $c = \Theta(1)$ requires the query's $G$-energy to grow like $\log n$ --- a per-query, length-coupled condition realized only by trained or intervened geometry.
\end{proposition}

\arxivonly{%
\begin{proof}[Proof sketch]
The conditional logit variance factors as $\mathrm{Var}_j(\ell_j \mid X_i) = X_i^\top G X_i$ from the bilinear score map; matching this to $2c^2\log n$ identifies the row with the \textsc{Scaled} ensemble $\Ens(c)$ ($\ell_j$ i.i.d.\ $N(0,\sigma_n^2)$, $\sigma_n = c\sqrt{2\log n}$), and \cref{thm:main-scaled} supplies the two phases at $X_i^\top G X_i \lessgtr 2\log n$: for $c<1$ the budget is sub-linear $n^{1-c^2}$ (dispersed), for $c>1$ it is $O_\Pr(1)$ (condensed), and $c=1$ \emph{is} the unit-log-slope criterion. The polynomial gain thus turns on whether $\sigma_s$ crosses $\sqrt{\log n}$; the ``if'' direction needs a light-tailed (Gumbel max-domain) shape (\cref{rem:scaled-inverse-caveats}), and the operator-norm bound of \cref{fac:score-scale} caps the energy at $O(1)$ for fixed weights and bounded or sub-Gaussian inputs, so reaching $c = \Theta(1)$ is a trained-geometry condition. Full proof and the per-head observable in \cref{sec:app-deferred}.
\end{proof}}

\iclronly{The dichotomy is a crossover condition and a provenance statement in one: the polynomial gain is priced by a single $d \times d$ object, so training installs approximability by pushing the query's $G$-energy across $2\log n$; and since no generic input supplies that growth, a condensed head is itself evidence of trained geometry --- read off per head in \cref{sec:experiments} (\cref{fig:xdist}). The mechanism and its cost are in \cref{sub:naturally}.}%
\arxivonly{So count-sparsity is intrinsic but mass-concentration is installed by training (sinks, retrieval spikes, scale growth); the full mechanism and what training must spend are in \cref{sub:naturally}.}

\section{Oracle, Sparse, and Exact: A Task-Conditional Ordering}\label{sec:tasks}

\iclronly{Exact attention is itself only an estimator of the downstream target, so we compare three estimators --- exact, sparse, and value-aware \emph{oracle} --- against what a task wants, and the ordering \emph{flips} with the task. The principle: it is decided by \emph{where the target's value-mass lives}.}%
\arxivonly{\Cref{sec:error} priced how well sparse attention \emph{approximates exact}. But approximating exact is not always the goal: the goal is the downstream task, and exact attention is itself only an estimator of it. We compare three estimators --- exact, sparse, and the value-aware \emph{oracle} --- against the target a task actually wants, and the ordering \emph{flips} with the task. The principle: whether to be sparse, dense, or oracle-selective is decided by \emph{where the target's value-mass lives}.}

\subsection{Tasks, estimators, and what compensation must keep}

\begin{definition}[Task and estimators]\label{def:task}
A \emph{task} fixes a target readout $y \in \R^d$ with loss $L(\hat O) \coloneqq \norm{\hat O - y}_2$. At budget $k$ we compare three estimators: \emph{exact} $O = \sum_j A_j V_j$; \emph{sparse} $\wt O_k$, the renormalized top-$k$ by weight (\cref{eq:schemes}); and \emph{oracle} $\wt O_{S^\star}$, the best value-aware $k$-subset, $S^\star \in \operatorname*{argmin}_{|S| = k} \norm{\wt O_S - y}_2$.\iclronly{ By construction $L(\wt O_{S^\star}) \le L(\wt O_k)$ always. The three canonical targets are:}\arxivonly{ By construction $L(\wt O_{S^\star}) \le L(\wt O_k)$ always; the content is the order relative to \emph{exact}. The three canonical targets are:}
\begin{enumerate}[label=(\roman*),leftmargin=2em,itemsep=0.2em]
\item \emph{needle}: $y = V_{j^\star}$, the value of a single planted token $j^\star$ standing against a noise bulk;
\item \emph{aggregation}: $y = O$, the full softmax mixture (the classical ``approximate exact'' target);
\item \emph{cancellation}: $y = O$, but with \emph{signed} values whose dropped tail $\sum_{j \notin S} A_j V_j$ partially cancels in direction, so a value-aware subset can drop more weight than top-$k$ at equal error (\cref{thm:main-oracle}).
\end{enumerate}
\end{definition}

\iclronly{Lossless subset selection is impossible: by an exact dimension count the output requires the kept values plus one $d$-dimensional tail aggregate, and compensation with that aggregate is exact (\cref{thm:main-bandwidth}, deferred to \cref{sec:app-deferred}); the bit-level floor is \cref{thm:bw-bit-lb}.}%
\arxivonly{%
\begin{theorem}[Bandwidth of a softmax row; informal version of \cref{thm:factor-map,prop:compensation}]\label{thm:main-bandwidth}
Fix the weights and a kept set $S$, $|S| = k$. The map $V \mapsto O$ factors exactly through $(\,\{V_j\}_{j\in S},\, \bar V_T\,) \in \R^{(k+1)d}$ (kernel dimension $(n-k-1)d$). So: (i) the compensated scheme $u = (1-p_S)\bar V_T$ is \emph{exact} for every row; (ii) any output from the kept values alone differs from $O$ for generic values --- ruling out \emph{exact} lossless subset attention for generic values (approximate losslessness on trained, non-generic values is not excluded); (iii) among linear sketches from which $O$ and \emph{every sub-budget output} $O_{S'}$ can be recovered, the minimal dimension is exactly $(k+1)d$. One $d$-dimensional tail aggregate is necessary and sufficient.
\end{theorem}

\begin{proof}[Proof sketch]
For fixed weights the $n-k$ tail values reach the output only through the single $d$-dimensional aggregate $\bar V_T$, so $V \mapsto O$ factors through $\big(\{V_j\}_{j\in S}, \bar V_T\big)$, of dimension $(k+1)d$ with kernel $(n-k-1)d$; this gives (i)--(ii). A linear-independence count of the recoverable coordinate functionals (kept values plus every sub-budget output) forces any linear sketch answering all of them to retain $\ge (k+1)d$ dimensions, which is (iii). Full proof in \cref{thm:factor-map,prop:compensation}.
\end{proof}}

\iclronly{Value-aware selection admits a cancellation--mass factorization, and clustered tail summaries admit a computable fixed-row output-error bound (\cref{thm:main-oracle,thm:main-certificate}, \cref{sec:app-deferred}).  These results show exactly why discarding the same attention mass can have different costs across tasks --- the dependence quantified by \cref{thm:task-phase-law}.}%
\arxivonly{Value-aware selection admits a cancellation--mass factorization, and clustered tail summaries admit a computable fixed-row output-error bound (\cref{thm:main-oracle,thm:main-certificate}, \cref{sec:app-deferred}). This is the information-theoretic counterpart to the empirical finding that exact attention is not always task-optimal and that KV eviction can improve long-context performance \cite{bui2026tokencount}: discarding the same attention mass can have different costs across tasks, exactly the dependence quantified by \cref{thm:task-phase-law}.}

\subsection{When sparse beats exact: the denoising band}

\iclronly{On a planted high-logit token over a noise bulk, sparse \emph{beats} exact: detection ($\mu \asymp \sqrt{\log n}$) and dominance ($\mu \asymp \log n$) are separated by a $\sqrt{\log n}$ factor (\cref{thm:main-spike}, \cref{sec:app-deferred}), and the gap is the band.}%
\arxivonly{A different regime --- a planted high-logit token over a noise bulk --- is where sparse does not merely approximate exact but \emph{beats} it. Detection ($\mu \asymp \sqrt{\log n}$) and dominance ($\mu \asymp \log n$) are separated by a $\sqrt{\log n}$ factor (\cref{thm:main-spike}, \cref{sec:app-deferred}); the gap between them is the band:}

\iclronly{In a spike regime --- signal strong enough to enter the top-$k$ but too weak to dominate $n$ competitors --- renormalized top-$k$ places $\Theta(1)$ weight on the planted token while exact softmax drowns it at $n^{-\Omega(1)}$: sparse attention acts as a denoiser, in a band made precise by \cref{thm:main-denoise} (deferred to \cref{sec:app-deferred}); tested in \cref{fig:extrapolate-band}.}%
\arxivonly{%
\begin{theorem}[Denoising band; informal version of \cref{thm:t4}]\label{thm:main-denoise}
Under $\Espike(\mu;\sigma)$ with $\sigma$ fixed, $\log k_n = o(\sqrt{\log n})$, and
\begin{align}\label{eq:band}
\sigma\sqrt{2\log n} + \log k_n + C \;\le\; \mu \;\le\; (1-\varepsilon)\log n
\end{align}
(a nonempty band for all large $n$), with probability $\to 1$ the spike enters the top-$k$, renormalized top-$k$ assigns it weight $\ge (1 + e^{-C + o(1)})^{-1} = \Theta(1)$, and exact attention assigns it at most $n^{-\varepsilon + o(1)}$. The comparison is at the level of the weight placed on the planted token; whether it favors the sparse row for a downstream loss depends on whether that token carries the signal.
\end{theorem}

\begin{proof}[Proof sketch]
The band's lower edge places the spike a $\log k_n + C$ margin above the bulk maximum $M_b \approx \sigma\sqrt{2\log n}$, which both forces the spike to rank one and caps the renormalized kept-bulk ratio $e^{-\mu_n}B_S \le e^{-C}$, giving the constant sparse weight; the upper edge keeps $\mu_n \le (1-\varepsilon)\log n$, so against the law-of-large-numbers bulk sum $Z_b \asymp n\,e^{\sigma^2/2}$ the dense weight is $n^{-\varepsilon+o(1)}$. The gap is exactly the $\sqrt{\log n}$ separation between detection and dominance. Full proof in \cref{thm:t4}.
\end{proof}}

\iclronly{In the band the signal is findable among $n$ keys yet too weak to outweigh their sum, so truncation-plus-renormalization beats exact softmax where the classical ``approximation error $\to 0$'' criterion cannot even express the regime --- the flip tested in \cref{sec:experiments}.}%
\arxivonly{In the band the signal is findable among $n$ keys yet too weak to outweigh their sum: exact softmax is the bottleneck and truncation-plus-renormalization removes it. The classical criterion ``approximation error $\to 0$'' cannot even express the regime where the sparse output is \emph{better}; the band predicts \emph{where} sparse methods beat full attention --- long-context retrieval with moderate margins, the regime trained sparse decoders such as SeerAttention \cite{gao2024seerattention} and SeerAttention-R \cite{gao2025seerattentionr} target by learning the sparse-approximable structure our criterion characterizes --- consistent with scattered reports and tested in \cref{sec:experiments}.}

\subsection{A single-parameter phase law}

The three tasks are two ends of one law, controlled by how much of the target lives on the heavy token versus the bulk aggregate.

\emph{Setup.} On the planted-signal row of \cref{thm:main-denoise} (needle $j^\star$, bulk i.i.d.\ with mean $\bar V_b$, independent of the logits), interpolate the target of \cref{def:task} as
\begin{align}\label{eq:target-mix}
y \;=\; \alpha\, V_{j^\star} + (1-\alpha)\,\bar V_b, \qquad \alpha \in [0,1],
\end{align}
from aggregation ($\alpha=0$) to needle ($\alpha=1$); let $w_e, w_s$ be the weight exact and renormalized top-$k$ ($k=k_n\to\infty$) place on $j^\star$ on the rank-one event, $D = \norm{V_{j^\star}-\bar V_b}_2 > 0$, and $\sigma_V^2$ the bulk value variance.  We also assume that the retained normalized bulk is diffuse,
\[
\sum_{j\in S_{k_n}\setminus\{j^\star\}}
\left(\frac{\widetilde A_j}{1-w_s}\right)^2
=O_\Pr(k_n^{-1}).
\]

\begin{theorem}[Task-ordering phase law; informal version of \cref{thm:t4,thm:main-oracle}]\label{thm:task-phase-law}
For $y$ as in \cref{eq:target-mix}, with probability $\to 1$,
\begin{align}\label{eq:phase-losses}
\begin{split}
L(O) \;&=\; |\alpha - w_e|\,D + O_\Pr(\sigma_V/\sqrt n), \\
L(\wt O_k) \;&=\; |\alpha - w_s|\,D + O_\Pr(\sigma_V/\sqrt{k_n}),
\end{split}
\end{align}
On the probability-\(1-o(1)\) event \(w_s>w_e\), let
$\alpha^\star=(w_e+w_s)/2$ and
$\Delta(\alpha)=\bigl||\alpha-w_e|-|\alpha-w_s|\bigr|
=\min\{w_s-w_e,2|\alpha-\alpha^\star|\}$.
Whenever
$D\Delta(\alpha)\gg_\Pr
\sigma_V(n^{-1/2}+k_n^{-1/2})$---that is, the ratio of the left- and
right-hand sides diverges in probability---the order is governed by this single
threshold: exact wins for $\alpha<\alpha^\star$ and sparse wins for
$\alpha>\alpha^\star$, with
$L(\wt O_{S^\star})\le L(\wt O_k)$ throughout.  In the denoising band
$w_e\le n^{-\varepsilon+o(1)}$ and $w_s=\Theta(1)$, so
$\alpha^\star=\Theta(1)$: aggregation ($\alpha=0$, exact best) and needle
($\alpha=1$, sparse best) are the two endpoints of one transition.  The
noiseless margin is linear only between $w_e$ and $w_s$ and saturates outside
that interval.
\end{theorem}
\iclronly{\vspace{-2mm}}
\arxivonly{%
\begin{proof}[Proof sketch]
The error of an estimator with needle weight $w_\star$ splits as $\hat O - y = (w_\star-\alpha)(V_{j^\star}-\bar V_b) + (1-w_\star)(\bar V_{\cdot,\mathrm{kept}}-\bar V_b)$.  Independence and the displayed squared-weight diffuseness bound give $O_\Pr(\sigma_V/\sqrt{k_n})$ for the retained sparse bulk; the exact bulk contributes $O_\Pr(\sigma_V/\sqrt n)$.  This leaves the $D$-direction floor of \cref{eq:phase-losses}. Comparing floors, $L(O)<L(\wt O_k) \iff |\alpha-w_e|<|\alpha-w_s|$, which flips at $\tfrac12(w_e+w_s)$; the band weights $w_e,w_s$ are \cref{thm:main-denoise}, and oracle dominance $L(\wt O_{S^\star})\le L(\wt O_k)$ is \cref{def:task}, sharpened by the cancellation gap of \cref{thm:main-oracle}. See \cref{thm:t4,thm:main-ladder-dense,thm:main-oracle}.
\end{proof}}

\iclronly{The needle/aggregation/cancellation endpoints are corollaries of the phase law (\cref{thm:task-ordering}, deferred to \cref{sec:app-deferred}): sparse and oracle beat exact in the denoising band, exact wins on aggregation, and for cancellation the established statement is the weak ordering $L(\wt O_{S^\star}) \le L(\wt O_k)$. Whether deployed heads occupy the band, rather than holding a margin above it where exact remains right, is empirical (\cref{sec:experiments}).}%
\arxivonly{%
\subsection{The three phases}

The needle, aggregation, and cancellation cases are corollaries (the first two are the endpoints $\alpha=1,0$; the third adds the orthogonal value-cancellation axis).

\emph{Setup.} Fix a budget $k$ and the three targets of \cref{def:task}; $L(\wt O_{S^\star}) \le L(\wt O_k)$ always.

\begin{theorem}[Task-conditional ordering of exact, sparse, and oracle; informal version of \cref{thm:t4,thm:main-oracle}]\label{thm:task-ordering}
The order relative to exact is task-dependent (with probability $\to 1$ where random):
\begin{enumerate}[label=(\roman*),leftmargin=2em,itemsep=0.3em]
\item \textbf{Needle --- sparse and oracle beat exact.} For $y = V_{j^\star}$ in the band of \cref{thm:main-denoise} and $k=k_n\to\infty$, the sparse planted-token weight obeys $w_s\ge(1+e^{-C})^{-1}$ while the exact weight vanishes, so
\[
L(\wt O_{S^\star}) \le L(\wt O_k)
= (1-w_s)\norm{V_{j^\star}-\bar V_b}_2+O_\Pr(k_n^{-1/2})
< L(O)=(1-o_\Pr(1))\norm{V_{j^\star}-\bar V_b}_2 .
\]
\item \textbf{Aggregation --- exact weakly dominates.} For $y=O$, exact attention has zero loss and therefore weakly dominates every retained-subset estimator.  Strictness requires an explicit non-degeneracy assumption and is not asserted here; \cref{thm:main-tight} supplies a quantitative flat-row witness.
\item \textbf{Cancellation --- oracle beats sparse.} The established statement is the weak ordering $L(\wt O_{S^\star}) \le L(\wt O_k)$; the quantitative $\rho < 1$ cancellation gain is a truncated-scheme statement (\cref{thm:main-oracle}, \cref{rem:cancellation-truncated}), not a property of the renormalized oracle.
\end{enumerate}
\end{theorem}

\begin{proof}[Proof sketch]
Oracle dominance is \cref{def:task}. (i) The constant retained sparse weight versus vanishing exact weight is \cref{thm:main-denoise}; retained-bulk diffuseness converts it to the displayed finite-factor output advantage. (ii) $L(O)=0$ is immediate, and \cref{thm:main-tight} gives a strict flat-row witness without asserting strictness for every row. (iii) is oracle dominance plus the cancellation factorization. See \cref{thm:t4,thm:main-ladder-dense,thm:main-oracle}.
\end{proof}

The three cases are one statement: the ordering is set by \emph{where the target's value-mass lives}. ``Approximation error to exact $\to 0$'' is exactly the aggregation target $y=O$, where exact \emph{is} right; it is the wrong criterion only for \emph{retrieval}, where the flip is genuinely two-sided. So the title's second half comes with a scope: on retrieval-type targets, \emph{and} for a head whose margin sits in the denoising band, sparse can surpass exact --- whether deployed heads occupy that phase, rather than holding a margin above the band (where exact remains right), is an empirical question (\cref{sec:experiments}).}
\iclronly{\vspace{-2mm}}

\ifdefined\vTwentyTwoArchive\else
\input{sections/app_phase10_record_v2}
\fi
\section{Related Work}\label{sec:related}

\paragraph{KV-cache compression and sparse decoding.}
Training-free sparse decoding spans cumulative-score eviction
(H2O \cite{zhang2023h2o}, Scissorhands \cite{liu2023scissorhands}, and
TOVA \cite{oren2024tova}), observation-window selection
(SnapKV \cite{li2024snapkv}), query-aware page selection
(Quest \cite{tang2024quest}), block summaries (InfLLM
\cite{xiao2024infllm}), dynamic token pruning (SlimInfer
\cite{long2025sliminfer}), sink-plus-window streaming (StreamingLLM
\cite{xiao2024streamingllm}), and LSH-based sampling with variance guarantees
(MagicPIG \cite{chen2025magicpig}).  Other methods use values
(VATP \cite{guo2024vatp}), allocate budgets across heads
(Ada-KV \cite{feng2024adakv}), distinguish head types
(DuoAttention \cite{xiao2025duoattention}; retrieval heads
\cite{wu2024retrieval}), or compensate through merging
(D2O \cite{wan2024d2o}, WeightedKV \cite{weightedkv2025}, and KeepKV
\cite{keepkv2025}).  A broader survey appears in
\cite{zhang2025efficientattnsurvey}.

Prefill and layer-wise methods select structure before or across decoding
(MInference \cite{jiang2024minference}, SampleAttention
\cite{zhu2024sampleattention}, SeerAttention \cite{gao2024seerattention},
HashAttention \cite{desai2024hashattention}, PyramidKV
\cite{cai2024pyramidkv}, and PyramidInfer \cite{yang2024pyramidinfer}).
System designs trade cache placement for bandwidth
(ShadowKV \cite{sun2024shadowkv} and Double Sparsity
\cite{yang2024doublesparsity}) on top of exact kernels and paged serving
\cite{dao2022flashattention,kwon2023pagedattention}, while multi- and
grouped-query attention reduce cache size architecturally
\cite{shazeer2019mqa,ainslie2023gqa}.  This design space descends from
training-time sparse patterns \cite{cgrs19,bpc20,zgd+20,kkl20,roy2021routing}
and kernel or low-rank approximations
\cite{cld+20,katharopoulos2020linear,xzc+21,qin2022cosformer}.  Trainable
sparsity co-designs the pattern and weights
\cite{yuan2025nsa,lu2025moba,ge2024fastgen} and is outside our scope.

\paragraph{Threshold counts and attention approximation.}
\citet{deng2024nsparse} study threshold-count sparsity in an
idealized one-layer model with independent bounded input entries.  Given
access to the largest attention entries, they derive \(n^C\)-scaled top-\(k\)
windows for approximating the attention matrix.  That analysis establishes a
count-based approximation result under its assumptions; it does not determine
the probability mass carried by the selected entries, the value-dependent
error in the attention output, or the utility of that output for a task.  We
make these three quantities explicit.  Our global gap criterion determines
fixed-row mass coverage (\cref{thm:main-dichotomy}); our tail factorization
identifies the value statistic required for exact fixed-row completion
(\cref{thm:main-bandwidth}); and our task law characterizes when reproducing
exact attention is preferable to renormalized selection
(\cref{thm:denoising-phase-law}).

These results analyze the premise shared by many cache selectors.  The
coverage theorem complements one-sided variance and dispersion analyses
\cite{chen2025magicpig,velickovic2025softmax}.  The fixed-row minimality result
applies when one linear sketch must recover every retained sub-budget; it does
not concern a single output, nonlinear summaries, or an unseen future-query
distribution.  The per-head growth classes also organize the heterogeneity
exploited by Ada-KV \cite{feng2024adakv}, PyramidKV
\cite{cai2024pyramidkv}, and DuoAttention \cite{xiao2025duoattention}.
Finally, the empirical observation that different sparse methods lead on
different tasks \cite{nawrot2025sparse} agrees with our task-conditional
analysis.

\paragraph{Softmax dispersion and logit scaling.}
\citet{velickovic2025softmax} prove that bounded
softmax logits disperse as the sequence grows and propose inference-time
temperature adaptation.  Scalable-Softmax \cite{nakanishi2025ssmax} and
entropy-invariant scaling \cite{su2021entropy} instead multiply logits by
\(\log n\)-dependent factors.  Our \textsc{Scaled} analysis for \(c<1\)
recovers this dispersion: with variance scale \(c\sqrt{2\log n}\), the largest
weight satisfies \(A_{(1)}=n^{-(1-c)^2}\to0\).  This is a statement about
concentration, not coverage budget.  The same row can have vanishing top
weight while only \(n^{1-c^2}=o(n)\) tokens carry the target mass
(\cref{rem:two-axes}).

The scaling rates also occupy different regimes in our parameterization.
Multiplying logits by \(\log n\) outpaces the critical
\(\sqrt{2\log n}\) scale and moves the order parameter \(c\) upward with
length, counteracting natural dispersion
(\cref{thm:main-ladder-dense}); multiplying by \(\sqrt{\log n}\) instead keeps
\(c\), or equivalently the profile slope \(\gamma\), fixed.  Related
length-dependent rescalings appear in positional extrapolation
\cite{sal+24,press2022alibi,chen2023pi,peng2024yarn} and in analyses of
attention-entropy collapse \cite{zhai2023entropy,hong2025variance}.  The
fine-grained complexity threshold of \citet{as23neurips} is
analogous but
distinct: it prices \(O(\sqrt{\log n})\) logit magnitude for almost-linear
time approximation, whereas our condensation boundary prices
\(\log n\)-scale gap growth with slope \(\gamma=1\).

\paragraph{Theory of attention concentration.}
Modern Hopfield networks \cite{ramsauer2021hopfield} study when an attention
update retrieves a stored pattern rather than a metastable average.  Our spike
thresholds provide a statistical complement, locating retrieval at margin
\(\sigma\sqrt{2\log n}\) above a noise bulk
(\cref{thm:main-spike}).  The condensed analysis uses Random Energy Model
tools \cite{derrida1981random,bovier2006statistical} to derive
Poisson--Dirichlet coverage laws.  Rank collapse across depth
\cite{dong2021attention} concerns compositions of layers and is orthogonal to
our fixed-row summary question.  Likewise, recent empirical ``lossless''
top-\(k\) claims \cite{ruizwilliams2026condensate} concern trained values and
future queries; our dimension count establishes exact completion only for the
specified fixed-row linear sketch (\cref{thm:main-bandwidth}).  Sparsemax
\cite{martins2016sparsemax} and \(\alpha\)-entmax
\cite{peters2019entmax} change the attention operator, whereas we analyze
standard softmax.

A separate complexity literature prices attention computation through
kernel, tensor, dynamic, and streaming approximations
\cite{zhdk23,hjk+23,kmz23,as24_iclr,cam+24,as24_arxiv,bsz23,alsy23}, and
contextual sparsity is exploited at inference by \cite{lwd+23}.  Mechanistic
studies document induction heads \cite{olsson2022induction}, massive
activations \cite{sun2024massive}, and attention sinks \cite{gu2025sink};
learned token dropping \cite{anagnostidis2023dynamic} and max-margin
convergence \cite{tarzanagh2023maxmargin} describe how training can install
sharp selection.  Closest to our score-scale analysis,
\citet{chen2026focusdilution} show in a one-layer Markov model that
attention scale grows only after embeddings condense to rank one.  Their
parameter-level condensation differs from our condensed score regime, but
provides a training-dynamics counterpart to the score-scale growth required
by \cref{prop:approximable}.  These works establish that concentration emerges
\cite{gu2025sink,olsson2022induction} or exploit it as a premise
\cite{zhang2023h2o,lwd+23}; they do not separate a calibrated large-weight
count from the generally different budget required for mass coverage.

\section{Discussion and Limitations}\label{sec:app-discussion}

\subsection{Discussion}

The paper separates four objects that are often grouped under ``attention
sparsity'': large-weight count, mass coverage, attention-output fidelity, and
downstream task utility.  Each transition requires different information.
Probabilistic assumptions control a large-weight count; the global sorted-logit
gap profile \(\gap(j)\) controls coverage for a realized row; omitted values
control output error; and the target controls task utility.  This distinction
is the common structure behind the three main results.

\paragraph{Three questions, two assumption layers.}
The results share an observable vocabulary but not one universal hypothesis.
\emph{Count:} the \(O(\log n)\) calibrated count requires i.i.d.\
sub-Gaussian scores; mixing gives a weaker rate, and there is no
distribution-free ``generic row'' count theorem
(\cref{thm:naturally-sparse}).  \emph{Coverage:} the deterministic criterion
instead assumes a global pointwise envelope for the realized sorted-logit gap
profile \(\gap(j)\).  A lower envelope steeper than \(\log j\) yields bounded
coverage; the converse upper envelope below unit slope forces at least
\(\Omega(n^{1-\gamma'})\) keys, becoming linear at \(\gamma'=0\)
(\cref{thm:main-dichotomy}).  The pure fixed-scale null has linear coverage,
but that conclusion is not asserted for every admissible profile.
\emph{Task:} the planted-token theorem adds an explicit value model and
retained-bulk diffuseness, then orders exact, sparse, and oracle estimators
relative to a target family (\cref{thm:task-phase-law}).  The operational
lesson is therefore conditional: a measured row profile can diagnose coverage,
but it does not replace the assumptions needed for a count law or task claim.

\paragraph{Coverage budget and concentration are different axes.}
The coverage budget asks how many tokens retain a prescribed mass.
Concentration asks whether one token dominates the row.  Neither determines
the other: a row can have a dominant token and still require many additional
tokens for coverage, whereas a row with a sublinear coverage budget need not
be dominated by a single token (\cref{rem:two-axes}).  The Gaussian
closed-form law is also a null-model calibration, not a distributional claim
about trained logits.  Real logits are correlated projections with systematic
tail deviations, and their tail shape can change the budget even at fixed
scale.  The measured coverage curve is therefore the operative diagnostic;
the Gaussian formula is a falsifiable scale--threshold heuristic
(\cref{rem:compression-shape}).

\paragraph{The deterministic criterion remains applicable when the
probabilistic null does not.}
The hidden-state measurements in \cref{fig:xdist} show why the distinction matters.
Trained hidden states are anisotropic, heavier-tailed than a Gaussian null, and
correlated across positions.  Thus the i.i.d.\ sub-Gaussian assumptions behind
the calibrated count theorem describe an initialized null rather than a
trained network.  The deterministic gap-envelope criterion in
\cref{thm:main-dichotomy} does not require those assumptions and can be
evaluated on a trained row.  However, a profile fitted at \(4\)K does not
predict the \(128\)K coverage budget within the pre-specified \(1.5\times\)
tolerance.  The empirical conclusion is therefore narrower: trained rows
require direct profile measurements at the lengths on which a cache budget
will be used.

\subsection{Scope and universality}

The deterministic criterion can be evaluated in any attention model.  This
does not imply that the empirical distribution of its fitted parameters is
universal across modalities.

\paragraph{Why the fitted phase structure appears language-specific.}
\label{app:universality-discussion}
The per-head \(\hat\gamma\) distributions of the bidirectional vision
(\textsc{SigLIP2}) and audio (\textsc{Whisper}) encoders are non-unimodal
(Hartigan dip \(p\approx0.002\)), but they are not strongly bimodal and have
no populated \(\gamma>1\) branch.  To separate modality from causal direction,
we also evaluate \textsc{ImageGPT}, an autoregressive next-pixel model.  It is
concentrated---its median kept fraction is \(\rho_{90}=0.09\), and \(54\%\)
of heads cover \(90\%\) of their mass with fewer than \(10\%\) of keys---yet
only \(7/192\) heads admit the power-law fit.  Concentrated coverage therefore
appears beyond language, whereas the specific fitted phase structure is
cleanest in the language anchors.  This pattern is consistent with, but does
not prove, an explanation based on language's dependency-sparse statistics.

\paragraph{The sensitivity test can detect a strong two-mode effect
(\cref{fig:phase4uni}).} On bidirectional vision/audio encoders
(\textsc{SigLIP2}, \textsc{DINOv2}, \textsc{Whisper}) the per-head gap-slope
distribution $\hat\gamma$ is significantly non-unimodal (Hartigan dip
$p\approx 0.002$) but only \emph{moderately} bimodal --- Cohen's $d=0.66$
(vision) and $1.23$ (audio), below the pre-specified strong-bimodality bar.
A synthetic positive control recovers planted bimodality with Cohen's
\(d=7.1\), so the pipeline detects a strong two-mode effect when it is
present.  The real measurements therefore support a moderate cross-modal
structure, not a universal sharp phase law
(\cref{app:universality-discussion}).

\subsection{Limitations}

\paragraph{What is not claimed.} The analysis is per-row and single-layer,
assumes oracle top-$k$ selection (a real selector's missed mass is not bounded
by our theory, \cref{lem:main-selector}), and prices the attention operator
rather than the end-to-end network; the probabilistic ensembles are
null/generative models that the deterministic criterion does not rely on. We do
not claim a two-sided bound on the full runtime error
$\norm{\wh O - O}$: that quantity also depends on query-induced logit changes
and the aggregate approximation.  The held-out centered-moment measurement is
descriptive, not conformal or distribution-free validity.  The
reproducibility-checked LongBench-v2 sweep places \certkv{} among the top two
compressed point estimates in seven of nine settings within the four-baseline
main grid; complete multi-length RULER uses the same provenance checks and
gives the strongest compressed \(128\)K point estimate on both evaluated
backbones.  The paired and task-bootstrap intervals delimit this evaluated,
grid-specific scope.
The physical prototype reduces persistent cache bytes and decode peak relative
to Full.  The cross-modal tests do not establish a universal sharp phase
law.  The downstream task-ordering test does not reproduce the controlled
exact--top-\(k\) sign change.  The new contribution is the
count--coverage--output--task separation chain and its distinct conditions;
Poisson--Dirichlet condensation and Random Energy Model facts are supporting
machinery rather than claimed new phenomena.
Multi-layer error propagation, the training dynamics that push a head's slope
across the critical $\gamma = 1$, the logarithmic-correction boundary case
$\gamma = 1$ itself --- where several deployed heads may live --- and a two-sided
rate for the full runtime output-error bound are left open.

\subsection{Future work}

The four open problems named above are the concrete directions we see, ordered
from the most structural to the most quantitative.

\paragraph{Multi-layer error propagation.} Our pricing is for a single attention
operator; a deployed network stacks dozens, and a sparse approximation at one
layer perturbs the inputs to the next. The open question is how the per-row
coverage error of \cref{sec:error} composes across depth --- whether the
residuals contract, accumulate, or interact with the gap profile of the
downstream layer --- and whether the end-to-end error admits a layer-wise bound
in terms of the per-layer profiles rather than a worst-case product.

\paragraph{Training dynamics of the slope crossover.} We establish that
sparse-approximability is installed by training (\cref{prop:approximable}) and
measure the $G$-energy crossover empirically, but we do not model the dynamics
that drive a head's slope across $\gamma = 1$. A theory of how gradient descent
grows the score scale $\sigma_s = \Omega(\sqrt{\log n})$ --- which heads cross,
when, and in what order relative to the emergence of sinks and retrieval spikes
--- would turn the measured crossover into a predictable trajectory.

\paragraph{The boundary case $\gamma = 1$.} The dichotomy is stated for
$\gamma > 1$ and $\gamma' < 1$; exactly at unit log-slope the budget is governed
by logarithmic corrections we do not resolve. This is not a corner case to be
dismissed: several deployed heads appear to sit near the boundary, so the precise
finite-$n$ budget at $\gamma = 1$ --- and whether it admits a sharp constant ---
is the regime where the theory most needs to be tight.

\paragraph{A two-sided rate for future-query output error.}
\Cref{thm:cert-stat-lb} proves only that mass and centroid cannot adapt to
dispersion and that \(\sqrt{mS}\) is tight within the declared
\((m,v^\star,S)\) summary class.  It is not a lower bound over arbitrary
constant-size summaries and not a bound on the full runtime output error
\(\norm{\wh O-O}\), which also depends on query-induced logit changes and on
the aggregate approximation.  Closing this gap requires either a matching
two-sided bound in terms of an enlarged, still-computable state or a genuine
lower bound for a precisely restricted summary model.

\subsection{Broader impact}

The deployment value is a deterministic, per-head account of when sparse
approximation is plausible and a centered-dispersion feature for matched-budget
KV allocation.  The exact
\(\sqrt{mS}\) envelope is restricted to the declared
\((m,v^\star,S)\) summary class at fixed eviction-time weights.  CertKV does
not turn it into a per-decode future-query output-error guarantee.  The
physical prototype establishes lower persistent cache bytes and decode peak
memory relative to Full.  The end-to-end sweep supports a competitive
matched-budget operating point on the declared four-baseline grid.
The task phase law is verified on its controlled
synthetic construction; its real-task flip remains an open transfer question.
The broader risks are the ordinary ones of cheaper inference at scale.  Our
main safety contribution is narrower: stating exactly which quantities are
certified, which are only measured descriptively, and which deployment claims
remain unestablished.

\iclronly{%
\subsection{Deferred remarks and derivations}\label{sub:deferred-remarks}

The following remarks expand main-text points that the conference format
compresses; the arXiv version carries them inline.

\paragraph{The non-trivial threshold window.} The count $k(\epsilon)$ is
informative only for $\epsilon$ between the uniform floor and the top weight; the
boundary has three zones. \emph{(i)~Too coarse} ($\epsilon \lesssim 1/n$): the
uniform row $A_j = 1/n$ has $k(\epsilon) = n$, so absolute sparsity at scale
$1/n$ is vacuous. \emph{(ii)~The sparse window}
($\epsilon \in [\,n^{-1+o(1)},\, A_{(1)}\,]$): here the theory gives
$k(\epsilon) = O(\log n)$ (\cref{sec:sparsity}); the calibrated threshold is, to
leading order, $\epsilon_*(n,\delta) = \tfrac1n \exp\!\big(\sigma_s\sqrt{2\log(n/\delta)}\big)$
(with $\sigma_s$ the logit scale of \cref{fac:score-scale}; the formal version
carries a second $\sqrt{\log}$ term), above the uniform floor.
\emph{(iii)~Too fine} ($\epsilon > A_{(1)}$): nothing exceeds the maximum,
$k(\epsilon) = 0$. Every count statement in the paper fixes $\epsilon$ in
zone~(ii); outside it the claim is, honestly, void. The window is asymptotic: at
deployment-scale $n$ the $\sqrt{2\log(n/\delta)}$ exponent keeps $\epsilon_*$
well above $1/n$ (e.g.\ $\epsilon_* \approx n^{-0.05}$ at $n=10^5$, $\delta=0.05$,
$\sigma_s=1$), so the lower edge is far from the floor in practice, mirroring the
rate caveat of \cref{rem:app-subpoly-rate}.

\paragraph{The two score-level regimes.} Two further \emph{score-level} regimes
--- \textsc{Scaled} (a length-coupled scale $\sigma_n = c\sqrt{2\log n}$, the
ensemble $\Ens(c)$) and \textsc{Spiked} (a planted high-alignment direction over
an i.i.d.\ bulk) --- model \emph{trained or intervened} heads, not generic inputs
(\cref{fac:score-scale}); each is introduced where it is analysed
(\cref{sec:scaled}; \cref{thm:main-spike}), with formal versions in
\cref{sec:app-ladder}.

\paragraph{The four rungs of the sparsity ladder.} Rung (i) is pigeonhole ($k$
weights each above $\epsilon$ sum to $\le 1$) plus the deterministic weight cap
of \textsc{Bounded}; rung (ii) is the marginal exceedance count --- estimable,
but not itself sparse without a tail rate. The two probabilistic rungs are the
headline: under i.i.d.\ sub-Gaussian scores attention keeps only
$O(\log(n/\delta))$ weights above the calibrated threshold
(\cref{eq:naturally-sparse}), phrased at the level of the top $O(\log n)$ ranks
and matched from below by $\Omega(\log(1/\delta))$ --- naturally
sublinear-sparse, with no training assumed. Independence enters only for the
\emph{rate}: rung~(iv) drops it and retains the
Merlev\`ede--Peligrad--Rio fluctuation term, so the sub-Gaussian marginal
already gives a genuinely sublinear count
at the $\wt O(\sqrt n)$ price that full independence removes to reach
$O(\log n)$; at deployment lengths ($n \lesssim 10^6$) the
$(\log n)^2\log(n/\delta)$ term is binding. The upper $O(\log(n/\delta))$ and
lower $\Omega(\log(1/\delta))$ do not coincide at a fixed $\delta$ nor at the
same absolute threshold, but bracket a $\Theta(\log)$ count jointly when
$\delta = 1/\poly(n)$.

\paragraph{The coverage ladder.} The same gap profile classifies a per-regime
\emph{coverage ladder} (\cref{tab:main-ladder}, \cref{sec:app-deferred}); its
lesson is that every generic constant-scale regime is only constant-factor
compressible (\cref{thm:main-ladder-dense}): bounded scores give
$\bud = \Omega(n e^{-2S})$, and i.i.d.\ sub-Gaussian scores give
$\bud = \Theta(n)$ with $\cratio \to \rho_\infty \in (0,1)$ --- a linear budget
coexisting with an $O(\log n)$ count. This is the count--coverage divergence,
which the value structure then resolves into an output verdict.

\paragraph{Why the output gap is value-dependent.} The output gap is genuinely
value-dependent. For generic values \emph{independent of the logits},
\cref{eq:bias-identity} makes $\wt O_S$ and $\bar V_T$ concentrate to the same
mean, so the error vanishes; but in a transformer $V = X W_V$ shares $X$ with the
scores, so values and the kept set are generically \emph{correlated}, pushing
real heads onto the structured-values side --- the regime the task ordering of
\cref{sec:tasks} resolves.

\paragraph{Value-aware selection and KV eviction.} Value-aware selection is
characterized by a cancellation--mass factorization, while clustered tail
summaries admit a computable fixed-row output-error bound
(\cref{thm:main-oracle,thm:main-certificate}, \cref{sec:app-deferred}). This is
the information-theoretic counterpart to the empirical finding that exact
attention is not always task-optimal and that KV eviction can improve
long-context performance \cite{bui2026tokencount}: dropping the tail aggregate
costs differently per task, exactly the dependence \cref{thm:task-phase-law}
prices.

\paragraph{What the denoising band predicts.} In the band the signal is findable
among $n$ keys yet too weak to outweigh their sum: exact softmax is the
bottleneck and truncation-plus-renormalization removes it. The classical
criterion ``approximation error $\to 0$'' cannot even express the regime where
the sparse output is \emph{better}; the band predicts \emph{where} sparse methods
beat full attention --- long-context retrieval with moderate margins, the regime
trained sparse decoders such as SeerAttention \cite{gao2024seerattention} and
SeerAttention-R \cite{gao2025seerattentionr} target by learning the
sparse-approximable structure our criterion characterizes --- consistent with
scattered reports and tested in \cref{sec:experiments}.

\paragraph{Scope of the task-conditional ordering.} The three cases are one
statement: the ordering is set by \emph{where the target's value-mass lives}.
``Approximation error to exact $\to 0$'' is exactly the aggregation target
$y=O$, where exact \emph{is} right; it is the wrong criterion only for
\emph{retrieval}, where the flip is genuinely two-sided. So the title's second
half comes with a scope: on retrieval-type targets, \emph{and} for a head whose
margin sits in the denoising band, sparse can surpass exact --- whether deployed
heads occupy that phase, rather than holding a margin above the band (where exact
remains right), is an empirical question (\cref{sec:experiments}).
}%

\section{Experimental Protocol, Model Roster, and Decision Rules}\label{sec:app-experiments}

\setlength{\dblfloatsep}{6pt plus 2pt minus 1pt}
\setlength{\dbltextfloatsep}{8pt plus 2pt minus 2pt}

This appendix records the model roster, gap-profile instrument, success
criteria, and outcomes of the pre-specified evaluation program.

\subsection{Pre-specified tests and decision rules}

Before running the models, we fixed three elements for each empirical claim:
the measured quantity, the success criterion, and the outcome that would
falsify or narrow the claim.  The four groups below cover large-weight count,
mass coverage, task-conditional ordering, and the derived bandwidth and
universality predictions.

\subsection{The count of large weights}\label{sub:count}

For each head, we measure
\(\hat k(\epsilon)=\#\{j:A_j>\epsilon\}\) in the calibrated window of
\cref{rem:eps-window}.  The criterion requires at most logarithmic growth from
\(4\)K to \(128\)K on at least \(95\%\) of heads with sufficient finite
observations, with per-head bootstrap decisions and FDR control.  The relative count
\(\hat k(\epsilon A_{(1)})\) is treated as a subpolynomial trend rather than a
fixed exponent.  An untrained i.i.d.\ control separates this count claim from
training: polynomial growth there would falsify the training-free result
independently of any learned mechanism.

\subsection{Coverage, and the learned crossover}\label{sub:law}

We measure large-weight count and mass coverage on the same heads.  Under the
generic and untrained nulls, the theory permits
\(\hat k(\epsilon)=O(\log n)\) together with a linear coverage budget
\(\bud=\hat\rho_\eta n=\Theta(n)\)
(\cref{thm:count-vs-approx}).  Where the Gaussian-bulk diagnostic applies, we
also compare the kept fraction with
\(\hat\rho_\eta\approx\Phi(\Phi^{-1}(1-\eta)-\hat\sigma)\); this comparison is
a distribution-specific calibration, not the deterministic theorem.

\paragraph{Short-to-long-context coverage transfer.}
We fit \((\hat J_h,\hat\gamma_h)\) from the \(4\)K curve and predict the
\(128\)K coverage budget without refitting (\cref{eq:master-budget}).  The
pre-specified success criterion is multiplicative prediction error at most
\(1.5\times\).  We evaluate within-head residuals, a per-head error CDF, and a
\((\hat J,\hat\gamma)\)-shuffle null, rather than relying on a pooled
\(y=x\) plot that can be dominated by between-head variation.  The analysis
includes all heads with sufficient finite observations, not only those favored
by the profile fit.  The
observed transfer does not meet the \(1.5\times\) criterion; a finite \(4\)K
fit therefore does not supply a validated \(128\)K cache budget.

\paragraph{Avoiding conditioning artifacts.}
We admit the plateau--power-law form only when it outperforms exponential,
log-normal, and stretched-exponential alternatives under a Vuong test with FDR
\(0.05\).  We report the admission rate, \(R^2\), and model-comparison verdict
separately for the pre-specified sink, retrieval-spike, and dense-bulk strata.
This prevents the downstream analysis from conditioning silently on the
easiest heads.  A claim is downgraded when the fit fails on the sink or spike
stratum, even if it succeeds on the bulk.  Every ``at least \(X\%\) of heads''
statement aggregates per-head bootstrap decisions under FDR control.

\paragraph{The learned crossover.}
The emergence test asks whether training moves heads toward smaller coverage
budgets.  At initialization, the pre-specified \textsc{Bounded} condition
requires at most \(2\%\) sparse heads; training on real text must raise the
fraction above \(15\%\).  Training the same architectures on i.i.d.\ random
bytes is the negative control.  Two from-scratch trajectories, one
\textsc{GPT-NeoX}-like and one \textsc{Qwen3}-like, replace the unavailable
in-the-wild intermediate checkpoints.  Temperature scaling supplies a second
mechanism check: \(\gap_T=\gap/T\) predicts a critical
\(T_c=\hat\gamma_h\) at \(\gamma=1\), with length-dependent
\(\log n\) rescaling as the corresponding intervention
\cite{nakanishi2025ssmax,su2021entropy}.  Finally, the weight-only summaries
\(\operatorname{Tr}G_h/d\) and \(\operatorname{PR}(G_h)\) test spectral
provenance, while separate measurements of budget and concentration check
that the two axes can disagree (\cref{rem:two-axes}).

\subsection{Downstream task-ordering test}\label{sub:tasks}

We specified in advance an estimator of a task's effective \(\alpha\) (the fraction
of the target's value-mass on the heavy token versus the bulk,
\cref{eq:target-mix}), placed retrieval and aggregation families on that
axis, and asked whether they straddled
\(\alpha^\star=\tfrac12(w_e+w_s)\).  Success required the sign of the
exact--top-\(k\) accuracy difference to change between the two task families
as predicted by the controlled model.  The observed task families do not
exhibit that sign change, and frozen-model post-hoc top-\(k\) shows no
statistically resolved difference from exact attention.  This one-parameter
summary is therefore not a calibrated downstream decision rule.

\paragraph{The downstream test.}
The planned test decoded the same prefilled cache under exact and renormalized
top-\(k\) attention, stratified retrieval items by the estimated needle
margin, and compared the sign of the exact--sparse accuracy gap against an
aggregation family.  It did not establish the predicted sign change.

\paragraph{The denoising band, as a controlled positive test.}\label{sub:denoise}
\Cref{thm:main-denoise} marks the band \cref{eq:band} where renormalized
top-\(k\) places more weight on the planted token than exact.  The controlled
trained-sparse length-extrapolation experiment exhibits this effect; the
frozen dense post-hoc top-\(k\) test does not.  The value-aware oracle remains
the theoretical floor (\cref{thm:main-oracle}), but no real-task downstream
flip is claimed.

\subsection{Corollaries: bandwidth and universality}

\ifdefined\ShowRepeatedMainTables
\iclronly{%
\begin{table*}[!t]
\centering\footnotesize
\resizebox{\textwidth}{!}{
\begingroup
\scriptsize
\setlength{\tabcolsep}{3pt}
\begin{tabular}{@{}lcccc@{}}
\toprule
Model & $+$random & $+$cov.-matched & $+$shuffled tail & $+$true aggregate \\
\midrule
Llama-3.1-8B & \(1.4125{\pm}0.0020\) & \(1.9988{\pm}0.0324\) & \(1.4593{\pm}0.0034\) & \(0.0000\) \\
Qwen3-8B & \(1.4135{\pm}0.0026\) & \(1.6893{\pm}0.1018\) & \(1.3906{\pm}0.0117\) & \(0.0000\) \\
Llama-3.3-70B & \(1.4131{\pm}0.0005\) & \(1.7958{\pm}0.2033\) & \(1.4533{\pm}0.0050\) & \(0.0000\) \\
Mistral-7B & \(1.4122{\pm}0.0020\) & \(2.9515{\pm}0.0766\) & \(1.4679{\pm}0.0055\) & \(0.0000\) \\
Gemma-2-9B & \(1.4128{\pm}0.0012\) & \(1.7967{\pm}0.1773\) & \(1.4321{\pm}0.0193\) & \(0.0000\) \\
\midrule
Avg. (5 anchors) & \(1.4128{\pm}0.0004\) & \(2.0464{\pm}0.3776\) & \(1.4406{\pm}0.0236\) & \(0.0000\) \\
\bottomrule
\end{tabular}
\endgroup
}
\caption{\textbf{Bandwidth reconstruction check} on five anchors.  Random
controls are estimate \({\pm}(U-L)/2\), where \([L,U]\) is the seed-\(0\),
\(B=5000\), 95\% document-bootstrap percentile interval; exact asymmetric
endpoints remain in the machine-readable results.  The true aggregate has
deterministically zero error in every entry, so no uncertainty interval applies.
Avg.\ is an equal-anchor macro-average; its seed-\(0\), \(B=5000\), 95\%
percentile interval bootstraps the five anchor-level estimates as the
resampling unit.  Lower is better; values are intentionally unranked because
the columns are controlled reconstruction variants, not related-work
baselines.}
\label{tab:bandwidth-repeat}
\end{table*}}
\else
The fixed-row reconstruction results, including all five anchors and their
equal-anchor Avg., are reported in \cref{tab:bandwidth-main}; we do not repeat
the identical table here.
\fi

\Cref{thm:main-bandwidth} identifies the true tail aggregate as the statistic
that reconstructs the fixed-row output.  The empirical test compares it with
norm-matched, covariance-matched, and cross-document shuffled summaries; only
the true aggregate reaches the \((k+1)d\) floor.  We separately test whether
the fitted gap-slope distribution has the same clean two-mode structure in
vision and audio.  Neither modality meets the pre-specified two-mode
criterion, so the data do not establish a universal sharp phase law.  The
released instrument is described in \cref{sub:instrument}, and
\cref{sub:setup} itemizes the compute budget.

\begin{figure*}[!t]\centering
\includegraphics[width=\textwidth]{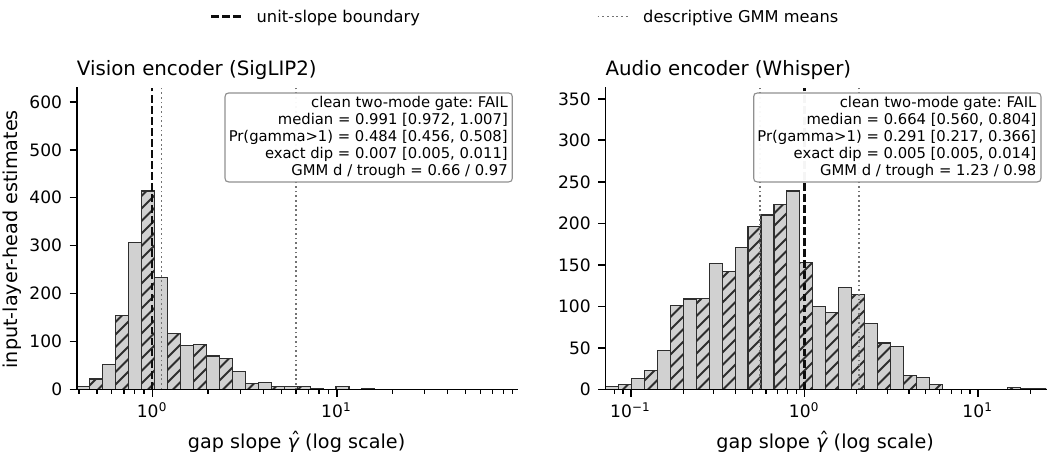}
\caption{\textbf{Cross-modal clean two-phase universality is not established.}
Per-input/layer/head gap-slope estimates are shown on a log axis; the dashed
line is the unit-slope boundary and dotted lines are descriptive two-Gaussian
means.  Neither modality meets the clean two-mode criterion because the fitted
mixtures have no sufficiently deep between-mean trough.  Brackets are \(95\%\)
percentile sensitivity intervals from resampling the four structured inputs
(seed \(0\), \(B=5000\)).  Exact Hartigan dip statistics are reported without
iid head-level \(p\)-values.}
\label{fig:phase4uni}
\end{figure*}

\subsection{Supplementary results}\label{sub:additional}

Two active supplementary measurements complete the empirical picture: the
from-scratch emergence experiment at a second, larger scale and resampling
error bars for the count law on real books.  A historical finite-precision
bandwidth panel remains in the source but is excluded from the submission
build because its per-document outcomes cannot be regenerated under the final
\(B=5000\) protocol.  The end-to-end KV-eviction evaluations are reported in
\cref{tab:ruler-main-v2,tab:lbv2-main-v2}; the algorithm itself is
\cref{alg:certkv}.

\ifdefined\ShowLegacyExploratoryTables
\paragraph{The tail aggregate at finite precision.} \Cref{thm:main-bandwidth}'s reconstruction is exact only with the real-valued $d$-dimensional aggregate (\cref{tab:bandwidth}); the deployable question is how close a \emph{quantized} aggregate gets. Quantizing the true tail centroid on an $\epsilon/\sqrt d$ grid --- the achievability behind the bit floor of \cref{thm:bw-bit-lb} --- and sweeping the kept budget $k/n$, an $8$-bit aggregate reconstructs each head's output to $\approx 2\%$ relative error at a $1\%$ kept budget and to $\lesssim 0.1\%$ at $16$ bits, an order of magnitude below the keep-only baseline, while a $2$-bit aggregate is under-resolved and can lose to keeping nothing extra (\cref{fig:e3bw}). The exact-reconstruction identity is thus a genuine, near-lossless compression at a handful of bits, not an artifact of infinite precision.

\paragraph{Count-sublinearity on real books, with resampling uncertainty.} We measure the count law on real \textsc{Project Gutenberg} books with six document/position resampling replicates from the same captured pool. The fraction of heads whose large-weight count grows sub-linearly in context length is $98.9\%$ (\textsc{Llama-3.1-8B}), $99.1\%$ (\textsc{Qwen3-8B}), and $99.4\%$ (\textsc{Mistral-7B}), with tight resampling intervals (\cref{fig:e5count}). This supports stability to the fixed resampling procedure, not six independent training or data-collection runs.

\paragraph{Emergence at a second, larger scale.} The from-scratch emergence experiment runs at two scales: the $4$M-parameter byte-level models of \cref{fig:emergence}, and a $152$M-parameter model with a real BPE tokenizer, trained on real \textsc{Project Gutenberg} books against a shuffled-token ablation over three seeds (\cref{fig:e4emerge}). At $152$M, over the plotted horizon, the real-text median kept-fraction falls from $0.67$ to $0.30$ while next-token accuracy rises to $0.41$; shuffled-token accuracy remains near its $0.05$ chance level. Because the shuffled-token kept-fraction also falls, sparsity alone is not evidence of functional learning. The supported observation is narrower: on real text, decreasing kept-fraction and increasing task accuracy recur together at this larger controlled scale.

\begin{figure*}[!t]\centering
\includegraphics[width=0.78\textwidth]{figures/e4_emergence}
\caption{\textbf{Controlled emergence at $152$M parameters} ($38\times$ the models in \cref{fig:emergence}; real \textsc{Gutenberg} books vs.\ a shuffled-token ablation; three training seeds, solid $=$ mean, band $=\pm$std). \emph{Left:} the real-text median kept-fraction $\rho_{\mathrm{med}}$ falls over the plotted horizon. \emph{Right:} next-token accuracy rises on real text but stays near chance on shuffled tokens. The omitted shuffled-token $\rho$ trajectory also falls, so attention sparsity alone does not identify functional learning; the supported signal is its conjunction with task accuracy.}
\label{fig:e4emerge}
\end{figure*}

\begin{figure}[tbp]\centering
\includegraphics[width=0.55\linewidth]{figures/e3_bandwidth}
\caption{\textbf{Recorded aggregate bandwidth diagnostic vs.\ kept budget} (\textsc{Llama-3.1-8B}, trained values; 12 documents, seed $0$, $B=1000$ document bootstrap). A $d$-dimensional tail aggregate quantized on an $\epsilon/\sqrt d$ grid (\cref{thm:bw-bit-lb}) drives the recorded reconstruction error below the keep-only baseline: $8$ bits is near-lossless, $16$ bits essentially exact, and $2$ bits is under-resolved. The per-document outcomes are not present in the local archive, so this interval has not been independently recomputed under the final $B=5000$ protocol.}
\label{fig:e3bw}
\end{figure}
\fi

\begin{figure}[tbp]\centering
\includegraphics[width=0.46\linewidth]{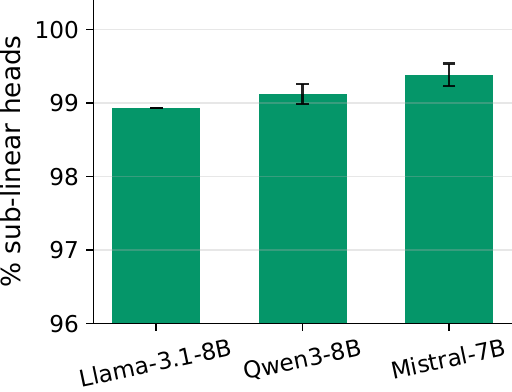}
\caption{\textbf{Count-sublinearity on real books, with resampling intervals} (real \textsc{Project Gutenberg}; fraction of heads with sub-linear large-weight count). Error bars are two-sided $95\%$ Student-$t$ intervals over six document/position resampling replicates (seed base $0$, eight documents), not six independent training runs.}
\label{fig:e5count}
\end{figure}

\newcommand{\renderScopeAndEmergenceFigures}{%
\begin{figure*}[!t]\centering
\includegraphics[width=0.90\textwidth]{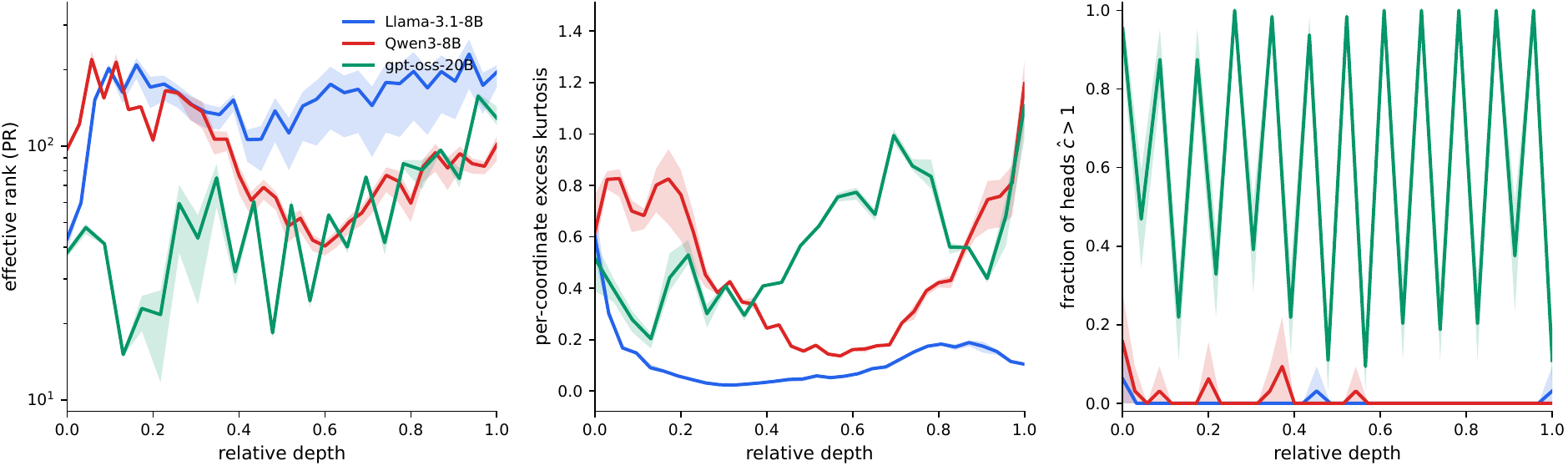}
\caption{\textbf{Real hidden states sit outside the clean
distributional regimes} (\cref{prop:scaled-inverse,cor:spectral-observable};
three architectures). \textbf{(a,b)} Effective rank and excess kurtosis, with
seed-\(0\), \(B=5000\), 95\% corpus-bootstrap bands over three trained corpora.
\textbf{(c)} Condensed-head fraction with corpus-then-head bootstrap; the
single-corpus untrained condition is omitted, and identical PG19/code inputs
are not treated as independent corpora.}
\label{fig:xdist}
\vspace{-1mm}
\end{figure*}
\begin{figure*}[!t]\centering
\includegraphics[width=0.68\textwidth]{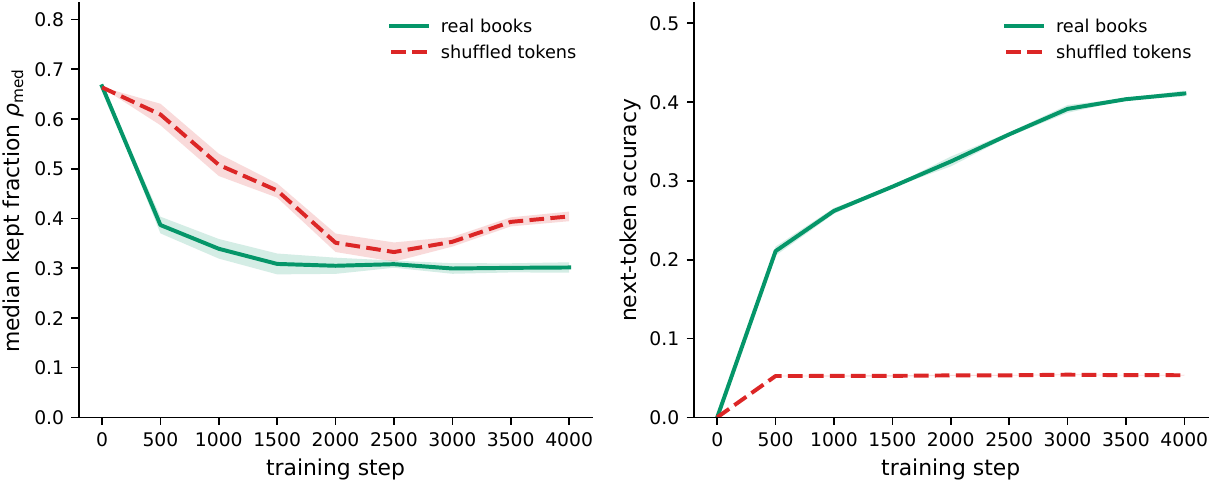}
\caption{\textbf{Sparsity accompanies learned alignment in a controlled
from-scratch replication} (\textsc{Qwen3}-architecture byte-LM).
\textbf{(a)} Median kept fraction, real versus random bytes;
\textbf{(b)} next-byte accuracy. Curves show mean \({\pm}\) sample standard
deviation over seeds \(0,1,2\); the single-checkpoint legacy control is omitted.}
\label{fig:fromscratch}
\vspace{-1mm}
\end{figure*}}
\ifdefined\vTwentyTwoArchive
\renderScopeAndEmergenceFigures
\else
\iclronly{\renderScopeAndEmergenceFigures}
\fi

\subsection{\certkv{} ablation and per-domain breakdown}\label{sub:certkv-ablation}

\paragraph{Long context: \textsc{RULER} at $128$K.}
The complete roster contains Llama and Qwen at
\(16/32/64/128\)K and Mistral at \(16\)K; the Qwen conversion has \(12\)
available tasks and is never imputed to \(13\).  The complete multi-length
table, including the \(128\)K slice, is already shown in
\cref{tab:ruler-main-v2}; it is not duplicated here.  Historical exploratory
values are excluded because their CertKV endpoint did not charge the aggregate
slot under the final matched-budget contract.  At \(128\)K, \certkv{} has the
strongest compressed point estimate on both evaluated backbones:
\(73.6{\pm}17.9\) on Llama and \(61.5{\pm}21.0\) on Qwen.  These rankings are
reported as point estimates because the task-bootstrap intervals remain wide.

\paragraph{One-factor ablations.}
The complete core and extension matrices vary the statistic,
allocator, tail summary, and per-head safeguard one factor at a time.  The
extension isolates true-uniform allocation, omitting the synthetic slot,
calibrating its proxy query, and the safeguard at both \(4\times\) and
\(10\times\).  Centered dispersion and Lagrangian-with-repair have the
strongest average point estimates in their blocks (\(25.0\%\)); the strongest
safeguard point is \(\alpha=0.1\) at \(25.1\%\).  Neither prespecified
nine-comparison family is significant after Holm correction.  The merged
table, including every two-ratio Avg., is shown once in
\cref{tab:certkv-ablation-v2}.

\begin{table}[!t]
\centering
\scriptsize
\setlength{\tabcolsep}{1.6pt}
\renewcommand{\arraystretch}{0.82}
\begin{adjustbox}{max width=\columnwidth}
\begin{tabular}{@{}llccc@{}}
\toprule
Component & Variant & $4\times$ & $10\times$ & Avg. \\
\midrule
Statistic & Tail mass & $24.7{\pm}3.7$ & $23.9{\pm}3.7$ & $24.3{\pm}3.5$ \\
 & Uncentered dispersion & $25.8{\pm}3.8$ & $23.5{\pm}3.7$ & $24.7{\pm}3.5$ \\
 & Centered dispersion & $26.0{\pm}3.8$ & $24.1{\pm}3.7$ & $25.0{\pm}3.6$ \\
\midrule
Allocator & True uniform & $26.2{\pm}3.8$ & $23.1{\pm}3.7$ & $24.7{\pm}3.5$ \\
 & Greedy marginal & $26.4{\pm}3.8$ & $23.1{\pm}3.7$ & $24.8{\pm}3.5$ \\
 & Lagrangian + repair & $26.0{\pm}3.8$ & $24.1{\pm}3.7$ & $25.0{\pm}3.6$ \\
\midrule
Tail summary & No slot & $26.0{\pm}3.8$ & $23.7{\pm}3.7$ & $24.9{\pm}3.5$ \\
 & Plain slot & $26.0{\pm}3.8$ & $24.1{\pm}3.7$ & $25.0{\pm}3.6$ \\
 & Calibrated $\bar q$ slot & $25.4{\pm}3.8$ & $24.5{\pm}3.7$ & $25.0{\pm}3.5$ \\
\midrule
Safeguard & $\alpha=0$ & $25.8{\pm}3.8$ & $24.3{\pm}3.7$ & $25.0{\pm}3.6$ \\
 & $\alpha=0.1$ & $26.0{\pm}3.8$ & $24.3{\pm}3.7$ & $25.1{\pm}3.6$ \\
 & $\alpha=0.2$ & $26.0{\pm}3.8$ & $24.1{\pm}3.7$ & $25.0{\pm}3.6$ \\
 & $\alpha=0.4$ & $25.8{\pm}3.8$ & $23.9{\pm}3.7$ & $24.9{\pm}3.5$ \\
\bottomrule
\end{tabular}

\end{adjustbox}
\caption{\textbf{\certkv{} ablations on Llama LongBench-v2.}
Accuracy is estimate \({\pm}\) 95\% example-bootstrap half-width (seed \(0\),
\(B=5000\)), conditional on one fixed checkpoint and configuration per row.
Rows are unranked one-factor changes; nine paired \(p\)-values form one family.}
\label{tab:certkv-ablation-v2}
\end{table}

\paragraph{LongBench-v2, aggregate and by domain.}
\Cref{tab:lbv2-main-v2} reports all \(503\) unique aligned examples at
\(2\times/4\times/10\times\) on three backbones, using the constrained A/B/C/D
parser.  Each random quantity is shown as estimate \({\pm}(U-L)/2\) for the
seed-\(0\), \(B=5000\), 95\% percentile-bootstrap interval; Full is an
unranked reference, while bold and underline mark the best and second distinct
compressed values.
\certkv{} ranks among the top two compressed methods in seven of the nine
model--budget settings within the four-baseline main grid.  This breadth
supports a competitive exact-budget operating point across the evaluated
backbones and ratios, not a universal method ranking.
After adjusting the \(36\) paired comparisons jointly, the Llama
\(2\times\) contrast with PyramidKV remains significant: \certkv{} leads by
\(8.75\) points (95\% paired CI \([3.58,13.92]\),
\(p_{\mathrm{Holm}}=0.0407\)).  The adjusted CertKV--Ada-KV comparisons
delimit pairwise uncertainty; the remaining point-estimate ranks are
descriptive.
\Cref{tab:lbv2-domain} begins the
six-domain breakdown.

The aggregate sweep is not duplicated in this appendix; the domain tables
below provide the additional breakdown.

\paragraph{Excluded legacy compression sweep.}\label{tab:crsweep}
The historical \(2\times/4\times/10\times/20\times\) \textsc{RULER} sweep is
retained in the source history for traceability but excluded from active
claims: its raw \texttt{\_ckv3} layer and paired-significance provenance are
unavailable, and \(20\times\) is outside the complete campaign.
This is a scope marker, not a result.

\ifdefined\ShowLegacyExploratoryTables
Per-domain samples are small ($n=33$--$175$), so the cells are noisy and no
method separates significantly within a domain; the table is included only
for historical completeness.

\begin{table*}[!t]
\centering\scriptsize
\resizebox{\textwidth}{!}{%
\begin{tabular}{@{}llcccccc@{}}
\toprule
Model & Method & Single-Doc & Multi-Doc & ICL & Dialogue & Code & Struct. \\
 & & {\scriptsize$n{=}175$} & {\scriptsize$n{=}125$} & {\scriptsize$n{=}81$} & {\scriptsize$n{=}39$} & {\scriptsize$n{=}50$} & {\scriptsize$n{=}33$} \\
\midrule
\textsc{Llama-3.1-8B} & Full cache (no eviction) & $30.3$ & $32.8$ & $23.5$ & $33.3$ & $30.0$ & $18.2$ \\
 & SnapKV \cite{li2024snapkv} & $\mathbf{34.3}$ & $28.8$ & $22.2$ & $\mathbf{33.3}$ & $28.0$ & $21.2$ \\
 & StreamingLLM \cite{xiao2024streamingllm} & $32.6$ & $29.6$ & $22.2$ & $30.8$ & $\mathbf{30.0}$ & $21.2$ \\
 & PyramidKV \cite{cai2024pyramidkv} & $24.0$ & $22.4$ & $\mathbf{28.4}$ & $30.8$ & $20.0$ & $\mathbf{36.4}$ \\
 & Ada-KV \cite{feng2024adakv} & $33.1$ & $\mathbf{31.2}$ & $21.0$ & $\mathbf{33.3}$ & $\mathbf{30.0}$ & $15.2$ \\
 & \textbf{\certkv{} (ours)} & $33.1$ & $29.6$ & $22.2$ & $\mathbf{33.3}$ & $\mathbf{30.0}$ & $15.2$ \\
\midrule
\textsc{Qwen3-8B} & Full cache (no eviction) & $32.6$ & $34.4$ & $32.1$ & $38.5$ & $34.0$ & $33.3$ \\
 & SnapKV \cite{li2024snapkv} & $\mathbf{34.3}$ & $\mathbf{33.6}$ & $27.2$ & $33.3$ & $\mathbf{34.0}$ & $33.3$ \\
 & StreamingLLM \cite{xiao2024streamingllm} & $33.7$ & $28.0$ & $\mathbf{29.6}$ & $25.6$ & $30.0$ & $39.4$ \\
 & PyramidKV \cite{cai2024pyramidkv} & $27.4$ & $26.4$ & $28.4$ & $23.1$ & $26.0$ & $\mathbf{45.5}$ \\
 & Ada-KV \cite{feng2024adakv} & $32.0$ & $30.4$ & $\mathbf{29.6}$ & $\mathbf{38.5}$ & $\mathbf{34.0}$ & $33.3$ \\
 & \textbf{\certkv{} (ours)} & $33.1$ & $28.8$ & $28.4$ & $\mathbf{38.5}$ & $\mathbf{34.0}$ & $33.3$ \\
\midrule
\textsc{Mistral-7B} & Full cache (no eviction) & $30.3$ & $30.4$ & $29.6$ & $30.8$ & $22.0$ & $27.3$ \\
 & SnapKV \cite{li2024snapkv} & $31.4$ & $29.6$ & $25.9$ & $25.6$ & $22.0$ & $21.2$ \\
 & StreamingLLM \cite{xiao2024streamingllm} & $\mathbf{33.1}$ & $\mathbf{31.2}$ & $\mathbf{29.6}$ & $\mathbf{28.2}$ & $20.0$ & $27.3$ \\
 & PyramidKV \cite{cai2024pyramidkv} & $26.3$ & $27.2$ & $25.9$ & $\mathbf{28.2}$ & $\mathbf{28.0}$ & $\mathbf{30.3}$ \\
 & Ada-KV \cite{feng2024adakv} & $32.0$ & $30.4$ & $28.4$ & $23.1$ & $22.0$ & $21.2$ \\
 & \textbf{\certkv{} (ours)} & $\mathbf{33.1}$ & $29.6$ & $28.4$ & $20.5$ & $20.0$ & $27.3$ \\
\bottomrule
\end{tabular}
}
\caption{\textbf{Legacy exploratory LongBench-v2 domain slice.}  Retained
verbatim for content provenance only.  It lacks the final three-ratio
interval and ranking protocol and is not used for a submission claim; the
complete replacements follow.}
\label{tab:lbv2-domain-legacy}
\end{table*}
\fi

\begin{table*}[!t]
\centering
\resizebox{\textwidth}{!}{}
\caption{\textbf{Complete LongBench-v2 by domain: Llama-3.1-8B.}
Accuracy (\%) is estimate \({\pm}(U-L)/2\) from seed-\(0\), \(B=5000\), 95\%
within-domain bootstrap; exact endpoints remain in the report.  Avg.\ is
503-example weighted; Full is unranked; bold/underline mark best/second
compressed methods.}
\label{tab:lbv2-domain}
\end{table*}

\begin{table*}[!t]
\centering
\resizebox{\textwidth}{!}{}
\caption{\textbf{Complete LongBench-v2 by domain: Qwen3-8B.}
Accuracy (\%) is estimate \({\pm}(U-L)/2\) from seed \(0\), \(B=5000\), 95\%
within-domain example bootstrap; exact endpoints remain in the report.
Avg.\ is 503-example weighted; Full is unranked; bold/underline mark the
best/second distinct compressed methods.}
\label{tab:lbv2-domain-qwen}
\end{table*}

\begin{table*}[!t]
\centering
\resizebox{\textwidth}{!}{}
\caption{\textbf{Complete LongBench-v2 by domain: Mistral-7B.}
Accuracy (\%) is estimate \({\pm}(U-L)/2\) from seed \(0\), \(B=5000\), 95\%
within-domain example bootstrap; exact endpoints remain in the report.
Avg.\ is 503-example weighted; Full is unranked; bold/underline mark the
best/second distinct compressed methods.}
\label{tab:lbv2-domain-mistral}
\end{table*}

\ifdefined\isarxiv\FloatBarrier\fi
\subsection{\certkv{}: implementation and exact-budget protocol}
\label{sub:efficiency}

\paragraph{The allocator.} \certkv{} turns the centered-dispersion statistic
into an eviction policy (\cref{alg:certkv}).  After prefill, each head is
scored by the unchanged SnapKV observation-window scorer
\cite{li2024snapkv}.  Because this scorer pools sums of softmax attention
weights, \(s_j>0\) in exact arithmetic; clipping is only a finite-precision
guard.  We normalize directly, rather than re-softmaxing:
\[
a_j=\frac{(s_j)_+}{\sum_i(s_i)_+}.
\]
For the score-sorted suffix \(E_h(b)\), define its mass, weighted value sum,
and weighted energy by
\[
\begin{aligned}
m_h(b)&=\textstyle\sum_{j\in E_h(b)} a_j,\\
W_h(b)&=\textstyle\sum_{j\in E_h(b)} a_j v_j,\\
Q_h(b)&=\textstyle\sum_{j\in E_h(b)} a_j\norm{v_j}^2.
\end{aligned}
\]
Its value centroid and centered second moment are
\[
\begin{aligned}
v_h^\star(b)&=\frac{W_h(b)}{m_h(b)},\\
S_h(b)&=\sum_{j\in E_h(b)}
a_j\norm{v_j-v_h^\star(b)}^2.
\end{aligned}
\]
The allocation curve is therefore
\[
\begin{aligned}
c_h(b)&=\sqrt{m_h(b)S_h(b)},\\[-1mm]
c_h^2(b)&=\bigl(m_h(b)Q_h(b)-\norm{W_h(b)}^2\bigr)_{\!+},
\end{aligned}
\]
the tight envelope within the declared \((m_h,v_h^\star,S_h)\) summary class
(\cref{thm:cert-stat-lb}).  All three ingredients are vectorized suffix scans
over the score-sorted axis.

Let \(n_{\mathrm{keep}}=\lfloor n(1-r)\rfloor\).  The matched live-slot
budget is \(B_{\mathrm{live}}=Hn_{\mathrm{keep}}\).  One aggregate slot is
charged to every head, leaving
\(B_{\mathrm{real}}=B_{\mathrm{live}}-H\) real-token slots.  The evaluated
allocator therefore solves the discrete objective
\[
\begin{aligned}
\min_{\,b_1,\dots,b_H}\quad
  &\sum_h c_h(b_h)\\
\text{s.t.}\quad
  &\sum_h b_h = B_{\mathrm{real}},\\
  &b_h\in\mathbb Z\cap[\,b_{\min},\,n-1\,],
\end{aligned}
\]
where
\(b_{\min}=\max\{1,\lfloor\alpha B_{\mathrm{real}}/H\rfloor\}\) and
\(\alpha=0.2\).  Feasibility requires \(n_{\mathrm{keep}}\ge2\).
For a shared multiplier \(\lambda\), every head globally minimizes its
penalized one-dimensional curve; bisection gives a candidate allocation and
deterministic marginal repair makes the integer budget exact.  Because the
curves may be nonconvex, this is a \emph{heuristic}, not a globally exact
solver.  Finally, the top-\(b_h\) real tokens and one charged proxy-weighted
key/value centroid remain live in each head.
For \(H\) heads, length \(L\), value width \(d\), and \(T\) bisection steps,
the implementation costs \(O(HL\log L+HLd+THL+H\Delta)\) time, where
\(\Delta\le HL\) is the number of one-token marginal repairs
(\(O(H^2L)\) in the worst case), and uses \(O(HL+HLd)\) transient auxiliary
storage for the curves and suffix moments.  All reported runs use \(T=40\).

\begin{algorithm}[tbp]
\caption{\certkv{} --- centered-dispersion allocation with one charged
aggregate slot per head (once, at end of prefill).}
\label{alg:certkv}
\begin{algorithmic}[1]
\Require scores $s_h\in\mathbb{R}^n$, keys/values
\(\{k_{h,j},v_{h,j}\}\), evicted fraction $r$, safeguard
\(\alpha=0.2\), steps \(T=40\)
\Ensure \(B_{\mathrm{live}}=H\lfloor n(1-r)\rfloor\) live slots: \(b_h\)
real tokens and one aggregate per head
\State \(B_{\mathrm{real}}\gets B_{\mathrm{live}}-H\);
\(b_{\min}\gets\max\{1,\lfloor\alpha B_{\mathrm{real}}/H\rfloor\}\)
\For{every head $h$ \textbf{in parallel}}
  \State sort by \(s_h\) descending;
  \(a\gets(s_h)_+/\sum_j(s_{h,j})_+\)
  \State suffix-scan \(m_h(b),W_h(b),Q_h(b)\) over ranks \(j>b\)
  \State \(c_h(b)\gets
  \sqrt{(m_h(b)Q_h(b)-\norm{W_h(b)}^2)_+}\) for \(b=0,\dots,n-1\)
\EndFor
\State \(\lambda_{\mathrm{lo}}\gets0\);
\(\lambda_{\mathrm{hi}}\gets2\max_{h,b}c_h(b)+1\)
\For{$t=1,\dots,T$} \Comment{bisection on the shared multiplier}
  \State \(\lambda\gets(\lambda_{\mathrm{lo}}+\lambda_{\mathrm{hi}})/2\)
  \State \(b_h(\lambda)\gets
  \arg\min_{b\in[b_{\min},n-1]}
  [c_h(b)+\lambda b]\)
  \State \textbf{if} \(\sum_h b_h(\lambda)>B_{\mathrm{real}}\)
  \textbf{then} \(\lambda_{\mathrm{lo}}\gets\lambda\)
  \textbf{else} \(\lambda_{\mathrm{hi}}\gets\lambda\)
\EndFor
\State \(b_h\gets b_h(\lambda_{\mathrm{hi}})\); repair one token at a time
until \(\sum_hb_h=B_{\mathrm{real}}\)
\State \Return per head: top-\(b_h\) real tokens plus one
proxy-weighted aggregate KV slot
\end{algorithmic}
\end{algorithm}

\paragraph{Protocol behind \cref{tab:ruler,tab:lbv2}.}
All reported end-to-end runs use the kvpress harness
\cite{nvidia2024kvpress}, greedy decoding, and inference seed \(42\).
LongBench-v2 uses all \(503\) unique aligned multiple-choice examples at
\(32\)K truncation for every \(2\times/4\times/10\times\) setting.  The
parser accepts only explicit A/B/C/D answer patterns; refusals and
non-answers remain invalid.  Example-level percentile-bootstrap intervals use
seed \(0\) and \(B=5000\); tables display estimate \({\pm}(U-L)/2\), while the
machine-readable results retain exact asymmetric endpoints.

The complete multi-length RULER roster is Llama and Qwen at
\(16/32/64/128\)K plus Mistral at \(16\)K, all at \(2\times\) matched
compression.  The fixed example fractions are
\((.10,.10,.10,.03)\) for Llama, \((.10,.10,.05,.03)\) for Qwen, and
\(.10\) for Mistral; every method within a model--length setting uses the same
sampled examples.  Qwen \(128\)K uses the fixed training-free YaRN
extension.  The sweep uses the fixed per-task generation caps and the
actual task denominators: \(13\) where all tasks are available and \(12\) for
the Qwen conversion, with no imputation.  Task-bootstrap intervals use seed
\(0\), \(B=5000\).  \Cref{tab:ruler-main-v2} is generated from the complete
nine-setting artifact set after checking exact live-token budgets, aligned
task denominators, prediction and score files, and per-file checksums.  Source
identity is established by the validated per-file hash list rather than the
runner's non-Git deployment metadata.

\paragraph{Physical cache capacity relative to Full.}
The physical prototype materializes a packed \texttt{bfloat16} cache for
\textsc{Llama-3.1-8B}.  At \(64\)K and \(10\times\), persistent KV state falls
from \(8.59\) GB for Full to \(0.86\) GB for \certkv{}, while decode peak
memory falls from \(23.11\) to \(16.34\) GiB.  Persistent bytes are exact; the
displayed peak is unchanged across five repetitions at the reported
precision.

\begin{table}[!t]
\centering
\small
\begin{tabular}{@{}lcc@{}}
\toprule
Method & Persistent KV & Decode peak \\
\midrule
Full cache (\(1\times\)) & \(8.59\) GB & \(23.11\) GiB \\
\certkv{} (\(10\times\)) & \(0.86\) GB & \(16.34\) GiB \\
\bottomrule
\end{tabular}

\caption{\textbf{Physical memory at \(64\)K.}  Full uses the uncompressed
cache; \certkv{} uses the \(10\times\) exact live-slot budget.  Persistent
bytes are exact, and decode peak is identical across five repetitions within
the displayed precision.}
\label{tab:physical-cache}
\label{tab:physical-memory}
\end{table}

\begin{figure*}[!t]\centering
\includegraphics[width=\textwidth]{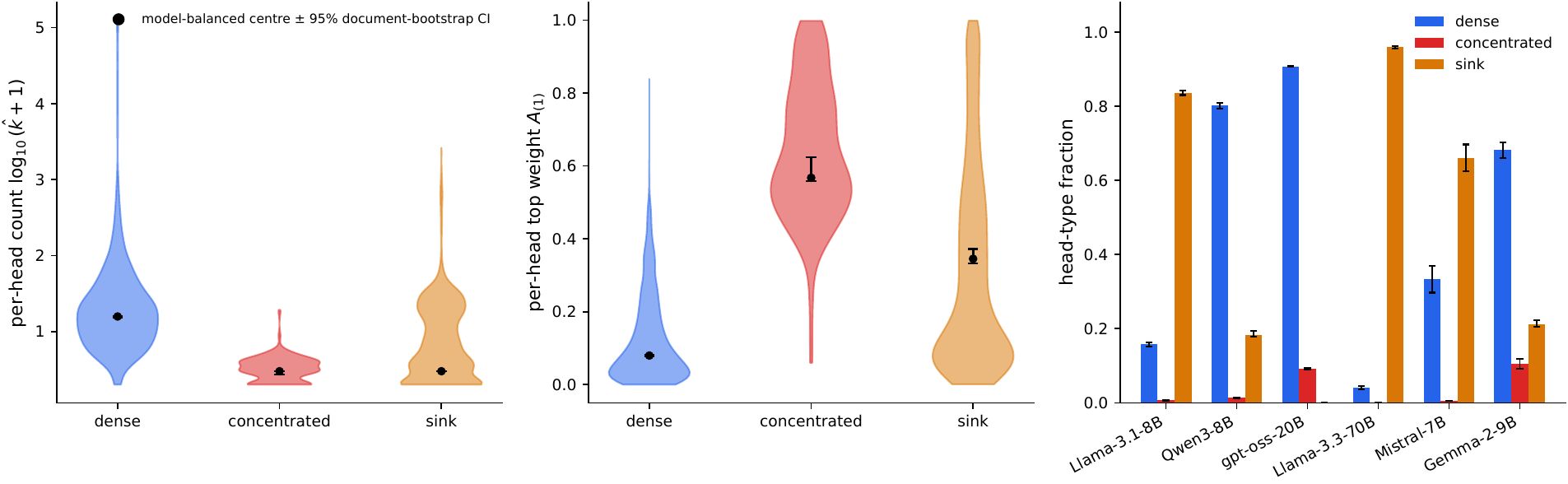}
\caption{\textbf{Per-head distributions by static type.} Left/middle:
violins show the finite observed distributions of per-head count \(\hat k\)
and top weight \(A_{(1)}\) at the longest context; black points and bars are
the model-balanced center and seed-\(0\), \(B=5000\), 95\% percentile
document-bootstrap interval.  Right: head-type composition per anchor with
the same document-bootstrap intervals.  The types summarize static profiles
and should not be read as functional head labels.}
\label{fig:perhead}
\end{figure*}

\subsection{Experimental setup}\label{sub:setup}
We run a pre-specified six-anchor evaluation.  This subsection records the
models, datasets, tasks, hyperparameters, numerical precision, software,
compute, and analysis used in the experiments.  The gap-profile instrument is
detailed in \cref{sub:instrument}, and the per-head stratification is defined
in \cref{sub:perhead}.

\paragraph{Models.} The program is anchored on six deployed language models.
Five --- \textsc{Llama-3.1-8B} \cite{grattafiori2024llama3},
\textsc{Qwen3-8B} \cite{yang2025qwen3}, \textsc{gpt-oss-20B}
\cite{openai2025gptoss}, \textsc{Llama-3.3-70B}
\cite{grattafiori2024llama3}, and \textsc{Mistral-7B} --- provide measured
profiles from \(4\)K through \(128\)K.  The recovered
\textsc{Gemma-2-9B} surface contains only \(4\)K and is used solely as a
single-length descriptive control, never for growth inference.  The
cross-modality universality test uses two bidirectional vision encoders,
\textsc{SigLIP2} \cite{tschannen2025siglip2} and \textsc{DINOv2}, an audio
encoder, \textsc{Whisper-large-v3} \cite{radford2023whisper}, and the causal
image control \textsc{ImageGPT}.  The emergence study trains a
\textsc{GPT-NeoX}-architecture model (\(3.22\)M parameters) and a
\textsc{Qwen3}-architecture model (\(4.26\)M parameters); the untrained
controls instantiate the same architecture configurations.

\paragraph{Datasets and tasks.} The controlling six-anchor profile records
contain one natural-text surface: Wikipedia fallback documents stored under a
legacy \texttt{pg19} key.  We retain that provenance literally rather than
calling the surface a three-domain atlas.  Older GovReport, mathematics, and
book summaries cover narrower model slices and are descriptive only; they are
not pooled into the six-anchor headline.  The synthetic \textsc{RULER}
\cite{hsieh2024ruler} corpus enters only the task experiments.  LongBench-v2
\cite{bai2024longbench2} supplies the 503-item multiple-choice evaluation;
decoding is greedy throughout.  The pre-specified NoLiMa
\cite{modarressi2025nolima} and real-task phase-transfer attempts remain
negative controls, not positive task evidence.

\paragraph{Hyperparameters.} Evaluated profile lengths are
\(n\in\{4\text{k},16\text{k},32\text{k},64\text{k},128\text{k}\}\) for
the five long-context anchors and \(4\)K only for Gemma.  Growth inference
requires at least three distinct lengths and at least two finite
document-level slopes per head; non-finite fits are excluded explicitly.  The
coverage budget is swept over
\(\eta\in\{0.2,0.1,0.05,0.02,0.01\}\), and the relative-count threshold over
\(\epsilon\in\{0.5,0.1,0.05\}\) as a fraction of \(A_{(1)}\).  Each gap
profile uses 256 log-spaced rank-weight samples, with the first four tokens
plus pre-identified sinks held out as an always-keep set.  Task experiments sweep
top-\(k\in\{4,8,16,32,64,128\}\); the from-scratch trajectories are logged
every 300 steps over three seeds.

\paragraph{Numerical precision.}
Models run in \texttt{bfloat16} by default, but the gap-profile instrument
recomputes row logits in \texttt{fp32}.  At half precision, logit
quantization error is approximately \(0.1\) nat, comparable to the resolution
required by the tail fits, and \(e^{-\gap}\) falls below \texttt{bf16}
resolution for \(\gap\gtrsim25\).  The captured rows match an eager
\texttt{output\_attentions} reference to approximately \(10^{-6}\).
Fitted curves are stored in \texttt{fp16}.

\paragraph{Software.} The production evaluation environment uses
\textsc{PyTorch}~2.8.0+cu128 and \textsc{HuggingFace
Transformers}~4.57.6, with \textsc{Datasets}, \textsc{NumPy},
\textsc{SciPy}, and \textsc{Matplotlib}.  The reproducibility bundle pins
each executable by SHA-256 and records the complete environment export.

\paragraph{Compute.} The program is an approximate multi-week single-GPU run
on one 96\,GB NVIDIA RTX PRO 6000 Blackwell (compute capability 12.0).  The
dominant cost is the 70B anchor and the five 128K profile surfaces, followed
by the development-checkpoint sweep.  No 120B anchor is part of the evaluated
roster.

\paragraph{Analysis.} The hardened growth report uses document bootstrap with
seed \(0\), \(B=5000\), followed by Benjamini--Hochberg correction at
\(q=0.05\) \cite{benjamini1995fdr}.  Underidentified or non-finite head fits
never enter the FDR denominator.  The plateau--power-law form is admitted only
where it beats exponential, log-normal, and stretched-exponential tails under
a Vuong test \cite{vuong1989tests} at FDR \(0.05\).  Heads are stratified a
priori into sink, spike, and dense types; the untrained control uses the same
architecture configuration.

\paragraph{Methodology notes.}
\emph{Task \(\alpha\).}  For each head, we estimate the task's effective
\(\alpha\) in \cref{eq:target-mix} by
\[
\hat\alpha=
\frac{\langle y-\bar V_b,V_{j^\star}-\bar V_b\rangle}
{\lVert V_{j^\star}-\bar V_b\rVert_2^2},
\]
then average over task items.  We estimate \(w_e\) and \(w_s\) independently
from the exact and top-\(k\) weights on \(j^\star\); the loss crossing is not
used to fit them.  \emph{Head strata.}  We assign each head a priori to the
attention-sink, retrieval-spike, or dense-bulk stratum and report admission
rate, \(R^2\), and Vuong verdicts separately.  A fit failure in the sink or
spike stratum downgrades the claim even when the bulk passes.
\emph{Per-head decisions.}  Each ``at least \(X\%\) of heads'' statement
aggregates per-head document-bootstrap intervals under FDR correction.

\begin{figure}[tbp]\centering
\includegraphics[width=0.7\linewidth]{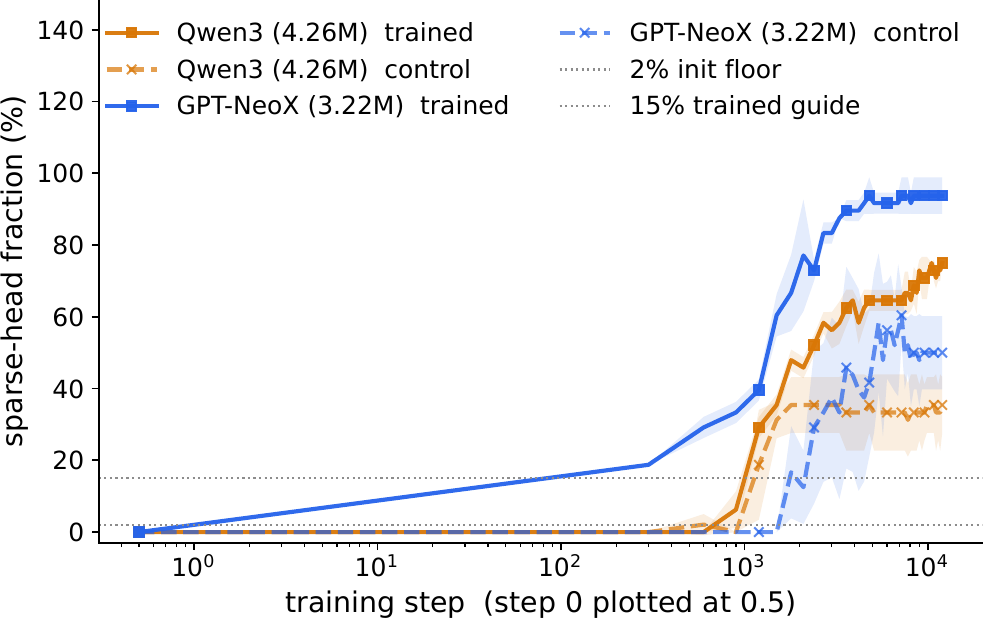}
\caption{\textbf{Sparse-head growth recurs in two controlled architectures.} Sparse-head fraction (kept-fraction $\rho_{\eta=0.1}\le0.1$) vs.\ training step for two from-scratch byte-level language models (\textsc{GPT-NeoX}-architecture, $3.22$M parameters; \textsc{Qwen3}-architecture, $4.26$M parameters; $3$ seeds, solid $=$ mean, band $=\pm$std). On real text both trajectories cross the $15\%$ guide (\textsc{GPT-NeoX} to ${\sim}93\%$, \textsc{Qwen3} to ${\sim}75\%$ of heads sparse); the i.i.d.-random-byte controls (dashed) are nonzero but remain lower.}
\label{fig:emergence}
\end{figure}


\paragraph{The instrument: gap-profile capture and fitting.}\label{sub:instrument}
A single prefill per document runs under the production kernel.  Hooks record
query and key states at sampled decode positions, after which the instrument
recomputes row logits in \texttt{fp32} and removes an estimated numerical
noise floor.  Model-specific adapters handle grouped-query attention
\cite{ainslie2023gqa}, Gemma-2 sliding windows, and learned sinks.

For each head, an always-keep set containing the first \(s_0=4\) tokens and
pre-identified sinks separates the profile into a plateau and a tail.  We estimate
the plateau exit \(\hat J_h\) and the tail slope \(\hat\gamma_h\) from
\(\log A_{(j)}\) versus \(\log j\) over
\([\hat J_h,j_{\max}]\), then pool positions by their per-head median.  A head
is admitted only when the plateau--power-law model beats exponential,
log-normal, and stretched-exponential alternatives under a Vuong test at FDR
\(0.05\) \cite{clauset2009powerlaw,vuong1989tests,benjamini1995fdr}.
Document bootstrap quantifies uncertainty, and layer-level partial pooling
regularizes the head estimates \cite{gelmanhill2007}.  The output contains one
record per model, layer, head, corpus, and length, including fitted parameters,
model-comparison verdicts, measured coverage budgets, and measured counts.

\paragraph{Per-head stratification with within-type pooling.}\label{sub:perhead}
Heads are not interchangeable, so each head is first assigned a static structural type and only then pooled within that type across the six anchors for the descriptive distributions in \cref{fig:perhead}. The three MInference patterns \cite{jiang2024minference} are: \emph{sink} (A-shape; high initial-token mass, the attention sinks of \cite{xiao2024streamingllm}), \emph{concentrated} (Vertical-Slash; a few dominant columns, high $A_{(1)}$, non-sink), and \emph{dense} (Block-Sparse; broad mass). The roster is sink-heavy on the Llama family ($80$--$84\%$ of heads) and dense-heavy on Qwen3/\textsc{gpt-oss} (whose learned sinks sit in $Z$, not on the first real tokens, so they read as dense). These are static profile classes, not functional retrieval or induction labels; the latter require task probes \cite{wu2024retrieval,olsson2022induction} and are out of scope here.

\textbf{Reproducibility records.} The release bundle records the profilometer,
figure generators, per-head profile records, protocol configurations, raw
predictions, item-level task outcomes, SHA-256 checksums, and a one-command
validation entry point.  It supports the results reported here; no broader
``Attention Atlas'' release is claimed.


\ifdefined\vTwentyTwoArchive\else
\input{sections/6_methods}
\fi
\section{Proof Roadmap: Dependency Graph and Reading Guide}\label{sec:app-roadmap}

This appendix is a map of the proofs. \Cref{fig:roadmap-dag} draws the dependency
graph of every fact, definition, lemma, proposition, theorem, and corollary in the
paper; the prose below walks the same structure cluster by cluster. An arrow
$A\!\to\!B$ means \emph{$A$ is used in the proof of $B$}. Each dashed box is one
source file (a main-text section or an appendix). Node shape/colour encodes the
result type (legend, top-left). Where a main-text statement is the informal twin of
a formal appendix result, the two are merged into a single node carrying both names
(e.g.\ ``summable-tail coverage / super-log-slope branch''); per-statement sub-labels and
remarks are omitted from the graph for legibility but are reached from their parent.

\begin{figure*}[!t]
\centering
\resizebox{\textwidth}{!}{%
\begin{tikzpicture}[>=Stealth, font=\footnotesize, every node/.style={align=center}]
\tikzset{
  fact/.style={rounded corners=2pt, draw=black!45, fill=black!7,  inner sep=2pt, minimum height=5.5mm},
  defn/.style={rounded corners=6pt, draw=black!35, fill=black!4,  inner sep=2pt, minimum height=5.5mm},
  lem/.style ={draw=blue!65,  fill=blue!7,   ellipse,   inner sep=1.5pt, minimum height=6mm},
  thm/.style ={draw=red!65,   fill=red!6,    rectangle, inner sep=2.5pt, minimum height=6mm},
  prop/.style={draw=orange!85,fill=orange!9, rectangle, inner sep=2.5pt, minimum height=6mm},
  cor/.style ={draw=green!55!black, fill=green!8, rectangle, inner sep=2.5pt, minimum height=6mm},
  main/.style={line width=1.1pt},
  dep/.style ={->, draw=black!50, thin},
  spine/.style={->, draw=black!80, line width=0.9pt},
  scoped/.style={->, draw=black!45, thin, dashed},
  clbox/.style={rounded corners, draw=black!30, dashed, line width=0.7pt, inner sep=7pt},
  cllab/.style={font=\scriptsize\bfseries, text=black!55},
}

\node[fact] (lgF) at (-1.2,15.4) {fact / def};
\node[lem]  (lgL) at (1.0,15.4)  {lemma};
\node[prop] (lgP) at (2.8,15.4)  {prop};
\node[thm]  (lgT) at (4.4,15.4)  {theorem};
\node[cor]  (lgC) at (6.1,15.4)  {corollary};
\node[thm,main] (lgM) at (8.3,15.4) {headline (bold)};
\node[align=left, font=\scriptsize] at (12.6,15.4)
  {arrow $A\!\to\!B$ : $A$ used in proof of $B$\\ dashed box = one source file};

\node[fact] (score)   at (0,12.5) {Score-scale\\bridge};
\node[defn] (gapprof) at (0,10.8) {Sorted-gap\\profile};
\node[defn] (cover)   at (0,9.2)  {Mass-coverage\\budget};
\node[lem]  (gaprep)  at (0,7.2)  {Gap-to-weight\\identity};
\node[lem]  (sumint)  at (0,5.6)  {Sum--integral\\comparison};

\node[thm] (s1)  at (3.1,8.0) {Summable-tail\\coverage};
\node[prop](s1t) at (3.1,6.4) {Matching profile\\witness};
\node[thm] (s2)  at (3.1,4.8) {Sub-log dense\\coverage};
\node[thm] (s3)  at (3.1,3.2) {Plateau--decay\\budget};

\node[lem] (cher)   at (6.5,12.8) {Dependent-count\\concentration};
\node[lem] (topk)   at (6.5,11.3) {Top-$k$ subset\\optimality};
\node[thm] (dich2)  at (6.5,9.8)  {Unit-slope coverage\\dichotomy};
\node[cor] (r1)     at (6.5,8.3)  {Bounded-score\\density};
\node[prop](r2)     at (6.5,6.8)  {Mixing-profile\\transfer};
\node[lem] (countas)at (6.5,5.0)  {Level-count\\asymptotics};
\node[prop](restrz) at (6.5,3.5)  {Restricted partition\\function};
\node[cor] (maxlog) at (6.5,2.0)  {Maximum-logit\\scale};
\node[thm] (r3a)    at (6.5,0.5)  {Constant-scale\\large-weight count};
\node[cor] (r3ad)   at (6.5,-1.0) {Constant-scale\\dense mass};

\node[thm,main](nat)  at (10.0,12.0) {Sublinear\\large-weight count};
\node[cor,main](sub)  at (10.0,10.5) {Relative-count\\subpolynomiality};
\node[thm,main](ldense)at(10.0,9.0)  {Generic dense\\coverage};
\node[thm,main](comp) at (10.0,7.5)  {Gaussian\\coverage law};
\node[thm,main](cva)  at (10.0,6.0)  {Count is not\\coverage};
\node[prop,main](appx)at (10.0,4.5)  {Training-induced\\approximability};
\node[prop](sinv)     at (10.0,3.0)  {Score-scale\\crossover};
\node[cor] (spec)     at (10.0,1.5)  {Spectral-scale\\diagnostic};

\node[lem] (rsw)   at (13.6,12.4) {Top-rank\\sandwich};
\node[thm] (t1)    at (13.6,10.9) {Dispersed-phase\\free energy};
\node[cor] (t1f)   at (13.6,9.4)  {Fixed-scale\\dense coverage};
\node[prop](vid)   at (13.6,7.9)  {Value-weight\\identity};
\node[cor] (vord)  at (13.6,6.4)  {Value-error\\ordering};
\node[lem] (tvec)  at (13.6,4.6)  {Extreme-weight\\process};
\node[thm] (t2pd)  at (13.6,2.7)  {Poisson--Dirichlet\\limit};
\node[lem] (atoml) at (13.6,1.2)  {Atomless\\coverage limit};
\node[thm] (t2cov) at (13.6,-0.3) {Condensed-phase\\coverage};
\node[thm,main](mscaled) at (13.6,-1.8){Score-scale\\phase transition};

\node[lem] (spc)   at (17.0,8.5)  {Spike exceedance\\count};
\node[thm] (t3)    at (17.0,7.0)  {Spike\\detectability};
\node[prop](t3d)   at (17.0,5.5)  {Spike\\dominance};
\node[thm] (t4)    at (17.0,4.0)  {Planted-signal\\denoising band};
\node[thm](mspike) at (17.0,2.5){Detection--dominance\\separation};
\node[thm](mden)   at (17.0,1.0){Stylized denoising\\consequence};

\node[prop](bias)  at (20.6,13.2) {Mass-to-output\\identity};
\node[prop](compn) at (20.6,11.7) {Exact tail\\compensation};
\node[thm] (factor)at (20.6,10.2) {Tail-aggregate\\bandwidth};
\node[thm] (bwlb)  at (20.6,8.7)  {Finite-precision\\bit floor};
\node[prop](flat)  at (20.6,7.2)  {Flat-row\\tightness};
\node[thm] (orex)  at (20.6,5.5)  {Value-aware subset\\existence};
\node[lem] (ordec) at (20.6,4.0)  {Cancellation--mass\\decomposition};
\node[thm] (orprop)at (20.6,2.5)  {Value-aware subset\\characterization};
\node[thm] (cert)  at (20.6,1.0)  {Fixed-row cluster\\compensation bound};
\node[cor] (certs) at (20.6,-0.5) {Small-radius\\specialization};
\node[lem] (selc)  at (20.6,-2.0) {Selector-error\\composition};

\node[thm](phase) at (24.4,4.2) {Stylized target-mixture\\phase law};
\node[thm](order) at (24.4,2.4) {Stylized exact--sparse\\ordering};

\begin{scope}[on background layer]
  \node[clbox, fit=(score)(gapprof)(cover)(gaprep)(sumint)] (cl0){};
  \node[cllab] at (0,14.2) {Definitions};
  \node[clbox, fit=(s1)(s1t)(s2)(s3)] (cl1){};
  \node[cllab] at (3.1,9.6) {Deterministic profile};
  \node[clbox, fit=(cher)(topk)(dich2)(r1)(r2)(countas)(restrz)(maxlog)(r3a)(r3ad)] (cl2){};
  \node[cllab] at (6.5,14.2) {Count and levels};
  \node[clbox, fit=(nat)(sub)(ldense)(comp)(cva)(appx)(sinv)(spec)] (cl3){};
  \node[cllab] at (10.0,13.4) {Count vs.\ coverage};
  \node[clbox, fit=(rsw)(t1)(t1f)(vid)(vord)(tvec)(t2pd)(atoml)(t2cov)(mscaled)] (cl4){};
  \node[cllab] at (13.6,13.8) {Dispersed / condensed};
  \node[clbox, fit=(spc)(t3)(t3d)(t4)(mspike)(mden)] (cl5){};
  \node[cllab] at (17.0,9.8) {Planted-signal branch};
  \node[clbox, fit=(bias)(compn)(factor)(bwlb)(flat)(orex)(ordec)(orprop)(cert)(certs)(selc)] (cl6){};
  \node[cllab] at (20.6,14.5) {Value reduction};
  \node[clbox, fit=(phase)(order)] (cl7){};
  \node[cllab] at (24.4,5.4) {Stylized tasks};
\end{scope}

\draw[dep] (gaprep) -- (s1); \draw[dep] (gaprep) -- (s2); \draw[dep] (gaprep) -- (s3);
\draw[dep] (sumint) -- (s1); \draw[dep] (sumint) -- (s2); \draw[dep] (s1) -- (s1t);
\draw[dep] (score) -- (r3a); \draw[dep] (cher) -- (r2); \draw[dep] (cher) -- (r3a);
\draw[dep] (topk) -- (dich2); \draw[dep] (s1) to[bend left=8] (dich2); \draw[dep] (s2) -- (r1);
\draw[dep] (countas) -- (restrz); \draw[dep] (restrz) -- (maxlog); \draw[dep] (maxlog) -- (r3a);
\draw[spine] (r3a) -- (nat); \draw[dep] (nat) -- (sub); \draw[dep] (r3ad) -- (ldense);
\draw[dep] (r3a) -- (r3ad); \draw[dep] (s2) to[bend right=6] (ldense);
\draw[spine] (nat) -- (cva); \draw[dep] (s1) to[bend right=10] (cva);
\draw[dep] (bias) to[bend left=4] (cva); \draw[dep] (comp) -- (appx); \draw[dep] (sinv) -- (appx);
\draw[dep] (sinv) -- (spec);
\draw[dep] (cher) to[bend left=4] (rsw); \draw[dep] (restrz) -- (t1); \draw[dep] (rsw) -- (t1);
\draw[dep] (t1) -- (t1f); \draw[dep] (vid) -- (vord); \draw[dep] (t1) -- (vord);
\draw[dep] (tvec) -- (t2pd); \draw[dep] (t2pd) -- (atoml); \draw[dep] (atoml) -- (t2cov);
\draw[dep] (t2pd) -- (t2cov); \draw[spine] (t1) -- (mscaled); \draw[spine] (t2pd) -- (mscaled);
\draw[dep] (sinv) to[bend left=6] (t1);
\draw[dep] (spc) -- (t3); \draw[dep] (spc) -- (t3d); \draw[dep] (t3) -- (t4); \draw[dep] (t3d) -- (t4);
\draw[dep] (t3) -- (mspike); \draw[scoped] (t4) -- (mden);
\draw[dep] (bias) -- (compn); \draw[dep] (bias) -- (factor); \draw[dep] (bias) -- (flat);
\draw[dep] (compn) -- (factor); \draw[dep] (factor) -- (bwlb); \draw[dep] (compn) -- (cert);
\draw[dep] (orex) -- (orprop); \draw[dep] (ordec) -- (orprop); \draw[dep] (cert) -- (certs);
\draw[scoped] (mden) -- (phase); \draw[scoped] (orprop) to[bend right=6] (order);
\draw[scoped] (factor) to[bend right=10] (order); \draw[scoped] (mden) -- (order);
\draw[scoped] (phase) -- (order);

\end{tikzpicture}}
\caption{Dependency graph of the paper's results. An arrow $A\!\to\!B$ means $A$ is
invoked in the proof of $B$; only principal dependencies are drawn (per-statement
sub-claims and remarks are suppressed). Dashed boxes group source clusters; bold nodes
form the load-bearing count--coverage--score-scale spine. Dashed arrows mark
scope-limited stylized consequences rather than real-task or deployment guarantees.
Node text uses semantic result titles; formal labels are given in the reading guide.
This is a deterministic dependency diagram, not a method leaderboard:
uncertainty, best/second ranking, and a method Avg.\ are not applicable, and
bold nodes denote logical load bearing rather than empirical rank.}
\label{fig:roadmap-dag}
\end{figure*}

\paragraph{How to read the graph (cluster by cluster).}
The argument flows left to right.
\emph{Preliminaries} (\cref{fac:score-scale}, \cref{def:gap-profile}, \cref{def:coverage})
fix the one object everything is about---the sorted-score \emph{gap profile} of a single
softmax row---and the two elementary tools, the gap-to-weight-to-coverage dictionary
\cref{lem:gap-rep} and the sum--integral comparison \cref{lem:sum-integral}.
\emph{Profile} (App~\ref{sec:app-profile}) turns those into the deterministic criterion:
a summable gap tail gives a length-independent budget (\cref{thm:s1-summable}, sharp by
\cref{prop:s1-tight}), a sub-logarithmic gap forces a dense row (\cref{thm:s2-sublog}),
and a plateau-then-decay profile gives a sublinear budget (\cref{thm:s3-plateau})---together
the log-slope dichotomy \cref{thm:main-dichotomy} and \cref{cor:main-plateau}.
\emph{Count ladder} (App~\ref{sec:app-ladder},~\ref{sec:app-level}) climbs the assumption
ladder---Chernoff/Bennett concentration (\cref{lem:chernoff}), the rule-agnostic optimality
of top-$k$ (\cref{lem:topk-optimal}) and two-sided dichotomy (\cref{thm:dichotomy-2sided}),
the Bounded density corollary (\cref{cor:r1-density}), Mixing counts (\cref{prop:r2-counts}),
and the level-decomposition free-energy machine (\cref{lem:count-asymptotics},
\cref{prop:restricted-z}, \cref{cor:max-logit})---to prove the i.i.d.\ count
(\cref{thm:r3a-count}) and its dense coverage (\cref{cor:r3a-dense}).
\emph{Count vs coverage} (\S\ref{sec:sparsity},~\ref{sec:error}) is the hinge: the count is
$O(\log n)$ (\cref{thm:naturally-sparse}, \cref{cor:subpoly-count}) yet the coverage budget is
$\Theta(n)$ (\cref{thm:main-ladder-dense}, \cref{thm:compression}), so count and coverage come
apart (\cref{thm:count-vs-approx}); approximability is recovered only as the score scale grows
(\cref{prop:approximable}, observable via \cref{cor:spectral-observable}).
\emph{Scaled ensemble} (App~\ref{sec:app-dispersed},~\ref{sec:app-condensed}) makes the
$c=1$ transition sharp: the dispersed phase $c<1$ (\cref{lem:rank-sandwich}, \cref{thm:t1-main},
\cref{cor:value-order}) and the condensed phase $c>1$ with its Poisson--Dirichlet limit
(\cref{lem:top-vector}, \cref{thm:t2-pd}, \cref{thm:t2-coverage}), unified in \cref{thm:main-scaled}.
\emph{Spiked \& denoising} (App~\ref{sec:app-spike}) analyzes a stylized planted signal---detection vs
dominance (\cref{lem:spike-count}, \cref{thm:t3-rank}, \cref{prop:t3-dominance})---and locates
the model-specific band where renormalized top-$k$ beats exact
(\cref{thm:t4}, \cref{thm:main-denoise}); this is an auxiliary formal branch, not a
claim that the ordering transfers to real retrieval tasks.
\emph{Reduction / oracle / bandwidth} (App~\ref{sec:app-reduction}) supplies the value-aware layer:
the mass-to-output bias identity (\cref{prop:bias-identity}), the mandatory single aggregate
(\cref{prop:compensation}, \cref{thm:factor-map}, \cref{thm:bw-bit-lb}), the flat-row barrier
(\cref{prop:flat-tight}), the optimal subset and its characterization (\cref{thm:oracle-existence},
\cref{lem:oracle-decomp}, \cref{thm:oracle-property}), and a fixed-row
cluster-compensation bound
(\cref{thm:certificate}, \cref{cor:certificate-small}).
\emph{Stylized task ordering} (\S\ref{sec:tasks}) is shown as a dashed auxiliary branch:
the single-parameter phase law \cref{thm:task-phase-law} and the
exact/sparse/oracle ordering \cref{thm:task-ordering} are consequences within
their planted-signal setup, jointly using the denoising band and oracle
characterization, but they are not load-bearing real-task or runtime claims.

\section{Reduction, Oracle Subsets, Tail Compensation, and Computable Error Bounds}\label{sec:app-reduction}

This appendix proves \cref{lem:main-mass,thm:main-tight,thm:main-oracle,thm:main-bandwidth,thm:main-certificate} of the main text: \cref{lem:main-mass} is \Cref{prop:bias-identity}, \cref{thm:main-bandwidth} is \Cref{prop:compensation,thm:factor-map}, \cref{thm:main-tight} is \Cref{prop:flat-tight}, \cref{thm:main-oracle} is covered by \Cref{lem:oracle-decomp,thm:oracle-existence,thm:oracle-property,prop:proxy-uncomp,rem:oracle-collapse,prop:flat-tight}, and \cref{thm:main-certificate} is \Cref{thm:certificate,cor:certificate-small}. \Cref{thm:future-query-lift} gives the a posteriori bridge from fixed-row compensation to a future-query synthetic slot. \Cref{prop:sampled-comp} prices the sampled compensation variant; \cref{lem:main-selector} is \Cref{lem:selector-comp}, and the \textsc{Free} error cell of \cref{tab:main-ladder} is \Cref{fac:orderstat-mass}.

\subsection{Reduction and sufficiency}\label{sub:app-t0}

All results in this subsection are exact algebraic identities for a fixed
logit vector and fixed values; no probability is involved. Throughout the
subsection, fix a softmax row as in \Cref{def:softmax-row}, a kept set $S$ as
in \Cref{def:schemes}, and recall $p_S \in (0,1)$, $T = [n] \setminus S$.

\begin{proposition}[Bias identity and mass bounds; formal version of \cref{lem:main-mass}]\label{prop:bias-identity}
The following hold.
\begin{enumerate}[label=(\roman*), ref=\ref*{prop:bias-identity}(\roman*)]
\item\label{prop:bias-identity:split} $O = p_S\, \wt{O}_S + (1 - p_S)\,
\bar{V}_T$, and consequently $\wt{O}_S - O = (1 - p_S)\big(\wt{O}_S -
\bar{V}_T\big)$.
\item\label{prop:bias-identity:trunc} $O_S - O = -(1 - p_S)\,\bar{V}_T$,
hence $\norm{O_S - O}_2 = (1 - p_S)\norm{\bar{V}_T}_2$.
\item\label{prop:bias-identity:mass} $\norm{\wt{O}_S - O}_2 \le 2\,\Vmax\,
(1 - p_S)$.
\end{enumerate}
\end{proposition}

\begin{proof}
Splitting the defining sum of $O$ over $S$ and $T$ and inserting the
definitions of $\wt{O}_S$ and $\bar{V}_T$ from \Cref{def:schemes},
\begin{align}\label{eq:o-split}
\begin{split}
O
&\;=\; \sum_{j \in S} A_j V_j + \sum_{j \in T} A_j V_j\\
&\;=\; p_S \cdot \frac{1}{p_S}
   \sum_{j \in S} A_j V_j\\
&\hspace{1.8em}
   + (1 - p_S) \cdot \frac{1}{1-p_S}
   \sum_{j \in T} A_j V_j\\
&\;=\; p_S\, \wt{O}_S + (1 - p_S)\, \bar{V}_T.
\end{split}
\end{align}
Subtracting Eq.~\eqref{eq:o-split} from $\wt{O}_S = p_S \wt{O}_S +
(1-p_S)\wt{O}_S$ gives
\begin{align*}
\wt{O}_S - O
&\;=\; (1 - p_S)\,\wt{O}_S
       - (1 - p_S)\,\bar{V}_T\\
&\;=\; (1 - p_S)
       \big(\wt{O}_S - \bar{V}_T\big),
\end{align*}
which proves (i). For (ii), Eq.~\eqref{eq:o-split} and $O_S = p_S \wt{O}_S$
give
\begin{align*}
O_S - O
&\;=\; p_S\,\wt{O}_S
 - \big(p_S\, \wt{O}_S
 + (1-p_S)\,\bar{V}_T\big)\\
&\;=\; -(1 - p_S)\,\bar{V}_T,
\end{align*}
and taking norms yields the stated equality. For (iii), both $\wt{O}_S$ and
$\bar{V}_T$ are convex combinations of value vectors
(\Cref{def:schemes}), so the triangle inequality gives
\begin{align*}
\norm{\wt{O}_S - O}_2
&\;=\; (1-p_S)\,
       \norm{\wt{O}_S - \bar{V}_T}_2\\
&\;\le\; (1-p_S)\,
 \big(\norm{\wt{O}_S}_2
 + \norm{\bar{V}_T}_2\big)\\
&\;\le\; 2\,\Vmax\,(1 - p_S),
\end{align*}
where the first step is (i) and the last step bounds each convex
combination by $\max_j \norm{V_j}_2 = \Vmax$. This completes the proof.
\end{proof}

\begin{proposition}[Compensation exactness; formal version of \cref{thm:main-bandwidth}]\label{prop:compensation}
For every $u \in \R^d$,
\begin{align*}
\norm{O_S + u - O}_2 \;=\; \norm{u - (1 - p_S)\,\bar{V}_T}_2,
\end{align*}
which vanishes if and only if $u = (1 - p_S)\,\bar{V}_T$. In particular,
$\min_{u \in \R^d} \norm{O_S + u - O}_2 = 0$, attained uniquely at the
$A$-weighted tail sum $u = \sum_{j \in T} A_j V_j$.
\end{proposition}

\begin{proof}
By \Cref{prop:bias-identity:trunc},
\begin{align*}
O_S + u - O
&\;=\; u + (O_S - O)\\
&\;=\; u - (1 - p_S)\,\bar{V}_T,
\end{align*}
and a Euclidean norm vanishes exactly at the zero vector; finally
$(1-p_S)\bar V_T = \sum_{j \in T} A_j V_j$ by \Cref{def:schemes}. This
completes the proof.
\end{proof}

\begin{theorem}[A posteriori future-query lift and proxy mismatch]
\label{thm:future-query-lift}
Let \(S,T\) partition the original cache indices, with \(T\neq\varnothing\).
For a fixed future query, let \(A\) be the exact attention distribution on
\(S\cup T\), and let \(\widehat A\) be the compressed attention distribution
on \(S\cup\{\star\}\).  Fix \(r\in\Delta(T)\), store
\(V^\star=\sum_{j\in T}r_jV_j\), and define the lifted distribution
\(\bar A\) on \(S\cup T\) by
\[
\bar A_i=\widehat A_i\quad(i\in S),
\qquad
\bar A_j=\widehat A_\star r_j\quad(j\in T).
\]
Then:
\begin{enumerate}[label=(\roman*), ref=\ref*{thm:future-query-lift}(\roman*)]
\item\label{thm:future-query-lift:identity}
The compressed output is exactly the original values evaluated under the
lifted distribution:
\[
\widehat O
=\sum_{i\in S}\widehat A_iV_i+\widehat A_\star V^\star
=\sum_{j\in S\cup T}\bar A_jV_j.
\]
\item\label{thm:future-query-lift:l1}
If \(\lVert V_j\rVert_2\le V_{\max}\) for all \(j\), then
\[
\lVert O-\widehat O\rVert_2
\le V_{\max}\lVert A-\bar A\rVert_1,
\]
and the constant is worst-case tight over all such value vectors.
\item\label{thm:future-query-lift:decomp}
Put \(p=A(T)\), \(q=\widehat A_\star\),
\(a=A_T/p\) when \(p>0\), and let
\(s=A_S/(1-p)\), \(\widehat s=\widehat A_S/(1-q)\) whenever the respective
masses are nonzero.  There exist nonnegative shape residuals \(P_S,P_T\)
such that
\[
\lVert A-\bar A\rVert_1
=2|p-q|+P_S+P_T,
\]
\begin{align*}
P_S&\le \min\{1-p,1-q\}
       \lVert s-\widehat s\rVert_1,\\
P_T&\le \min\{p,q\}\lVert a-r\rVert_1.
\end{align*}
The zero-mass cases follow by taking the corresponding conditional
distribution arbitrary.
\item\label{thm:future-query-lift:retained}
If compression leaves every retained future-query logit unchanged, then
\(s=\widehat s\), \(P_S=0\), and
\begin{align*}
\lVert O-\widehat O\rVert_2
&\le V_{\max}\!\left(2|p-q|\right.\\[-2pt]
&\hspace{3.2em}\left.
+\min\{p,q\}\lVert a-r\rVert_1\right).
\end{align*}
\item\label{thm:future-query-lift:logit}
Under the same unchanged-retained-logit condition, let
\[
L_T=\log\sum_{j\in T}e^{z_j}
\quad\text{and}\quad z_\star
\]
be the exact tail log-partition and synthetic-slot logit.  Then
\[
|p-q|
\le\tanh\!\left(\frac{|L_T-z_\star|}{4}\right)
\le\frac14|L_T-z_\star|.
\]
\end{enumerate}
\end{theorem}

\begin{proof}
The lifted weights are nonnegative and sum to one.  Expanding the stored
synthetic value gives \cref{thm:future-query-lift:identity}; hence
\[
O-\widehat O=\sum_j(A_j-\bar A_j)V_j,
\]
and the triangle inequality proves
\cref{thm:future-query-lift:l1}.  For tightness, fix a unit vector \(u\) and
choose
\(
V_j=V_{\max}\operatorname{sgn}(A_j-\bar A_j)u
\)
(arbitrarily when the coefficient vanishes).

For nonnegative vectors \(x,y\), write
\[
u_+=\sum_j(x_j-y_j)_+,\qquad
u_-=\sum_j(y_j-x_j)_+.
\]
Then
\[
\lVert x-y\rVert_1
=\left|\sum_jx_j-\sum_jy_j\right|
+2\min\{u_+,u_-\}.
\]
Applying this identity on \(S\) and \(T\) defines \(P_S\) and \(P_T\) and
gives the exact decomposition in
\cref{thm:future-query-lift:decomp}.  If \(p\ge q\), the smaller one-sided
tail excess satisfies
\[
\sum_{j\in T}(qr_j-pa_j)_+
\le q\sum_{j\in T}(r_j-a_j)_+
=\frac q2\lVert r-a\rVert_1;
\]
the case \(q\ge p\) is symmetric.  Multiplying by two proves the stated
bound for \(P_T\); the retained-set bound is identical.

Unchanged retained logits preserve all within-\(S\) softmax ratios, proving
\cref{thm:future-query-lift:retained}.  Finally, with
\(Z_S=\sum_{i\in S}e^{z_i}\),
\begin{align*}
p&=\operatorname{sigmoid}(L_T-\log Z_S),\\
q&=\operatorname{sigmoid}(z_\star-\log Z_S).
\end{align*}
The maximum displacement of the sigmoid under a shift of magnitude \(d\)
is \(\tanh(d/4)\le d/4\), which proves
\cref{thm:future-query-lift:logit}.
\end{proof}

The bound is query-conditioned and a posteriori: evaluating \(p\), \(a\), or
\(L_T\) after eviction requires the omitted keys.  It therefore isolates the
two quantities a future-query compressor must estimate---synthetic tail-mass
error and conditional tail-proxy error.  \certkv{} approximates these
quantities from the observation window; the theorem itself does not make them
available before eviction.

We now show that the pair (kept values, tail aggregate) is not only
sufficient for the output but dimension-minimal for the family of all
sub-budget outputs. For $S' \subseteq S$ write $O_{S'} \coloneqq
\sum_{j \in S'} A_j V_j$ (the truncated output at budget $S'$; $O_\emptyset
= 0$).

\begin{theorem}[Factor map: sufficiency and minimality; formal version of \cref{thm:main-bandwidth}]\label{thm:factor-map}
Fix the weights $\{A_j\}_{j \in [n]}$ (all strictly positive) and the set
$S$, $|S| = k < n$, and regard $V = (V_1, \dots, V_n) \in \R^{nd}$ as the
variable. Define the linear map
\begin{align*}
\Phi &\colon \R^{nd}
  \to \R^{kd} \times \R^{d}
  = \R^{(k+1)d},\\
\Phi(V)
&\;\coloneqq\;
\left(
\begin{aligned}
&(V_j)_{j \in S},\\[-0.2em]
&\bar{V}_T
\end{aligned}
\right).
\end{align*}
Then:
\begin{enumerate}[label=(\roman*), ref=\ref*{thm:factor-map}(\roman*)]
\item\label{thm:factor-map:rank} $\Phi$ is linear and surjective, and
$\dim \ker \Phi = (n - k - 1)\,d$ exactly.
\item\label{thm:factor-map:suff} The output and all sub-budget outputs
factor through $\Phi$: with $F\big((v_j)_{j \in S}, w\big) \coloneqq
\sum_{j \in S} A_j v_j + (1 - p_S)\,w$ and $F_{S'}\big((v_j)_{j \in S},
w\big) \coloneqq \sum_{j \in S'} A_j v_j$,
\begin{align*}
O &= F(\Phi(V)),\\
&\text{and}\qquad
O_{S'} = F_{S'}(\Phi(V)),\\
&\hspace{-2em}\text{for all } V \in \R^{nd}
\text{ and all } S' \subseteq S.
\end{align*}
\item\label{thm:factor-map:min} (Minimality.) Let $\Psi \colon \R^{nd} \to
\R^m$ be any linear map for which there exist functions $G$ and
$(G_{S'})_{S' \subseteq S}$ --- not assumed linear or continuous --- with
$G(\Psi(V)) = O(V)$ and $G_{S'}(\Psi(V)) = O_{S'}(V)$ for all
$V \in \R^{nd}$. Then $m \ge \rank \Psi \ge (k+1)d$. The map $\Phi$ attains
this bound.
\end{enumerate}
\end{theorem}

\begin{proof}
\textbf{Step 1 (proof of (i): rank and surjectivity).} Linearity is immediate from \Cref{def:schemes} since
$\bar V_T$ is a fixed linear combination of $(V_j)_{j \in T}$. For
surjectivity, given any target $\big((v_j)_{j \in S}, w\big) \in
\R^{kd} \times \R^d$, choose
\begin{align*}
V_j &\;=\; v_j \quad (j \in S),\\
V_j &\;=\; w \quad (j \in T),\\
&\Longrightarrow\\[-0.2em]
\bar{V}_T
&\;=\; \frac{1}{1-p_S}
       \sum_{j \in T} A_j\, w
\;=\; w,
\end{align*}
because the coefficients $A_j/(1-p_S)$, $j \in T$, sum to one. Rank--nullity
then gives
\begin{align*}
\dim \ker \Phi
&\;=\; nd - \dim \mathrm{im}\, \Phi\\
&\;=\; nd - (k+1)d\\
&\;=\; (n - k - 1)\,d.
\end{align*}

\textbf{Step 2 (proof of (ii): factorisation).} The first identity is Eq.~\eqref{eq:o-split} rewritten
as $O = \sum_{j \in S} A_j V_j + (1-p_S)\bar V_T$; the second is the
definition of $O_{S'}$, which reads off the kept coordinates of $\Phi(V)$.

\textbf{Step 3 (proof of (iii): minimality).} We first record a kernel criterion: a linear functional
$\theta \colon \R^{nd} \to \R$ admits a (possibly nonlinear) factorisation
$\theta = \wt{G} \circ \Psi$ if and only if $\ker \Psi \subseteq \ker
\theta$. Indeed, if some $V_0 \in \ker \Psi \setminus \ker \theta$ existed,
then for every candidate $\wt G$,
\begin{align*}
\wt{G}(\Psi(V_0))
&\;=\; \wt{G}(\Psi(0)),\\
&\text{but}\\[-0.2em]
\theta(V_0) &\;\ne\; \theta(0) = 0,
\end{align*}
a contradiction; conversely, if $\ker \Psi \subseteq \ker \theta$ then
$\theta$ is constant on the fibres of $\Psi$ and $\wt G$ may be defined
fibre-wise. Now let $\mathcal{F}$ be the span of all coordinate functionals
of the recoverable quantities: for $j_0 \in S$ the singleton budget
$S' = \{j_0\}$ recovers $O_{\{j_0\}} = A_{j_0} V_{j_0}$, and subtracting
the recoverable $O_S$ from $O$ recovers the tail part, so $\mathcal{F}$
contains, for every coordinate $i \in [d]$,
\begin{align*}
\theta_{j_0, i}(V)
&\;\coloneqq\; A_{j_0} V_{j_0}^{(i)}
 \quad (j_0 \in S),\\
\theta_{T, i}(V)
&\;\coloneqq\; (1 - p_S)\,\bar{V}_T^{(i)}\\
&\;=\; \sum_{j \in T} A_j V_j^{(i)}.
\end{align*}
The $kd$ functionals $\theta_{j_0,i}$ are supported on pairwise disjoint
coordinate blocks and are nonzero ($A_{j_0} > 0$); the $d$ functionals
$\theta_{T,i}$ are supported on the $T$-blocks, disjoint from all
$S$-blocks, are nonzero ($A_j > 0$ for $j \in T \ne \emptyset$), and act on
disjoint coordinates $i$ of the $T$-blocks, so the whole family of
$(k+1)d$ functionals is linearly independent:
\begin{align*}
\dim \mathrm{span}\,\mathcal{F} \;\ge\; kd + d \;=\; (k+1)d.
\end{align*}
By the kernel criterion applied to each functional in $\mathcal{F}$,
\begin{align*}
\ker \Psi &\;\subseteq\; \bigcap_{\theta \in \mathcal{F}} \ker \theta,
\qquad\text{hence}\\
\rank \Psi
&\;=\; nd - \dim\ker\Psi\\
&\;\ge\; nd - \big(nd - (k+1)d\big)\\
&\;=\; (k+1)d,
\end{align*}
where $\dim \bigcap_{\theta} \ker \theta = nd - \dim
\mathrm{span}\,\mathcal{F} \le nd - (k+1)d$ by duality. Finally
$m \ge \rank \Psi$ always, and part (ii) shows $\Phi$ (with
$m = (k+1)d$) recovers all the required quantities. This completes the
proof.
\end{proof}

\begin{remark}[Information collapse of the tail]\label{rem:factor-map}
\Cref{thm:factor-map} formalises two statements at once. \emph{Sufficiency:}
for fixed weights, the $n - k$ tail values influence the output only
through the $d$-dimensional aggregate $\bar{V}_T$, so the
$(k+1)d$ numbers $\Phi(V)$ determine $O$ exactly; the
$(n-k-1)d$-dimensional kernel measures how much of $V$ is invisible.
\emph{Minimality:} any storage scheme that takes a linear sketch of the
values and must be able to answer the full output \emph{and} every
sub-budget truncated output $O_{S'}$, $S' \subseteq S$ (the natural anytime
guarantee for a budgeted sparse-attention pipeline), must retain at least
$(k+1)d$ dimensions --- the decoder may be arbitrary, only the sketch is
assumed linear. For a single fixed output $O$ alone, $d$ dimensions
trivially suffice (store $O$ itself), so the quantification over
sub-budgets in \Cref{thm:factor-map:min} is what the lower bound is about.
\end{remark}

\subsection{A bit-complexity floor on the dispersed trained-value tail}\label{sub:app-bitlb}

\Cref{rem:factor-map} leaves two escape hatches: it assumes a \emph{linear} sketch and
\emph{exact} reproduction. The next theorem closes both on the trained-value regime
$V=XW_V$ of \eqref{eq:score-bridge}: even an arbitrary (nonlinear) bit-sketch that need only
reproduce the single output $O$ to additive error $\eps$ must store $\Omega(d\log(1/\eps))$
bits about a dispersed tail.

\begin{theorem}[Bit-complexity bandwidth floor on a dispersed trained-value tail]\label{thm:bw-bit-lb} Fix $d,n$, accuracy $\eps\in(0,\tfrac14)$, a tail-mass floor $\beta\in(0,1]$, and a kept budget $k=o(n)$ with $n-k\ge d$. Consider the trained instantiation \eqref{eq:score-bridge}: a fixed query direction $u_i\in\R^d$, logits $\ell_j=\inner{u_i}{X_j}$, values $V_j=X_jW_V$ with $W_V\in\R^{d\times d}$ of full rank and $W_V^\top u_i\neq 0$, and $\norm{V_j}_2\le\Vmax$. A bit-compression scheme $(S,\mathrm{Sketch},\mathrm{Dec})$ consists of a kept set $S$, $|S|=k$, chosen by any rule reading only the public logits/query before the tail values; a map $\mathrm{Sketch}\colon(\R^d)^{n-k}\to\{0,1\}^m$; and an arbitrary decoder $\mathrm{Dec}\colon((V_j)_{j\in S},\sigma,A,\mathrm{query})\mapsto\widehat O\in\R^d$; only the $m$ tail bits are charged. Then for any such scheme with $m\le(1-\delta)\,(d-1)\,\log_2\!\big(\tfrac{\beta\Vmax}{4\eps}\big)$ for some $\delta\in(0,1)$, there exists an admissible trained row on the dispersed side ($1-p_S\ge\beta$) --- namely the uniform-logit instance $\ell_j=0$ (whence $V_j$ ranges freely over the token-independent subspace $U_i:=u_i^\perp W_V$ of dimension $d-1$ and the tail aggregate sweeps the radius-$(1-p_S)\Vmax\ge\beta\Vmax$ ball in $U_i$) --- on which $\norm{\widehat O-O}_2>\eps$. Equivalently $\max_{\mathrm{row}}\norm{\widehat O-O}_2>\eps$, so guaranteeing worst-case error $\le\eps$ requires $m\ge(d-1)\log_2\!\big(\tfrac{\beta\Vmax}{4\eps}\big)=\Omega(d\log(1/\eps))$. A matching quantized-aggregate scheme uses $m=O(d\log(\Vmax/\eps))$ bits, so the tight order is $m=\Theta(d\log(\Vmax/\eps))$ (lower bound at rank $d-1$, upper at rank $d$). The floor binds only the constant-mass dispersed regime; on concentrated $\gamma>1$ rows ($1-p_S\le\eta\to0$) it degrades to $\sim\eta\Vmax$ and is vacuous.\end{theorem}

\begin{proof}[Proof sketch]
Pin the hard instance to the uniform-logit witness $\ell_j=0$, so $A_j=1/n$, $p_S=k/n=o(1)$,
and each $V_j=X_jW_V$ ranges freely over $U_i:=u_i^\perp W_V$ (dimension $d-1$, since
$W_V^\top u_i\neq0$ removes one direction); the tail aggregate $\bar V_T$ then sweeps the
radius-$(1-p_S)\Vmax\ge\beta\Vmax$ ball $B\subset U_i$. By \cref{thm:factor-map},
$O=O_S+(1-p_S)\bar V_T$ with $O_S$ fixed by the kept values, so two tail configurations whose
aggregates are $>2\eps/(1-p_S)$ apart give outputs $>2\eps$ apart; a single sketch value decodes
to one $\widehat O$, which $\eps$-covers at most one. A volume/packing bound on $B$ yields a
$(\beta\Vmax/4\eps)^{\,d-1}$-packing, forcing $m\ge(d-1)\log_2(\beta\Vmax/4\eps)$ bits; quantising
one $d$-dimensional aggregate matches at $O(d\log(\Vmax/\eps))$. The floor binds only where
$1-p_S=\Theta(1)$. The metric-entropy constant inside the log is not optimised and the
lower/upper rank gap is $d-1$ vs.\ $d$ (both order-preserving).
\end{proof}

\subsection{A tight centered-moment envelope for the evicted values}\label{sub:app-certstatlb}

The bit-floor of \Cref{thm:bw-bit-lb} bounds the \emph{bandwidth} a tail
sketch must spend to reproduce the output.  Here we establish a different,
more limited statement about a particular scalar functional of the evicted
values.  Mass and centroid alone cannot adapt to the tail's realized
dispersion, whereas the centered second moment gives a sharp
Cauchy--Schwarz envelope.  This is neither a lower bound over arbitrary
constant-size summaries nor a two-sided bound on the runtime output error
$\norm{\wh O-O}_2$ of \eqref{eq:certificate-bound}; see
\Cref{rem:cert-stat-scope}.

\textbf{Setting.} Fix one head and one decode step. Let $E$ be the evicted
tail, with eviction-time weights $a_j \ge 0$ ($j \in E$) and values $v_j \in
\R^d$ obeying $\norm{v_j}_2 \le \Vmax$. Write
\begin{align}\label{eq:cert-stat-summaries}
\begin{split}
m &\;\coloneqq\; \sum_{j \in E} a_j,\\
v^\star &\;\coloneqq\; \frac{1}{m}
                  \sum_{j \in E} a_j v_j,\\
S &\;\coloneqq\; \sum_{j \in E}
                  a_j \norm{v_j - v^\star}_2^2\\
&\;=\; \sum_{j \in E} a_j \norm{v_j}_2^2
       - m\,\norm{v^\star}_2^2
       \;\ge\; 0.
\end{split}
\end{align}
We assume $m>0$; when $m=0$, set $v^\star=0$ and $S=0$, and the statements
below are trivial.  The functional bounded by the centered moment is
\begin{align}\label{eq:cert-stat-err}
\mathrm{Err}_1(C) \;\coloneqq\; \sum_{j \in E} a_j\, \norm{v_j - v^\star}_2.
\end{align}
A \emph{value-summary bound} is a summary together with a nonnegative upper
bound on \(\mathrm{Err}_1\).  Such a bound is
\emph{$(m,v^\star)$-only and valid}
on a class of tails if it is a single function
\(\beta=\beta(m,v^\star)\) satisfying
\(\beta(m,v^\star)\ge\mathrm{Err}_1(C)\) for every tail \(C\) in that
class.  Let
\(\mathcal A(m,v^\star;\Vmax)\) be the class of finite evicted tails with
prescribed mass \(m\), centroid \(v^\star\), and
\(\norm{v_j}_2\le\Vmax\).  We work with \(d\ge2\) and
\(\norm{v^\star}_2\le\Vmax\).

\begin{theorem}[Exact dispersion envelopes for two declared summary classes]\label{thm:cert-stat-lb}
In the setting of \eqref{eq:cert-stat-summaries}--\eqref{eq:cert-stat-err},
the following hold.
\begin{enumerate}[label=(\alph*), ref=\ref*{thm:cert-stat-lb}(\alph*)]
\item\label{thm:cert-stat-lb:nec} \textbf{Mass and centroid are
insufficient for dispersion adaptation.} The exact worst case over their
equivalence class is
\begin{align}\label{eq:cert-stat-mass-centroid-envelope}
\begin{split}
&\sup_{C\in\mathcal A(m,v^\star;\Vmax)}
\mathrm{Err}_1(C)\\
&\qquad=m\sqrt{\Vmax^2-\norm{v^\star}_2^2}.
\end{split}
\end{align}
Consequently every $(m,v^\star)$-only valid bound must be at least
this large.
The same \((m,v^\star)\) also admits a zero-dispersion tail with
\(\mathrm{Err}_1=0\).  Hence a bound that reads only mass and centroid cannot
shrink with the realized dispersion.
\item\label{thm:cert-stat-lb:rate} \textbf{Tightness within the
\((m,v^\star,S)\) summary class.} A centered moment is feasible exactly when
\[
0\le S\le m\bigl(\Vmax^2-\norm{v^\star}_2^2\bigr).
\]
For every such fixed \(S\),
\begin{align}\label{eq:cert-stat-envelope}
\sup_{\substack{C\in\mathcal A(m,v^\star;\Vmax)\\ S(C)=S}}
\mathrm{Err}_1(C)
\;=\; \sqrt{mS},
\end{align}
and the supremum is attained.  Therefore the smallest uniformly valid
envelope restricted to the three summaries \((m,v^\star,S)\) is
\(\sqrt{mS}\).
\end{enumerate}
\end{theorem}

\begin{proof}
\textbf{(a).} Normalize the weights by \(p_j=a_j/m\).  The variance identity
and the value-norm constraint give
\begin{align*}
\frac{1}{m}\mathrm{Err}_1(C)
&=\sum_j p_j\norm{v_j-v^\star}_2\\
&\le
\sqrt{\sum_j p_j\norm{v_j-v^\star}_2^2}\\
&=\sqrt{\sum_j p_j\norm{v_j}_2^2-\norm{v^\star}_2^2}\\
&\le\sqrt{\Vmax^2-\norm{v^\star}_2^2}.
\end{align*}
To attain the bound, let \(u\perp v^\star\) be a unit vector and put
\(\delta=\sqrt{\Vmax^2-\norm{v^\star}_2^2}\).  The two-point tail
\begin{align*}
v_1&=v^\star+\delta u,&
v_2&=v^\star-\delta u,\\
a_1&=m/2,&a_2&=m/2.
\end{align*}
has mass \(m\), centroid \(v^\star\), and
\[
\norm{v_{1,2}}_2^2
=\norm{v^\star}_2^2+\delta^2
=\Vmax^2.
\]
It is therefore admissible and attains
\eqref{eq:cert-stat-mass-centroid-envelope}.  Conversely, the one-point tail
\(v_1=v^\star,a_1=m\) has the same mass and centroid but
\(S=\mathrm{Err}_1=0\).  Thus the two summaries cannot adapt the bound to
realized dispersion.

\textbf{(b).} The same variance identity proves the necessary feasibility
condition
\[
S\le m\bigl(\Vmax^2-\norm{v^\star}_2^2\bigr).
\]
Weighted Cauchy--Schwarz gives
\begin{align*}
\mathrm{Err}_1(C)
&=\sum_{j\in E}\sqrt{a_j}
\bigl(\sqrt{a_j}\norm{v_j-v^\star}_2\bigr)\\
&\le
\sqrt{\sum_j a_j}\,
\sqrt{\sum_j a_j\norm{v_j-v^\star}_2^2}\\
=\sqrt{mS}.
\end{align*}
For equality, let \(u\perp v^\star\) be a unit vector,
\(\delta=\sqrt{S/m}\le
\sqrt{\Vmax^2-\norm{v^\star}_2^2}\), and take
\(v_{1,2}=v^\star\pm\delta u\), \(a_1=a_2=m/2\).  The witness is admissible,
has second moment \(S\), and satisfies
\(\mathrm{Err}_1=m\delta=\sqrt{mS}\).  This proves both the supremum and the
smallest valid envelope within the declared three-summary class, as well as
the sufficiency of the stated feasibility interval.
\end{proof}

\begin{remark}[Honest scope: a summary-restricted envelope, not a bound on
$\norm{\wh O-O}_2$]\label{rem:cert-stat-scope}
\Cref{thm:cert-stat-lb} compares two declared summary classes.  It gives the
exact worst-case envelope for \((m,v^\star)\), shows why those two summaries
cannot adapt to realized dispersion, and proves that \(\sqrt{mS}\) is the
tight envelope when the bound is restricted to \((m,v^\star,S)\).  It
does \emph{not} prove optimality over arbitrary constant-size summaries: for
example, a summary could store \(\mathrm{Err}_1\) itself.  Nor is it a
two-sided bound on the runtime output error
\(\norm{\wh O-O}_2\) of \eqref{eq:certificate-bound}; that error also depends
on query-induced logit changes and on the aggregate approximation, and can
vary at fixed \((m,v^\star,S)\).  The theorem therefore justifies
\(\sqrt{mS}\) as a principled, additive centered-moment feature for
allocation, not as an information-theoretically minimal statistic or a
future-query accuracy guarantee.
\end{remark}

\subsection{Tightness on flat rows}\label{sub:app-tight}

This subsection proves \cref{thm:main-tight}. The witness is the
\emph{flat row} $\ell_1 = \cdots = \ell_n$, for which $A_j = 1/n$ for
every $j$. Its scores are bounded ($|\ell_j| \le S$ for any
$S \ge |\ell_1|$, so the row lies in the \textsc{Bounded} regime of
\cref{sec:app-ladder} for every $S \ge 0$ when $\ell_j \equiv 0$), and it
is realisable in the transformer instantiation of
\Cref{fac:transformer-bounds}: identical token embeddings
$x_j = B\,\mathbf{1}_d/\sqrt d$ and projections $W_Q = W_K = W I_d$ give
$\ell_j = B^2 W^2$ for every $j$, while the values may be chosen freely
subject to $\norm{V_j}_2 \le \Vmax$ (the construction below uses only
this norm bound). Part (i) is the uniform-row witness for the
renormalised scheme; part (ii) is the all-equal-values extension that
pins down oracle selection.

\begin{proposition}[Tightness on flat rows; formal version of \cref{thm:main-tight}]\label{prop:flat-tight}
Fix $n \ge 2$, $k \in [n-1]$, and $\Vmax > 0$, and consider the flat row
$A_j = 1/n$, $j \in [n]$.
\begin{enumerate}[label=(\roman*), ref=\ref*{prop:flat-tight}(\roman*)]
\item\label{prop:flat-tight:renorm} (\emph{Weight-based selection;
renormalised and truncated schemes.}) Let $S_k$ be the kept set returned
by any selection rule that reads only the weights (e.g.\ top-$k$ with
deterministic tie-breaking); on the flat row $S_k$ is a fixed $k$-subset,
and after relabelling indices we may take $S_k = \{1, \dots, k\}$.
Choosing $V_j \coloneqq -\Vmax\, e_1$ for $j \in S_k$ and $V_j \coloneqq
\Vmax\, e_1$ for $j \notin S_k$ (so $\norm{V_j}_2 = \Vmax$ for all $j$),
\begin{align*}
\norm{\wt O_{S_k} - O}_2
&\;=\; \frac{2\,\Vmax\,(n-k)}{n},\\
\norm{O_{S_k} - O}_2
&\;=\; \frac{\Vmax\,(n-k)}{n}.
\end{align*}
\item\label{prop:flat-tight:oracle} (\emph{All values equal; every kept
set, oracle selection included.}) Choosing instead $V_j \coloneqq
\Vmax\, e_1$ for every $j \in [n]$, \emph{every} kept set $S$ with $|S|
= k$ --- in particular the error-minimising oracle subset of
\Cref{thm:oracle-existence} below --- satisfies
\begin{align*}
\norm{O_S - O}_2
&\;=\; (1 - p_S)\,\Vmax\\
&\;=\; \frac{\Vmax\,(n-k)}{n}.
\end{align*}
\item\label{prop:flat-tight:budget} (\emph{Budget consequence.}) For a
target error $\eta\,\Vmax$ with $\eta \in (0,1)$: instance (i) forces
$k \ge n(1 - \eta/2)$ for the renormalised scheme under weight-based
selection, and instance (ii) forces $k \ge n(1 - \eta)$ for the truncated
scheme under \emph{every} selection rule, the oracle included. In both
cases $k = \Theta(n)$, so the linear \textsc{Free}--\textsc{Bounded} budgets of
\cref{tab:main-ladder} are not an artifact of analysis.
\end{enumerate}
\end{proposition}

\begin{proof}
The strategy is direct computation on the flat row; throughout, $p_S =
\sum_{j \in S} A_j = k/n$ for every $|S| = k$.

\textbf{Step 1 (proof of (i): two-sided values witness).} The selection rule sees only the flat weight vector,
so its output is a $k$-set fixed before the values are chosen; relabel
so that $S_k = \{1, \dots, k\}$. With the stated values,
\begin{align*}
O &\;=\; \frac{1}{n}\Big[
k\,(-\Vmax e_1) + (n-k)\,\Vmax e_1
\Big]\\
&\;=\; \frac{n - 2k}{n}\,\Vmax\, e_1,\\
\wt O_{S_k}
&\;=\; \frac{1/n}{k/n} \cdot k \,(-\Vmax e_1)
\;=\; -\Vmax\, e_1,
\end{align*}
so $\norm{\wt O_{S_k} - O}_2 = \Vmax \big| \tfrac{n-2k}{n} - (-1) \big|
= \tfrac{2(n-k)}{n} \Vmax$. For the truncated scheme, $O_{S_k} = p_S\,
\wt O_{S_k} = -\tfrac{k}{n} \Vmax e_1$, hence $\norm{O_{S_k} - O}_2 =
\Vmax\big| \tfrac{n-2k}{n} + \tfrac{k}{n} \big| = \tfrac{n-k}{n}\Vmax$.

\textbf{Step 2 (proof of (ii): equal-values witness).} With all values equal, $O = \Vmax e_1$ and, for every
$|S| = k$, $O_S = p_S\, \Vmax e_1 = \tfrac{k}{n}\Vmax e_1$; therefore
$\norm{O_S - O}_2 = (1 - p_S)\Vmax = \tfrac{n-k}{n}\Vmax$,
independently of which $k$-subset is kept --- the minimum over subsets
equals the common value.

\textbf{Step 3 (proof of (iii): budget rearrangement).} Requiring $\tfrac{2(n-k)}{n}\Vmax \le \eta\Vmax$ in
(i) rearranges to $k \ge n(1 - \eta/2)$; requiring
$\tfrac{n-k}{n}\Vmax \le \eta \Vmax$ in (ii) rearranges to $k \ge
n(1-\eta)$. This completes the proof.
\end{proof}

\begin{remark}[Sharpness of the two constants]\label{rem:flat-tight-sharp}
Each constant in \Cref{prop:flat-tight} is extremal for its scheme. For
the truncated scheme, on \emph{every} row the top-$k$ set satisfies
$p(k) \ge k/n$ (the $k$ largest entries of a probability vector carry at
least a $k/n$ fraction of its mass), so the oracle truncated error never
exceeds $(1 - p(k))\,\Vmax \le \Vmax (n-k)/n$: instance
\ref*{prop:flat-tight}(ii) is exactly worst-case for oracle selection,
and the factor $2$ in the headline rate of \cref{thm:main-tight} is
contributed by the renormalised scheme of part (i), whose error can
reach the mass ceiling $2\Vmax(1-p_S)$ of
\Cref{prop:bias-identity:mass}. Conversely, no single witness can pin
\emph{all} scheme--selector combinations at the factor-$2$ level
simultaneously: on instance (ii) the renormalised error vanishes for
every $S$. On flat rows the mass barrier thus binds every
compensation-free scheme --- at constant $2$ for renormalisation under
weight-based selection and at the extremal constant $1$ for truncation
under arbitrary (oracle) selection --- and compensation
(\Cref{prop:compensation}) is the only exit.
\end{remark}

\subsection{Oracle subsets: the cancellation characterisation}\label{sub:app-oracle}

This subsection proves parts (i) and (iii) of \cref{thm:main-oracle};
parts (ii) and (iv) are \Cref{prop:proxy-uncomp} below and
\Cref{prop:flat-tight}\ref*{prop:flat-tight:oracle} above. The top-$k$
rule keeps the indices with the largest weights $A_j$ and minimises the
dropped softmax mass $1 - p_S$. But the truncated error depends on $V$
as well as $A$: the dropped tokens contribute $\sum_{j \notin S} A_j
V_j$, and this sum can have small norm either because each $A_j V_j$ is
individually small (which top-$k$ targets) \emph{or} because the $A_j
V_j$'s cancel in direction (which top-$k$ ignores). This raises a
natural question, posed in the MagicPIG work \cite{chen2025magicpig}:
\emph{is there a $k$-subset that beats top-$k$?} We answer in the
affirmative and characterise the property. Throughout, note that every
strictly positive probability vector is a softmax row (take $\ell_j =
\log A_j$, so $Z = 1$); the examples below are stated directly in terms
of the weights.

\begin{definition}[Truncated error of a kept set]\label{def:trunc-error}
For a subset $\mathcal S \subseteq [n]$, the truncated approximation
error is
\begin{align}\label{eq:trunc-err}
E(\mathcal S) \;\coloneqq\; \norm{O - O_{\mathcal S}}_2
\;=\; \Big\| \sum_{j \notin \mathcal S} A_j V_j \Big\|_2,
\end{align}
the equality by \Cref{prop:bias-identity:trunc}.
\end{definition}

\begin{theorem}[Optimal $k$-subset for truncated sparse attention; formal version of \cref{thm:main-oracle}]\label{thm:oracle-existence}
Fix any $k \in [n]$. Let
\begin{align}\label{eq:oracle-S-star}
\mathcal S^\star \;\in\; \arg\min_{|\mathcal S| = k} E(\mathcal S),
\end{align}
and let $\mathcal S^{\text{top-}k}$ be the top-$k$ subset by $A_j$ (with
arbitrary tie-breaking). Then:
\begin{enumerate}[label=(\roman*), ref=\ref*{thm:oracle-existence}(\roman*)]
\item\label{thm:oracle-existence:opt} \textbf{Optimality.}
$E(\mathcal S^\star) \le E(\mathcal S^{\text{top-}k})$.
\item\label{thm:oracle-existence:strict} \textbf{Strict improvement is
achievable.} For every $k \ge 2$, there exist values $V_1, \dots, V_n$
and a weight row $(A_j)_{j\in[n]}$ such that the inequality in (i) is
strict; in fact, one can have $E(\mathcal S^\star) = 0$ while
$E(\mathcal S^{\text{top-}k})$ is bounded away from zero.
\item\label{thm:oracle-existence:aligned} \textbf{Sufficient condition
for top-$k$ optimality.} If $V_j = c_j u$ for a fixed unit vector $u \in
\R^d$ and scalars $c_j \ge 0$, then $\mathcal S^\star$ is the top-$k$
subset by $A_j c_j$. In particular, when $c_j \equiv c$ is constant (all
values aligned with the same magnitude), $\mathcal S^\star = \mathcal
S^{\text{top-}k}$.
\end{enumerate}
\end{theorem}

\begin{proof}
The proof has three independent steps.

\medskip\noindent\textbf{Step 1 (proof of (i): feasibility).}
Immediate from the definition Eq.~\eqref{eq:oracle-S-star}: $\mathcal
S^{\text{top-}k}$ is one element of the feasible set $\{\mathcal S
\subseteq [n] : |\mathcal S| = k\}$ over which $E$ is minimised at
$\mathcal S^\star$.

\medskip\noindent\textbf{Step 2 (proof of (ii): explicit witness).}
Take $n = 4$, $k = 2$, $d = 2$, weight row $A = (0.4,\, 0.4,\, 0.15,\,
0.05)$, and values
\begin{align*}
V_1 &= e_1, & V_2 &= -e_1,\\
V_3 &= e_2, & V_4 &= -10\, e_2,
\end{align*}
where $e_1, e_2$ is the standard basis of $\R^2$. Then $O = 0.4\, e_1 -
0.4\, e_1 + 0.15\, e_2 - 0.5\, e_2 = -0.35\, e_2$.
\begin{enumerate}[label=(\roman*),leftmargin=2em,itemsep=0.3em]
\item \textbf{Top-$2$ subset.} $\mathcal S^{\text{top-}k} =
\{1, 2\}$ (the two largest weights). The dropped mass is $\sum_{j \notin
\{1,2\}} A_j V_j = 0.15\, e_2 - 0.5\, e_2 = -0.35\, e_2$, so
$E(\{1,2\}) = 0.35$.
\item \textbf{An alternative $2$-subset.} $\mathcal S =
\{3, 4\}$: the dropped mass is $\sum_{j \notin \{3,4\}} A_j V_j = 0.4\,
e_1 - 0.4\, e_1 = 0$, so $E(\{3,4\}) = 0$.
\end{enumerate}
Hence $E(\mathcal S^\star) \le E(\{3,4\}) = 0 < 0.35 = E(\mathcal
S^{\text{top-}k})$. The construction extends to any $k \ge 2$ by padding
additional zero-contributing tokens.

\medskip\noindent\textbf{Step 3 (proof of (iii): aligned values).}
If $V_j = c_j u$, then $E(\mathcal S) = \norm{u \sum_{j \notin \mathcal
S} A_j c_j}_2 = \sum_{j \notin \mathcal S} A_j c_j$ (since $A_j c_j \ge
0$). Minimising this over $|\mathcal S| = k$ is equivalent to maximising
$\sum_{j \in \mathcal S} A_j c_j$, which by definition picks the top-$k$
indices ranked by $A_j c_j$. When $c_j \equiv c$, the ranking by $A_j
c_j$ coincides with the ranking by $A_j$. This completes the proof.
\end{proof}

The witness in the proof of
\Cref{thm:oracle-existence}\ref*{thm:oracle-existence:strict} reveals
the mechanism: $\mathcal S^\star = \{3,4\}$ wins not because $\{1,2\}$
have small attention weights (they have the largest), but because the
dropped vectors $A_1 V_1$ and $A_2 V_2$ cancel each other exactly. We
formalise this geometric quantity.

\begin{definition}[Cancellation index]\label{def:cancellation-index}
For a subset $\mathcal S \subseteq [n]$, define the \emph{(value-mass)
tail} and the \emph{cancellation index} of $\mathcal S$ by
\begin{align}\label{eq:rho-def}
M(\mathcal S)
&\;\coloneqq\; \sum_{j \notin \mathcal S}
A_j \norm{V_j}_2,\\
\rho(\mathcal S)
&\;\coloneqq\; 1 -
\frac{E(\mathcal S)}{M(\mathcal S)}
\ \in\ [0, 1].
\notag
\end{align}
with the convention $\rho(\mathcal S) = 1$ when $M(\mathcal S) = 0$.
\end{definition}

The cancellation index $\rho(\mathcal S) \in [0,1]$ measures the
directional alignment of the dropped vectors: $\rho(\mathcal S) = 0$
when all dropped $A_j V_j$ point in the same direction (no cancellation,
worst case), and $\rho(\mathcal S) = 1$ when they sum to zero (perfect
cancellation, best case).

\begin{lemma}[Two-factor decomposition of the truncated error; formal version of \cref{thm:main-oracle}]\label{lem:oracle-decomp}
For every $\mathcal S \subseteq [n]$,
\begin{align}\label{eq:oracle-decomp}
E(\mathcal S) \;=\; \big( 1 - \rho(\mathcal S) \big) \cdot M(\mathcal S).
\end{align}
\end{lemma}

\begin{proof}
Direct rearrangement of Eq.~\eqref{eq:rho-def}: $E(\mathcal S) =
M(\mathcal S) - \rho(\mathcal S) M(\mathcal S) = (1 - \rho(\mathcal S))
M(\mathcal S)$. The decomposition is well-defined: by the triangle
inequality, $E(\mathcal S) \le M(\mathcal S)$, hence $\rho(\mathcal S)
\ge 0$; and $E(\mathcal S) \ge 0$ gives $\rho(\mathcal S) \le 1$. This
completes the proof.
\end{proof}

\Cref{lem:oracle-decomp} factorises the truncated error into two
interpretable axes: the \emph{tail mass} $M(\mathcal S)$ (the total
``size'' of dropped contributions, weighted by attention and value
norm) and the \emph{cancellation} $\rho(\mathcal S)$ (the directional
alignment among the dropped vectors). Three concrete subset rules sit on
this two-axis plane: (i) top-$k$ by $A_j$, which minimises the
dropped softmax mass and ignores $\norm{V_j}_2$ and $\rho$; (ii)
top-$k$ by $A_j \norm{V_j}_2$, which minimises $M(\mathcal S)$ and still
ignores $\rho$; and (iii) the optimal $\mathcal S^\star$, which
jointly minimises the product $M(\mathcal S)(1 - \rho(\mathcal S))$ and
is combinatorial.

\begin{theorem}[Property characterising $\mathcal S^\star$; formal version of \cref{thm:main-oracle}]\label{thm:oracle-property}
Let $\mathcal S^\star$ be any optimum of Eq.~\eqref{eq:oracle-S-star}
and $\mathcal S^{(2)}$ the top-$k$ subset by $A_j \norm{V_j}_2$. Then:
\begin{enumerate}[label=(\roman*), ref=\ref*{thm:oracle-property}(\roman*)]
\item\label{thm:oracle-property:sandwich} \textbf{Sandwich.}
$E(\mathcal S^\star) \le E(\mathcal S^{(2)}) \le M(\mathcal S^{(2)})
\le M(\mathcal S^{\text{top-}k})$.
\item\label{thm:oracle-property:strict} \textbf{Strict improvement is
driven by cancellation.} $E(\mathcal S^\star) < E(\mathcal S^{(2)})$ if
and only if $(1 - \rho(\mathcal S^\star))\, M(\mathcal S^\star) < (1 -
\rho(\mathcal S^{(2)}))\, M(\mathcal S^{(2)})$. Since $M(\mathcal
S^\star) \ge M(\mathcal S^{(2)})$ by definition of $\mathcal S^{(2)}$,
this is equivalent to
$\rho(\mathcal S^\star) > 1 - \dfrac{M(\mathcal S^{(2)})}{M(\mathcal
S^\star)}\,\big(1 - \rho(\mathcal S^{(2)})\big)$.
\item\label{thm:oracle-property:local} \textbf{Local-swap optimality.}
$\mathcal S^\star$ admits no improving $1$-swap: for every
$j_{\text{in}} \in \mathcal S^\star$ and $j_{\text{out}} \notin \mathcal
S^\star$, the swapped subset $\mathcal S' = (\mathcal S^\star \setminus
\{j_{\text{in}}\}) \cup \{j_{\text{out}}\}$ satisfies $E(\mathcal S')
\ge E(\mathcal S^\star)$.
\end{enumerate}
\end{theorem}

\begin{proof}
The three claims are independent.

\medskip\noindent\textbf{Step 1 (proof of (i): sandwich).}
By \Cref{thm:oracle-existence}\ref*{thm:oracle-existence:opt},
$E(\mathcal S^\star) \le E(\mathcal S^{(2)})$. By
\Cref{lem:oracle-decomp}, $E(\mathcal S^{(2)}) \le M(\mathcal S^{(2)})$.
By definition of $\mathcal S^{(2)}$, $M(\mathcal S^{(2)}) \le M(\mathcal
S)$ for every $|\mathcal S| = k$, in particular $M(\mathcal S^{(2)})
\le M(\mathcal S^{\text{top-}k})$.

\medskip\noindent\textbf{Step 2 (proof of (ii): rearrangement).}
By \Cref{lem:oracle-decomp}, $E(\mathcal S^\star) < E(\mathcal
S^{(2)})$ iff $(1 - \rho(\mathcal S^\star)) M(\mathcal S^\star) < (1 -
\rho(\mathcal S^{(2)})) M(\mathcal S^{(2)})$. Dividing both sides by
$M(\mathcal S^\star) > 0$ and rearranging,
\begin{align*}
1 - \rho(\mathcal S^\star)
\;<\; \frac{M(\mathcal S^{(2)})}{M(\mathcal S^\star)}\,
\big( 1 - \rho(\mathcal S^{(2)}) \big),
\\[-0.2em]
\text{i.e.}\qquad
\rho(\mathcal S^\star)
\;>\; 1 - \frac{M(\mathcal S^{(2)})}{M(\mathcal S^\star)}\,
\big( 1 - \rho(\mathcal S^{(2)}) \big).
\end{align*}

\medskip\noindent\textbf{Step 3 (proof of (iii): local optimality).}
If some $1$-swap strictly decreased $E$, then $\mathcal S^\star$ would
not minimise $E$ over $|\mathcal S| = k$, contradicting
Eq.~\eqref{eq:oracle-S-star}. This completes the proof.
\end{proof}

\begin{remark}[On the scope of \ref*{thm:oracle-property}(iii): necessary
but not sufficient]\label{rem:oracle-local}
\Cref{thm:oracle-property}\ref*{thm:oracle-property:local} is a
\emph{necessary} condition for optimality of $\mathcal S^\star$: every
$\mathcal S$ admitting an improving $1$-swap is sub-optimal. It is
\emph{not sufficient}: $E$ is a non-convex combinatorial objective on
$\binom{n}{k}$ subsets, and local minima of the $1$-swap graph need not
be global minima --- \Cref{rem:negative} below exhibits an explicit
instance where $1$-swap local search stalls at twice the optimum. We
therefore present (iii) as a \emph{local} characterisation of $\mathcal
S^\star$; the global characterisation is by definition
(Eq.~\eqref{eq:oracle-S-star}). A matching sufficient condition for
global optimality is open in this regime: it would require additional
structural properties of $E$ (e.g., sub-modularity or a
convex-relaxation hierarchy) beyond the multiplicative factorisation $E
= (1 - \rho) M$ of \Cref{lem:oracle-decomp}.
\Cref{thm:oracle-existence}\ref*{thm:oracle-existence:strict}
accordingly gives only existence --- strictly beating top-$k$ --- not a
polynomial-time algorithm.
\end{remark}

\begin{remark}[The oracle gap under exchangeable tails --- open]\label{rem:oracle-collapse}
The factor map of \Cref{thm:factor-map} shows that, for fixed weights,
the only feature of the dropped tail visible to the output is its
aggregate $\sum_{j \notin \mathcal S} A_j V_j$, so the cancellation that
$\mathcal S^\star$ exploits is a property of that single $d$-dimensional
aggregate. It is tempting to conclude that under the generic i.i.d.\
score regimes \textsc{Iid}--\textsc{Scaled} of \cref{sec:app-ladder},
with values independent of the scores, the oracle gap to the
value-weighted ranking (rule (ii) above) becomes asymptotically
negligible. We do \emph{not} claim this, and we suspect it is false: the
oracle error $E(\mathcal S^\star) = \min_{|\mathcal S| = k}
\norm{\sum_{j \notin \mathcal S} A_j V_j}_2$ is a \emph{minimum over the
$\binom{n}{k}$ correlated dropped aggregates}, which the per-set
sufficiency of \Cref{thm:factor-map} (a fixed-$\mathcal S$ statement)
does not control --- distinct near-optimal subsets drop distinct token
sets, hence distinct aggregates. For centred values in the dispersed
regime $n - k \gg d$, the combinatorial freedom to balance $n - k$
vectors in $\R^d$ should let the oracle drive its dropped aggregate well
below the value-weighted one, keeping the gap open. What \emph{is}
established is the sandwich $E(\mathcal S^\star) \le E(\mathcal S^{(2)})
\le M(\mathcal S^{(2)})$
(\Cref{thm:oracle-property}\ref*{thm:oracle-property:sandwich}: the
value-weighted rule is proxy-optimal and the oracle can only improve on
it) and the strict-improvement witness of
\Cref{thm:oracle-existence}\ref*{thm:oracle-existence:strict}.
Quantifying the oracle gap under exchangeable values --- in particular
deciding whether it vanishes or persists --- is left as an open problem.
\end{remark}
\subsection{Selection rules: proxy optimality and its limits}\label{sub:app-selection}

This subsection is deterministic: fix a softmax row $(A_j)_{j \in [n]}$ and
values $(V_j)_{j \in [n]}$ as in \Cref{def:softmax-row}, and a budget $1
\le k < n$. We compare \emph{selection rules}: procedures that choose the
kept set $S$, $|S| = k$. Throughout, recall from
\Cref{prop:bias-identity:trunc} that the truncation error is exactly
$\norm{O_S - O}_2 = \norm{\sum_{j \notin S} A_j V_j}_2$. Note that every
strictly positive probability vector is a softmax row (take $\ell_j =
\log A_j$, so $Z = 1$); the examples below are stated directly in terms
of the weights.

\subsubsection{Two proxy-optimal rankings}

\begin{proposition}[Proxy optimality without compensation]\label{prop:proxy-uncomp}
Define the scores $s_j \coloneqq A_j \norm{V_j}_2$ and the proxy
$M(S) \coloneqq \sum_{j \notin S} s_j$. Then:
\begin{enumerate}[label=(\roman*), ref=\ref*{prop:proxy-uncomp}(\roman*)]
\item\label{prop:proxy-uncomp:bound} $\norm{O_S - O}_2 \le M(S)$ for
every $S$;
\item\label{prop:proxy-uncomp:opt} a set $S$ with $|S| = k$ minimises
$M$ over all $k$-subsets if and only if $\min_{j \in S} s_j \ge
\max_{j \notin S} s_j$; in particular every set of indices of $k$
largest scores is a minimiser;
\item\label{prop:proxy-uncomp:aligned} (aligned case) if $V_j = c_j u$
for a single unit vector $u$ and scalars $c_j \ge 0$, then
$\norm{O_S - O}_2 = M(S)$ for every $S$, so the top-$k$-by-$s_j$ rule
minimises the \emph{true} truncation error.
\end{enumerate}
\end{proposition}

\begin{proof}
The strategy is the triangle inequality for (i), an exchange argument for
(ii), and exact computation of the norm of a conic combination for (iii).
For (i), by \Cref{prop:bias-identity:trunc} and the triangle inequality,
\begin{align}\label{eq:proxy-bound}
\norm{O_S - O}_2
&\;=\; \Big\| \sum_{j \notin S}
 A_j V_j \Big\|_2\\
&\;\le\; \sum_{j \notin S}
 A_j \norm{V_j}_2\notag\\
&\;=\; M(S).
\notag
\end{align}
For (ii), since $M(S) = \sum_{j \in [n]} s_j - \sum_{j \in S} s_j$,
minimising $M$ over $k$-subsets is equivalent to maximising $\Sigma(S)
\coloneqq \sum_{j \in S} s_j$. If $\min_{j \in S} s_j \ge \max_{j \notin
S} s_j$, then for any other $k$-subset $S'$, pairing the sets' symmetric
differences (which satisfy $|S \setminus S'| = |S' \setminus S|$),
\begin{align*}
\Sigma(S) - \Sigma(S')
&\;=\; \sum_{j \in S \setminus S'} s_j
 - \sum_{j \in S' \setminus S} s_j\\
&\;\ge\; |S \setminus S'| \cdot
\Big( \min_{j \in S} s_j
 - \max_{j \notin S} s_j \Big)\\
&\;\ge\; 0,
\end{align*}
so $S$ is a maximiser of $\Sigma$, i.e.\ a minimiser of $M$. Conversely,
if $s_l > s_j$ for some $j \in S$, $l \notin S$, then the swap $S'' =
(S \setminus \{j\}) \cup \{l\}$ has $\Sigma(S'') - \Sigma(S) = s_l - s_j
> 0$, so $S$ is not a minimiser. Sets of $k$ largest scores satisfy the
criterion by construction. For (iii), with $V_j = c_j u$, $c_j \ge 0$,
$\norm{u}_2 = 1$,
\begin{align*}
\Big\| \sum_{j \notin S} A_j V_j \Big\|_2
&\;=\; \Big( \sum_{j \notin S}
A_j c_j \Big) \norm{u}_2\\
&\;=\; \sum_{j \notin S} A_j c_j\\
&\;=\; \sum_{j \notin S}
A_j \norm{V_j}_2\\
&\;=\; M(S),
\end{align*}
because the coefficients $A_j c_j$ are nonnegative and
$\norm{V_j}_2 = c_j$. Combining with (ii), the proxy minimiser minimises
the true error. This completes the proof.
\end{proof}

\begin{proposition}[Proxy optimality with compensation]\label{prop:proxy-comp}
Fix an estimate $\wh V \in \R^d$ and compensate with $u_S \coloneqq
\big( \sum_{j \notin S} A_j \big) \wh V = (1 - p_S)\,\wh V$. Define the
scores $s_j' \coloneqq A_j \norm{V_j - \wh V}_2$ and the proxy
$M_{\wh V}(S) \coloneqq \sum_{j \notin S} s_j'$. Then:
\begin{enumerate}[label=(\roman*), ref=\ref*{prop:proxy-comp}(\roman*)]
\item\label{prop:proxy-comp:identity} the compensated error is exactly
\begin{align*}
\norm{O_S + u_S - O}_2
&\;=\; \Big\| \sum_{j \notin S}
A_j \big( V_j - \wh V \big) \Big\|_2\\
&\;\le\; M_{\wh V}(S),
\end{align*}
the inequality by the triangle inequality;
\item\label{prop:proxy-comp:opt} $M_{\wh V}$ is minimised over
$k$-subsets exactly by the sets with $\min_{j \in S} s_j' \ge
\max_{j \notin S} s_j'$ (top-$k$ by $s_j'$);
\item\label{prop:proxy-comp:aligned} if $V_j - \wh V = c_j u$ with
$c_j \ge 0$ and $\norm{u}_2 = 1$, the inequality in (i) is an equality
and the rule is exactly optimal for the true compensated error.
\end{enumerate}
\end{proposition}

\begin{proof}
The strategy is to reduce to the previous proposition with shifted
values. By \Cref{prop:bias-identity:trunc} and $(1-p_S) = \sum_{j \notin
S} A_j$,
\begin{align*}
O_S + u_S - O
&\;=\; \Big( \sum_{j \notin S} A_j \Big)
\wh V - \sum_{j \notin S} A_j V_j\\
&\;=\; - \sum_{j \notin S}
A_j \big( V_j - \wh V \big),
\end{align*}
and taking norms plus the triangle inequality proves (i). For (ii) and
(iii), substitute the shifted values: with
\begin{align*}
V_j' &\;\coloneqq\; V_j - \wh V,\\
s_j' &\;=\; A_j \norm{V_j'}_2,\\
M_{\wh V}(S)
&\;=\; \sum_{j \notin S}
A_j \norm{V_j'}_2,
\end{align*}
the proxy $M_{\wh V}$ is exactly the proxy $M$ of
\Cref{prop:proxy-uncomp} for the value family $(V_j')_j$, so
\Cref{prop:proxy-uncomp:opt} gives (ii) verbatim, and
\Cref{prop:proxy-uncomp:aligned} applied to $V_j' = c_j u$ gives the
equality
\begin{align*}
\Big\| \sum_{j \notin S} A_j V_j' \Big\|_2 \;=\; M_{\wh V}(S)
\qquad \text{for every } S,
\end{align*}
which is (iii). This completes the proof.
\end{proof}

\subsubsection{An explicit separation}

Per the scheme class of \Cref{def:schemes}, both rules are
runtime-implementable; they genuinely differ, and the compensated
ranking can strictly win. Following our convention for the estimate, we
take $\wh V$ to be the unweighted mean of \emph{all} $n$ values --- a
fixed, pre-selection statistic.

\begin{proposition}[Separation of the two rankings]\label{prop:separation}
There is an instance with $n = 4$, $k = 2$, $d = 1$ on which, with
$\wh V \coloneqq \tfrac1n \sum_{j=1}^n V_j$:
\begin{enumerate}[label=(\roman*), ref=\ref*{prop:separation}(\roman*)]
\item the ranking by $s_j = A_j|V_j|$ and the ranking by $s_j' =
A_j|V_j - \wh V|$ select different sets $S_{\mathrm{unc}} \ne
S_{\mathrm{comp}}$;
\item compensating both with $u_S = (1-p_S)\wh V$, the
$s'$-selected set has strictly smaller error:
\begin{align*}
\left\|
\begin{aligned}
&O_{S_{\mathrm{comp}}}
+ u_{S_{\mathrm{comp}}}\\[-0.2em]
&\hspace{1.2em}- O
\end{aligned}
\right\|_2
&\;=\; 2
\;<\; \frac{11}{4}\\
&\;=\;
\left\|
\begin{aligned}
&O_{S_{\mathrm{unc}}}
+ u_{S_{\mathrm{unc}}}\\[-0.2em]
&\hspace{1.2em}- O
\end{aligned}
\right\|_2.
\end{align*}
\end{enumerate}
\end{proposition}

\begin{proof}
The strategy is exhaustive arithmetic on a hand-built instance. Take
\begin{align*}
A &\;=\; \Big(
\frac{2}{5},\ \frac{3}{10},\
\frac{1}{5},\ \frac{1}{10}\Big),\\
V &\;=\; \big( 10,\ 10,\ -10,\ 0 \big),\\
\wh V
&\;=\; \frac{10 + 10 - 10 + 0}{4}
\;=\; \frac{5}{2}.
\end{align*}
The two score vectors are
\begin{align*}
\big( s_j \big)_{j}
&\;=\; \big( A_j |V_j| \big)_j\\
&\;=\; \Big( 4,\ 3,\ 2,\ 0 \Big),\\
\big( s_j' \big)_j
&\;=\; \big( A_j |V_j - \tfrac52| \big)_j\\
&\;=\; \Big( 3,\ \frac94,\
\frac52,\ \frac14 \Big),
\end{align*}
where, explicitly, $s_3' = \tfrac15 \cdot \tfrac{25}{2} = \tfrac52$ and
$s_2' = \tfrac{3}{10}\cdot\tfrac{15}{2} = \tfrac94$. Hence the two
top-$2$ sets are
\begin{align*}
S_{\mathrm{unc}}
&\;=\; \{1, 2\}
\quad (4 > 3 > 2 > 0),\\
S_{\mathrm{comp}}
&\;=\; \{1, 3\}
\quad \Big(3 > \frac52
> \frac94 > \frac14\Big),
\end{align*}
which differ, proving (i). For (ii), by
\Cref{prop:proxy-comp:identity} the compensated error of a kept set $S$
is $\big| \sum_{j \notin S} A_j ( V_j - \tfrac52 ) \big|$, and the
residuals $A_j(V_j - \tfrac52)$ are
\begin{multline*}
\Big(
\frac{2}{5}\cdot\frac{15}{2},\
\frac{3}{10}\cdot\frac{15}{2},\
\frac15 \cdot \Big({-\frac{25}{2}}\Big),\\
\frac{1}{10}\cdot\Big({-\frac52}\Big)
\Big)
\;=\; \Big(
3,\ \frac94,\ -\frac52,\ -\frac14
\Big),
\end{multline*}
so the two errors evaluate to
\begin{align*}
\left\|
\begin{aligned}
&O_{S_{\mathrm{unc}}}
+ u_{S_{\mathrm{unc}}}\\[-0.2em]
&\hspace{1.2em}- O
\end{aligned}
\right\|_2
&\;=\; \Big| -\frac52 - \frac14 \Big|
\;=\; \frac{11}{4},\\
\left\|
\begin{aligned}
&O_{S_{\mathrm{comp}}}
+ u_{S_{\mathrm{comp}}}\\[-0.2em]
&\hspace{1.2em}- O
\end{aligned}
\right\|_2
&\;=\; \Big| \frac94 - \frac14 \Big|
\;=\; 2,
\end{align*}
and $2 < 11/4$. This completes the proof.
\end{proof}

\begin{remark}[The combinatorial gap: what the propositions do
\emph{not} claim]\label{rem:negative}
\Cref{prop:proxy-uncomp,prop:proxy-comp} optimise \emph{proxies}: upper
bounds that ignore cancellation among the dropped residuals. Minimising
the \emph{true} error $E(S) \coloneqq \norm{\sum_{j\notin S} A_j (V_j -
\wh V)}_2$ over the $\binom{n}{k}$ kept sets is a combinatorial
problem for which we claim neither a polynomial-time algorithm nor
NP-hardness. One concrete obstruction: $1$-swap local optimality is
necessary but not sufficient for global optimality. Take $n = 4$, $k =
2$, $d = 2$, $A_j = \tfrac14$ for all $j$, $\wh V = 0$, and $V_j = 4
u_j$ with
\begin{align*}
u_1 &= (1, 0),
&u_2 &= \Big({-1}, \frac{1}{10}\Big),\\
u_3 &= (0, 1),
&u_4 &= \Big(\frac15, -1\Big),
\end{align*}
so that $E(S) = \norm{\sum_{j \notin S} u_j}_2$. The kept set $S =
\{1,2\}$ has $E = \norm{u_3 + u_4}_2 = \norm{(\tfrac15, 0)}_2 =
\tfrac15$, and each of its four $1$-swaps is strictly worse:
\begin{align*}
\norm{u_2 + u_4}_2
&= \frac{\sqrt{145}}{10},
&\norm{u_2 + u_3}_2
&= \frac{\sqrt{221}}{10},\\
\norm{u_1 + u_4}_2
&= \frac{\sqrt{61}}{5},
&\norm{u_1 + u_3}_2
&= \sqrt{2},
\end{align*}
all at least $\sqrt{145}/10 > 1 > \tfrac15$ (the four values are the
dropped-pair sums after swapping exactly one kept index). Yet the
$2$-swap $S = \{3,4\}$ achieves $E = \norm{u_1 + u_2}_2 =
\norm{(0,\tfrac{1}{10})}_2 = \tfrac1{10} < \tfrac15$. Thus local search
over single swaps can stall at twice the optimum, and the aligned case
of \Cref{prop:proxy-uncomp:aligned} (where the proxy is exact) is the
honest boundary of what the ranking rules guarantee.
\end{remark}
\subsection{A deterministic output-error bound for clustered tail summaries}\label{sub:app-certificate}

This subsection is fully deterministic and self-contained. It quantifies a
compensation scheme that stores, instead of the exact tail
(\Cref{prop:compensation} would require the exact aggregate $\sum_{j\in
T} A_j V_j$), only \emph{per-cluster summaries} of the tail --- and it
does so with an error bound every quantity of which is computable at
runtime from the stored summaries.

\textbf{Setting.} Fix a logit vector $(\ell_j)_{j\in[n]}$ and values
$(V_j)_{j\in[n]}$ as in \Cref{def:softmax-row}, a kept set $S$, and a
partition of the tail $T = [n] \setminus S$ into nonempty clusters $T =
\bigsqcup_{g=1}^{m} G_g$. For each cluster, store the count, a
representative logit, and a representative value,
\begin{align}\label{eq:cluster-summaries}
n_g \coloneqq |G_g|,
\qquad
\bar\ell_g \in \R,
\qquad
\bar V_g \in \R^d,
\end{align}
together with radii $r_g \ge 0$, $\rho_g \ge 0$ required to satisfy
\begin{align}\label{eq:cluster-radii}
\big| \ell_j - \bar\ell_g \big|
&\;\le\; r_g,\\
&\text{and}
\notag\\[-0.2em]
\norm{V_j - \bar V_g}_2
&\;\le\; \rho_g,
\qquad \text{for all } j \in G_g,\ g \in [m].
\notag
\end{align}
The scheme outputs the ratio estimator built from exact kept terms plus
cluster summaries:
\begin{align}\label{eq:certificate-estimator}
\wh O
&\;\coloneqq\; \frac{\wh N}{\wh Z},\\
\wh N
&\;\coloneqq\; \sum_{j \in S} e^{\ell_j} V_j
+ \sum_{g=1}^m n_g e^{\bar\ell_g} \bar V_g,
\notag\\
\wh Z
&\;\coloneqq\;
\underbrace{\sum_{j \in S} e^{\ell_j}}_{\eqqcolon\, Z_S}
\notag\\
&\hspace{1.8em}
+ \sum_{g=1}^m
\underbrace{n_g e^{\bar\ell_g}}_{\eqqcolon\, \wh Z_g}.
\notag
\end{align}
to be compared with the exact $O = N/Z$, $N \coloneqq \sum_j e^{\ell_j}
V_j$, $Z = \sum_j e^{\ell_j}$.

\begin{lemma}[Ratio perturbation]\label{lem:ratio-perturbation}
Let $Z > 0$, $\wh Z > 0$, $N, \wh N \in \R^d$, and write $\Delta N
\coloneqq \wh N - N$, $\Delta Z \coloneqq \wh Z - Z$. Then, exactly,
\begin{align}\label{eq:ratio-identity}
\frac{\wh N}{\wh Z} - \frac{N}{Z}
\;=\; \frac{1}{\wh Z} \Big( \Delta N - \frac{N}{Z}\, \Delta Z \Big),
\end{align}
and consequently, on the condition $|\Delta Z| \le Z/2$,
\begin{align}\label{eq:ratio-bound}
\Big\| \frac{\wh N}{\wh Z} - \frac{N}{Z} \Big\|_2
\;\le\; \frac{2}{Z} \norm{\Delta N}_2
+ \frac{2 \norm{N}_2}{Z^2}\, |\Delta Z|.
\end{align}
\end{lemma}

\begin{proof}
The strategy is a single algebraic rearrangement. Over the common
denominator $\wh Z Z$,
\begin{align*}
\frac{\wh N}{\wh Z} - \frac{N}{Z}
&\;=\; \frac{\wh N Z - N \wh Z}{\wh Z\, Z}
\\
&\;=\; \frac{(N + \Delta N) Z
- N (Z + \Delta Z)}{\wh Z\, Z}\\
&\;=\; \frac{Z\, \Delta N - N\, \Delta Z}{\wh Z\, Z}
\\
&\;=\; \frac{1}{\wh Z}\Big(
\Delta N - \frac{N}{Z} \Delta Z \Big),
\end{align*}
which is Eq.~\eqref{eq:ratio-identity}. If $|\Delta Z| \le Z/2$ then
$\wh Z = Z + \Delta Z \ge Z/2 > 0$, so taking norms and the triangle
inequality,
\begin{align*}
\Big\| \frac{\wh N}{\wh Z} - \frac{N}{Z} \Big\|_2
&\;\le\; \frac{1}{\wh Z} \Big(
\norm{\Delta N}_2
+ \frac{\norm{N}_2}{Z}|\Delta Z|
\Big)\\
&\;\le\; \frac{2}{Z} \norm{\Delta N}_2
+ \frac{2\norm{N}_2}{Z^2}|\Delta Z|.
\end{align*}
This completes the proof.
\end{proof}

\begin{theorem}[Runtime-computable output-error bound for clustered tail
summaries; formal version of \cref{thm:main-certificate}]\label{thm:certificate}
In the setting of Eq.~\eqref{eq:cluster-summaries},
Eq.~\eqref{eq:cluster-radii}, and Eq.~\eqref{eq:certificate-estimator},
define the three runtime-computable quantities
\begin{align}
D_Z
&\;\coloneqq\; \sum_{g=1}^m
\big( e^{r_g} - 1 \big)e^{r_g}\wh Z_g,
\label{eq:certificate-quantities}\\
D_N
&\;\coloneqq\; \sum_{g=1}^m
e^{r_g}\wh Z_g
\Big[
\big(e^{r_g} - 1\big)\norm{\bar V_g}_2
+ \rho_g \Big],
\notag\\
V'_{\max}
&\;\coloneqq\; \max\Big(
\max_{j \in S} \norm{V_j}_2,
\notag\\
&\hspace{4.1em}
\max_{g \in [m]}
\big( \norm{\bar V_g}_2 + \rho_g \big)
\Big).
\nonumber
\end{align}
If the runtime-checkable condition $D_Z \le \wh Z / 3$ holds, then
\begin{align}\label{eq:certificate-bound}
\big\| \wh O - O \big\|_2
\;\le\; \frac{3}{\wh Z} \Big( D_N + V'_{\max}\, D_Z \Big).
\end{align}
Moreover, with the computable kept-mass estimate $\wh p_S \coloneqq
Z_S / \wh Z$, the bound has the structural form
\begin{align}\label{eq:certificate-structure}
\big\| \wh O - O \big\|_2
&\;\le\; 3\, \big( 1 - \wh p_S \big)
\max_{g \in [m]} e^{r_g}\\
&\hspace{1.8em}\cdot
\Big[
2\, V'_{\max} \big( e^{r_g} - 1 \big)
+ \rho_g \Big],
\notag
\end{align}
which vanishes as the cluster radii $(r_g, \rho_g)$ shrink, at a rate
modulated by the (estimated) dropped mass $1 - \wh p_S$.
\end{theorem}

\begin{proof}
The strategy is to bound the numerator and denominator perturbations
cluster by cluster using the radii, convert the unknown cluster masses
into the stored $\wh Z_g$, and finish with
\Cref{lem:ratio-perturbation}. Write $Z_g \coloneqq \sum_{j \in G_g}
e^{\ell_j}$ (not stored; used only inside the proof).

\textbf{Step 1 (pointwise exponential perturbation).} For $j \in G_g$,
Eq.~\eqref{eq:cluster-radii} gives $\ell_j - r_g \le \bar\ell_g \le
\ell_j + r_g$, so
\begin{align}\label{eq:exp-perturbation}
e^{-r_g} e^{\ell_j}
&\;\le\; e^{\bar\ell_g}
\;\le\; e^{r_g} e^{\ell_j},\\
&\text{hence}
\notag\\[-0.2em]
\big| e^{\bar\ell_g} - e^{\ell_j} \big|
&\;\le\; \big( e^{r_g} - 1 \big)e^{\ell_j},
\notag
\end{align}
using $\max(e^{r}-1,\, 1 - e^{-r}) = e^r - 1$ for $r \ge 0$. Summing the
left chain of Eq.~\eqref{eq:exp-perturbation} over $j \in G_g$ also
yields the two-sided mass conversion
\begin{align}\label{eq:mass-conversion}
e^{-r_g} Z_g
&\;\le\; \wh Z_g
\;=\; n_g e^{\bar \ell_g}
\;\le\; e^{r_g} Z_g,\\
&\text{in particular}
\notag\\[-0.2em]
Z_g &\;\le\; e^{r_g}\wh Z_g.
\notag
\end{align}

\textbf{Step 2 (denominator perturbation).} The kept terms of $\wh Z$ and
$Z$ coincide, so by Eq.~\eqref{eq:exp-perturbation} and
Eq.~\eqref{eq:mass-conversion},
\begin{align}\label{eq:deltaz-bound}
\begin{split}
|\Delta Z|
&\;=\; \Big| \sum_{g} \sum_{j \in G_g}
\big( e^{\bar\ell_g} - e^{\ell_j} \big)
\Big|\\
&\;\le\; \sum_g \big( e^{r_g} - 1 \big) Z_g\\
&\;\le\; \sum_g \big( e^{r_g} - 1 \big)
e^{r_g}\wh Z_g
\;=\; D_Z.
\end{split}
\end{align}

\textbf{Step 3 (numerator perturbation).} For $j \in G_g$, inserting
$e^{\bar\ell_g} \bar V_g - e^{\ell_j} V_j = (e^{\bar\ell_g} -
e^{\ell_j}) \bar V_g + e^{\ell_j} (\bar V_g - V_j)$ and using
Eq.~\eqref{eq:exp-perturbation} and Eq.~\eqref{eq:cluster-radii},
\begin{align*}
\big\| e^{\bar\ell_g} \bar V_g - e^{\ell_j} V_j \big\|_2
&\;\le\; \big( e^{r_g} - 1 \big)e^{\ell_j}
\norm{\bar V_g}_2\\
&\hspace{1.8em}+ e^{\ell_j}\rho_g,
\end{align*}
so, summing over $j \in G_g$ and $g \in [m]$ and applying
Eq.~\eqref{eq:mass-conversion},
\begin{align}\label{eq:deltan-bound}
\begin{split}
\norm{\Delta N}_2
&\;\le\; \sum_g Z_g \Big[
\big( e^{r_g} - 1 \big)\norm{\bar V_g}_2
+ \rho_g \Big]\\
&\;\le\; \sum_g e^{r_g}\wh Z_g \Big[
\big( e^{r_g} - 1 \big)\norm{\bar V_g}_2
+ \rho_g \Big]\\
&\;=\; D_N.
\end{split}
\end{align}

\textbf{Step 4 (the condition licenses the ratio lemma).} Assume $D_Z
\le \wh Z/3$. By Eq.~\eqref{eq:deltaz-bound}, $|\Delta Z| \le \wh Z /
3$, hence
\begin{align}\label{eq:z-zhat}
Z
&\;=\; \wh Z - \Delta Z
\;\ge\; \tfrac{2}{3}\wh Z,\\
&\text{and}
\notag\\[-0.2em]
|\Delta Z|
&\;\le\; \tfrac13 \wh Z
\;\le\; \tfrac13 \cdot \tfrac32 Z
\;=\; \tfrac{Z}{2}.
\notag
\end{align}
Moreover $O$ is a convex combination of the $V_j$, and every $\norm{V_j}_2
\le V'_{\max}$: for $j \in S$ directly, and for $j \in G_g$ by
$\norm{V_j}_2 \le \norm{\bar V_g}_2 + \norm{V_j - \bar V_g}_2 \le
\norm{\bar V_g}_2 + \rho_g$; hence
\begin{align}\label{eq:o-bound}
\frac{\norm{N}_2}{Z} \;=\; \norm{O}_2 \;\le\; \max_{j\in[n]}
\norm{V_j}_2 \;\le\; V'_{\max}.
\end{align}
Combining \Cref{lem:ratio-perturbation} (applicable by
Eq.~\eqref{eq:z-zhat}) with Eq.~\eqref{eq:deltaz-bound},
Eq.~\eqref{eq:deltan-bound}, and Eq.~\eqref{eq:o-bound},
\begin{align*}
\big\| \wh O - O \big\|_2
&\;\le\; \frac{2}{Z}\norm{\Delta N}_2
+ \frac{2\norm{N}_2}{Z^2}|\Delta Z|\\
&\;\le\; \frac{2}{Z}
\Big( D_N + V'_{\max} D_Z \Big)\\
&\;\le\; \frac{3}{\wh Z}
\Big( D_N + V'_{\max} D_Z \Big),
\end{align*}
where the last step uses $Z \ge \tfrac23 \wh Z$. This is
Eq.~\eqref{eq:certificate-bound}. For
Eq.~\eqref{eq:certificate-structure}, bound each cluster's contribution
by the maximal bracket: since $\norm{\bar V_g}_2 \le V'_{\max}$,
\begin{gather*}
D_N + V'_{\max} D_Z
\;\le\; \sum_g e^{r_g}\wh Z_g
\\
\hspace{1.8em}\cdot
\Big[
\big(e^{r_g}-1\big)
\big(\norm{\bar V_g}_2+V'_{\max}\big)
+ \rho_g \Big]\\
\;\le\; \Big( \sum_g \wh Z_g \Big)\\
\hspace{1.8em}\cdot \max_g e^{r_g}\\
\hspace{1.8em}\cdot
\Big[
2 V'_{\max}\big( e^{r_g} - 1 \big)
+ \rho_g \Big],
\end{gather*}
and $\sum_g \wh Z_g = \wh Z - Z_S = (1 - \wh p_S)\, \wh Z$ cancels the
$\wh Z$ in Eq.~\eqref{eq:certificate-bound}. This completes the proof.
\end{proof}

\begin{corollary}[Small-radius form; formal version of \cref{thm:main-certificate}]\label{cor:certificate-small}
If additionally $\max_g r_g \le 1$, then under the condition of
\Cref{thm:certificate},
\begin{align*}
\big\| \wh O - O \big\|_2
&\;\le\; 30\, \big( 1 - \wh p_S \big)
\max_{g \in [m]}\\
&\hspace{1.8em}\cdot
\Big( V'_{\max} r_g + \rho_g \Big).
\end{align*}
\end{corollary}

\begin{proof}
The strategy is to linearise the exponentials in
Eq.~\eqref{eq:certificate-structure}. For $r \in [0,1]$, convexity of
$e^x$ gives $e^r - 1 \le (e-1) r$ and $e^r \le e$, so for every $g$,
\begin{gather*}
e^{r_g} \Big[
2 V'_{\max} \big( e^{r_g} - 1 \big)
+ \rho_g \Big]
\\
\;\le\; e \Big[
2 (e-1) V'_{\max} r_g + \rho_g
\Big]\\
\;\le\; 10\, V'_{\max}\, r_g
+ 3\, \rho_g\\
\;\le\; 10 \big( V'_{\max} r_g + \rho_g \big),
\end{gather*}
using $2e(e-1) \le 10$ and $e \le 3$; multiplying by $3 (1-\wh p_S)$ and
inserting into Eq.~\eqref{eq:certificate-structure} gives the claim. The
resulting constant $30$ is deliberately conservative: the dominant term
already obeys $2e(e-1) \le 10$, so it could be sharpened, but we keep the
round value because sharpening it does not change the runtime error rate.
This completes the proof.
\end{proof}

\begin{remark}[Runtime checkability, and the radii in the attention
application]\label{rem:certificate-runtime}
Every quantity in Eq.~\eqref{eq:certificate-quantities} through
Eq.~\eqref{eq:certificate-structure}
--- $\wh Z_g$, $\wh Z$, $D_Z$, $D_N$, $V'_{\max}$, $\wh p_S$, and the
condition $D_Z \le \wh Z/3$ --- is computable from the stored summaries
$(n_g, \bar\ell_g, \bar V_g, r_g, \rho_g)$ and the kept entries alone,
without touching the $n - |S|$ dropped entries.  The right-hand side can
therefore be evaluated at inference time, and the system can determine
whether the resulting upper bound meets its target error.
inference time. In the attention application the tail logits are inner
products $\ell_j = \inner{q}{k_j}/\sqrt{d_k}$ with the current query
$q$: choosing $\bar\ell_g \coloneqq \inner{q}{\bar k_g}/\sqrt{d_k}$ for
a stored key centroid $\bar k_g$ gives, by Cauchy--Schwarz,
\begin{align*}
\big| \ell_j - \bar\ell_g \big|
&\;=\; \frac{\big| \inner{q}{k_j - \bar k_g}
\big|}{\sqrt{d_k}}\\
&\;\le\; \frac{\norm{q}_2}{\sqrt{d_k}}\,
\max_{j \in G_g} \norm{k_j - \bar k_g}_2,
\end{align*}
so a valid $r_g$ is the product of the query norm (known at runtime) and
the stored key-cluster radius --- tight key clusters give tight
logit radii for \emph{every} query of bounded norm. The value radii
$\rho_g$ are query-independent and stored once.  The theorem thus turns
geometric clusterability of the dropped keys and values into a deterministic
output-error bound, in the structural form $(1 - \wh
p_S) \cdot O(\text{logit radius} + \text{value radius})$ anticipated by
the exact compensation identity of \Cref{prop:compensation}.
\end{remark}

\subsection{Sampled tail compensation}\label{sub:app-sampled}

The bound in \Cref{thm:certificate} applies to
\emph{clustered} tail summaries; the following proposition prices the
\emph{sampled} alternative quoted after \cref{thm:main-certificate} in
the main text: estimate the exact compensation vector of
\Cref{prop:compensation} from $m$ sampled tail tokens, using the fact
that the tail \emph{weights} are known at runtime even when the tail
\emph{values} are not retained.

\begin{proposition}[Sampled tail compensation]\label{prop:sampled-comp}
Fix a softmax row as in \Cref{def:softmax-row} and a kept set $S$ with
$|S| = k < n$, write $T = [n] \setminus S$ for the tail and
\begin{align*}
u^\star
&\;\coloneqq\; \sum_{j \in T} A_j V_j
\;=\; (1 - p_S)\bar V_T,\\
\wt A_j
&\;\coloneqq\; \frac{A_j}{1 - p_S}
\quad (j \in T),\\
Q_T \;\coloneqq\; \sum_{j \in T} A_j^2,
\end{align*}
for the exact compensation vector (\Cref{prop:compensation}) and the
normalised tail weights. Let the values $V_1, \dots, V_n$ be i.i.d.\
random vectors in $\R^d$, independent of the sampling below, with
$\E \norm{V_1}_2^2 < \infty$, mean $\mu_V$, and coordinate variance
$\sigma_V^2 \coloneqq \Tr\big(\operatorname{Cov}(V_1)\big)/d$, and fix
$1 \le m \le n - k$. All statements are conditional on the row (the
weights are fixed throughout).
\begin{enumerate}[label=(\roman*), ref=\ref*{prop:sampled-comp}(\roman*)]
\item\label{prop:sampled-comp:weighted} (\emph{Weight-proportional
sampling.}) Let $j_1, \dots, j_m$ be i.i.d.\ samples from $T$ with
$\Pr(j_r = j) = \wt A_j$ --- implementable at runtime because the tail
weights are known --- and set
$\wh u \coloneqq \frac{1 - p_S}{m} \sum_{r=1}^{m} V_{j_r}$. Then
$\E[\, \wh u \mid V \,] = u^\star$ for every realisation of the values
(unbiasedness for the tail sum over the sampling alone), and
\begin{multline}\label{eq:sampled-mse}
\E\big[ \norm{\wh u - u^\star}_2^2 \big]
\\
\;=\; \frac{d\, \sigma_V^2\, (1 - p_S)^2}{m}
\Big(
1 - \frac{Q_T}{(1-p_S)^2}
\Big)\\
\;\le\; \frac{d\, \sigma_V^2}{m}
(1 - p_S)^2,
\end{multline}
where the expectation is the MSE over values and sampling, conditional on
the fixed weight row, so the root-mean-square error is at most
$(1 - p_S)\, \sigma_V \sqrt{d/m}$.
\item\label{prop:sampled-comp:uniform} (\emph{Uniform sampling.}) Let
$M$ be a uniformly random $m$-subset of $T$, independent of the values,
and $\wh u_{\mathrm{unif}} \coloneqq \frac{1 - p_S}{m} \sum_{j \in M}
V_j$. Then $\E[\, \wh u_{\mathrm{unif}} - u^\star \mid M \,] = 0$ for
every sample (unbiasedness over the values), and exactly
\begin{multline}\label{eq:sampled-mse-unif}
\E\left[
\begin{aligned}
&\left\|
\begin{aligned}
&\wh u_{\mathrm{unif}}\\[-0.2em]
&\hspace{0.7em}{}- u^\star
\end{aligned}
\right\|_2^2
\end{aligned}
\right]
\\
\;=\; d\, \sigma_V^2\, (1-p_S)^2\\
\hspace{1.8em}\cdot
\Big[
\frac{1}{m} - \frac{2}{n-k}
+ \frac{Q_T}{(1-p_S)^2}
\Big],
\end{multline}
which is at most $2 d \sigma_V^2 (1-p_S)^2/m$ --- hence RMSE $\le
(1-p_S)\,\sigma_V \sqrt{2d/m}$ --- whenever the dropped tail is balanced
in the sense $Q_T \le (1 - p_S)^2/m$.
\end{enumerate}
In either case, by \Cref{prop:compensation} the compensated output
$O_S + \wh u$ approximates $O$ with the same conditional RMSE.
\end{proposition}

\begin{proof}
Both estimators have the form $\wh u = (1 - p_S) \sum_{j \in T} w_j
V_j$ with sampling-measurable coefficients $w_j \ge 0$ summing to one:
$w_j = m_j/m$ with $m_j \coloneqq \#\{r : j_r = j\}$ in (i), and $w_j =
\1\{j \in M\}/m$ in (ii). Since $u^\star = (1-p_S) \sum_{j \in T} \wt
A_j V_j$ and $\sum_{j \in T} \wt A_j = 1$,
\begin{align}\label{eq:sampled-coeff}
\wh u - u^\star
&\;=\; (1 - p_S) \sum_{j \in T} c_j V_j,\\
c_j &\;\coloneqq\; w_j - \wt A_j,
\notag\\
\sum_{j \in T} c_j \;=\; 1 - 1 \;=\; 0.
\notag
\end{align}
\emph{Unbiasedness.} In (i), $m_j \sim \Bin(m, \wt A_j)$, so $\E[w_j]
= \wt A_j$ and $\E[\wh u \mid V] = (1-p_S)\sum_j \wt A_j V_j =
u^\star$. In (ii), conditionally on $M$ the coefficients are constants,
so by Eq.~\eqref{eq:sampled-coeff} and $\E V_j = \mu_V$,
$\E[\wh u_{\mathrm{unif}} - u^\star \mid M] = (1-p_S) (\sum_j c_j)
\mu_V = 0$.
\emph{Variance.} Conditionally on the sample, write $V_j = \mu_V + (V_j
- \mu_V)$ in Eq.~\eqref{eq:sampled-coeff}; the $\mu_V$ part vanishes
because $\sum_j c_j = 0$, and the centred parts are i.i.d., so the cross
terms vanish and
\begin{multline*}
\E\left[
\begin{aligned}
&\norm{\wh u - u^\star}_2^2\\[-0.2em]
&\mid\, \text{sample}
\end{aligned}
\right]
\\
\;=\; (1-p_S)^2 \sum_{j \in T} c_j^2\,
\E\norm{V_1 - \mu_V}_2^2\\
\;=\; d\, \sigma_V^2\, (1 - p_S)^2
\sum_{j \in T} c_j^2 .
\end{multline*}
It remains to average $\sum_j c_j^2$ over the sampling. In (i), $\E
c_j^2 = \operatorname{Var}(m_j/m) = \wt A_j (1 - \wt A_j)/m$, and
summing over $j \in T$ gives $\big( 1 - \sum_j \wt A_j^2 \big)/m = \big(
1 - Q_T/(1-p_S)^2 \big)/m$, which proves Eq.~\eqref{eq:sampled-mse}. In
(ii), $\E[\1\{j \in M\}] = m/(n-k)$ and $\1\{j\in M\}^2 = \1\{j \in
M\}$, so
\begin{align*}
\E\, c_j^2
\;=\; \frac{m/(n-k)}{m^2}
- \frac{2 \wt A_j}{n-k}
+ \wt A_j^2,\\
\sum_{j \in T} \E\, c_j^2
\;=\; \frac{1}{m} - \frac{2}{n-k}
+ \frac{Q_T}{(1-p_S)^2},
\end{align*}
which proves Eq.~\eqref{eq:sampled-mse-unif}; under $Q_T \le
(1-p_S)^2/m$ the bracket is at most $2/m$. The final claim is
\Cref{prop:compensation}: $\norm{O_S + \wh u - O}_2 = \norm{\wh u -
u^\star}_2$. This completes the proof.
\end{proof}

\begin{remark}[Which sampler the rate charges]\label{rem:sampled-balance}
Part \ref*{prop:sampled-comp}(i) attains the rate $(1-p_S)\sigma_V
\sqrt{d/m}$ with constant $1$ for \emph{every} weight configuration;
part \ref*{prop:sampled-comp}(ii) shows that the simpler uniform sampler
pays exactly the tail-imbalance term $Q_T/(1-p_S)^2$, the price of
estimating a weighted aggregate by an unweighted sample mean. For the
dispersed ensembles of \cref{sec:app-dispersed} the top-$k$ tail is
polynomially flat ($Q_T/(1-p_S)^2 = n^{-\Omega(1)}$ with probability
$\to 1$; cf.\ Eq.~\eqref{eq:qsqt-exponents} and
Eq.~\eqref{eq:ps-exponent}), so uniform sampling matches the
weight-proportional rate for every $m \le n^{\Omega(1)}$ there.
\end{remark}

\subsection{Order-statistic mass bound and selector composition}\label{sub:app-orderstat-selector}

The following two short statements are cited from the main text (the \textsc{Free} cell of \cref{tab:main-ladder} and \cref{lem:main-selector}); we record their proofs for completeness.

\begin{fact}[Order-statistic mass bound, \textsc{Free}]\label{fac:orderstat-mass}
For every row and every $k \in [n-1]$: $A_{(k+1)} \le \frac{1}{k+1}$ and
\begin{align*}
1 - \cov(k) \;\le\; (n-k)\, A_{(k+1)} \;\le\; \frac{n-k}{k+1},
\\
\text{hence}\\[-0.2em]
\norm{\wt O_{S^{*}_k} - O}_2 \;\le\; \frac{2\Vmax (n-k)}{k+1}
\end{align*}
by \Cref{prop:bias-identity}. The bound is informative only for $k > (n-1)/2$ (below that it exceeds the trivial $2\Vmax$) and is tight in order exactly in the regime $k = \Theta(n)$: on the uniform row $A_{(j)} \equiv 1/n$ one has $1 - \cov(k) = \frac{n-k}{n} = \frac{n-k}{k+1}\cdot\frac{k+1}{n}$, matching up to the factor $\frac{k+1}{n} \le 1$, which is $\Theta(1)$ iff $k = \Theta(n)$.
\end{fact}

\begin{proof}
The top $k+1$ sorted weights each dominate $A_{(k+1)}$ and sum to at most the row total:
\begin{align*}
(k+1) A_{(k+1)}
&\;\le\; \sum_{j=1}^{k+1} A_{(j)}
\;\le\; 1,\\
&\Longrightarrow\\[-0.2em]
A_{(k+1)}
&\;\le\; \frac{1}{k+1}.
\end{align*}
The dropped tail has $n-k$ entries, each at most $A_{(k+1)}$:
\begin{align*}
1 - \cov(k)
&\;=\; \sum_{j=k+1}^{n} A_{(j)}\\
&\;\le\; (n-k) A_{(k+1)}
\;\le\; \frac{n-k}{k+1}.
\end{align*}
The output bound follows from the mass bound of \Cref{prop:bias-identity}. This completes the proof.
\end{proof}

\begin{lemma}[One-sided selector composition; formal version of \cref{lem:main-selector}]\label{lem:selector-comp}
Let $S^{*}_k$ be an oracle top-$k$ set, $\widehat S$ any set with $|\widehat S| = k$, and
$M_{\mathrm{miss}} \coloneqq \sum_{j \in S^{*}_k \setminus \widehat S} A_j$. Then
\begin{align*}
1 - p_{\widehat S} \;\le\; \big( 1 - p_{S^{*}_k} \big) + M_{\mathrm{miss}},
\\
\text{and consequently}\\[-0.2em]
\norm{\wt O_{\widehat S} - O}_2
\;\le\; 2 \Vmax \Big[
\big(1 - p_{S^{*}_k}\big) + M_{\mathrm{miss}}
\Big].
\end{align*}
The inequality is one-sided by design: $M_{\mathrm{miss}}$ depends on the selector and its history and is not bounded by any functional of the sorted profile.
\end{lemma}

\begin{proof}
Splitting the kept mass of $\widehat S$ on membership in $S^{*}_k$ and dropping the non-negative second part,
\begin{align*}
p_{\widehat S}
&\;=\; \sum_{j \in \widehat S \cap S^{*}_k} A_j
+ \sum_{j \in \widehat S \setminus S^{*}_k} A_j\\
&\;\ge\; \sum_{j \in \widehat S \cap S^{*}_k} A_j\\
&\;=\; p_{S^{*}_k}
- \sum_{j \in S^{*}_k \setminus \widehat S} A_j\\
&\;=\; p_{S^{*}_k} - M_{\mathrm{miss}},
\end{align*}
where the second equality holds because $\widehat S \cap S^{*}_k$ and $S^{*}_k \setminus \widehat S$ partition $S^{*}_k$. Rearranging gives the mass inequality, and the output bound follows by applying the mass bound of \Cref{prop:bias-identity} to the kept set $\widehat S$. This completes the proof.
\end{proof}

\section{Deterministic Profile Theorems}\label{sec:app-profile}

This appendix states and proves the deterministic log-slope criterion \cref{thm:main-dichotomy} and its plateau refinement \cref{cor:main-plateau} --- the engine used as machinery throughout the main text: \cref{thm:main-dichotomy}(i) is \cref{thm:s1-summable} (sharp by \cref{prop:s1-tight}), \cref{thm:main-dichotomy}(ii) is \cref{thm:s2-sublog}, and \cref{cor:main-plateau} is \cref{thm:s3-plateau}.

All results in this section are deterministic statements about a single
fixed logit vector; as in \Cref{sec:app-reduction}, no probability is
involved. Fix a softmax row as in \Cref{def:softmax-row}, write
$\ell_{(1)} \ge \ell_{(2)} \ge \cdots \ge \ell_{(n)}$ for the logits
sorted in nonincreasing order (ties broken by index), recall the coverage
profile $p(k)$ and coverage budget $k_\eta$ of \Cref{def:coverage}, and
let $S_k \subseteq [n]$ denote the set of indices of the $k$ largest
weights (ties broken by index); the multiset of kept weights is then
exactly $\{A_{(1)}, \dots, A_{(k)}\}$, so the kept mass is $\sum_{j \in
S_k} A_j = \sum_{j \le k} A_{(j)} = p(k)$ (\Cref{def:coverage}). The probabilistic sections control coverage through
ensemble-level events; this section asks instead which conditions on the
\emph{shape} of one row force sparsity or density, and answers in terms
of the growth rate of the gaps $\ell_{(1)} - \ell_{(j)}$: gap growth at
least $\gamma \log j$ with $\gamma > 1$ forces budgets free of $n$
(\Cref{thm:s1-summable}, sharp by \Cref{prop:s1-tight}); gap growth at
most $\gamma' \log j$ with $\gamma' < 1$ forces $k_\eta$ to grow at
least at the rate $n^{1-\gamma'}$ (\Cref{thm:s2-sublog}); and a width-$J$ plateau
followed by decay yields budgets of order $J$ (\Cref{thm:s3-plateau}).
The sparse/dense dichotomy sits exactly at unit log-slope
(\Cref{rem:s2-dichotomy}).

We first record the digest used as machinery in the main text --- the log-slope dichotomy and its plateau refinement --- and then prove it through the lemmas and theorems of this appendix.

\begin{theorem}[Log-slope criterion: length-independent vs.\ length-growing budget; informal version of \cref{thm:s1-summable,thm:s2-sublog,prop:s1-tight}]\label{thm:main-dichotomy}
Fix a row with gap profile $\gap$ as in \cref{def:gap-profile}.
\begin{enumerate}[label=(\roman*),leftmargin=2em,itemsep=0.2em]
\item \textbf{Super-logarithmic decay $\Rightarrow$ $n$-free sparsity.} If $\gap(j) \ge \gamma \log j - b$ for all $j \in [n]$, with $\gamma > 1$ and $b \ge 0$, then for every $k \ge 1$ and every $n$,
\begin{align}\label{eq:s1-mass}
\begin{aligned}
1 - \cov(k)
&\;\le\; \frac{e^{b}}{\gamma - 1}\,
k^{-(\gamma-1)},\\
&\text{hence}\\
\bud
&\;\le\; \Big\lceil
\Big( \frac{e^{b}}{(\gamma-1)\,\eta} \Big)^{
\frac{1}{\gamma-1}} \Big\rceil ,
\end{aligned}
\end{align}
and the row is relatively $(\epsilon, k)$-sparse with $k = (e^b/\epsilon)^{1/\gamma}$ for every $\epsilon \in (0,1]$. All three bounds are independent of $n$, and they are \emph{matched from below} (sharp up to constants): on the witness profile $\gap(j) = \gamma\log j$, which satisfies the hypothesis, $1 - \cov(k) \ge \tfrac{(1-2^{1-\gamma})\,2^{1-\gamma}}{\gamma}\, k^{-(\gamma-1)}$ for all $1 \le k \le (n-1)/2$, and hence $\bud \ge c_\gamma\,\eta^{-1/(\gamma-1)}$ with $c_\gamma = \big(\tfrac{(1-2^{1-\gamma})2^{1-\gamma}}{\gamma}\big)^{1/(\gamma-1)}$ --- so both the tail-mass exponent and the ($n$-free) budget rate are tight (\cref{prop:s1-tight}).
\item \textbf{Sub-logarithmic profile $\Rightarrow$ length-growing budget.} If instead $\gap(j) \le \gamma' \log j + b'$ for all $j \in [n]$, with $0 \le \gamma' < 1$, then for every $\eta \in (0,1)$ there is $c_0 = c_0(\gamma', b', \eta) > 0$ with
\begin{align}\label{eq:s2-lower}
c_0\, n^{\,1 - \gamma'} \;\le\; \bud \;\le\; \lceil (1-\eta)\, n \rceil ,
\end{align}
the upper bound being the universal pigeonhole cap (\cref{fac:pigeonhole}). The lower bound forces a \emph{length-growing} (non-$n$-free) budget; at $\gamma' = 0$ (a constant-width band) the two ends match at $\Theta(n)$ --- as incompressible as the uniform row. For $\gamma' \in (0,1)$ the pure slope witness $\gap(j) = \gamma'\log j$ has the larger budget $\bud = n(1-\eta)^{1/(1-\gamma')}$, while the one-sided hypothesis itself guarantees only the displayed polynomial lower bound. Thus a sub-logarithmic upper envelope rules out an \(n\)-independent budget but does not determine a unique growth rate; sub-linear rates require additional profile structure.
\end{enumerate}
\end{theorem}

The hypothesis of (i) is a one-plot condition: on a log--log plot of sorted weights against rank, it asks that the curve eventually falls with slope steeper than $-1$. Equivalently, for a fixed profile shape the budget is length-independent precisely when the normalized weights are summable, $\sum_{j\ge1} e^{-\gap(j)} < \infty$ --- a $p$-series test whose boundary is the unit slope. At $\gamma = 1$ the verdict is decided by the logarithmic correction: $\gap(j) = \log j + p\log\log j$ is length-independent iff $p > 1$ (convergence of $\sum_j 1/(j\log^p j)$), sharpening the boundary the slope alone leaves open.

\begin{corollary}[Plateau sets the sub-linear scale; informal version of \cref{thm:s3-plateau}]\label{cor:main-plateau}
Suppose there exist $J \in [n]$, $\gamma > 1$, and $b, b_0 \ge 0$ such that the profile has a genuine plateau, $\gap(j) \le b_0$ for all $j \le J$, followed by super-logarithmic decay, $\gap(j) \ge \gamma \log(j/J) - b$ for all $j > J$. Then for every $k \ge J$,
\begin{align}\label{eq:s3-mass}
\begin{aligned}
1 - \cov(k)
&\;\le\; \frac{e^{b + b_0}}{\gamma-1}
\Big(\frac{k}{J}\Big)^{-(\gamma-1)},\\
\bud
&\;\le\; J \cdot \Big\lceil
\Big( \frac{e^{b+b_0}}{(\gamma-1)\,\eta} \Big)^{
\frac{1}{\gamma-1}} \Big\rceil .
\end{aligned}
\end{align}
Both halves of the hypothesis are load-bearing: without the plateau-height condition the tail-mass bound picks up an extra factor of $J$ --- and the budget a factor $J^{1/(\gamma-1)}$ (\cref{thm:s3-plateau}).
The coverage budget is the plateau width times an $n$-free factor, so it inherits the plateau's growth class exactly: constant plateaus recover \cref{thm:main-dichotomy}(i), poly-logarithmic plateaus give poly-logarithmic budgets, and polynomially sub-linear plateaus give polynomially sub-linear budgets --- the row is sparse at whatever scale its plateau is narrow.
\end{corollary}

\subsection{The gap profile and two comparison lemmas}

\begin{definition}[Gap profile]\label{def:gap-profile}
The \emph{gap profile} of the row is
\begin{align*}
G(j) \;\coloneqq\; \ell_{(1)} - \ell_{(j)}, \qquad j \in [n].
\end{align*}
Since $(\ell_{(j)})_{j \in [n]}$ is nonincreasing, $G$ is nondecreasing
with $G(1) = 0$ and $G(j) \ge 0$ for all $j \in [n]$.
\end{definition}

\begin{lemma}[Gap representation of weights and coverage]\label{lem:gap-rep}
Let $D \coloneqq \sum_{l=1}^{n} e^{-G(l)}$. Then $1 \le D \le n$, and:
\begin{enumerate}[label=(\roman*), ref=\ref*{lem:gap-rep}(\roman*)]
\item\label{lem:gap-rep:weights} $A_{(j)} = e^{-G(j)}/D$ for every $j \in
[n]$; in particular $A_{(j)}/A_{(1)} = e^{-G(j)}$;
\item\label{lem:gap-rep:coverage} $1 - p(k) = \frac{1}{D}
\sum_{j=k+1}^{n} e^{-G(j)}$ for every $k \in \{0, 1, \dots, n\}$.
\end{enumerate}
\end{lemma}

\begin{proof}
The strategy is to normalise every weight by the top exponential
$e^{\ell_{(1)}}$. The sum defining $Z$ in \Cref{def:softmax-row} is
invariant under reordering its summands, and $x \mapsto e^x$ is strictly
increasing, so $e^{\ell_{(1)}} \ge \cdots \ge e^{\ell_{(n)}}$ is the
nonincreasing rearrangement of $(e^{\ell_j})_{j \in [n]}$; dividing by
$Z > 0$ preserves this order, so the sorted weights of
\Cref{def:coverage} are
\begin{align}\label{eq:sorted-weight-formula}
\begin{aligned}
A_{(j)}
&\;=\; \frac{e^{\ell_{(j)}}}{Z}\\
&\;=\; \frac{e^{\ell_{(j)}}}{
\sum_{l=1}^{n} e^{\ell_{(l)}}}\\
&\;=\; \frac{e^{\ell_{(j)} - \ell_{(1)}}}{
\sum_{l=1}^{n} e^{\ell_{(l)} - \ell_{(1)}}}\\
&\;=\; \frac{e^{-G(j)}}{D},
\end{aligned}
\end{align}
where the second equality reorders the summands of $Z$, the third
multiplies numerator and denominator by $e^{-\ell_{(1)}} > 0$, and the
fourth inserts \Cref{def:gap-profile}. Since $G(1) = 0$ and $0 \le
G(l)$ for every $l$ (\Cref{def:gap-profile}),
\begin{align*}
1 \;=\; e^{-G(1)} \;\le\; D \;\le\; \sum_{l=1}^{n} 1 \;=\; n.
\end{align*}
The ratio form in (i) is Eq.~\eqref{eq:sorted-weight-formula} at $j$
divided by Eq.~\eqref{eq:sorted-weight-formula} at $1$, using $e^{-G(1)}
= 1$. For (ii), summing Eq.~\eqref{eq:sorted-weight-formula} over $j >
k$ and using $p(n) = \sum_{j=1}^n A_{(j)} = 1$ (\Cref{def:coverage}),
\begin{align*}
1 - p(k)
\;=\; \sum_{j=k+1}^{n} A_{(j)}
\;=\; \frac{1}{D} \sum_{j=k+1}^{n} e^{-G(j)}.
\end{align*}
This completes the proof.
\end{proof}

\begin{lemma}[Sum--integral comparison]\label{lem:sum-integral}
Let $h \colon [1, \infty) \to [0, \infty)$ be nonincreasing and let
$a \le b$ be integers with $a \ge 1$. Then:
\begin{enumerate}[label=(\roman*), ref=\ref*{lem:sum-integral}(\roman*)]
\item\label{lem:sum-integral:upper} $\displaystyle \sum_{j=a+1}^{b} h(j)
\;\le\; \int_{a}^{b} h(x)\,dx$;
\item\label{lem:sum-integral:lower} $\displaystyle \sum_{j=a}^{b} h(j)
\;\ge\; \int_{a}^{b+1} h(x)\,dx$.
\end{enumerate}
\end{lemma}

\begin{proof}
Both parts compare each summand with the integral of $h$ over an
adjacent unit interval; $h$ is Riemann integrable on bounded
subintervals of $[1, \infty)$ because it is monotone. For (i):
monotonicity gives $h(j) \le h(x)$ for all $x \in [j-1, j]$ (any integer
$j \ge 2$), hence
\begin{align*}
\begin{aligned}
h(j)
&\;=\; \int_{j-1}^{j} h(j)\,dx\\
&\;\le\; \int_{j-1}^{j} h(x)\,dx,\\
&\text{so} \\
\sum_{j=a+1}^{b} h(j)
&\;\le\; \sum_{j=a+1}^{b}
\int_{j-1}^{j} h(x)\,dx\\
&\;=\; \int_{a}^{b} h(x)\,dx,
\end{aligned}
\end{align*}
the last equality because the intervals $[j-1, j]$, $a + 1 \le j \le b$,
tile $[a, b]$. For (ii): monotonicity gives $h(j) \ge h(x)$ for all
$x \in [j, j+1]$, hence
\begin{align*}
\begin{aligned}
h(j)
&\;=\; \int_{j}^{j+1} h(j)\,dx\\
&\;\ge\; \int_{j}^{j+1} h(x)\,dx,\\
&\text{so} \\
\sum_{j=a}^{b} h(j)
&\;\ge\; \sum_{j=a}^{b}
\int_{j}^{j+1} h(x)\,dx\\
&\;=\; \int_{a}^{b+1} h(x)\,dx.
\end{aligned}
\end{align*}
This completes the proof.
\end{proof}

\subsection{Summable logarithmic decay}

\begin{theorem}[Summable decay --- $n$-free budgets; formal version of \cref{thm:main-dichotomy}(i)]\label{thm:s1-summable}
Fix $\gamma > 1$ and $b \ge 0$, and suppose the gap profile satisfies
\begin{align}\label{eq:s1-hyp}
G(j) \;\ge\; \gamma \log j - b \qquad \text{for all } j \in [n].
\end{align}
Then the following hold; the right-hand sides of (i), (ii), and (iv)
are independent of $n$, and that of (iii) involves the instance only
through $\Vmax$.
\begin{enumerate}[label=(\roman*), ref=\ref*{thm:s1-summable}(\roman*)]
\item\label{thm:s1-summable:tail} (tail mass) for every $k \in [n]$,
\begin{align*}
1 - p(k) \;\le\; \frac{e^{b}\, k^{1-\gamma}}{\gamma - 1}.
\end{align*}
\item\label{thm:s1-summable:budget} (budget) for every $\eta \in (0,1)$,
\begin{align*}
k_\eta \;\le\; \bigg\lceil \Big( \frac{e^{b}}{(\gamma-1)\,\eta}
\Big)^{\frac{1}{\gamma-1}} \bigg\rceil.
\end{align*}
\item\label{thm:s1-summable:output} (output error) for every $k \in
[n-1]$ and any values $V_1, \dots, V_n \in \R^d$, the renormalised
scheme of \Cref{def:schemes} on the kept set $S_k$ satisfies
\begin{align*}
\big\| \wt{O}_{S_k} - O \big\|_2
\;\le\; \frac{2\, \Vmax\, e^{b}\, k^{1-\gamma}}{\gamma - 1}.
\end{align*}
\item\label{thm:s1-summable:count} (relative count) $A_{(j)} \le e^{b}
j^{-\gamma} A_{(1)}$ for every $j \in [n]$, and for every $\varepsilon
\in (0, 1]$,
\begin{align*}
\#\big\{ j \in [n] : A_j > \varepsilon A_{(1)} \big\}
\;\le\; \Big( \frac{e^{b}}{\varepsilon} \Big)^{1/\gamma}.
\end{align*}
\end{enumerate}
\end{theorem}

\begin{proof}
The strategy is to insert the hypothesis Eq.~\eqref{eq:s1-hyp} into the
gap representation of \Cref{lem:gap-rep} and to compare the resulting
power sums with integrals via \Cref{lem:sum-integral}.

\textbf{Step 1 (proof of (i): tail-mass bound).} By \Cref{lem:gap-rep:coverage}, the bound $D \ge 1$
of \Cref{lem:gap-rep}, and Eq.~\eqref{eq:s1-hyp},
\begin{align}\label{eq:s1-master}
\begin{aligned}
1 - p(k)
&\;=\; \frac{1}{D}
\sum_{j=k+1}^{n} e^{-G(j)}\\
&\;\le\; \sum_{j=k+1}^{n} e^{-G(j)}\\
&\;\le\; e^{b}
\sum_{j=k+1}^{n} j^{-\gamma}.
\end{aligned}
\end{align}
The map $x \mapsto x^{-\gamma}$ is nonincreasing on $[1, \infty)$, so
\Cref{lem:sum-integral:upper} with $h(x) = x^{-\gamma}$, $a = k$,
$b = n$, then nonnegativity of the integrand, and the antiderivative
$x^{1-\gamma}/(1-\gamma)$ (here $\gamma > 1$, so the integral over
$[k, \infty)$ converges) give
\begin{align}\label{eq:power-tail}
\begin{aligned}
\sum_{j=k+1}^{n} j^{-\gamma}
&\;\le\; \int_{k}^{n} x^{-\gamma}\,dx\\
&\;\le\; \int_{k}^{\infty} x^{-\gamma}\,dx\\
&\;=\; \frac{k^{1-\gamma}}{\gamma - 1}.
\end{aligned}
\end{align}
Combining Eq.~\eqref{eq:s1-master} and Eq.~\eqref{eq:power-tail} proves
(i).

\textbf{Step 2 (proof of (ii): budget inversion).} Set
\begin{align*}
\begin{aligned}
k^\circ
&\;\coloneqq\; \bigg\lceil
\Big( \frac{e^{b}}{(\gamma-1)\,\eta}
\Big)^{\frac{1}{\gamma-1}} \bigg\rceil\\
&\;\ge\; 1.
\end{aligned}
\end{align*}
If $k^\circ > n$, then $k_\eta \le n < k^\circ$, since $p(n) = 1 \ge
1 - \eta$ gives $k_\eta \le n$ (\Cref{def:coverage}). If $k^\circ \le
n$, then part (i) at $k = k^\circ$, the lower bound $k^\circ \ge
(e^b/((\gamma-1)\eta))^{1/(\gamma-1)}$ from the ceiling, and the fact
that $t \mapsto t^{1-\gamma}$ is nonincreasing on $(0, \infty)$ give
\begin{align*}
\begin{aligned}
1 - p(k^\circ)
&\;\le\; \frac{e^{b}\,
(k^\circ)^{1-\gamma}}{\gamma-1}\\
&\;\le\; \frac{e^{b}}{\gamma-1}
\Big( \frac{e^{b}}{(\gamma-1)\,\eta}
\Big)^{\frac{1-\gamma}{\gamma-1}}\\
&\;=\; \frac{e^{b}}{\gamma-1}
\cdot \frac{(\gamma-1)\,\eta}{e^{b}}\\
&\;=\; \eta,
\end{aligned}
\end{align*}
where the first equality uses $(1-\gamma)/(\gamma-1) = -1$, i.e.\ the
power is the reciprocal. Hence $p(k^\circ) \ge 1 - \eta$, and $k_\eta
\le k^\circ$ by the minimality in \Cref{def:coverage}.

\textbf{Step 3 (proof of (iii): output error).} The set $S_k$ collects the $k$ largest weights, so
$p_{S_k} = \sum_{j \in S_k} A_j = p(k)$ by \Cref{def:coverage}, and
$0 < k < n$ makes $S_k$ admissible in \Cref{def:schemes}. The mass
bound \Cref{prop:bias-identity:mass} and part (i) give
\begin{align*}
\begin{aligned}
\big\| \wt{O}_{S_k} - O \big\|_2
&\;\le\; 2\,\Vmax\,
\big(1 - p_{S_k}\big)\\
&\;=\; 2\,\Vmax\,\big(1 - p(k)\big)\\
&\;\le\; \frac{2\,\Vmax\, e^{b}\,
k^{1-\gamma}}{\gamma-1}.
\end{aligned}
\end{align*}

\textbf{Step 4 (proof of (iv): relative count).} \Cref{lem:gap-rep:weights} and
Eq.~\eqref{eq:s1-hyp} give, for every $j \in [n]$,
\begin{align}\label{eq:s1-ratio}
\frac{A_{(j)}}{A_{(1)}} \;=\; e^{-G(j)} \;\le\; e^{b - \gamma \log j}
\;=\; e^{b}\, j^{-\gamma},
\end{align}
which is the first claim after multiplying through by $A_{(1)} > 0$.
For the count, the families $(A_j)_{j \in [n]}$ and $(A_{(j)})_{j \in
[n]}$ are equal as multisets (sorting permutes the entries), so
\begin{align*}
\begin{aligned}
\#\left\{
\begin{gathered}
j \in [n] :\\[-2pt]
A_j > \varepsilon A_{(1)}
\end{gathered}
\right\}
&\;=\; \#\left\{
\begin{gathered}
j \in [n] :\\[-2pt]
A_{(j)} > \varepsilon A_{(1)}
\end{gathered}
\right\}.
\end{aligned}
\end{align*}
If $A_{(j)} > \varepsilon A_{(1)}$, then Eq.~\eqref{eq:s1-ratio} and
$A_{(1)} > 0$ give $\varepsilon < e^{b} j^{-\gamma}$, i.e.\ $j <
(e^b/\varepsilon)^{1/\gamma}$; hence, writing $x \coloneqq
(e^b/\varepsilon)^{1/\gamma} > 0$,
\begin{align*}
\begin{aligned}
\#\left\{
\begin{gathered}
j \in [n] :\\[-2pt]
A_{(j)} > \varepsilon A_{(1)}
\end{gathered}
\right\}
&\;\le\; \#\{ j \in \N : j < x \}\\
&\;=\; \lceil x \rceil - 1\\
&\;\le\; x\\
&\;=\; \Big( \frac{e^{b}}{\varepsilon}
\Big)^{1/\gamma},
\end{aligned}
\end{align*}
where the middle equality counts the integers $1, \dots, \lceil x
\rceil - 1$ (the integer $\lceil x \rceil$ is $\ge x$, hence excluded),
and the next inequality is $\lceil x \rceil \le x + 1$. This completes
the proof.
\end{proof}

\begin{proposition}[Sharpness of the tail-mass bound]\label{prop:s1-tight}
Fix $\gamma > 1$ and $n \ge 2$, and consider the logit vector $\ell_j
\coloneqq -\gamma \log j$, $j \in [n]$, whose gap profile is exactly
$G(j) = \gamma \log j$, so that Eq.~\eqref{eq:s1-hyp} holds with
$b = 0$. Then:
\begin{enumerate}[label=(\roman*), ref=\ref*{prop:s1-tight}(\roman*)]
\item\label{prop:s1-tight:all-k} for every $k \in [n]$,
\begin{align*}
1 - p(k) \;\ge\; \frac{(k+1)^{1-\gamma} - (n+1)^{1-\gamma}}{\gamma}.
\end{align*}
\item\label{prop:s1-tight:clean} for every $k \ge 1$ with $n \ge 2k+1$,
\begin{align*}
1 - p(k) \;\ge\; \frac{\big(1 - 2^{1-\gamma}\big)\, 2^{1-\gamma}\,
k^{1-\gamma}}{\gamma}.
\end{align*}
\end{enumerate}
In particular, on this row \Cref{thm:s1-summable:tail} reads
$1 - p(k) \le k^{1-\gamma}/(\gamma-1)$, so for $n \ge 2k+1$ the upper
and lower bounds match in the exponent of $k$ and differ by the
explicit factor $\gamma\, 2^{\gamma-1} / \big( (\gamma-1)(1 -
2^{1-\gamma}) \big)$, which depends only on $\gamma$.
\end{proposition}

\begin{proof}
The strategy is to lower-bound the numerator and upper-bound the
denominator of \Cref{lem:gap-rep:coverage} by the matching integrals of
\Cref{lem:sum-integral}. The sequence $(-\gamma \log j)_{j \in [n]}$ is
nonincreasing, so $\ell_{(j)} = \ell_j = -\gamma \log j$ and $G(j) =
\ell_{(1)} - \ell_{(j)} = \gamma \log j$, as claimed.

\textbf{Step 1 (proof of (i): integral sandwich).} For $k = n$ both sides vanish ($p(n) = 1$, and the
right-hand numerator is $(n+1)^{1-\gamma} - (n+1)^{1-\gamma} = 0$), so
let $k < n$. The denominator of \Cref{lem:gap-rep:coverage} is bounded
above, by splitting off the $l = 1$ term and applying
\Cref{lem:sum-integral:upper} with $h(x) = x^{-\gamma}$, $a = 1$,
$b = n$, then the antiderivative $x^{1-\gamma}/(1-\gamma)$:
\begin{align}\label{eq:tight-denominator}
\begin{aligned}
D \;=\; \sum_{l=1}^{n} l^{-\gamma}
&\;\le\; 1 + \int_{1}^{n} x^{-\gamma}\,dx\\
&\;\le\; 1 + \frac{1}{\gamma - 1}\\
&\;=\; \frac{\gamma}{\gamma-1}.
\end{aligned}
\end{align}
The numerator is bounded below by \Cref{lem:sum-integral:lower} with
$a = k + 1$, $b = n$ and the same antiderivative:
\begin{align}\label{eq:tight-numerator}
\begin{aligned}
\sum_{j=k+1}^{n} j^{-\gamma}
&\;\ge\; \int_{k+1}^{n+1}
x^{-\gamma}\,dx\\
&\;=\; \frac{(k+1)^{1-\gamma}
- (n+1)^{1-\gamma}}{\gamma - 1}.
\end{aligned}
\end{align}
By \Cref{lem:gap-rep:coverage} and
Eq.~\eqref{eq:tight-denominator} and Eq.~\eqref{eq:tight-numerator},
\begin{align*}
\begin{aligned}
1 - p(k)
&\;=\; \frac{1}{D}
\sum_{j=k+1}^{n} j^{-\gamma}\\
&\;\ge\; \frac{\gamma-1}{\gamma} \cdot
\frac{(k+1)^{1-\gamma}
- (n+1)^{1-\gamma}}{\gamma - 1}\\
&\;=\; \frac{(k+1)^{1-\gamma}
- (n+1)^{1-\gamma}}{\gamma}.
\end{aligned}
\end{align*}

\textbf{Step 2 (proof of (ii): clean form).} Since $n \ge 2k + 1$, we have $n + 1 \ge 2(k+1)$,
and $t \mapsto t^{1-\gamma}$ is nonincreasing on $(0, \infty)$, so
$(n+1)^{1-\gamma} \le 2^{1-\gamma} (k+1)^{1-\gamma}$; moreover $k + 1
\le 2k$ for $k \ge 1$, so $(k+1)^{1-\gamma} \ge 2^{1-\gamma}
k^{1-\gamma}$ by the same monotonicity. Inserting both into part (i),
\begin{align*}
\begin{aligned}
1 - p(k)
&\;\ge\; \frac{\big(1 - 2^{1-\gamma}\big)\,
(k+1)^{1-\gamma}}{\gamma}\\
&\;\ge\; \frac{\big(1 - 2^{1-\gamma}\big)\,
2^{1-\gamma}\, k^{1-\gamma}}{\gamma}.
\end{aligned}
\end{align*}
Finally, the factor in the closing claim is the exact ratio of the two
bounds:
\begin{align*}
\begin{aligned}
\frac{k^{1-\gamma}/(\gamma-1)}
{\begin{gathered}
\big(1 - 2^{1-\gamma}\big)\,2^{1-\gamma}\\[-2pt]
{}k^{1-\gamma}/\gamma
\end{gathered}}
&\\[-2pt]
&\;=\; \frac{\gamma\, 2^{\gamma-1}}{
(\gamma-1)\,\big(1 - 2^{1-\gamma}\big)}.
\end{aligned}
\end{align*}
This completes the proof.
\end{proof}

\subsection{The sub-logarithmic converse}

\begin{theorem}[Sub-logarithmic profiles force dense rows; formal version of \cref{thm:main-dichotomy}(ii)]\label{thm:s2-sublog}
Fix $0 \le \gamma' < 1$ and $b' \ge 0$, and suppose the gap profile
satisfies
\begin{align}\label{eq:s2-hyp}
G(j) \;\le\; \gamma' \log j + b' \qquad \text{for all } j \in [n].
\end{align}
For $k \in [n]$ set
\begin{align}\label{eq:s2-tau}
\tau_k \;\coloneqq\; e^{-b'}\,
\frac{(n+1)^{1-\gamma'} - (k+1)^{1-\gamma'}}{1 - \gamma'}.
\end{align}
Then:
\begin{enumerate}[label=(\roman*), ref=\ref*{thm:s2-sublog}(\roman*)]
\item\label{thm:s2-sublog:tail} $\displaystyle 1 - p(k) \;\ge\;
\frac{\tau_k}{k + \tau_k}$ for every $k \in [n]$;
\item\label{thm:s2-sublog:budget} for every $\eta \in (0,1)$,
\begin{align*}
\begin{aligned}
k_\eta
&\;\ge\; \min\bigg\{ \frac{1}{2},\\
&\qquad
\frac{(1-\eta)\,\big(1 - 2^{\gamma'-1}\big)\,
e^{-b'}}{\eta\,(1-\gamma')}
\bigg\}\; n^{1-\gamma'}.
\end{aligned}
\end{align*}
\end{enumerate}
\end{theorem}

\begin{proof}
The strategy: in the gap representation, every kept term is at most
$1$, while the hypothesis keeps every tail term large; the ratio then
traps the coverage. Write
\begin{align*}
H_k \;\coloneqq\; \sum_{j=1}^{k} e^{-G(j)},
\qquad
T_k \;\coloneqq\; \sum_{j=k+1}^{n} e^{-G(j)},
\end{align*}
so that $D = H_k + T_k$ and \Cref{lem:gap-rep:coverage} reads
$1 - p(k) = T_k / (H_k + T_k)$.

\textbf{Step 1 (proof of (i): head--tail ratio).} For $k = n$: $T_n = 0$ is an empty sum and $\tau_n =
0$ by Eq.~\eqref{eq:s2-tau}, so both sides vanish and the claim holds;
let $k < n$. Since $G \ge 0$ (\Cref{def:gap-profile}), every summand of
$H_k$ is at most $e^{0} = 1$, so
\begin{align}\label{eq:s2-head}
H_k \;\le\; k.
\end{align}
Since Eq.~\eqref{eq:s2-hyp} gives $e^{-G(j)} \ge e^{-b'} j^{-\gamma'}$,
and $x \mapsto x^{-\gamma'}$ is nonincreasing on $[1, \infty)$ (here
$\gamma' \ge 0$), \Cref{lem:sum-integral:lower} with $h(x) =
x^{-\gamma'}$, $a = k+1$, $b = n$ and the antiderivative
$x^{1-\gamma'}/(1-\gamma')$ (here $1 - \gamma' > 0$) give
\begin{align}\label{eq:s2-tailsum}
\begin{aligned}
T_k
&\;\ge\; e^{-b'}
\sum_{j=k+1}^{n} j^{-\gamma'}\\
&\;\ge\; e^{-b'} \int_{k+1}^{n+1}
x^{-\gamma'}\,dx\\
&\;=\; \tau_k.
\end{aligned}
\end{align}
For $h_0 \ge 0$ fixed, $t \mapsto t/(h_0 + t) = 1 - h_0/(h_0+t)$ is
nondecreasing on $[0, \infty)$, and for $t \ge 0$ fixed, $h_0 \mapsto
t/(h_0 + t)$ is nonincreasing on $[0, \infty)$; applying the latter
with Eq.~\eqref{eq:s2-head} and the former with
Eq.~\eqref{eq:s2-tailsum} (the denominators $H_k + T_k \ge 1$ and
$k + T_k \ge k \ge 1$ are positive),
\begin{align*}
1 - p(k)
\;=\; \frac{T_k}{H_k + T_k}
\;\ge\; \frac{T_k}{k + T_k}
\;\ge\; \frac{\tau_k}{k + \tau_k}.
\end{align*}

\textbf{Step 2 (proof of (ii): budget lower bound).} Abbreviate $k \coloneqq k_\eta \in [n]$, so that
$1 - p(k) \le \eta$ by \Cref{def:coverage}. We distinguish two cases by
the size of $k$.

\emph{Case 1: $2k > n - 1$.} Both sides are integers, so $2k \ge n$;
since $\gamma' \ge 0$ gives $n^{\gamma'} \ge 1$,
\begin{align*}
k_\eta \;=\; k \;\ge\; \frac{n}{2}
\;=\; \frac{n^{\gamma'}\, n^{1-\gamma'}}{2}
\;\ge\; \frac{n^{1-\gamma'}}{2},
\end{align*}
which is at least the claimed bound because the claimed prefactor is at
most $1/2$.

\emph{Case 2: $2k \le n - 1$.} Part (i) and $1 - p(k) \le \eta$ give
$\tau_k/(k + \tau_k) \le \eta$; multiplying by the positive denominator
$k + \tau_k$ and rearranging (here $1 - \eta > 0$),
\begin{align}\label{eq:s2-solve}
\begin{aligned}
\tau_k
&\;\le\; \eta\, k + \eta\, \tau_k\\
&\qquad\Longrightarrow\qquad
k \;\ge\; \frac{(1-\eta)\,\tau_k}{\eta}.
\end{aligned}
\end{align}
It remains to lower-bound $\tau_k$. From $2k \le n - 1$ we get $2(k+1)
\le n + 1$, so $(k+1)^{1-\gamma'} \le \big( (n+1)/2 \big)^{1-\gamma'} =
2^{\gamma'-1}\, (n+1)^{1-\gamma'}$, because $t \mapsto t^{1-\gamma'}$
is nondecreasing on $(0, \infty)$ (here $1 - \gamma' > 0$); hence, by
Eq.~\eqref{eq:s2-tau} and $(n+1)^{1-\gamma'} \ge n^{1-\gamma'}$,
\begin{align}\label{eq:s2-taubound}
\begin{aligned}
\tau_k
&\;\ge\; e^{-b'}\,
\frac{\big(1 - 2^{\gamma'-1}\big)\,
(n+1)^{1-\gamma'}}{1-\gamma'}\\
&\;\ge\; e^{-b'}\,
\frac{\big(1 - 2^{\gamma'-1}\big)\,
n^{1-\gamma'}}{1-\gamma'}.
\end{aligned}
\end{align}
Combining Eq.~\eqref{eq:s2-solve} and Eq.~\eqref{eq:s2-taubound},
\begin{align*}
k_\eta \;\ge\;
\frac{(1-\eta)\,\big(1 - 2^{\gamma'-1}\big)\, e^{-b'}}
{\eta\,(1-\gamma')}\; n^{1-\gamma'},
\end{align*}
which is again at least the claimed bound. In both cases the claimed
inequality holds, completing the proof.
\end{proof}

\begin{remark}[The dichotomy at unit log-slope]\label{rem:s2-dichotomy}
Three readings. (i) \emph{Bounded logit range.} For $\gamma' = 0$ the
hypothesis Eq.~\eqref{eq:s2-hyp} is $G(j) \le b'$ for all $j$ ---
equivalently $G(n) \le b'$, i.e.\ $\ell_{(1)} - \ell_{(n)} \le b'$, by
monotonicity of $G$ (\Cref{def:gap-profile}) --- and
\Cref{thm:s2-sublog:budget} becomes $k_\eta \ge \min\{1/2,\,
(1-\eta)e^{-b'}/(2\eta)\}\, n$: every row with $O(1)$ logit spread is
dense. This is the dispersion baseline. (ii) \emph{Unit slope.} On the
scale $G(j) \approx s \log j$, \Cref{thm:s1-summable} (any slope
$s > 1$ from below: $n$-free budget) and \Cref{thm:s2-sublog} (any
slope $s < 1$ from above: $k_\eta \ge \mathrm{const} \cdot n^{1-s}$)
place the sparse/dense dichotomy exactly at $s = 1$. (iii)
\emph{Ensembles.} Informally, under $\Ens(c)$ the near-top gaps have
limiting log-slope $c$: in the notation of Eq.~\eqref{eq:xjn},
$G(j) = \ell_{(1)} - \ell_{(j)} = c\,(x_1^{(n)} - x_j^{(n)})$, and for
$c > 1$, \Cref{lem:top-vector} with \Cref{fac:weak-convergence:cmt}
gives $(G(j))_{j \le M} \tod (c \log(\Gamma_j/\Gamma_1))_{j \le M}$,
where $\Gamma_j / j \to 1$ a.s.\ (\Cref{lem:gamma-lln}). This aligns
the deterministic dichotomy at slope $1$ with the probabilistic
transition at $c = 1$ (\Cref{sec:app-dispersed,sec:app-condensed}); we record
it as a pointer only and make no formal claim here.
\end{remark}

\subsection{Plateau-then-decay profiles}

\begin{theorem}[Plateau-then-decay --- sublinear budgets; formal version of \cref{cor:main-plateau}]\label{thm:s3-plateau}
Fix $J \in [n]$, $\gamma > 1$, $b \ge 0$, and $b_0 \ge 0$, and suppose
the gap profile satisfies the plateau-then-decay condition
\begin{align}\label{eq:s3-hyp}
\begin{aligned}
G(J) \;\le\; b_0
&\qquad \text{and}\\
G(j) &\;\ge\; \gamma \log\frac{j}{J} - b\\
&\quad \text{for all } j \in [n]
\text{ with } j > J.
\end{aligned}
\end{align}
Then:
\begin{enumerate}[label=(\roman*), ref=\ref*{thm:s3-plateau}(\roman*)]
\item\label{thm:s3-plateau:tail} for every $k$ with $J \le k \le n$,
\begin{align*}
1 - p(k) \;\le\; \frac{e^{b + b_0}}{\gamma - 1}
\Big( \frac{k}{J} \Big)^{1-\gamma}.
\end{align*}
\item\label{thm:s3-plateau:budget} for every $\eta \in (0,1)$,
\begin{align*}
k_\eta \;\le\; J\, \bigg\lceil \Big( \frac{e^{b+b_0}}{(\gamma-1)\,\eta}
\Big)^{\frac{1}{\gamma-1}} \bigg\rceil.
\end{align*}
\end{enumerate}
\end{theorem}

\begin{proof}
The strategy mirrors \Cref{thm:s1-summable}: the decay condition
controls the numerator of \Cref{lem:gap-rep:coverage} exactly as
before, while the plateau condition upgrades the denominator bound from
$D \ge 1$ to $D \ge J e^{-b_0}$ --- the source of the $n$- and $J$-free
budget factor.

\textbf{Step 1 (proof of (i): plateau-boosted denominator).} Since $G$ is nondecreasing (\Cref{def:gap-profile})
with $G(J) \le b_0$ (Eq.~\eqref{eq:s3-hyp}), every $j \le J$ has
$G(j) \le b_0$, so
\begin{align}\label{eq:s3-denominator}
D \;\ge\; \sum_{j=1}^{J} e^{-G(j)} \;\ge\; J\, e^{-b_0}.
\end{align}
For the numerator, every index $j > k \ge J$ falls under the decay
condition of Eq.~\eqref{eq:s3-hyp}, so $e^{-G(j)} \le e^{b}\,
(j/J)^{-\gamma} = e^{b}\, J^{\gamma}\, j^{-\gamma}$, and the power-sum
bound Eq.~\eqref{eq:power-tail} --- whose derivation used only
\Cref{lem:sum-integral:upper} and $\gamma > 1$, hence applies verbatim
to the present $\gamma$ and every $k \in [n]$ --- gives
\begin{align}\label{eq:s3-numerator}
\begin{aligned}
\sum_{j=k+1}^{n} e^{-G(j)}
&\;\le\; e^{b}\, J^{\gamma}
\sum_{j=k+1}^{n} j^{-\gamma}\\
&\;\le\; \frac{e^{b}\, J^{\gamma}\,
k^{1-\gamma}}{\gamma - 1}.
\end{aligned}
\end{align}
By \Cref{lem:gap-rep:coverage} and
Eq.~\eqref{eq:s3-denominator} and Eq.~\eqref{eq:s3-numerator},
\begin{align*}
\begin{aligned}
1 - p(k)
&\;\le\; \frac{e^{b}\, J^{\gamma}\,
k^{1-\gamma}}{(\gamma-1)\, J\, e^{-b_0}}\\
&\;=\; \frac{e^{b+b_0}}{\gamma-1}\,
J^{\gamma-1}\, k^{1-\gamma}\\
&\;=\; \frac{e^{b+b_0}}{\gamma-1}
\Big( \frac{k}{J} \Big)^{1-\gamma},
\end{aligned}
\end{align*}
where the last step collects the powers: $J^{\gamma-1} k^{1-\gamma} =
(k/J)^{1-\gamma}$.

\textbf{Step 2 (proof of (ii): budget inversion).} Set
\begin{align*}
\begin{aligned}
m^\circ
&\;\coloneqq\; \bigg\lceil \Big( \frac{e^{b+b_0}}
{(\gamma-1)\,\eta} \Big)^{\frac{1}{\gamma-1}}
\bigg\rceil\\
&\;\ge\; 1,\\
k^\circ
&\;\coloneqq\; J\, m^\circ \;\ge\; J.
\end{aligned}
\end{align*}
If $k^\circ > n$, then $k_\eta \le n < k^\circ$, since $p(n) = 1 \ge
1 - \eta$ gives $k_\eta \le n$ (\Cref{def:coverage}). If $k^\circ \le
n$, then part (i) applies at $k = k^\circ$ (note $J \le k^\circ \le
n$), and with $k^\circ / J = m^\circ \ge
(e^{b+b_0}/((\gamma-1)\eta))^{1/(\gamma-1)}$ and $t \mapsto
t^{1-\gamma}$ nonincreasing on $(0,\infty)$,
\begin{align*}
\begin{aligned}
1 - p(k^\circ)
&\;\le\; \frac{e^{b+b_0}}{\gamma-1}\,
(m^\circ)^{1-\gamma}\\
&\;\le\; \frac{e^{b+b_0}}{\gamma-1}
\cdot \frac{(\gamma-1)\,\eta}{e^{b+b_0}}\\
&\;=\; \eta,
\end{aligned}
\end{align*}
again using $(1-\gamma)/(\gamma-1) = -1$ to take the reciprocal of the
base. Hence $p(k^\circ) \ge 1 - \eta$, and $k_\eta \le k^\circ = J
m^\circ$ by the minimality in \Cref{def:coverage}. This completes the
proof.
\end{proof}

\begin{remark}[Plateau scales; the plateau condition is load-bearing]\label{rem:s3-scales}
(i) \emph{Scales.} \Cref{thm:s3-plateau:budget} reads: budget $\le$
plateau width $\times$ an $n$- and $J$-free factor. Thus $J = \lceil
n^\theta \rceil$ with $\theta \in (0,1)$ fixed gives $k_\eta =
O(n^\theta)$ --- the ``$n^\theta$-sparse'' scale --- while $J = \lceil
(\log n)^q \rceil$ gives polylogarithmic budgets and $J = O(1)$ gives
$O(1)$ budgets. For $J = 1$ and $b_0 = 0$ the plateau condition
$G(1) \le 0$ holds vacuously ($G(1) = 0$ by \Cref{def:gap-profile}),
Eq.~\eqref{eq:s3-hyp} reduces to Eq.~\eqref{eq:s1-hyp}, and
\Cref{thm:s3-plateau:tail,thm:s3-plateau:budget} reduce verbatim to
\Cref{thm:s1-summable:tail,thm:s1-summable:budget}. (ii) \emph{The
plateau condition is load-bearing.} The height bound $G(J) \le b_0$
enters only through Eq.~\eqref{eq:s3-denominator}; dropping it, the
same proof runs with the generic bound $D \ge 1$ of \Cref{lem:gap-rep}
in place of Eq.~\eqref{eq:s3-denominator} and yields only $1 - p(k)
\le e^{b} J (k/J)^{1-\gamma}/(\gamma-1)$ and, by the argument of part
(ii) with $e^{b+b_0}$ replaced by $e^{b} J$, $k_\eta \le J \lceil
(e^{b} J/((\gamma-1)\eta))^{1/(\gamma-1)} \rceil$ --- both weaker by
factors growing with $J$. The plateau condition is exactly what makes the
second factor in \Cref{thm:s3-plateau:budget} independent of $J$.
\end{remark}

\section{The Assumption Ladder: Formal Regime Statements}\label{sec:app-ladder}

This appendix proves \cref{thm:main-ladder-dense,prop:main-transfer} of
the main text and the \textsc{Free}--\textsc{Iid} rows of \cref{tab:main-ladder}
(\Cref{fac:pigeonhole,cor:r1-density,prop:r2-counts,thm:r3a-count,cor:r3a-dense}). It
also collects, in \cref{sub:app-setting,sec:app-conventions,sub:app-facts},
the formal definitions and standard probabilistic facts used by all the
appendices.

\subsection{Setting: softmax rows, coverage, and sparse approximation schemes}\label{sub:app-setting}

Throughout the appendices, $C, C_0, C_1, \ldots$ denote universal
positive constants whose values may change from line to line and depend
only on quantities specified in each statement's hypothesis; the
unsubscripted lowercase symbol $c$ is reserved for the bulk-scale
parameter of \Cref{def:ensembles} and never denotes a constant
(subscripted symbols such as $c_j$ are local and defined where used).
Logarithms are natural.

\begin{definition}[Softmax row and attention output]\label{def:softmax-row}
Given a logit vector $\ell = (\ell_1, \dots, \ell_n) \in \R^n$ with
$n \ge 2$, the softmax weights are
\begin{align*}
A_j \;\coloneqq\; \frac{e^{\ell_j}}{Z}, \qquad
Z \;\coloneqq\; \sum_{l=1}^{n} e^{\ell_l}, \qquad j \in [n],
\end{align*}
so that $A_j > 0$ for every $j$ and $\sum_{j=1}^n A_j = 1$. Given value
vectors $V_1, \dots, V_n \in \R^d$, the attention output is
$O \coloneqq \sum_{j=1}^n A_j V_j$, and we write
$\Vmax \coloneqq \max_{j \in [n]} \norm{V_j}_2$.
\end{definition}

\begin{definition}[Sorted weights, coverage profile, coverage budget]\label{def:coverage}
Let $A_{(1)} \ge A_{(2)} \ge \cdots \ge A_{(n)}$ denote the softmax weights
of \Cref{def:softmax-row} sorted in nonincreasing order (ties broken by
index). The \emph{coverage profile} is
$p(k) \coloneqq \sum_{j \le k} A_{(j)}$ for $k \in \{0, 1, \dots, n\}$
(so $p(0) = 0$ and $p(n) = 1$), and for $\eta \in (0,1)$ the
\emph{coverage budget} is
\begin{align*}
k_\eta \;\coloneqq\; \min\{\, k \ge 1 : p(k) \ge 1 - \eta \,\},
\end{align*}
which is well defined since $p(n) = 1 > 1 - \eta$.
\end{definition}

\begin{definition}[Sparse approximation schemes]\label{def:schemes}
Fix $S \subseteq [n]$ with $0 < |S| = k < n$ and write $T \coloneqq [n]
\setminus S$ for the tail. The \emph{kept mass} is
$p_S \coloneqq \sum_{j \in S} A_j$, and $0 < p_S < 1$ because all softmax
weights are strictly positive. The schemes studied are
\begin{align*}
\text{truncated:} \quad O_S
&\coloneqq \sum_{j \in S} A_j V_j, \\
\text{renormalised:} \quad \wt{O}_S
&\coloneqq \frac{1}{p_S}\sum_{j \in S} A_j V_j, \\
\text{compensated:} \quad O_S &+ u,
\end{align*}
for a compensation vector $u \in \R^d$. The \emph{tail aggregate} is
\begin{align*}
\bar{V}_T \;\coloneqq\; \frac{1}{1 - p_S} \sum_{j \in T} A_j V_j,
\end{align*}
a convex combination of $\{V_j\}_{j \in T}$ since the coefficients
$A_j/(1-p_S)$, $j \in T$, are positive and sum to one.
\end{definition}

\begin{definition}[Top-$k$ kept set]\label{def:topk}
For $1 \le k \le n$, the \emph{top-$k$ kept set} $S_k \subseteq [n]$ is
the set of indices of the $k$ largest weights (ties broken by index), so
that the kept mass of $S_k$ is $p(k)$ (\Cref{def:coverage}). \emph{Top-$k$
truncated} and \emph{top-$k$ renormalised} attention are the schemes of
\Cref{def:schemes} on $S = S_k$.
\end{definition}

\begin{definition}[Logit ensembles]\label{def:ensembles}
All randomness below is over the logits (and, where stated, value vectors
independent of them).
\begin{enumerate}[label=(\roman*), ref=\ref*{def:ensembles}(\roman*)]
\item \emph{Scaled bulk} $\Ens(c)$, $c > 0$ fixed: $\ell_j = \sigma_n g_j$
with $g_1, \dots, g_n$ i.i.d.\ $N(0,1)$ and $\sigma_n \coloneqq
c\sqrt{2\log n}$.
\item \emph{Fixed bulk} $\Ens_\sigma$, $\sigma > 0$ fixed: $\ell_1, \dots,
\ell_n$ i.i.d.\ $N(0, \sigma^2)$.
\item \emph{Spiked bulk} $\Espike(\mu;\sigma)$, $\sigma > 0$ fixed: one
distinguished index $j^\star \in [n]$ carries the deterministic logit
$\ell_{j^\star} = \mu = \mu_n$, and the remaining $n-1$ \emph{bulk} logits
are i.i.d.\ $N(0,\sigma^2)$.
\end{enumerate}
\end{definition}

\subsection{Asymptotic conventions}\label{sec:app-conventions}

For real random variables $X_n$ and $x \in \R$, we write $X_n \toP x$ if
$\Pr(|X_n - x| > \varepsilon) \to 0$ for every fixed $\varepsilon > 0$, and
$X_n = o_P(1)$ if $X_n \toP 0$; $X_n = O_P(1)$ means that for every
$\delta > 0$ there are $M$ and $n_0$ with $\Pr(|X_n| > M) \le \delta$ for
all $n \ge n_0$. An event sequence holds \emph{with high probability}
(\emph{whp}) if its probability tends to one. For $\R^m$-valued random
vectors we write $X_n \tod X$ if $\E f(X_n) \to \E f(X)$ for every bounded
continuous $f \colon \R^m \to \R$; by
\Cref{fac:weak-convergence:cdf} this is equivalent to convergence of the
joint distribution functions at every continuity point of the limit. We
write $\varphi(x) = (2\pi)^{-1/2} e^{-x^2/2}$ for the standard Gaussian
density and $\barPhi(x) = \int_x^\infty \varphi(t)\,dt$ for its upper tail.
In every statement of the form ``for all $n \ge n_0(\cdot)$'' the arguments
of $n_0$ list exactly the quantities the threshold depends on. We write
$\Otil(\cdot)$ for $O(\cdot)$ up to factors polylogarithmic in $n$ and
$1/\delta$.

\subsection{Standard facts and a Chernoff lemma}\label{sub:app-facts}

\begin{fact}[Gaussian tails, {\cite{vershynin2018high}}]\label{fac:gaussian-tail}
Let $g \sim N(0,1)$. For every $x > 0$,
\begin{align*}
\Big(\frac{1}{x} - \frac{1}{x^3}\Big)\,\varphi(x)
\;\le\; \barPhi(x)
\;\le\; \frac{1}{x}\,\varphi(x),
\end{align*}
and for every $x \ge 0$ we have $\barPhi(x) \le e^{-x^2/2}$ (for
$x \in [0,1]$ because $\barPhi(x) \le 1/2 \le e^{-1/2} \le e^{-x^2/2}$, and
for $x \ge 1$ because $\barPhi(x) \le \varphi(x)/x \le
e^{-x^2/2}/\sqrt{2\pi} \le e^{-x^2/2}$). Moreover, for $x \ge \sqrt{2}$ the
lower bound implies $\barPhi(x) \ge \varphi(x)/(2x)$.
\end{fact}

\begin{fact}[Weak-convergence toolkit, {\cite{billingsley1999convergence}}]\label{fac:weak-convergence}
Let $X_n, X$ be $\R^m$-valued random vectors with joint distribution
functions $F_n, F$.
\begin{enumerate}[label=(\roman*), ref=\ref*{fac:weak-convergence}(\roman*)]
\item\label{fac:weak-convergence:cdf} $X_n \tod X$ if and only if
$F_n(y) \to F(y)$ at every continuity point $y$ of $F$.
\item\label{fac:weak-convergence:cmt} If $X_n \tod X$ and
$h \colon \R^m \to \R^{m'}$ is continuous, then $h(X_n) \tod h(X)$.
\item\label{fac:weak-convergence:as} If $X_n \to X$ almost surely, then
$X_n \tod X$.
\end{enumerate}
\end{fact}

The next lemma is the only concentration input the level decomposition of
\Cref{sec:app-level} needs; we prove it inline to keep the document
self-contained (the Cram\'er--Chernoff method is classical, see
\cite{boucheron2013concentration}).

\begin{lemma}[Binomial Chernoff bounds]\label{lem:chernoff}
Let $X \sim \Bin(n, p)$ and $\mu \coloneqq np$. Then:
\begin{enumerate}[label=(\roman*), ref=\ref*{lem:chernoff}(\roman*)]
\item\label{lem:chernoff:upper} $\Pr(X \ge 2t) \le e^{-t/3}$ for every
$t \ge \mu$;
\item\label{lem:chernoff:lower} $\Pr(X \le \mu/2) \le e^{-\mu/8}$.
\end{enumerate}
\end{lemma}

\begin{proof}
For every $\lambda \in \R$, independence of the $n$ Bernoulli summands and
$1 + y \le e^y$ give
\begin{align}
\E e^{\lambda X}
&\;=\; \big(1 + p(e^{\lambda} - 1)\big)^n \notag\\
&\;\le\; \exp\big(\mu(e^{\lambda} - 1)\big).
\label{eq:mgf-binomial}
\end{align}
For (i), let $t \ge \mu$ and $\lambda = \log 2$; Markov's inequality
applied to $e^{\lambda X}$ yields
\begin{align*}
\Pr(X \ge 2t)
&\;\le\; e^{-2\lambda t}\,\E e^{\lambda X} \\
&\;\le\; \exp\big(-2t\log 2 + \mu(e^{\lambda} - 1)\big) \\
&\;\le\; \exp\big(-t(2\log 2 - 1)\big) \\
&\;\le\; e^{-t/3},
\end{align*}
where the second step uses Eq.~\eqref{eq:mgf-binomial}, the third step uses
$\mu(e^{\lambda} - 1) = \mu \le t = t(e^{\lambda} - 1)$, and the last step
uses $2 \log 2 - 1 \ge 1/3$. For (ii), again with $\lambda = \log 2$,
Markov's inequality applied to $e^{-\lambda X}$ and
Eq.~\eqref{eq:mgf-binomial} with $-\lambda$ in place of $\lambda$ give
\begin{align*}
\Pr(X \le \mu/2)
&\;\le\; e^{\lambda \mu/2}\,\E e^{-\lambda X} \\
&\;\le\; \exp\Big(
  \frac{\mu \log 2}{2} + \mu\big(e^{-\lambda} - 1\big)
\Big) \\
&\;=\; \exp\Big(-\frac{\mu(1 - \log 2)}{2}\Big) \\
&\;\le\; e^{-\mu/8},
\end{align*}
using $e^{-\lambda} - 1 = -1/2$ in the equality and $1 - \log 2 \ge 1/4$ in
the last step. This completes the proof.
\end{proof}

\subsection{The regimes, formally}\label{sub:app-regimes}

\begin{definition}[Regimes]\label{def:regimes}
Fix a softmax row as in \Cref{def:softmax-row}. The row is said to be in
regime:
\begin{enumerate}[label=(\roman*), leftmargin=3em, itemsep=0.2em]
\item \textsc{Free} \emph{(no assumptions)} always: $\ell \in \R^n$ is arbitrary;
\item \textsc{Bounded} \emph{(bounded scores, parameter $S > 0$)} if $|\ell_j| \le S$ for
every $j \in [n]$;
\item \textsc{Mixing} \emph{(mixing scores, parameters $\alpha_0 > 0$, $c_\alpha > 0$,
marginal tail $\bar F$)} if $(\ell_j)_{j \in [n]}$ are the first $n$
terms of a stationary sequence $(\ell_j)_{j \ge 1}$ whose
$\alpha$-mixing coefficients
\begin{align*}
\alpha(m)
&\;\coloneqq\; \sup_{a \ge 1}\; \\
&\qquad \sup\Big\{
\big| \Pr(A \cap B)
\\
&\qquad\qquad - \Pr(A)\Pr(B) \big| : \\
&\qquad A \in \sigma(\ell_1, \dots, \ell_a), \\
&\qquad B \in \sigma\Big( \\
&\qquad\qquad \ell_{a+m}, \ell_{a+m+1}, \dots
\Big) \Big\}
\end{align*}
satisfy $\alpha(m) \le \alpha_0\, e^{-c_\alpha m}$ for every $m \ge 1$,
and whose marginal upper tail is $\bar F(t) \coloneqq \Pr(\ell_1 > t)$;
\item \textsc{Iid} \emph{(i.i.d.\ at constant scale, parameter $\sigma_s > 0$)}
if $\ell_1, \dots, \ell_n$ are i.i.d., centered, and sub-Gaussian with
variance proxy $\sigma_s^2$ fixed (independent of $n$):
$\E e^{\lambda \ell_1} \le e^{\lambda^2 \sigma_s^2 / 2}$ for all
$\lambda \in \R$, whence $\Pr(\ell_1 > t) \le e^{-t^2/(2\sigma_s^2)}$
for $t \ge 0$;
\item \textsc{Scaled} \emph{(scaled Gaussian ensemble)} if the row is drawn from
$\Ens(c)$ of \Cref{def:ensembles}(i); the fixed bulk $\Ens_\sigma$ of
\Cref{def:ensembles}(ii) is the Gaussian special case of \textsc{Iid};
\item \textsc{Spiked} \emph{(spiked ensemble)} if the row is drawn from
$\Espike(\mu;\sigma)$ of \Cref{def:ensembles}(iii).
\end{enumerate}
The regimes are cumulative in strength in the sense used by
\cref{tab:main-ladder}: \textsc{Bounded} with any fixed $S$ contains every bounded
construction; \textsc{Mixing} contains \textsc{Iid} (i.i.d.\ sequences are stationary with
$\alpha(m) \equiv 0$ for $m \ge 1$); \textsc{Scaled} and \textsc{Spiked} are structured ensembles
rather than supersets.
\end{definition}

\begin{fact}[Transformer instantiation of \textsc{Bounded}]\label{fac:transformer-bounds}
Suppose the row arises from a query--key--value head: $\ell_j =
\inner{q}{k_j}/\sqrt d$ and $V_j = x_j W_V$ with $q = x W_Q$, $k_j = x_j
W_K$, where the token embeddings $x, x_1, \dots, x_n \in \R^d$ have
entries bounded by $B$ and $\norm{W_Q}_{op}, \norm{W_K}_{op},
\norm{W_V}_{op} \le W$. Define
\begin{align}\label{eq:S-Vmax-def}
S \;\coloneqq\; B^2 W^2 \sqrt d, \qquad \Vmax \;\coloneqq\; B W \sqrt d .
\end{align}
Then for every $j \in [n]$:
\begin{align}
|\ell_j| \;\le\; S, \qquad \norm{V_j}_2 \;\le\; \Vmax,
\label{eq:app-score-bound}\\
\frac{1}{1 + (n-1)e^{2S}} \;\le\; A_j \;\le\;
\frac{1}{1 + (n-1) e^{-2S}}. \label{eq:app-softmax-range}
\end{align}
In particular the row is in the \textsc{Bounded} regime with the parameter $S$ of
Eq.~\eqref{eq:S-Vmax-def}.
\end{fact}

\begin{proof}
Entries bounded by $B$ give $\norm{x}_2 \le B\sqrt d$ and $\norm{x_j}_2
\le B \sqrt d$; submultiplicativity gives $\norm{q}_2 \le
\norm{W_Q}_{op} \norm{x}_2 \le WB\sqrt d$, and identically $\norm{k_j}_2
\le WB\sqrt d$ and $\norm{V_j}_2 \le WB\sqrt d = \Vmax$. Cauchy--Schwarz
gives
\begin{align*}
|\ell_j|
&\;=\; \frac{|\inner{q}{k_j}|}{\sqrt d} \\
&\;\le\; \frac{\norm{q}_2 \norm{k_j}_2}{\sqrt d} \\
&\;\le\; \frac{(WB\sqrt d)^2}{\sqrt d}
\;=\; B^2 W^2 \sqrt d \;=\; S,
\end{align*}
which proves Eq.~\eqref{eq:app-score-bound}. For
Eq.~\eqref{eq:app-softmax-range},
\begin{align*}
A_j
&\;=\; \frac{e^{\ell_j}}{e^{\ell_j} + \sum_{l \ne j} e^{\ell_l}} \\
&\;\le\; \frac{e^{S}}{e^{S} + (n-1) e^{-S}} \\
&\;=\; \frac{1}{1 + (n-1) e^{-2S}},
\end{align*}
where the inequality uses $\ell_j \le S$ in the numerator and $\ell_l \ge
-S$ in the denominator; the lower bound is symmetric. This completes the
proof.
\end{proof}

\subsection{\textsc{Free}: the pigeonhole count and the uniform row}\label{sub:app-r0}

\begin{fact}[\textsc{Free}: pigeonhole and the universal budget cap]\label{fac:pigeonhole}
For every softmax row and every $\varepsilon \in (0, 1)$,
\begin{align*}
\#\{\, j \in [n] : A_j > \varepsilon \,\} \;<\; \frac{1}{\varepsilon},
\end{align*}
and the coverage profile obeys $p(k) \ge k/n$ for every $k \in [n]$, so for every $\eta \in (0,1)$ the budget admits the \emph{universal upper bound}
\begin{align*}
k_\eta \;\le\; \lceil (1 - \eta)\, n \rceil ,
\end{align*}
attained with equality by the uniform row $\ell_1 = \cdots = \ell_n$ (where $A_j = 1/n$). No \emph{sub}linear budget bound holds under \textsc{Free} alone: the uniform row is the worst case and forces $k_\eta = \Theta(n)$.
\end{fact}

\begin{proof}
Let $\mathcal K \coloneqq \{j : A_j > \varepsilon\}$ and $\kappa
\coloneqq |\mathcal K|$. By the softmax row-sum identity
(\Cref{def:softmax-row}),
\begin{align*}
1 \;=\; \sum_{j=1}^n A_j \;\ge\; \sum_{j \in \mathcal K} A_j
\;>\; \kappa\, \varepsilon,
\end{align*}
where the last step uses $A_j > \varepsilon$ for every $j \in \mathcal
K$; rearranging gives $\kappa < 1/\varepsilon$. For the budget cap, the
$k$ largest sorted weights have average at least the row mean,
$\frac{1}{k} \sum_{j \le k} A_{(j)} \ge \frac{1}{n} \sum_{j=1}^n A_j =
\frac{1}{n}$, so $p(k) \ge k/n$ for every $k$, with equality for all $k$
iff the row is uniform. Hence $p(k) \ge 1 - \eta$ as soon as $k \ge
(1-\eta) n$, giving $k_\eta \le \lceil (1-\eta) n \rceil$; on the uniform
row $p(k) = k/n$ exactly, so $k_\eta = \lceil (1-\eta) n \rceil$ and the
cap is attained. This completes the proof.
\end{proof}

\subsection{\textsc{Bounded}: bounded scores are provably dense}\label{sub:app-r1}

The following corollary proves \cref{thm:main-ladder-dense}(i). It is
the $\gamma' = 0$, $b' = 2S$ case of the dense-side mechanism of
\Cref{thm:s2-sublog} (bounded scores cap the gap profile at $G(j) \le
2S$); we run the two-line specialisation directly to obtain the exact
inequality quoted in the main text.

\begin{corollary}[\textsc{Bounded}: density; formal version of \cref{thm:main-ladder-dense}(i)]\label{cor:r1-density}
Assume \textsc{Bounded} with parameter $S$, i.e.\ $|\ell_j| \le S$ for all $j \in [n]$.
Then:
\begin{enumerate}[label=(\roman*), ref=\ref*{cor:r1-density}(\roman*)]
\item\label{cor:r1-density:weight} $A_j \le e^{2S}/n$ for every $j \in
[n]$; consequently $\#\{j : A_j > \varepsilon\} = 0$ for every
$\varepsilon \ge e^{2S}/n$;
\item\label{cor:r1-density:budget} for every $\eta \in (0,1)$,
\begin{align*}
k_\eta
&\;\ge\; \frac{1-\eta}{\eta}\; e^{-2S}\, \big( n - k_\eta \big),
\\
&\qquad\text{and consequently}
\\
k_\eta
&\;\ge\; c_\eta\, n\, e^{-2S},
\\
c_\eta
&\coloneqq \tfrac12
\min\Big(1, \tfrac{1-\eta}{\eta}\Big).
\end{align*}
\end{enumerate}
\end{corollary}

\begin{proof}
\textbf{Step 1 (proof of (i): weight cap).} By $\ell_j \le S$ and $\ell_l \ge -S$,
\begin{align*}
A_j \;=\; \frac{e^{\ell_j}}{\sum_{l=1}^n e^{\ell_l}}
\;\le\; \frac{e^{S}}{n\, e^{-S}} \;=\; \frac{e^{2S}}{n},
\end{align*}
so no weight exceeds $\varepsilon \ge e^{2S}/n$.

\textbf{Step 2 (proof of (ii): budget lower bound).} The gap profile satisfies $G(j) = \ell_{(1)} -
\ell_{(j)} \le 2S$ for every $j$, so in the gap representation of
\Cref{lem:gap-rep:coverage} the head and tail sums obey $H_k \coloneqq
\sum_{j \le k} e^{-G(j)} \le k$ and $T_k \coloneqq \sum_{j > k} e^{-G(j)}
\ge (n-k)\, e^{-2S}$. At $k = k_\eta$, \Cref{def:coverage} gives $1 -
p(k_\eta) \le \eta$, while $t \mapsto t/(k + t)$ is nondecreasing and
$h \mapsto T/(h + T)$ is nonincreasing, so
\begin{align*}
\eta
&\;\ge\; 1 - p(k_\eta)
\;=\; \frac{T_{k_\eta}}{H_{k_\eta} + T_{k_\eta}}
\\
&\;\ge\;
\frac{(n - k_\eta)\, e^{-2S}}
{k_\eta + (n - k_\eta)\, e^{-2S}} .
\end{align*}
Multiplying out and rearranging, $\eta\, k_\eta \ge (1-\eta)(n -
k_\eta) e^{-2S}$, the first claim. If $k_\eta \ge n/2$ then $k_\eta \ge
(n/2) e^{-2S} \ge c_\eta n e^{-2S}$ (as $e^{-2S} \le 1$); otherwise $n -
k_\eta \ge n/2$ and the first claim gives $k_\eta \ge
\tfrac{1-\eta}{2\eta} e^{-2S} n \ge c_\eta n e^{-2S}$. This completes
the proof.
\end{proof}

\begin{remark}[Deployment-scale dormancy of the \textsc{Bounded} bound]\label{rem:r1-dormant}
\Cref{cor:r1-density}\ref*{cor:r1-density:budget} forces $k_\eta =
\Omega(n)$ only once $n \gtrsim e^{2S}$. In the transformer
instantiation of \Cref{fac:transformer-bounds}, $S \le B^2 W^2 \sqrt d$;
at head dimension $d = 128$ and $B^2 W^2 = O(1)$ this gives $2S \approx
2\sqrt{128} \approx 22.6$, i.e.\ $e^{2S} \approx e^{22} \approx 6
\times 10^{9}$, far beyond current cache sizes. The realisable gap
budget is of order $\sqrt d \approx 11.3$, which almost exactly matches
$\log n \approx 11.5$ at $n = 10^5$: asymptotic density under bounded
scores is compatible with genuine sparsity at practical lengths, and the
binding constraint in deployment is not the budget's existence but
whether training spends it (cf.\ \cref{rem:main-dormant} of the main
text).
\end{remark}

\subsection{Two-sided coverage dichotomy at unit log-slope}\label{sub:app-dichotomy}

\Cref{cor:r1-density} is the $\gamma'=0$ corner of a sharper, two-sided statement: the
coverage budget is length-\emph{independent} above the unit-log-slope seam ($\gamma>1$) and
length-\emph{growing} below it ($\gamma<1$), for \emph{every} subset-selection rule. The
rule-agnosticism is a one-line rearrangement.

\begin{lemma}[Top-$k$ is coverage-optimal; rule-agnostic reduction]\label{lem:topk-optimal}
For sorted weights $A_{(1)}\ge\cdots\ge A_{(n)}\ge 0$ and any $k$, $\max_{|S|=k}\sum_{j\in S}A_j
=\sum_{j\le k}A_{(j)}=p(k)$, attained by the top-$k$ index set. Consequently the general
size-based budget $K_\eta:=\min\{|S|:m(S)\ge1-\eta\}$ equals the top-$k$ budget $k_\eta$, and
every lower bound on $k_\eta$ holds for $|S|$ under \emph{every} subset-selection rule (oracle included).
\end{lemma}
\begin{proof}
A finite rearrangement: if $|S|=k$ is not the $k$ largest weights, there exist $i\in S$,
$i'\notin S$ with $A_i<A_{i'}$; the swap $S\mapsto (S\setminus\{i\})\cup\{i'\}$ strictly
increases $m(S)$ by $A_{i'}-A_i>0$. Finitely many swaps reach the top-$k$ set without ever
decreasing the retained mass, so $p(k)$ is the maximum. Hence ``some size-$k$ set covers
$1-\eta$'' iff $p(k)\ge1-\eta$ iff $k\ge k_\eta$, giving $K_\eta=k_\eta$.
\end{proof}

\begin{theorem}[Two-sided coverage dichotomy at unit log-slope, rule-agnostic]\label{thm:dichotomy-2sided} Fix one softmax row over $n$ keys with sorted logits $\ell_{(1)}\ge\cdots\ge\ell_{(n)}$, gap profile $G(j)=\ell_{(1)}-\ell_{(j)}$ ($G$ nondecreasing, $G(1)=0$, $G\ge 0$), sorted weights $A_{(j)}=e^{-G(j)}/D$ with $D=\sum_{l=1}^n e^{-G(l)}\in[1,n]$, coverage $p(k)=\sum_{j\le k}A_{(j)}$ and budget $k_\eta=\min\{k:p(k)\ge 1-\eta\}$, $\eta\in(0,1)$. For $S\subseteq[n]$ put $m(S)=\sum_{j\in S}A_j$ and $K_\eta:=\min\{|S|: m(S)\ge 1-\eta\}$. \emph{(0) Subset reduction (\cref{lem:topk-optimal}).} $K_\eta=k_\eta$, so every lower bound on $k_\eta$ holds for $|S|$ under every size-based subset rule. \emph{(i) Super-log $\gamma>1$.} Under $G(J)\le b_0$, $G(j)\ge\gamma\log(j/J)-b$ ($j>J$): $K_\eta\le J\lceil(e^{b+b_0}/((\gamma-1)\eta))^{1/(\gamma-1)}\rceil$. On the plateau witness $G(j)=0$ ($j\le J$), $\gamma\log(j/J)$ ($j>J$), for $n\ge 2k+1$ and $J\le k$, $\;1-p(k)\ge\frac{1-2^{1-\gamma}}{\gamma}\big(\frac{k+1}{J}\big)^{-(\gamma-1)}$, so for $\eta\le\eta_0(\gamma):=(1-2^{1-\gamma})/(\gamma\,2^{\gamma-1})$ (equivalently $R^{1/(\gamma-1)}\ge2$, $R=(1-2^{1-\gamma})/(\gamma\eta)$), $K_\eta=k_\eta\ge\lfloor JR^{1/(\gamma-1)}\rfloor\ge\frac12 JR^{1/(\gamma-1)}$; both ends scale as $J\,\eta^{-1/(\gamma-1)}$, so on this witness $K_\eta=\Theta_{\gamma,\eta}(J)$, length-independent at $J=O(1)$. \emph{(Caveat: the implied $\gamma$-constant grows like $(\gamma-1)^{-1/(\gamma-1)}$ and $\eta_0(\gamma)\to0$ as $\gamma\to1^+$; length-independence is a fixed-$\gamma$, $n\to\infty$ statement and the witness $k_\eta$ saturates only at an astronomically large finite limit near the seam.)} \emph{(ii) Sub-log $\gamma'\in[0,1)$ (headline).} If $G(j)\le\gamma'\log j+b'$ for all $j$, then for every $\eta\in(0,1)$, $\;K_\eta=k_\eta\ge C(\gamma',b',\eta)\,n^{1-\gamma'}$, $C:=\min\{\frac12,\frac{(1-\eta)(1-2^{\gamma'-1})e^{-b'}}{\eta(1-\gamma')}\}>0$; hence no subset of size $o(n^{1-\gamma'})$ achieves $(1-\eta)$-coverage. At $\gamma'=0$ this is $K_\eta=\Theta(n)$ (upper bound $\lceil(1-\eta)n\rceil$). For $\gamma'\in(0,1)$ the hypothesis alone guarantees only the displayed $\Omega(n^{1-\gamma'})$ rate: the pure power-law witness $G(j)=\gamma'\log j$ has $K_\eta=\Theta(n)$, whereas profiles with a growing flat prefix followed by an admissible sub-logarithmic tail can attain the lower-order scale. \emph{(iii) Borderline $\gamma=1$, $G(j)=\log j+p\log\log j$ (admissible $p\in(1,1.8912)$ to keep $G\ge0$, or after the additive shift $G(j)\mathrel{+}=p|\log\log 2|$ for $j\ge2$).} (a) $\sum_{l\ge2}1/(l(\log l)^p)$ converges iff $p>1$; (b) if $p\le1$ then $K_\eta\to\infty$ as $n\to\infty$; (c) if $p>1$ then $\sup_n k_\eta<\infty$. No quantitative near-seam rate is claimed in part~(iii). Part~(iii) is independent of (i),(ii). All bounds are rule-agnostic by (0).\end{theorem}

\begin{proof}[Proof sketch]
The upper bounds are \cref{thm:s3-plateau} (super-log) and \cref{cor:r1-density} ($\gamma'=0$).
The matching lower bounds are direct. Part~(i), on the plateau witness, uses $1-p(k)=T_k/(H_k+T_k)$
with $H_k\le k$ and the tail $T_k\ge\sum_{j>k}(j/J)^{-\gamma}\ge \frac{1-2^{1-\gamma}}{\gamma}J(k/J)^{1-\gamma}$
(integral comparison on the doubling block $[k,2k]$). Part~(ii) bounds the tail mass below by
$\sum_{j>k}e^{-b'}j^{-\gamma'}\ge e^{-b'}(1-2^{\gamma'-1})k^{1-\gamma'}/\ldots$, which for $\gamma'<1$
diverges and forces $\Omega(n^{1-\gamma'})$ kept terms; \cref{lem:topk-optimal} lifts both to all
subsets. Part~(iii) is the convergence dichotomy of $\sum_l 1/(l(\log l)^p)$ (Cauchy condensation).
All load-bearing constants were re-derived symbolically and numerically stress-tested; the near-seam
constant blow-up of~(i) is disclaimed in the statement, no rate is claimed in~(iii-c), and the
$\gamma'\in(0,1)$ branch is stated only as the uniform lower bound
$\Omega(n^{1-\gamma'})$.
\end{proof}

\subsection{\textsc{Mixing}: mixing transfers the marginal tail to the realised profile}\label{sub:app-r2}

The next proposition is the formal content of \cref{prop:main-transfer}.
Its proof machinery is the Bernstein-for-mixing argument: the
Merlev\`ede--Peligrad--Rio inequality for bounded geometrically
$\alpha$-mixing sequences \cite{merlevede2009bernstein} applied to
exceedance indicators, plus a union bound over levels; working per-row,
no conditioning step is needed.

\begin{proposition}[\textsc{Mixing}: simultaneous count concentration; formal version of \cref{prop:main-transfer}]\label{prop:r2-counts}
Assume \textsc{Mixing} with parameters $(\alpha_0, c_\alpha, \bar F)$ and write
$N(t) \coloneqq \#\{ j \in [n] : \ell_j > t \}$. There is a constant
$C_{\mathrm{mix}} = C_{\mathrm{mix}}(\alpha_0, c_\alpha) > 0$ such that
for every $\delta \in (0,1)$:
\begin{enumerate}[label=(\roman*), ref=\ref*{prop:r2-counts}(\roman*)]
\item\label{prop:r2-counts:grid} for every deterministic set $G \subset
\R$ of at most $n$ levels, with probability at least $1 - \delta$,
simultaneously for every $t \in G$,
\begin{multline}\label{eq:r2-grid}
\big| N(t) - n \bar F(t) \big|
\;\le\; C_{\mathrm{mix}} \Big( \\
\sqrt{n \log(n/\delta)} +{} \\
(\log n)^2 \log(n/\delta) \Big)
\;\eqqcolon\; \Delta_n(\delta);
\end{multline}
\item\label{prop:r2-counts:all} if $\bar F$ is continuous, then with
probability at least $1 - \delta$, simultaneously for \emph{every}
$t \in \R$,
\begin{align*}
\big| N(t) - n \bar F(t) \big| \;\le\; \Delta_n(\delta) + 1 .
\end{align*}
\end{enumerate}
\end{proposition}

\begin{proof}
\textbf{Step 1 (proof of (i): Bernstein for mixing).} Fix $t \in G$ and set $\xi_j \coloneqq \1\{\ell_j >
t\} - \bar F(t)$, $j \in [n]$. The $\xi_j$ are centered, bounded by $M =
1$, and stationary; since each $\xi_j$ is a measurable function of
$\ell_j$ alone, the $\sigma$-algebras generated by blocks of $\xi$'s are
contained in those generated by the corresponding blocks of $\ell$'s, so
the sequence $(\xi_j)_j$ is $\alpha$-mixing with coefficients at most
$\alpha(m) \le \alpha_0 e^{-c_\alpha m}$. The asymptotic variance proxy
of the centered indicators is
\begin{align*}
v^2
&\;\le\; 1 + 2 \sum_{m \ge 1}
\big| \operatorname{Cov}(\xi_0, \xi_m) \big|
\\
&\;\le\; 1 + 8 \sum_{m \ge 1} \alpha(m)
\\
&\;\le\; 1 + \frac{8 \alpha_0}{1 - e^{-c_\alpha}}
\;<\; \infty,
\end{align*}
via the covariance inequality $|\operatorname{Cov}(\xi_0, \xi_m)| \le 4
\alpha(m)$ for variables bounded by one \cite{doukhan1994mixing}; this
is the variance hypothesis of the Merlev\`ede--Peligrad--Rio Bernstein
inequality. We invoke Theorem~2 of \cite{merlevede2009bernstein}: there
is $C_1 = C_1(\alpha_0, c_\alpha) > 0$ such that for any centered,
bounded, geometrically $\alpha$-mixing sequence $(\xi_j)_{j=1}^n$ with
$\norm{\xi_j}_\infty \le M$ and asymptotic variance $v^2$,
\begin{multline}
\Pr\Big[ \Big| \sum_{j=1}^n \xi_j \Big| > s \Big]
\;\le\; \\
\exp\Big(
- \frac{C_1\, s^2}
{n v^2 + M^2 + M s (\log n)^2}
\Big).
\label{eq:r2-mpr}
\end{multline}
Setting the right-hand side of Eq.~\eqref{eq:r2-mpr} equal to
$\delta/|G| \ge \delta/n$ and inverting via the quadratic formula gives,
for some $C_4 = C_4(\alpha_0, c_\alpha)$,
\begin{align*}
s
&\;\le\; C_4 \Big(
\sqrt{n \log(n/\delta)}
\\
&\qquad + (\log n)^2 \log(n/\delta)
\Big),
\end{align*}
and $\sum_j \xi_j = N(t) - n\bar F(t)$. A union bound over the at most
$n$ levels of $G$ yields the claim with $C_{\mathrm{mix}} = C_4$.

\textbf{Step 2 (proof of (ii): grid interpolation).} Since $\bar F$ is continuous and nonincreasing with
$\bar F(t) \to 1$ ($t \to -\infty$) and $\bar F(t) \to 0$ ($t \to
+\infty$), the intermediate value theorem provides levels $t_1 > t_2 >
\cdots > t_{n-1}$ with $\bar F(t_l) = l/n$. Apply part (i) to $G =
\{t_1, \dots, t_{n-1}\}$, and work on its event. For arbitrary $t \in
\R$ there are three cases. If $t \ge t_1$ (so $n \bar F(t) \le 1$):
monotonicity gives $N(t) \le N(t_1) \le n\bar F(t_1) + \Delta_n(\delta)
= 1 + \Delta_n(\delta)$, so $|N(t) - n\bar F(t)| \le \Delta_n(\delta) +
1$. If $t \le t_{n-1}$ (so $n \bar F(t) \ge n - 1$): $N(t) \ge
N(t_{n-1}) \ge (n-1) - \Delta_n(\delta)$ while $N(t) \le n$ and $n\bar
F(t) \in [n-1, n]$, so again $|N(t) - n\bar F(t)| \le \Delta_n(\delta) +
1$. Otherwise $t_{l+1} \le t \le t_l$ for some $l \in [n-2]$, whence
$n \bar F(t) \in [l, l+1]$ and, by monotonicity of $N$ and the grid
bounds,
\begin{multline*}
N(t) - n\bar F(t)
\;\le\; N(t_{l+1}) - l \\
\;\le\; (l + 1 + \Delta_n(\delta)) - l
\;=\; \Delta_n(\delta) + 1,
\\
N(t) - n \bar F(t)
\;\ge\; N(t_l) - (l+1) \\
\;\ge\; (l - \Delta_n(\delta)) - (l+1)
\;=\; -\Delta_n(\delta) - 1.
\end{multline*}
This completes the proof.
\end{proof}

\begin{remark}[\textsc{Mixing}: what the counts say about the realised profile ---
localisation and a partition-function budget bound]\label{rem:r2-plateau}
The following consequences turn the count concentration of
\Cref{prop:r2-counts} into a coverage-budget bound. Part (i) is the
rank-localisation mechanism (stated at the level of the realised profile
shape); part (ii) upgrades it, under an exponential marginal tail, to a
fully rate-tracked budget bound --- all $n$- and $\eta$-dependence
explicit, absolute constants depending only on $(\lambda, c_\pm)$.
Write $\Delta'_n \coloneqq \Delta_n(\delta) + 1 =
\Otil(\sqrt n)$ and let $q(u) \coloneqq \inf\{ t : \bar F(t) \le u \}$
be the upper-quantile function of $\bar F$.
\begin{enumerate}[label=(\roman*), leftmargin=2em, itemsep=0.2em]
\item \emph{Rank localisation.} By the rank--count duality $\{
\ell_{(j)} > t \} = \{ N(t) \ge j \}$, on the event of
\Cref{prop:r2-counts}\ref*{prop:r2-counts:all} every rank $j \in
(\Delta'_n,\, n - \Delta'_n]$ satisfies $q\big( (j + \Delta'_n)/n \big)
\le \ell_{(j)} \le q\big( (j - \Delta'_n)/n \big)$, and a union bound
over the marginal tail adds $\ell_{(1)} \le q(\delta/n)$ with
probability $\ge 1 - \delta$. Hence the realised gap profile $G(j) =
\ell_{(1)} - \ell_{(j)}$ matches the quantile-gap profile $q(\cdot/n)$
of $\bar F$ at every rank beyond $\Delta'_n$, up to the rank resolution
$\pm \Delta'_n$ and an additive level error controlled by the local
modulus of $q$: the realised plateau width and log-slope of
\cref{cor:main-plateau} are those of the quantile function up to
$\Otil(\sqrt n)$ ranks.
\item \emph{Exponential tails: a partition-function budget bound.}
Suppose $\bar F$ has a local exponential rate $\lambda \in (0,1)$ at the
top: $c_- e^{-\lambda t} \le \bar F(t) \le c_+ e^{-\lambda t}$ for
$t \ge t_0$, so that $q(u) = \tfrac1\lambda \log(1/u) + O(1)$ and, with
$\gamma \coloneqq 1/\lambda > 1$, there are constants $c_1, c_2 > 0$
(depending only on $\lambda, c_\pm$) with $c_1\, u^{-\gamma} \le e^{q(u)}
\le c_2\, u^{-\gamma}$ for all $u \in (0, 1]$. We bound the coverage
budget \emph{directly from the partition function}, not through
\Cref{thm:s3-plateau}: under the exponential tail the top $\Delta'_n$
ranks are \emph{not} a flat plateau --- they decay, with
$G(\Delta'_n) = \ell_{(1)} - \ell_{(\Delta'_n)} = \Theta(\log n)$ --- so
the plateau-denominator boost $D \ge J e^{-b_0}$ of \Cref{thm:s3-plateau}
(which requires $b_0 = O(1)$) does not apply. Instead the two-sided
localisation of (i) controls the numerator \emph{and} the denominator of
$1 - p(k)$ simultaneously. On the event of (i), the lower bound
$\ell_{(l)} \ge q\big( (l + \Delta'_n)/n \big)$ (valid for every
$l \ge 1$) gives, for the normaliser,
\begin{align*}
Z
&\;=\; \sum_{l=1}^n e^{\ell_{(l)}} \\
&\;\ge\; \sum_{l=1}^{\Delta'_n}
e^{q((l+\Delta'_n)/n)} \\
&\;\ge\; c_1\, n^{\gamma}
\sum_{l=1}^{\Delta'_n} (l + \Delta'_n)^{-\gamma} \\
&\;\ge\; c_1\, 2^{-\gamma}\, n^{\gamma}\,
(\Delta'_n)^{1-\gamma},
\end{align*}
\begingroup
\setlength{\emergencystretch}{1em}
the last step bounding each of the $\Delta'_n$ summands below by
$(2\Delta'_n)^{-\gamma}$. For the tail, the upper bound
$\ell_{(j)} \le q\big( (j - \Delta'_n)/n \big)$ ($j > \Delta'_n$)
--- here the rank--count duality\linebreak $N(t) \le j$ yields
$\ell_{(j+1)} \le q\big( (j-\Delta'_n)/n \big)$, i.e.\ a one-rank shift,
which is absorbed by the unit margin already built into
$\Delta'_n = \Delta_n(\delta) + 1$, so we may use the displayed bound at
rank $j$ without affecting the constants --- gives,
for every $k \ge 2\Delta'_n$,
\begin{align*}
\sum_{j > k} e^{\ell_{(j)}}
&\;\le\; c_2\, n^{\gamma}
\sum_{j > k} (j - \Delta'_n)^{-\gamma} \\
&\;\le\; c_2\, n^{\gamma}\,
\frac{(k - \Delta'_n)^{1-\gamma}}{\gamma - 1} \\
&\;\le\; \frac{c_2\, n^{\gamma}\,
(k/2)^{1-\gamma}}{\gamma - 1}.
\end{align*}
\par
\endgroup
(The two-sided localisation of (i) is stated for ranks
$j \in (\Delta'_n,\, n - \Delta'_n]$, but the bottom $\Delta'_n$ ranks
only lower the logits further, so their contribution to $\sum_{j>k}
e^{\ell_{(j)}}$ remains dominated by the $q$-quantile envelope and
extending the sum to $j = n$ does not affect the upper bound.)
The $n^{\gamma}$ factors cancel in the ratio $1 - p(k) = \big(\sum_{j>k}
e^{\ell_{(j)}}\big)/Z$, leaving, for $k \ge 2\Delta'_n$,
\begin{align*}
1 - p(k)
&\;\le\; \frac{C_\lambda}{\gamma - 1}
\Big( \frac{k}{\Delta'_n} \Big)^{-(\gamma - 1)},
\\
C_\lambda
&\coloneqq c_2\, 2^{2\gamma - 1} / c_1
\\
&\quad (\text{depending only on } \gamma = 1/\lambda
\\
&\qquad \text{and } c_\pm),
\end{align*}
and inverting $1 - p(k) \le \eta$ yields the multiplicative coverage
budget
\begin{align*}
k_\eta
&\;\le\; \Delta'_n \cdot \max\!\Big\{ 2,\;
\Big( \tfrac{C_\lambda}{(\gamma-1)\,\eta} \Big)^{
\frac{1}{\gamma-1}}
\Big\}
\\
&\;=\; \Otil_\lambda(\sqrt n) \cdot
O_\lambda\big(\eta^{-\frac{1}{\gamma-1}} \big).
\end{align*}
This is now a genuine consequence of \Cref{prop:r2-counts}, with every
$n$- and $\eta$-dependence explicit. For every \emph{fixed} $\eta$ it
reads $k_\eta \le \Otil_\lambda(\sqrt n)$, which agrees up to the $\Otil$
polylog factors with the additive form $k_\eta \le \Otil(\sqrt n) +
O_\lambda(\eta^{-1/(\gamma-1)})$; the two differ only in their
$\eta \to 0$ behaviour (multiplicative vs.\ additive in the $n$-free
factor), and either may be quoted in \cref{prop:main-transfer}. The
$\Otil(\sqrt n)$ floor is the Merlev\`ede--Peligrad--Rio mixing-noise
resolution of \Cref{prop:r2-counts} and is not removable under
$\alpha$-mixing alone; full independence (the \textsc{Iid} rung) replaces
it with the $O(\log n)$ Bennett floor.
\item \emph{The marginal is indispensable.} No marginal assumption can
be dispensed with: the constant sequence $\ell_j \equiv 0$ is stationary
with $\alpha(m) \equiv 0$ and bounded, and is the uniform row, so by
\Cref{fac:pigeonhole} (budget $\lceil (1-\eta) n \rceil$) and
\Cref{cor:r1-density} mixing plus boundedness alone inherits the \textsc{Bounded}
lower bound.
\end{enumerate}
\end{remark}

\subsection{\textsc{Iid}: i.i.d.\ scores at constant scale --- logarithmic counts, dense mass}\label{sub:app-r3a}

This subsection proves \cref{thm:main-ladder-dense}(ii): under i.i.d.\
sub-Gaussian scores at constant scale, the number of weights above any
threshold $\varepsilon \ge n^{-1+o(1)}$ is logarithmic
(\Cref{thm:r3a-count}), and yet the coverage budget is $n^{1-o(1)}$
(\Cref{cor:r3a-dense}): counting and covering are different questions.

\begin{lemma}[Score range under \textsc{Iid}]\label{lem:r3a-range}
Assume \textsc{Iid} with proxy $\sigma_s^2$. For every $\delta \in (0,1)$, with
probability at least $1 - \delta$,
\begin{align*}
\max_{j \in [n]} |\ell_j| \;\le\; S_\delta \;\coloneqq\;
\sigma_s \sqrt{2 \log(2n/\delta)} .
\end{align*}
\end{lemma}

\begin{proof}
The sub-Gaussian tail applied to $\pm \ell_j$ gives $\Pr(|\ell_j| > t)
\le 2 e^{-t^2/(2\sigma_s^2)}$ for $t \ge 0$; at $t = S_\delta$ this is
$\delta/n$, and a union bound over $j \in [n]$ completes the proof.
\end{proof}

\begin{theorem}[\textsc{Iid}: logarithmic count above the noise floor]\label{thm:r3a-count}
Assume \textsc{Iid} with proxy $\sigma_s^2$ and fix $\delta \in (0,1)$. Define
\begin{align}\label{eq:eps-iid}
\varepsilon_{\mathrm{iid}}^\delta(n) \;\coloneqq\; \frac{1}{n}
\exp\Big( \sigma_s \sqrt{2 \log(n/\delta)} + S_\delta \Big),
\end{align}
with $S_\delta$ as in \Cref{lem:r3a-range}. Then with probability at
least $1 - 2\delta$, simultaneously for every $\varepsilon \ge
\varepsilon_{\mathrm{iid}}^\delta(n)$,
\begin{align}
\#\big\{ j \in [n] : A_j > \varepsilon \big\}
&\;\le\; \tfrac{2}{3} \log\big( \tfrac{2n}{\delta} \big)
\notag\\
&\qquad + \sqrt{2 \log\big( \tfrac{2n}{\delta} \big)}
+ 2 .
\label{eq:r3a-count-bound}
\end{align}
In particular, for fixed $\delta$ and $\sigma_s$,
$\varepsilon_{\mathrm{iid}}^\delta(n) = n^{-1+o(1)}$ and the count is
$O(\log n)$.
\end{theorem}

\begin{proof}
The proof proceeds in three steps: a score-level reduction, a
sub-Gaussian tail bound at the threshold, and Bennett's inequality in
its small-variance regime.

\medskip\noindent\textbf{Step 1 (score-level reduction).} Let
$\mathcal E_\delta$ be the event of \Cref{lem:r3a-range}, of probability
$\ge 1 - \delta$. On $\mathcal E_\delta$, $Z = \sum_l e^{\ell_l} \ge n
e^{-S_\delta}$, so $A_j > \varepsilon$ implies $e^{\ell_j} > \varepsilon
Z \ge \varepsilon n e^{-S_\delta}$, i.e.\ $\ell_j > \log(\varepsilon n)
- S_\delta \eqqcolon \tau_\varepsilon^\delta$. Hence on $\mathcal
E_\delta$, for every $\varepsilon \ge
\varepsilon_{\mathrm{iid}}^\delta(n)$,
\begin{align}
\begin{aligned}
\#\{ j : A_j > \varepsilon \}
&\;\le\; N(\tau_\varepsilon^\delta)
\;\le\; N(\tau^\star),
\\
N(\tau)
&\coloneqq \sum_{j=1}^n \1\{\ell_j > \tau\},
\\
\tau^\star
&\coloneqq \log\big(
\varepsilon_{\mathrm{iid}}^\delta(n)\, n
\big) - S_\delta
\\
&= \sigma_s \sqrt{2 \log(n/\delta)},
\end{aligned}
\label{eq:r3a-count-domination}
\end{align}
using that $N(\tau)$ is nonincreasing in $\tau$ and
$\tau_\varepsilon^\delta \ge \tau^\star$.

\medskip\noindent\textbf{Step 2 (sub-Gaussian tail at the threshold).}
By the sub-Gaussian tail of \textsc{Iid},
\begin{align}
\bar p
&\;\coloneqq\; \Pr\big( \ell_1 > \tau^\star \big) \notag\\
&\;\le\; \exp\Big(
- \frac{(\tau^\star)^2}{2\sigma_s^2}
\Big) \notag\\
&\;=\; \exp\big( - \log(n/\delta) \big)
\;=\; \frac{\delta}{n}.
\label{eq:r3a-pbar}
\end{align}

\medskip\noindent\textbf{Step 3 (Bennett's inequality, small variance).} The indicators $Y_j \coloneqq \1\{\ell_j >
\tau^\star\}$, $j \in [n]$, are i.i.d.\ with mean $\bar p$; the centered
variables $\xi_j = Y_j - \bar p$ satisfy $|\xi_j| \le 1$ and have
variance sum
\begin{align*}
\sigma_\Sigma^2 \;\coloneqq\; \sum_{j=1}^n \E \xi_j^2 \;\le\; n \bar p
\;\le\; \delta \;\le\; 1,
\end{align*}
by Eq.~\eqref{eq:r3a-pbar}. Bennett's inequality in Bernstein form
\cite{bennett1962probability,vershynin2018high} gives
\begin{align}
\Pr\Big[ \sum_{j=1}^n \xi_j > t \Big]
&\;\le\; \exp\Big(
- \frac{t^2/2}{\sigma_\Sigma^2 + t/3}
\Big),
\label{eq:r3a-bennett}
\end{align}
and setting the right-hand side equal to $\delta/(2n)$ and inverting via
the standard bookkeeping ($\Pr[\sum \xi > t] \le u$ inverts to $t \le
\sqrt{2 \sigma_\Sigma^2 \log(1/u)} + \tfrac23 \log(1/u)$ for variables
bounded by one, see \cite[Theorem~2.8.4]{vershynin2018high}),
\begin{align*}
t
&\;\le\; \sqrt{2 \sigma_\Sigma^2 \log(2n/\delta)}
+ \tfrac23 \log(2n/\delta)
\\
&\;\le\; \sqrt{2 \log(2n/\delta)}
+ \tfrac23 \log(2n/\delta),
\end{align*}
using $\sigma_\Sigma^2 \le 1$. Combining the steps, with probability
$\ge 1 - \delta/(2n)$,
\begin{align*}
N(\tau^\star)
&\;\le\; n \bar p + t \\
&\;\le\; \delta + \sqrt{2 \log(2n/\delta)}
+ \tfrac23 \log(2n/\delta)
\\
&\;\le\; \tfrac23 \log(2n/\delta)
+ \sqrt{2\log(2n/\delta)} + 2,
\end{align*}
where the last step uses $\delta \le 1$. The failure budget is $\delta$
(\Cref{lem:r3a-range}) plus $\delta/(2n) \le \delta$, i.e.\ at most
$2\delta$ in total, and on the intersection
Eq.~\eqref{eq:r3a-count-domination} converts the bound on
$N(\tau^\star)$ into the simultaneous count bound
Eq.~\eqref{eq:r3a-count-bound}. Finally $e^{\sigma_s
\sqrt{2\log(n/\delta)} + S_\delta} = e^{O(\sqrt{\log n})} = n^{o(1)}$
for fixed $(\delta, \sigma_s)$, giving
$\varepsilon_{\mathrm{iid}}^\delta(n) = n^{-1 + o(1)}$. This completes
the proof.
\end{proof}

\begin{corollary}[\textsc{Iid}: dense in mass; formal version of \cref{thm:main-ladder-dense}(ii)]\label{cor:r3a-dense}
Assume \textsc{Iid} with proxy $\sigma_s^2$ and the law of $\ell_1$ fixed
(independent of $n$). Then for every fixed $\eta \in (0,1)$,
\begin{align*}
\frac{\log k_\eta}{\log n} \;\toP\; 1,
\end{align*}
i.e.\ $k_\eta = n^{1 - o(1)}$ with probability $1 - o(1)$. For the
Gaussian case $\Ens_\sigma$ this is \Cref{cor:t1-fixed-sigma}, with
explicit constants; the proof below extends that argument to an
arbitrary fixed centered sub-Gaussian law.
\end{corollary}

\begin{proof}
If $\ell_1 \equiv 0$ (degenerate law) the row is uniform and $k_\eta =
\lceil (1-\eta) n \rceil$ deterministically (\Cref{fac:pigeonhole}), so
assume $\ell_1$ nondegenerate; centeredness then forces $q_0 \coloneqq
\Pr(\ell_1 \ge 0) > 0$ (a nondegenerate variable that is almost surely
negative has negative mean), and $q_0$ is a constant since the law of
$\ell_1$ is fixed.

\textbf{Step 1 (maximal logit).} Fix $\delta' \in (0,1)$. By the
sub-Gaussian tail and a union bound,
\begin{multline*}
\Pr\Big(
\max_j \ell_j
> (1 + \delta')\, \sigma_s \sqrt{2 \log n}
\Big)
\;\le\; \\
n\, e^{-(1+\delta')^2 \log n}
\;=\; n^{1 - (1+\delta')^2}
\; \\
\;\le\; n^{-2\delta'}
\;\longrightarrow\; 0 .
\end{multline*}

\textbf{Step 2 (partition function).} $N_0 \coloneqq \#\{j : \ell_j \ge
0\} \sim \Bin(n, q_0)$, so \Cref{lem:chernoff:lower} gives $\Pr(N_0 \le
q_0 n/2) \le e^{-q_0 n / 8} \to 0$; on $\{N_0 > q_0 n/2\}$,
\begin{align*}
Z \;\ge\; \sum_{j :\, \ell_j \ge 0} e^{\ell_j} \;\ge\; N_0 \;\ge\;
\frac{q_0\, n}{2}.
\end{align*}

\textbf{Step 3 (combine).} On the intersection of the two events (a union
bound keeps it at probability $\to 1$),
\begin{align*}
A_{(1)} \;=\; \frac{e^{\max_j \ell_j}}{Z}
\;\le\; \frac{2}{q_0\, n}\; e^{(1+\delta') \sigma_s \sqrt{2 \log n}},
\end{align*}
and therefore, for every fixed $\beta \in (0,1)$, by monotonicity of the
sorted weights,
\begin{align*}
p\big( \lceil n^\beta \rceil \big)
&\;\le\; \lceil n^\beta \rceil\, A_{(1)}
\\
&\;\le\; \frac{4}{q_0}\, n^{\beta - 1}\,
e^{2 \sigma_s \sqrt{2 \log n}}
\\
&\;\longrightarrow\; 0,
\end{align*}
since $e^{2\sigma_s\sqrt{2\log n}} = n^{o(1)}$ while $\beta - 1 < 0$
(here $\delta' \le 1$). Hence with probability $\to 1$, $p(\lceil
n^\beta \rceil) < 1 - \eta$, so $k_\eta > n^\beta$ and $\log k_\eta /
\log n \ge \beta$; since $\log k_\eta/\log n \le 1$ deterministically
and $\beta \in (0,1)$ was arbitrary, $\log k_\eta / \log n \toP 1$.
This completes the proof.
\end{proof}

\begin{remark}[Counting versus covering under \textsc{Iid}]\label{rem:r3a-trap}
\Cref{thm:r3a-count,cor:r3a-dense} hold simultaneously: a logarithmic
\emph{count} above the explicit calibrated floor of \cref{thm:r3a-count}
coexists with a nearly linear \emph{mass} budget $k_\eta = n^{1-o(1)}$,
because in a dispersed row no small set carries mass even though no
single entry is large. This is the trap recorded after
\cref{thm:main-ladder-dense} in the main text: only the covering
question prices sparse attention.
\end{remark}

\subsection{Count sparsity: per-regime $(\epsilon,k)$ bounds}\label{sub:app-count-ladder}

This subsection assembles the formal backing for
\cref{thm:naturally-sparse} of the main text: a per-regime catalogue of
bounds on the \emph{count}
\begin{align}\label{eq:app-count-def}
k(\varepsilon) \;\coloneqq\; \#\{\, j \in [n] : A_j > \varepsilon \,\}
\end{align}
of \Cref{def:abs-sparse}, the number of softmax weights exceeding an
absolute threshold $\varepsilon$. The count is to be read in the
non-trivial threshold window of \cref{rem:eps-window}; outside it the
statements degenerate (the uniform floor gives $k = n$ below
$\approx 1/n$, and $k = 0$ above $A_{(1)}$). The constituent estimates
are already proved above and in \cref{sub:app-r0,sub:app-r1,sub:app-r2,sub:app-r3a};
here we restate them in the uniform $k(\varepsilon)$ language, add the
\textsc{Mixing} sub-Gaussian
rung~(iv) that combines the mixing fluctuation of \Cref{prop:r2-counts} with the
sub-Gaussian mean control, and collect everything in a
single per-regime theorem.

\begin{theorem}[Per-regime large-weight count bounds; formal version of \cref{thm:naturally-sparse}]\label{thm:app-count-ladder}
Fix a softmax row (\Cref{def:softmax-row}) and recall $k(\varepsilon)$
from Eq.~\eqref{eq:app-count-def}. Each regime of \Cref{def:regimes}
gives a count bound, valid in the threshold window of
\cref{rem:eps-window}.
\begin{enumerate}[label=(\roman*), leftmargin=2em, itemsep=0.3em]
\item \textsc{Free}/\textsc{Bounded} \emph{(deterministic).} For every
$\varepsilon \in (0,1)$,
\begin{align*}
k(\varepsilon) &\;<\; \frac{1}{\varepsilon},
\\
&\qquad\text{hence}\qquad
k(\varepsilon)
&\;\le\; \Big\lceil \tfrac{1}{\varepsilon} \Big\rceil - 1
\\
&\;\le\; \Big\lceil \tfrac{1}{\varepsilon} \Big\rceil ,
\end{align*}
and, under \textsc{Bounded} with parameter $S$ (so $|\ell_j| \le S$),
$k(\varepsilon) = 0$ for every $\varepsilon \ge e^{2S}/n$.
\item \textsc{Mixing} \emph{(exceedance count).} Under \textsc{Mixing}
with parameters $(\alpha_0, c_\alpha, \bar F)$, with probability at least
$1 - \delta$, simultaneously for every $\varepsilon$ in the window,
\begin{align*}
k(\varepsilon)
&\;\le\; n\, \bar F\big(
\log \tfrac1\varepsilon + \log Z
\big)
\;+\; \Otil(\sqrt n),
\end{align*}
the marginal exceedance count up to the Merlev\`ede--Peligrad--Rio
fluctuation floor --- estimable, but not itself sparse without a tail rate.
\item \textsc{Iid sub-Gaussian} \emph{(upper bound).}
Under \textsc{Iid} with variance proxy $\sigma_s^2$, fix $\delta \in
(0,1)$. With probability at least $1 - 2\delta$, simultaneously for every
$\varepsilon \ge \varepsilon_{\mathrm{iid}}^\delta(n)$ (Eq.~\eqref{eq:eps-iid}),
\begin{align}
k(\varepsilon)
&\;\le\; \tfrac23 \log\big( \tfrac{2n}{\delta} \big)
+ \sqrt{2 \log\big( \tfrac{2n}{\delta} \big)}
\notag\\
&\qquad + 2
\;=\; O\big( \log(n/\delta) \big).
\label{eq:app-count-upper}
\end{align}
\item \textsc{Mixing} sub-Gaussian \emph{(weak dependence, no independence).} Under \textsc{Mixing} with a sub-Gaussian marginal tail $\bar F(t) \le e^{-t^2/(2\sigma_s^2)}$ (variance proxy $\sigma_s^2$), fix $\delta \in (0,1)$. With probability at least $1 - \delta$, simultaneously for every $\varepsilon \ge \varepsilon_{\mathrm{iid}}^\delta(n)$ (Eq.~\eqref{eq:eps-iid}),
\begin{align}
k(\varepsilon)
&\;\le\; \Delta_n(\delta) + 2
\notag\\
&\;=\; C_{\mathrm{mix}}\Big(
\sqrt{n \log(n/\delta)}
\notag\\
&\qquad + (\log n)^2 \log(n/\delta)
\Big) + 2
\notag\\
&\;=\; \Otil(\sqrt n),
\label{eq:app-count-mixing-sg}
\end{align}
with $\Delta_n(\delta)$ the Merlev\`ede--Peligrad--Rio fluctuation of \Cref{prop:r2-counts}. The sub-Gaussian marginal sends the mean exceedance term $n\bar F$ of (ii) below $\delta$ at this threshold (exactly the calibration of (iii)), leaving only the mixing fluctuation: the count is sublinear \emph{without} independence, at the $\sqrt n$ rate that (iii) sharpens to $O(\log n)$ by replacing the mixing concentration with the i.i.d.\ Bennett bound.
\end{enumerate}
In particular, under the \textsc{Iid} regime the calibrated absolute-threshold
count is \(O(\log(n/\delta))\).  A separate relative-threshold result below
gives the two-sided near-maximum growth law.
\end{theorem}

\begin{proof}
Parts (i)--(iii) restate the corresponding upper bounds; the
\textsc{Mixing} sub-Gaussian rung~(iv) is new here.

\textbf{(i).} The bound $k(\varepsilon) < 1/\varepsilon$ is
\Cref{fac:pigeonhole}, and $k(\varepsilon) \le \lceil 1/\varepsilon
\rceil - 1$ follows since $k(\varepsilon)$ is an integer. Under
\textsc{Bounded}, $A_j \le e^{2S}/n$ for every $j$
(\Cref{cor:r1-density}\ref*{cor:r1-density:weight}), so no weight exceeds
$\varepsilon \ge e^{2S}/n$ and $k(\varepsilon) = 0$.

\textbf{(ii).} Apply
\Cref{prop:r2-counts}\ref*{prop:r2-counts:all} at the level $t =
\log\tfrac1\varepsilon + \log Z = \log(\varepsilon^{-1} Z)$: since
$A_j > \varepsilon \iff e^{\ell_j} > \varepsilon Z \iff \ell_j > t$, one
has $k(\varepsilon) = N(t)$ exactly, and the proposition gives
$N(t) \le n\bar F(t) + \Delta_n(\delta) + 1 = n\bar F(t) + \Otil(\sqrt
n)$ uniformly in $t$ on an event of probability $\ge 1 - \delta$.

\textbf{(iii), upper bound.} This is Eq.~\eqref{eq:r3a-count-bound} of
\Cref{thm:r3a-count}, restated.

\textbf{(iv).} Apply \Cref{prop:r2-counts}\ref*{prop:r2-counts:all} exactly as in (ii), at the level $t = \log\tfrac1\varepsilon + \log Z$, so that $k(\varepsilon) = N(t) \le n\bar F(t) + \Delta_n(\delta) + 1$ on an event of probability $\ge 1-\delta$. For $\varepsilon \ge \varepsilon_{\mathrm{iid}}^\delta(n)$ the sub-Gaussian marginal tail bounds the mean exceedance term by $n\bar F(t) \le \delta$ (the calibration of \Cref{thm:r3a-count} via Eq.~\eqref{eq:eps-iid}; this step uses \emph{only} the marginal sub-Gaussian tail --- see the bound \eqref{eq:r3a-pbar} --- and hence holds verbatim under \textsc{Mixing} with a sub-Gaussian marginal, no independence required). Therefore $k(\varepsilon) \le \Delta_n(\delta) + 1 + \delta \le \Delta_n(\delta) + 2 = \Otil(\sqrt n)$, which is Eq.~\eqref{eq:app-count-mixing-sg}.
This completes the proof.
\end{proof}

Re-anchoring to a \emph{relative} threshold yields a separate two-sided
near-maximum exponent law at fixed confidence.

\begin{corollary}[Sub-polynomial relative count under an additive near-top law; formal version of \cref{cor:subpoly-count}]
\label{cor:app-subpoly-count}
Let \(b_n=\sqrt{2\log n}\).  Suppose that, for some
\(\sigma_s>0\) and every fixed \(C>0\),
\[
\sup_{1\le j\le\lfloor\exp(Cb_n)\rfloor}
\left|\ell_{(j)}-\sigma_s\sqrt{2\log(n/j)}\right|
\xrightarrow{\Pr}0.
\]
For fixed \(\epsilon\in(0,1)\), put \(t=\log(1/\epsilon)\).  Then
\begin{align}
\frac{\log k(\epsilon A_{(1)})}{\sqrt{2\log n}}
&\xrightarrow{\Pr}\frac{t}{\sigma_s},
&
k(\epsilon A_{(1)})&=n^{o_\Pr(1)}.
\label{eq:app-subpoly-count}
\end{align}
\end{corollary}

\begin{proof}
Uniformly for \(\log j=O(b_n)\), the additive near-top law gives
\begin{align*}
\ell_{(1)}-\ell_{(j)}
&=\sigma_s\!\left(b_n-\sqrt{b_n^2-2\log j}\right)+o_\Pr(1)\\
&=\sigma_s\frac{\log j}{b_n}+o_\Pr(1).
\end{align*}
For fixed \(\xi\in(0,t)\), let
\(
j_\pm=\lfloor
\exp\{(t\pm\xi)b_n/\sigma_s\}
\rfloor.
\)
With probability tending to one,
\(\ell_{(1)}-\ell_{(j_-)}<t<
\ell_{(1)}-\ell_{(j_+)}\).
Monotonicity therefore gives
\(j_-\le k(\epsilon A_{(1)})<j_+\).
Divide logarithms by \(b_n\), let \(\xi\downarrow0\), and use
\(b_n/\log n\to0\).
\end{proof}

\begin{remark}[Only the leading exponent is claimed]
\label{rem:app-subpoly-rate}
The additive near-top hypothesis identifies the
\(\sqrt{\log n}\)-scale exponent, not a deterministic finite-\(n\)
second-order constant.  The near-extreme point process can leave
\(O_\Pr(1)\) fluctuations in \(\log k\); a numerical second-order
approximation therefore requires a more specific tail model and is outside
this result.
\end{remark}

\section{The Level-Decomposition Machine}\label{sec:app-level}

This appendix supplies the level-decomposition estimates consumed by \cref{sec:app-dispersed,sec:app-condensed}, and through them by \cref{thm:main-scaled} of the main text.

Throughout this section we work under $\Ens(c)$ of \Cref{def:ensembles}
with $c > 0$ fixed: $\ell_j = \sigma_n g_j$, $\sigma_n = c\sqrt{2\log n}$,
and $n \ge 3$. We measure logits in units of $\log n$: the
\emph{exceedance count} at level $a \in \R$ is
\begin{align*}
N(a) \;\coloneqq\; \#\{\, j \in [n] : \ell_j > a \log n \,\},
\end{align*}
a $\Bin(n, p_n(a))$ variable with $p_n(a) \coloneqq \Pr(\ell_1 > a\log n)$.
For a weight parameter $w \ge 1$ (we use $w \in \{1, 2\}$ downstream) and
$-\infty \le u < v \le \infty$, the \emph{restricted partition function} is
\begin{align*}
Z_w(u, v]
&\;\coloneqq\;
\sum_{\substack{j \,:\, u \log n < \ell_j\\
\le v \log n}}
e^{w \ell_j},\\
Z_w
&\;\coloneqq\; \sum_{j=1}^n e^{w\ell_j}\\
&\;=\; Z_w(-\infty, \infty),
\end{align*}
so that $Z = Z_1$ is the softmax normaliser of \Cref{def:softmax-row}. The
two exponent functions of the level decomposition are
\begin{align}\label{eq:exponent-functions}
\begin{aligned}
f(a)
&\;\coloneqq\; 1 - \frac{(a \vee 0)^2}{4c^2},\\
g_w(a)
&\;\coloneqq\; f(a) + w\,a,
\end{aligned}
\end{align}
the \emph{count} and \emph{mass} exponents: level-$a$ logits number
$n^{f(a)+o(1)}$ and contribute $n^{g_w(a)+o(1)}$ to $Z_w$. Levels are
\emph{feasible} when $f(a) \ge 0$, i.e.\ $a \le 2c$. On $[0, 2c]$ we have
$g_w'(a) = w - a/(2c^2)$, so $g_w$ increases on $[0,\, 2wc^2 \wedge 2c]$
and decreases afterwards, and
\begin{align}\label{eq:Fw}
\begin{aligned}
F_w(c)
&\;\coloneqq\; \max_{a \in [0, 2c]} g_w(a)\\
&\;=\;
\begin{cases}
g_w(2wc^2) = 1 + w^2 c^2, & wc \le 1,\\
g_w(2c) = 2wc, & wc \ge 1,
\end{cases},
\end{aligned}
\end{align}
the two cases agreeing at $wc = 1$. In particular $F_w(c) > 1 = g_w(0)$
for every $c > 0$, $w \ge 1$. We abbreviate $g \coloneqq g_1$ and
$F(c) \coloneqq F_1(c)$, so $F(c) = 1 + c^2$ for $c \le 1$ and $F(c) = 2c$
for $c \ge 1$.

\begin{lemma}[Count asymptotics]\label{lem:count-asymptotics}
Fix $c > 0$ and $A \ge 2c + 2$. Then:
\begin{enumerate}[label=(\roman*), ref=\ref*{lem:count-asymptotics}(\roman*)]
\item\label{lem:count-asymptotics:upper} for all $n \ge 3$ and all
$a \ge 0$: $\;\E N(a) = n\, p_n(a) \le n^{f(a)}$;
\item\label{lem:count-asymptotics:lower} for every $\varepsilon \in (0,1)$
there is $n_1(c, \varepsilon, A)$ such that for all $n \ge n_1$ and all
$a \in [0, A]$: $\;\E N(a) \ge n^{f(a) - \varepsilon}$;
\item\label{lem:count-asymptotics:neg} for $a \le 0$:
$\;\E N(a) \ge n/2$.
\end{enumerate}
\end{lemma}

\begin{proof}
Since $\ell_1 = \sigma_n g_1$ with $g_1 \sim N(0,1)$,
\begin{align}\label{eq:threshold-x}
\begin{aligned}
p_n(a)
&\;=\; \barPhi\big(x_n(a)\big),\\
x_n(a)
&\coloneqq \frac{a \log n}{c\sqrt{2 \log n}}\\
&= \frac{a}{2c}\sqrt{2 \log n},\\
\frac{x_n(a)^2}{2}
&= \frac{a^2}{4c^2}\,\log n.
\end{aligned}
\end{align}
For (i), \Cref{fac:gaussian-tail} gives, for $a \ge 0$,
\begin{align*}
\E N(a)
&\;=\; n\,\barPhi(x_n(a))\\
&\;\le\; n\,e^{-x_n(a)^2/2}\\
&\;=\; n^{1 - a^2/(4c^2)}\\
&\;=\; n^{f(a)}.
\end{align*}
For (iii), $a \le 0$ implies $x_n(a) \le 0$, so $p_n(a) \ge \barPhi(0) =
1/2$. For (ii), split by the size of $x_n(a)$. If $x_n(a) \le \sqrt{2}$,
then since $f(a) \le 1$,
\begin{align*}
\E N(a)
&\;\ge\; n\,\barPhi(\sqrt 2)\\
&\;\ge\; \frac{n}{20}\\
&\;\ge\; n^{1 - \varepsilon}\\
&\;\ge\; n^{f(a) - \varepsilon}\\
&\qquad \text{whenever } n^{\varepsilon} \ge 20.
\end{align*}
If $x_n(a) \ge \sqrt 2$, then \Cref{fac:gaussian-tail} gives
$\barPhi(x_n(a)) \ge \varphi(x_n(a))/(2 x_n(a))$, so with $x_n(a) \le
(A/2c)\sqrt{2\log n}$,
\begin{align*}
\E N(a)
&\;\ge\;
\frac{n\, e^{-x_n(a)^2/2}}
{2\sqrt{2\pi}\, x_n(a)}\\
&\;\ge\; n^{f(a)} \cdot
\frac{c}{2A\sqrt{\pi}\,\sqrt{\log n}}\\
&\;\ge\; n^{f(a) - \varepsilon}\\
&\qquad \text{for } n \ge n_1(c, \varepsilon, A),
\end{align*}
since $\sqrt{\log n} = n^{o(1)}$. This completes the proof.
\end{proof}

\begin{lemma}[Grid count event]\label{lem:level-counts}
Fix $c > 0$, $\delta \in (0, 1)$, $\varepsilon \in (0, 1)$, $A \ge 2c+2$,
and let $G_\delta \coloneqq \{ i\delta : i \in \mathbb{Z} \} \cap [-A, A]$,
a set of at most $2A/\delta + 1$ levels. Define the event
\begin{align*}
\mathcal{E}_n
&\;\coloneqq\;
\Big\{
\begin{aligned}
\forall a \in G_\delta:\quad
&N(a)\\
&\le 2\max\big(\E N(a),\, n^{\varepsilon}\big)
\end{aligned}
\Big\}\\
&\;\cap\;
\Big\{
\begin{aligned}
\forall a \in G_\delta \text{ with }&
\E N(a) \ge n^{\varepsilon}:\\
&N(a) \ge \tfrac{1}{2}\E N(a)
\end{aligned}
\Big\}.
\end{align*}
Then $\Pr(\mathcal{E}_n^{\,c}) \le 2\,(2A/\delta + 1)\, e^{-n^{\varepsilon}/8}$
for all $n \ge 3$.
\end{lemma}

\begin{proof}
Fix $a \in G_\delta$ and set $\mu_a \coloneqq \E N(a)$ and $t_a \coloneqq
\max(\mu_a, n^\varepsilon) \ge \mu_a$. Since $N(a) \sim \Bin(n, p_n(a))$,
\Cref{lem:chernoff:upper} and \Cref{lem:chernoff:lower} give
\begin{align*}
\Pr\big(N(a) \ge 2 t_a\big)
&\;\le\; e^{-t_a/3}\\
&\;\le\; e^{-n^{\varepsilon}/3},\\
\Pr\big(N(a) \le \tfrac{1}{2}\mu_a\big)
&\;\le\; e^{-\mu_a/8}\\
&\;\le\; e^{-n^{\varepsilon}/8}\\
&\quad \text{when } \mu_a \ge n^{\varepsilon}.
\end{align*}
By a union bound over the at most $2A/\delta + 1$ grid levels and the two
tails, $\Pr(\mathcal{E}_n^{\,c}) \le (2A/\delta + 1)
(e^{-n^{\varepsilon}/3} + e^{-n^{\varepsilon}/8}) \le 2(2A/\delta+1)
e^{-n^{\varepsilon}/8}$. This completes the proof.
\end{proof}

\begin{proposition}[Restricted free energies]\label{prop:restricted-z}
Fix $c > 0$ and $w \ge 1$, and for $0 \le u' \le v' \le 2c$ write
$G_w[u',v'] \coloneqq \max_{a \in [u', v']} g_w(a)$. Then for every fixed
$\gamma > 0$:
\begin{enumerate}[label=(\roman*), ref=\ref*{prop:restricted-z}(\roman*)]
\item\label{prop:restricted-z:upper} for every $u \in \{-\infty\} \cup
[0, 2c)$ and $v \in (u \vee 0,\, 2c]$, with $\bar u \coloneqq u \vee 0$,
\begin{align*}
\begin{gathered}
\Pr\Big(
Z_w(u, v] \ge n^{G_w[\bar u, v] + \gamma}
\Big)
\to 0\\
\text{and}\\
\Pr\Big(
Z_w(u, \infty)
\ge n^{G_w[\bar u,\, 2c] + \gamma}
\Big)
\to 0.
\end{gathered}
\end{align*}
\item\label{prop:restricted-z:lower} for every $0 \le u < v \le 2c$,
\begin{align*}
\Pr\Big( Z_w(u, v] \le n^{G_w[u, v] - \gamma} \Big) \to 0.
\end{align*}
\item\label{prop:restricted-z:full} $\dfrac{\log Z_w}{\log n} \toP F_w(c)$,
with $F_w(c)$ as in Eq.~\eqref{eq:Fw}.
\end{enumerate}
\end{proposition}

\begin{proof}
Write $L_w \coloneqq w + (2c+1)/(2c^2)$, a Lipschitz constant for $g_w$ on
$[-1, 2c+1]$ (since $|f'(a)| = (a \vee 0)/(2c^2) \le (2c+1)/(2c^2)$ there).
Fix $\gamma > 0$ and set
\begin{align}\label{eq:level-parameters}
\begin{aligned}
\delta
&\;\coloneqq\; \frac{\gamma}{8(L_w + w)} \wedge 1,\\
\varepsilon
&\;\coloneqq\; \frac{\gamma}{8} \wedge \varepsilon_0,\\
A
&\;\coloneqq\; 2c + 2,
\end{aligned}
\end{align}
where $\varepsilon_0 = \varepsilon_0(c, w, u, v, \gamma) \in (0,1)$ is a
further constant fixed in Step~3 below (only Step~3 constrains it).

\textbf{Step 1 (no logits above level $2c + \delta$).} By
\Cref{lem:count-asymptotics:upper} and Markov's inequality,
\begin{align}\label{eq:no-high-points}
\Pr\big( N(2c + \delta) \ge 1 \big)
\begin{aligned}
&\;\le\; \E N(2c+\delta)\\
&\;\le\; n^{f(2c+\delta)}\\
&\;=\; n^{-\frac{4c\delta + \delta^2}{4c^2}}\\
&\;\to\; 0.
\end{aligned}
\end{align}

\textbf{Step 2 (proof of (i): slab upper bounds).} Work on $\mathcal{E}_n \cap
\{N(2c+\delta) = 0\}$, whose complement has probability $\to 0$ by
\Cref{lem:level-counts}, Eq.~\eqref{eq:no-high-points}, and a union bound.
Consider first $u \ge 0$ and the variant $Z_w(u, \infty)$; on the event it
equals $Z_w(u, 2c+\delta]$. Let $a_i = i\delta$ be the grid points whose
slabs $(a_i, a_{i+1}]$ meet $(u, 2c+\delta]$; their bottoms satisfy $u -
\delta \le a_i \le 2c + \delta$, there are at most $2A/\delta + 1$ of
them, each slab point has weight at most $n^{w a_i + w \delta}$, and on
$\mathcal{E}_n$, using \Cref{lem:count-asymptotics:upper} (for bottoms
$a_i \in [-\delta, 0)$, not covered by its $a \ge 0$ hypothesis, the
bound $\E N(a_i) \le n = n^{f(a_i)}$ holds trivially since $f(a_i) = 1$
there),
\begin{align}\label{eq:slab-upper}
Z_w(u, \infty)
\begin{aligned}
&\;\le\; \sum_{i} N(a_i)\,
n^{w a_i + w\delta}\\
&\;\le\; 2\Big(\frac{2A}{\delta} + 1\Big)\\
&\qquad{}\cdot
n^{\max_i \left[
\max(f(a_i),\, \varepsilon) + w a_i
\right] + w\delta}.
\end{aligned}
\end{align}
For each contributing $a_i$ let $a_i' \coloneqq (a_i \vee u) \wedge 2c \in
[u, 2c]$, so that $|a_i - a_i'| \le \delta$. If the inner maximum is
$f(a_i)$, then $f(a_i) + w a_i = g_w(a_i) \le g_w(a_i') + L_w \delta \le
G_w[u, 2c] + L_w\delta$. If the inner maximum is $\varepsilon$, then,
since $f(a_i') \ge 0$,
\begin{align*}
\varepsilon + w a_i
&\;\le\; \varepsilon + w a_i' + w\delta\\
&\;\le\; \varepsilon + g_w(a_i') + w\delta\\
&\;\le\; G_w[u, 2c] + \varepsilon + w\delta.
\end{align*}
Substituting into Eq.~\eqref{eq:slab-upper} and using
Eq.~\eqref{eq:level-parameters},
\begin{align*}
\frac{\log Z_w(u, \infty)}{\log n}
&\;\le\; G_w[u, 2c]
+ (L_w + 2w)\delta + \varepsilon\\
&\qquad{}
+ \frac{\log\big(2(2A/\delta + 1)\big)}{\log n}\\
&\;\le\; G_w[u, 2c] + \gamma,
\end{align*}
for all $n \ge n_2(c, w, \gamma)$, on an event of probability $\to 1$.
The bounded-window variant $Z_w(u, v]$ is identical with the slabs
restricted to those meeting $(u, v]$ (bottoms in $[u - \delta, v]$, and
$a_i' \coloneqq (a_i \vee u) \wedge v$), giving the exponent
$G_w[u, v] + \gamma$. For $u = -\infty$, the additional contribution of
logits $\le 0$ is $Z_w(-\infty, 0] \le n \cdot 1 = n^{g_w(0)} \le
n^{G_w[0, v]}$ (at most $n$ terms, each $e^{w\ell_j} \le 1$, and $0 \in
[0, v]$), so
\begin{align*}
Z_w(-\infty, v]
&\;\le\; n^{G_w[0,v]}
+ n^{G_w[0,v] + \gamma/2}\\
&\;\le\; n^{G_w[0,v] + \gamma},
\end{align*}
for $n$ large, after running the $u \ge 0$ argument at $\gamma/2$ on
$Z_w(0, v]$. This proves (i).

\textbf{Step 3 (proof of (ii): window lower bound).} Let $a^\star \in [u, v]$ attain
$G_w[u,v] = g_w(a^\star)$; since $g_w$ is strictly increasing on
$[0,\, 2wc^2 \wedge 2c] \supseteq [0,\, 2c^2 \wedge v]$, we have
$a^\star \ge 2c^2 \wedge v > 0$. Choose the window
\begin{align*}
(b_1, b_2]
&\;\coloneqq\;
\left\{
\begin{aligned}
&\big(\max(u,\, a^\star - \rho_0),\;
a^\star\big],\\
&\quad a^\star > u, \quad
\rho_0 \coloneqq \delta \wedge \tfrac{a^\star}{2},\\
&\big(u,\; u + (\delta \wedge (v - u))\big],\\
&\quad a^\star = u,
\end{aligned}
\right.
\end{align*}
and write $\rho \coloneqq b_2 - b_1 > 0$. In both cases: $(b_1, b_2]
\subseteq (u, v]$; $\rho$ and $b_1$ are constants depending only on
$(c, w, u, v, \gamma)$; $b_1 \ge a^\star/2 > 0$ (in the first case $b_1 \ge
a^\star - \rho_0 \ge a^\star/2$; in the second $b_1 = u = a^\star$); every
$b \in [b_1, b_2]$ has $|b - a^\star| \le \delta$ and hence
\begin{align}\label{eq:window-near-max}
g_w(b) \;\ge\; G_w[u,v] - L_w \delta
\qquad \text{for all } b \in [b_1, b_2].
\end{align}
Moreover $b_1 = b_2 - \rho \le v - \rho \le 2c - \rho$, so
\begin{align}\label{eq:f-window-floor}
\begin{aligned}
f(b_1)
&\;=\; \frac{(2c - b_1)(2c + b_1)}{4c^2}\\
&\;\ge\; \frac{\rho \cdot 2c}{4c^2}\\
&\;=\; \frac{\rho}{2c}\\
&\;=:\; 2\kappa_1 \;>\; 0,
\end{aligned}
\end{align}
a constant. Now fix $\varepsilon_0 \coloneqq \frac{\kappa_1}{2} \wedge
\frac{b_1 \rho}{4c^2}$ in Eq.~\eqref{eq:level-parameters}. The window
count $W \coloneqq N(b_1) - N(b_2) \sim \Bin\big(n,\, p_n(b_1) -
p_n(b_2)\big)$ has mean
\begin{align*}
\E W
&\;=\; \E N(b_1)\,\Big(
1 - \frac{\E N(b_2)}{\E N(b_1)} \Big)\\
&\;\ge\; \E N(b_1)\,\big(
1 - n^{f(b_2) - f(b_1) + \varepsilon} \big)\\
&\;\ge\; \tfrac{1}{2}\, n^{f(b_1) - 2\varepsilon}\\
&\;\ge\; n^{\varepsilon},
\end{align*}
for all $n \ge n_3(c, w, u, v, \gamma)$, where the second step uses
\Cref{lem:count-asymptotics:upper,lem:count-asymptotics:lower}, the third
step uses \Cref{lem:count-asymptotics:lower} again together with
\begin{align*}
f(b_2) - f(b_1)
&\;=\; -\frac{b_2^2 - b_1^2}{4c^2}\\
&\;\le\; -\frac{2 b_1 \rho}{4 c^2}\\
&\;=\; -\frac{b_1\rho}{2c^2}\\
&\;\le\; -2\varepsilon\\
&\qquad (\text{by } \varepsilon \le \varepsilon_0
\le b_1\rho/(4c^2)),
\end{align*}
and the last step holds because Eq.~\eqref{eq:f-window-floor} and
$\varepsilon \le \varepsilon_0 \le \kappa_1/2$ give $f(b_1) - 2\varepsilon
\ge 2\kappa_1 - \kappa_1 = \kappa_1 \ge 2\varepsilon$, whence $\tfrac12
n^{f(b_1) - 2\varepsilon} \ge \tfrac12 n^{2\varepsilon} \ge
n^{\varepsilon}$ once $n^{\varepsilon} \ge 2$. By
\Cref{lem:chernoff:lower}, $\Pr(W \le \E W/2) \le e^{-\E W/8} \le
e^{-n^{\varepsilon}/8} \to 0$. On $\{W \ge \E W/2\}$, every window point
carries weight at least $n^{w b_1}$, so by
Eq.~\eqref{eq:window-near-max} and Eq.~\eqref{eq:level-parameters},
\begin{align*}
Z_w(u, v]
&\;\ge\; W \cdot n^{w b_1}\\
&\;\ge\; \tfrac14\,
n^{f(b_1) - 2\varepsilon + w b_1}\\
&\;=\; \tfrac14\, n^{g_w(b_1) - 2\varepsilon}\\
&\;\ge\; \tfrac14\,
n^{G_w[u,v] - L_w\delta - 2\varepsilon}\\
&\;\ge\; n^{G_w[u,v] - \gamma},
\end{align*}
for $n \ge n_4(c, w, u, v, \gamma)$. This proves (ii).

\textbf{Step 4 (proof of (iii): full sums).} Apply (i) with $(u, v] = (-\infty,
\infty)$ --- on the Step-1 event $Z_w = Z_w(-\infty, 2c+\delta]$ and the
slab bound gives exponent $\le G_w[0, 2c] + \gamma = F_w(c) + \gamma$ ---
and (ii) with $(u, v] = (0, 2c]$, for which $G_w[0, 2c] = F_w(c)$ and
$Z_w \ge Z_w(0, 2c]$. Combining, $\Pr(|\log Z_w/\log n - F_w(c)| > \gamma)
\to 0$ for every $\gamma > 0$. This completes the proof.
\end{proof}

\begin{corollary}[Maximal logit]\label{cor:max-logit}
Under $\Ens(c)$, $\ell_{(1)} \coloneqq \max_{j \in [n]} \ell_j$ satisfies
$\ell_{(1)}/\log n \toP 2c$.
\end{corollary}

\begin{proof}
Fix $\delta \in (0, \min(1, 2c))$ (it suffices to treat arbitrarily small
$\delta$, so this restriction is without loss for the limit). For the upper
bound, Eq.~\eqref{eq:no-high-points}
gives $\Pr(\ell_{(1)} > (2c+\delta)\log n) = \Pr(N(2c+\delta) \ge 1) \to
0$. For the lower bound, \Cref{lem:count-asymptotics:lower} with
$\varepsilon = f(2c-\delta)/2$ gives $\E N(2c - \delta) \ge
n^{f(2c-\delta)/2} \to \infty$, since
\begin{align*}
f(2c - \delta) \;=\; \frac{4c\delta - \delta^2}{4c^2} \;>\; 0,
\end{align*}
so by \Cref{lem:chernoff:lower}, $\Pr(N(2c-\delta) = 0) \le
\Pr\big(N(2c-\delta) \le \tfrac12 \E N(2c - \delta)\big) \le e^{-\E
N(2c-\delta)/8} \to 0$; on $\{N(2c - \delta) \ge 1\}$ we have
$\ell_{(1)} > (2c-\delta)\log n$. This completes the proof.
\end{proof}

\section{Dispersed Phase: Free Energy, Max Weight, Coverage}\label{sec:app-dispersed}

This appendix proves \cref{thm:main-scaled}(i) and the \textsc{Iid} rows of \cref{thm:main-ladder-dense,tab:main-ladder} of the main text.

Throughout this section we work under $\Ens(c)$ of \Cref{def:ensembles} with
$0 < c < 1$ fixed (except in \Cref{cor:t1-fixed-sigma}, which concerns
$\Ens_\sigma$), and we freely use the notation of \Cref{sec:app-level}:
$N(a)$, $Z_w(u,v]$, $f$, $g_w$, $g = g_1$, $F_w(c)$, $F(c) = F_1(c)$. Since
$c < 1$, the free energy is $F(c) = 1 + c^2$, attained at the interior
maximiser $a^\circ = 2c^2 < 2c$ of $g$ on $[0, 2c]$. For $\beta \in (0,1)$
the \emph{coverage level} is
\begin{align}\label{eq:abeta}
\begin{split}
a_\beta \;\coloneqq\; 2c\sqrt{1 - \beta},
\\
&\text{so that}\\
&f(a_\beta) \;=\; 1 - \frac{4c^2(1-\beta)}{4c^2}\\
&\hspace{8em} \;=\; \beta,\\
&a_\beta \in (0, 2c),
\end{split}
\end{align}
and the comparison with the maximiser is
\begin{align}\label{eq:abeta-vs-interior}
\begin{split}
a_\beta < 2c^2
&\;\Longleftrightarrow\; \sqrt{1-\beta} < c\\
&\;\Longleftrightarrow\; \beta > 1 - c^2.
\end{split}
\end{align}
We write $\ell_{(1)} \ge \ell_{(2)} \ge \cdots \ge \ell_{(n)}$ for the
sorted logits; the logits are a.s.\ pairwise distinct, and since
$x \mapsto e^x/Z$ is strictly increasing, the weight order of
\Cref{def:coverage} agrees a.s.\ with the logit order, i.e.\
$A_{(j)} = e^{\ell_{(j)}}/Z$ for all $j$.

\subsection{A rank sandwich for top-\texorpdfstring{$k$}{k} sets}

The following lemma localises the top-$\lceil n^\beta \rceil$ set between
two deterministic logit levels; it drives both the coverage threshold in
\Cref{thm:t1-main} and the value-error exponents in \Cref{cor:value-order}.

\begin{lemma}[Rank sandwich]\label{lem:rank-sandwich}
Fix $c > 0$, $\beta \in (0,1)$, and $\rho \in (0, a_\beta)$. Let
$k_n \coloneqq \lceil n^\beta \rceil$ and let $S_n \subseteq [n]$ be the set
of indices of the $k_n$ largest weights (ties broken by index; a.s.\
immaterial). Then there is $n_5(c, \beta, \rho)$ such that for all
$n \ge n_5$, with probability at least $1 - 2e^{-n^{\beta}/8}$,
simultaneously for every $w \ge 0$:
\begin{align}
\begin{aligned}
Z_w(a_\beta + \rho,\, \infty)
&\;\le\; \sum_{j \in S_n} e^{w \ell_j}\\
&\;\le\; Z_w(a_\beta - \rho,\, \infty),
\end{aligned}
\label{eq:sandwich-kept}\\
\begin{aligned}
Z_w(-\infty,\, a_\beta - \rho]
&\;\le\; \sum_{j \notin S_n} e^{w \ell_j}\\
&\;\le\; Z_w(-\infty,\, a_\beta + \rho].
\end{aligned}
\label{eq:sandwich-tail}
\end{align}
\end{lemma}

\begin{proof}
The strategy is to show that the rank-$k_n$ logit lies between the two
levels with the stated probability; the four inequalities are then
deterministic consequences. Since $N(a) = \#\{j : \ell_j > a \log n\}$,
for every $a \in \R$ and $k \in [n]$,
\begin{align}\label{eq:rank-count-duality}
\begin{gathered}
\{ N(a) \ge k \}\\
{}=\{ \ell_{(k)} > a \log n \},
\end{gathered}
\end{align}
both events stating that at least $k$ logits exceed $a \log n$. Consider
the event
\begin{align*}
\begin{aligned}
\mathcal{R}_n \;\coloneqq\;&
\{ N(a_\beta + \rho) < k_n \}\\
&{}\cap \{ N(a_\beta - \rho) \ge k_n \}\\
={}& \big\{ (a_\beta - \rho)\log n < \ell_{(k_n)}\\
&\hspace{3em} \le (a_\beta+\rho)\log n \big\}.
\end{aligned}
\end{align*}
On $\mathcal{R}_n$: every index $j$ with $\ell_j > (a_\beta + \rho)\log n
\ge \ell_{(k_n)}$ has rank $< k_n$, hence lies in $S_n$, and every kept
index $j \in S_n$ has $\ell_j \ge \ell_{(k_n)} > (a_\beta - \rho)\log n$;
therefore, for every $w \ge 0$,
\begin{align*}
\begin{aligned}
\sum_{j \in S_n} e^{w\ell_j}
&\;\ge\; \sum_{\substack{j:\, \ell_j >
(a_\beta+\rho)\log n}} e^{w\ell_j}\\
&\;=\; Z_w(a_\beta + \rho, \infty),
\\
\sum_{j \in S_n} e^{w\ell_j}
&\;\le\; \sum_{\substack{j:\, \ell_j >
(a_\beta-\rho)\log n}} e^{w\ell_j}\\
&\;=\; Z_w(a_\beta - \rho, \infty),
\end{aligned}
\end{align*}
which is Eq.~\eqref{eq:sandwich-kept}; subtracting each side from the full
sum $Z_w = \sum_{j} e^{w\ell_j}$ gives Eq.~\eqref{eq:sandwich-tail}. It
remains to bound $\Pr(\mathcal{R}_n^{\,c})$. For the upper level, since
$a_\beta + \rho > a_\beta \ge 0$, \Cref{lem:count-asymptotics:upper} and
$f(a_\beta+\rho) < f(a_\beta) = \beta$ give
\begin{align*}
\begin{aligned}
\E N(a_\beta + \rho)
&\;\le\; n^{f(a_\beta + \rho)}
\;=\; n^{\beta - \zeta_1},\\
\zeta_1
&\coloneqq \beta - f(a_\beta + \rho)\\
&=\frac{2 a_\beta \rho + \rho^2}{4c^2}
\;>\; 0,
\end{aligned}
\end{align*}
so for $n \ge n_5$ we have $k_n/2 \ge n^{\beta}/2 \ge n^{\beta - \zeta_1}
\ge \E N(a_\beta+\rho)$, and \Cref{lem:chernoff:upper} with
$t = k_n/2$ yields
\begin{align*}
\Pr\big( N(a_\beta + \rho) \ge k_n \big)
\;\le\; e^{-k_n/6} \;\le\; e^{-n^{\beta}/6}.
\end{align*}
For the lower level, $f(a_\beta - \rho) - \beta = (2a_\beta \rho -
\rho^2)/(4c^2) > 0$ (as $0 < \rho < a_\beta$), so
\Cref{lem:count-asymptotics:lower} with $\varepsilon \coloneqq
(f(a_\beta-\rho) - \beta)/2$ and $A = 2c+2$ gives, for $n \ge n_5$,
\begin{align*}
\E N(a_\beta - \rho) &\;\ge\; n^{\beta + \varepsilon}
\\
\;\ge\; \max\big(2\,k_n,\; n^{\beta}\big),\\
\Pr\big( N(a_\beta - \rho) < k_n \big)
&\;\le\; \Pr\Big(
N(a_\beta-\rho)\\
&\hspace{3em}\le \tfrac12 \E N(a_\beta-\rho)
\Big)\\
&\;\le\; e^{-\E N(a_\beta-\rho)/8}
\\
\;\le\; e^{-n^{\beta}/8},
\end{align*}
where the second chain uses $k_n \le \E N(a_\beta-\rho)/2$ (first step),
\Cref{lem:chernoff:lower} (second step), and $\E N(a_\beta -\rho) \ge
n^\beta$ (last step). By a union bound over the two levels,
$\Pr(\mathcal{R}_n^{\,c}) \le e^{-n^\beta/6} + e^{-n^\beta/8} \le
2e^{-n^{\beta}/8}$. This completes the proof.
\end{proof}

\subsection{Main theorem of the dispersed phase}

\begin{theorem}[Free energy, maximal weight, sharp coverage threshold; formal version of \cref{thm:main-scaled}(i)]\label{thm:t1-main}
Let $0 < c < 1$ be fixed and consider $\Ens(c)$. Then:
\begin{enumerate}[label=(\roman*), ref=\ref*{thm:t1-main}(\roman*)]
\item\label{thm:t1-main:freeenergy} $\dfrac{\log Z}{\log n} \toP 1 + c^2$;
\item\label{thm:t1-main:maxweight} $\dfrac{\log A_{(1)}}{\log n} \toP -(1-c)^2$;
\item\label{thm:t1-main:coverage} for every fixed $\beta \in (0,1)$, with
$k_n = \lceil n^\beta \rceil$:
\begin{align*}
\begin{aligned}
p(k_n) &\toP 1 \ \text{ if } \beta > 1 - c^2,\\
p(k_n) &\toP 0 \ \text{ if } \beta < 1 - c^2,
\end{aligned}
\end{align*}
both directions of the sharp threshold;
\item\label{thm:t1-main:keta} for every fixed $\eta \in (0,1)$:
$\dfrac{\log k_\eta}{\log n} \toP 1 - c^2$.
\end{enumerate}
\end{theorem}

\begin{proof}
We prove the four parts in order; parts (iii)--(iv) combine
\Cref{lem:rank-sandwich} with the restricted free energies of
\Cref{prop:restricted-z}.

\textbf{Step 1 (proof of (i): free energy).} This is \Cref{prop:restricted-z:full} with $w = 1$,
since $F_1(c) = 1 + c^2$ for $c \le 1$ by Eq.~\eqref{eq:Fw}.

\textbf{Step 2 (proof of (ii): maximal weight).} Since $A_{(1)} = e^{\ell_{(1)}}/Z$,
\begin{align*}
\frac{\log A_{(1)}}{\log n}
&\;=\; \frac{\ell_{(1)}}{\log n}
- \frac{\log Z}{\log n}\\
&\;\toP\; 2c - (1 + c^2)\\
&\;=\; -(1-c)^2,
\end{align*}
where the convergence holds because the two summands converge in
probability separately, by \Cref{cor:max-logit} and part (i).

\textbf{Step 3 (proof of (iii), case $\beta > 1 - c^2$: dropped mass vanishes).} By
Eq.~\eqref{eq:abeta-vs-interior} we have $a_\beta < 2c^2$; set
\begin{align*}
\rho \;\coloneqq\; \frac{2c^2 - a_\beta}{2} \wedge \frac{a_\beta}{2},
\\
a' \;\coloneqq\; a_\beta + \rho \;<\; 2c^2,
\\
\kappa \;\coloneqq\; F(c) - g(a') \;>\; 0,
\end{align*}
where $\kappa > 0$ because $g$ is strictly increasing on $[0, 2c^2]$
(\,$g'(a) = 1 - a/(2c^2) > 0$ for $a < 2c^2$\,) and $g(2c^2) = F(c)$. On
the event of \Cref{lem:rank-sandwich} (with this $\rho$, $w = 1$),
Eq.~\eqref{eq:sandwich-tail} gives
\begin{align*}
1 - p(k_n)
\;=\; \sum_{j \notin S_n} A_j
\\
\;=\; \frac{1}{Z}\sum_{j \notin S_n} e^{\ell_j}
\\
\;\le\; \frac{Z_1(-\infty,\, a']}{Z}.
\end{align*}
By \Cref{prop:restricted-z:upper} with $\gamma = \kappa/4$ and window
$(-\infty, a']$, using $G_1[0, a'] = g(a')$ ($g$ increasing on
$[0,a'] \subseteq [0, 2c^2]$), and by \Cref{prop:restricted-z:lower} on
$(0, 2c]$,
\begin{align*}
Z_1(-\infty,\, a'] \;\le\; n^{g(a') + \kappa/4}
\\
\text{and}\quad
Z \;\ge\; Z_1(0, 2c] \;\ge\; n^{F(c) - \kappa/4}
\\
\text{with probability} \to 1,
\end{align*}
so that, by a union bound over the three events (rank sandwich and the two
displays), with probability $\to 1$,
\begin{align*}
1 - p(k_n)
\;\le\; n^{g(a') - F(c) + \kappa/2}
\\
\;=\; n^{-\kappa/2}
\;\longrightarrow\; 0,
\end{align*}
which proves $p(k_n) \toP 1$; moreover the dropped mass is bounded by the
explicit polynomial rate $n^{-\kappa(\beta,c)/2}$ on that event.

\textbf{Step 4 (proof of (iii), case $\beta < 1 - c^2$: kept mass vanishes).} Now
$a_\beta > 2c^2$; set
\begin{align*}
\rho \;\coloneqq\; \frac{a_\beta - 2c^2}{2},
\\
a'' \;\coloneqq\; a_\beta - \rho \;=\; \frac{a_\beta + 2c^2}{2} \;\in\;
(2c^2,\, a_\beta),
\\
\kappa' \;\coloneqq\; F(c) - g(a'') \;>\; 0,
\end{align*}
where $\kappa' > 0$ because $g$ is strictly decreasing on $[2c^2, 2c]$ and
$a'' > 2c^2$. On the event of \Cref{lem:rank-sandwich},
Eq.~\eqref{eq:sandwich-kept} gives, with probability $\to 1$,
\begin{align*}
p(k_n)
&\;=\; \frac{1}{Z} \sum_{j \in S_n} e^{\ell_j}
\\
\;\le\; \frac{Z_1(a'',\, \infty)}{Z}
\\
\;\le\; n^{G_1[a'',\, 2c] - F(c) + \kappa'/2}\\
&\;=\; n^{g(a'') - F(c) + \kappa'/2}
\\
\;=\; n^{-\kappa'/2}
\;\longrightarrow\; 0,
\end{align*}
where the second inequality applies \Cref{prop:restricted-z:upper}
(window $(a'', \infty)$, $\gamma = \kappa'/4$) together with $Z \ge
n^{F(c) - \kappa'/4}$ as in the previous case, again after a union bound
over the three events, and the following equality uses
$G_1[a'', 2c] = g(a'')$ ($g$ decreasing on $[a'', 2c] \subseteq
[2c^2, 2c]$).

\textbf{Step 5 (proof of (iv): budget exponent).} Fix $\eta \in (0,1)$ and $\varepsilon \in
(0,\, \min(1-c^2,\, c^2))$, and set $\beta_\pm \coloneqq 1 - c^2 \pm
\varepsilon \in (0, 1)$. By part (iii) at $\beta_+ > 1-c^2$, with
probability $\to 1$,
\begin{align*}
p\big(\lceil n^{\beta_+} \rceil\big) \;\ge\; 1 - \eta
\\
\Longrightarrow\qquad
k_\eta \;\le\; \lceil n^{\beta_+} \rceil \;\le\; 2 n^{\beta_+},
\end{align*}
using that $p(\cdot)$ is nondecreasing and $k_\eta$ is the minimal
budget reaching $1 - \eta$ (\Cref{def:coverage}). By part (iii) at
$\beta_- < 1 - c^2$, with probability $\to 1$,
\begin{align*}
p\big(\lceil n^{\beta_-} \rceil\big) \;<\; 1 - \eta
\\
\Longrightarrow\qquad
k_\eta \;>\; \lceil n^{\beta_-} \rceil \;\ge\; n^{\beta_-}.
\end{align*}
On the intersection of the two events, for all $n$ with $\log 2/\log n \le
\varepsilon$,
\begin{align*}
1 - c^2 - \varepsilon
\;\le\; \frac{\log k_\eta}{\log n}
\\
\;\le\; \beta_+ + \frac{\log 2}{\log n}
\\
\;\le\; 1 - c^2 + 2\varepsilon.
\end{align*}
Since $\varepsilon \in (0, \min(1-c^2, c^2))$ was arbitrary and the bound
for smaller $\varepsilon$ implies that for larger, $\log k_\eta / \log n
\toP 1 - c^2$. This completes the proof.
\end{proof}

\begin{remark}[Effective support $n^{1-c^2}$]\label{rem:t1-interpretation}
\Cref{thm:t1-main:coverage,thm:t1-main:keta} say that the coverage
profile has a sharp threshold at budget $n^{1-c^2}$: keeping any
polynomially larger budget captures all but an $n^{-\Omega(1)}$ fraction of
the mass, keeping any polynomially smaller budget captures only an
$n^{-\Omega(1)}$ fraction. In the dispersed phase a softmax row is
``effectively dense'': its effective support is a power of $n$ that
approaches $n$ as $c \downarrow 0$ and shrinks to $n^{0}$ as
$c \uparrow 1$, connecting continuously to the condensed phase of
\Cref{sec:app-condensed}.
\end{remark}

\subsection{Fixed-variance corollary: generic rows are dense}

\begin{corollary}[Fixed bulk]\label{cor:t1-fixed-sigma}
Let $\sigma > 0$ be fixed and consider $\Ens_\sigma$. Then:
\begin{enumerate}[label=(\roman*), ref=\ref*{cor:t1-fixed-sigma}(\roman*)]
\item\label{cor:t1-fixed-sigma:max} for every fixed $\delta > 0$, with
probability $\to 1$,
$\;A_{(1)} \le n^{-1}\, e^{(1+\delta)\sigma\sqrt{2\log n}}$;
\item\label{cor:t1-fixed-sigma:keta} for every fixed $\eta \in (0,1)$:
$\dfrac{\log k_\eta}{\log n} \toP 1$.
\end{enumerate}
\end{corollary}

\begin{proof}
The strategy is elementary: bound the maximal logit by the Gaussian tail,
bound $Z$ below by the number of nonnegative logits, and combine. Write
$\ell_{(1)} = \max_j \ell_j$ and fix $\delta \in (0, 1)$ (the claim for
$\delta \ge 1$ follows from that for smaller $\delta$). By a union bound
over the $n$ logits and \Cref{fac:gaussian-tail} ($\barPhi(x) \le
e^{-x^2/2}$),
\begin{align}\label{eq:fixed-sigma-max}
\begin{split}
\Pr\Big(
\begin{aligned}
\ell_{(1)} >{}&
\big(1+\tfrac{\delta}{2}\big)\sigma\\[-2pt]
&\sqrt{2\log n}
\end{aligned}
\Big)
&\;\le\; n\, \barPhi\Big(
\big(1+\tfrac{\delta}{2}\big)\sqrt{2\log n}
\Big)\\
&\;\le\; n\, n^{-(1+\delta/2)^2}
\\
&\;\le\; n^{-\delta}
\;\longrightarrow\; 0.
\end{split}
\end{align}
Next, $N_0 \coloneqq \#\{j : \ell_j > 0\} \sim \Bin(n, 1/2)$, so
\Cref{lem:chernoff:lower} gives
\begin{align}\label{eq:fixed-sigma-z}
\begin{aligned}
\Pr\big( N_0 \le n/4 \big)
&\;\le\; e^{-n/16},\\
\text{and on } \{N_0 > n/4\}:\qquad
Z
&\;\ge\; \sum_{j:\,\ell_j>0} e^{\ell_j}\\
&\;\ge\; N_0
\;\ge\; \frac{n}{4}.
\end{aligned}
\end{align}
By a union bound over Eq.~\eqref{eq:fixed-sigma-max} and
Eq.~\eqref{eq:fixed-sigma-z}, with probability $\to 1$,
\begin{align*}
A_{(1)} \;=\; \frac{e^{\ell_{(1)}}}{Z}
\\
\;\le\; \frac{4}{n}\,
e^{(1+\delta/2)\sigma\sqrt{2\log n}}
\\
\;\le\; \frac{1}{n}\,
e^{(1+\delta)\sigma\sqrt{2\log n}},
\end{align*}
where the last step holds for all $n \ge n_6(\delta, \sigma)$ since
$e^{(\delta/2)\sigma\sqrt{2\log n}} \ge 4$ eventually. This proves (i).
For (ii), $k_\eta \le n$ always, so $\log k_\eta/\log n \le 1$
deterministically. For the lower bound fix $\beta \in (0,1)$; by part (i)
with $\delta = 1$ and monotonicity of the sorted weights, with
probability $\to 1$,
\begin{align*}
p\big(\lceil n^\beta \rceil\big)
&\;\le\; \lceil n^\beta \rceil\, A_{(1)}
\\
&\;\le\; 2 n^{\beta} \cdot
\frac{e^{2\sigma\sqrt{2\log n}}}{n}\\
&\;=\; 2\, n^{\beta - 1}\,
e^{2\sigma\sqrt{2\log n}}\\
&\;\longrightarrow\; 0,
\end{align*}
since $e^{2\sigma\sqrt{2\log n}} = n^{2\sigma\sqrt{2}/\sqrt{\log n}} =
n^{o(1)}$ while $\beta - 1 < 0$. Hence with probability $\to 1$,
$p(\lceil n^\beta\rceil) < 1 - \eta$, so $k_\eta > n^{\beta}$ and
$\log k_\eta / \log n \ge \beta$; as $\beta \in (0,1)$ was arbitrary,
$\log k_\eta/\log n \toP 1$. This completes the proof.
\end{proof}

\subsection{Value-error corollaries}

We now quantify what the coverage threshold means for the attention
\emph{output}. The proposition is an exact identity for any kept set; the
corollary evaluates its order under $\Ens(c)$ and exhibits the bias floor
of uncompensated truncation.

\begin{proposition}[Conditional value-error identities]\label{prop:value-identity}
Let the value vectors $V_1, \dots, V_n$ be i.i.d.\ random vectors in
$\R^d$, independent of the logits $\ell$, with $\E\norm{V_1}_2^2 < \infty$,
mean $\mu_V \coloneqq \E V_1$, covariance $\Sigma_V$, and
$\sigma_V^2 \coloneqq \Tr(\Sigma_V)/d$. Let $S$ be any
$\sigma(\ell)$-measurable kept set with $0 < |S| < n$. Then, almost
surely,
\begin{gather}
\begin{aligned}
\E\big[\, \wt{O}_S - O \,\big|\, \ell \,\big] &\;=\; 0,\\
\E\big[\, \norm{\wt{O}_S - O}_2^2 \,\big|\, \ell \,\big]
&\;=\; d\,\sigma_V^2
\left[ \frac{(1-p_S)^2}{p_S^2}\, Q_S \right.\\
&\hspace{8em}\left. {}+ Q_T \right],
\end{aligned}
\label{eq:value-identity}\\
\begin{aligned}
\E\big[\, \norm{O_S - O}_2^2 \,\big|\, \ell \,\big]
&\;=\; (1 - p_S)^2 \norm{\mu_V}_2^2\\
&\hspace{4em} {}+ d\,\sigma_V^2\, Q_T,
\end{aligned}
\label{eq:value-trunc}
\end{gather}
where $Q_S \coloneqq \sum_{j \in S} A_j^2$ and
$Q_T \coloneqq \sum_{j \notin S} A_j^2$.
\end{proposition}

\begin{proof}
The strategy is to write each error as a linear combination of the $V_j$
with $\sigma(\ell)$-measurable coefficients and integrate over $V$ only,
which is licensed by the independence of $(V_1, \dots, V_n)$ from $\ell$.
From \Cref{def:schemes},
\begin{align}\label{eq:value-cj}
\begin{aligned}
\wt{O}_S - O
&\;=\; \sum_{j=1}^n c_j V_j,\\
c_j \;\coloneqq\;
&\\[-2pt]
&
\begin{cases}
A_j \dfrac{1 - p_S}{p_S}, & j \in S,\\
-A_j, & j \notin S,
\end{cases}
\\
\sum_{j=1}^n c_j
&\;=\; \frac{1-p_S}{p_S}\, p_S - (1 - p_S)\\
&\;=\; 0.
\end{aligned}
\end{align}
Conditionally on $\ell$, the $c_j$ are constants and the $V_j$ are i.i.d.;
hence the conditional mean is $(\sum_j c_j)\mu_V = 0$, proving the first
claim of Eq.~\eqref{eq:value-identity}, and, writing $V_j = \mu_V +
(V_j - \mu_V)$ and using $\sum_j c_j = 0$,
\begin{align*}
\E\big[ \norm{\wt O_S - O}_2^2 \,\big|\, \ell \big]
&\;=\; \E\Bigg[
\begin{aligned}
\Big\| \sum_{j} c_j\\[-2pt]
(V_j - \mu_V) \Big\|_2^2
\end{aligned}
\,\Big|\, \ell \Bigg] \\
&\;=\; \sum_{j, j'} c_j c_{j'}\,
\\[-2pt]
&\hspace{2em}
\E\big[
(V_j - \mu_V)^\top\\[-2pt]
&\hspace{6em}(V_{j'} - \mu_V)
\big]\\
&\;=\; \Tr(\Sigma_V) \sum_{j} c_j^2,
\end{align*}
where the last step holds because the cross terms ($j \ne j'$) vanish by
independence and centring of the $V_j$, while each diagonal term equals
$\E\norm{V_1 - \mu_V}_2^2 = \Tr(\Sigma_V) = d\sigma_V^2$. The coefficient
sum evaluates, by Eq.~\eqref{eq:value-cj}, to
\begin{align*}
\sum_{j=1}^n c_j^2
&\;=\; \frac{(1-p_S)^2}{p_S^2}
\sum_{j \in S} A_j^2\\
&\hspace{4em} {}+ \sum_{j \notin S} A_j^2\\
&\;=\; \frac{(1-p_S)^2}{p_S^2}\, Q_S + Q_T,
\end{align*}
and substituting it into the preceding display proves
Eq.~\eqref{eq:value-identity}. For Eq.~\eqref{eq:value-trunc},
\Cref{prop:bias-identity:trunc} gives $O_S - O = -\sum_{j\notin S}A_jV_j$,
whose conditional mean is $-(1-p_S)\mu_V$, so by the bias--variance
decomposition $\E\norm{X}_2^2 = \norm{\E X}_2^2 + \E\norm{X - \E X}_2^2$
applied conditionally,
\begin{align*}
\E\big[ \norm{O_S - O}_2^2 \,\big|\, \ell \big]
&\;=\; (1-p_S)^2 \norm{\mu_V}_2^2
\\
&{}+ \E\Bigg[
\begin{aligned}
\Big\| \sum_{j \notin S} A_j\\[-2pt]
(V_j - \mu_V) \Big\|_2^2
\end{aligned}
\,\Big|\, \ell\Bigg] \\
&\;=\; (1-p_S)^2 \norm{\mu_V}_2^2\\
&{}+ d\sigma_V^2\, Q_T,
\end{align*}
where the variance term is computed exactly as above (diagonal terms
only). This completes the proof.
\end{proof}

\begin{corollary}[Value-error orders in the dispersed phase]\label{cor:value-order}
Assume the setting of \Cref{prop:value-identity} with $\sigma_V > 0$,
let $0 < c < 1$, work under $\Ens(c)$, and let $S_n$ be the top-$k$
set by weight. Write $\mathrm{MSE}_n \coloneqq \E[\norm{\wt O_{S_n} -
O}_2^2 \mid \ell]$, $F = F(c) = 1+c^2$, and $h \coloneqq g_2 = f + 2a$
(the $w=2$ mass exponent of Eq.~\eqref{eq:exponent-functions}). Then:
\begin{enumerate}[label=(\roman*), ref=\ref*{cor:value-order}(\roman*)]
\item\label{cor:value-order:fixedk} (Bounded budget.) For every fixed
$k \ge 1$, with probability $\to 1$,
\begin{align*}
\frac{d\,\sigma_V^2}{2k}
\;\le\; \mathrm{MSE}_n
\;\le\; 2\, d\,\sigma_V^2,
\end{align*}
so the conditional RMSE is of exact order $\sigma_V \sqrt{d}$ (with
$k$-dependent constants): a non-vanishing fluctuation floor.
\item\label{cor:value-order:beta} (Polynomial budget below threshold.)
For every fixed $\beta \in (0, 1-c^2)$, with $k_n = \lceil n^\beta
\rceil$ and $a_\beta$ as in Eq.~\eqref{eq:abeta},
\begin{align*}
\begin{aligned}
\frac{\log \mathrm{MSE}_n}{\log n}
&\;\toP\; -\theta(\beta, c),\\
\theta(\beta,c)
&\;\coloneqq\; \min\Big(
2g(a_\beta) - h^{*}_{+},\\
&\hspace{8em} 2F - h^{*}_{-} \Big),
\end{aligned}
\end{align*}
where $h^*_+ \coloneqq \max_{a \in [a_\beta, 2c]} h(a)$ and
$h^*_- \coloneqq \max_{a \in [0, a_\beta]} h(a)$. Moreover
\begin{align*}
\theta(\beta, c)
\;\ge\; \min\big(
g(a_\beta) - 2c,\; F - a_\beta
\big)
\\
\;>\; 0,
\\
\text{and}\qquad 0
\;<\; \theta(\beta,c)
\;\le\; 2\big(g(a_\beta) - 2c\big)
\\
\;\longrightarrow\; 0 \ \text{ as } \beta \downarrow 0,
\end{align*}
so the renormalised error vanishes at a polynomial rate whose exponent
degrades to $0$ as the budget exponent shrinks, matching the bounded-$k$
floor of part (i).
\item\label{cor:value-order:bias} (Bias floor of uncompensated
truncation.) If $\mu_V \ne 0$ then, both for fixed $k$ and for
$k_n = \lceil n^\beta\rceil$ with $\beta \in (0, 1-c^2)$,
\begin{align*}
\E\big[ \norm{O_{S_n} - O}_2^2 \,\big|\, \ell \big]
\\
\;\ge\; (1 - p_{S_n})^2 \norm{\mu_V}_2^2
\\
\;=\; \big(1 - o_P(1)\big) \norm{\mu_V}_2^2,
\end{align*}
a conditional RMSE floor of order $\norm{\mu_V}_2$ --- removed exactly by
compensation (\Cref{prop:compensation}) or renormalisation
(Eq.~\eqref{eq:value-identity}: $\wt O_S$ is conditionally unbiased).
\end{enumerate}
\end{corollary}

\begin{proof}
The strategy is: for (i), squeeze $Q_S/p_S^2$ between $1/k$ and $1$ and
kill $Q_T$ and $p_S$ with the max-weight exponent; for (ii), convert
$Q_S$, $Q_T$, $p_S$ into restricted partition functions via
\Cref{lem:rank-sandwich} and read off the exponents from
\Cref{prop:restricted-z}; for (iii), apply Eq.~\eqref{eq:value-trunc}.
Throughout, $Q_S, Q_T, p_S$ are those of $S = S_n$.

\textbf{Step 1 (proof of (i): participation squeeze).} Fix $k$. For positive reals, the Cauchy--Schwarz
inequality $p_S^2 = (\sum_{j \in S} A_j)^2 \le k \sum_{j \in S}A_j^2$ and
the crude bound $\sum_{j \in S} A_j^2 \le (\sum_{j \in S} A_j)^2$ give
\begin{align}\label{eq:participation-squeeze}
\begin{aligned}
\frac{1}{k} \;\le\; \frac{Q_S}{p_S^2} \;\le\; 1,
\\
\text{and}\qquad
Q_T \;=\; \sum_{j \notin S_n} A_j^2
\\
\;\le\; A_{(1)} \sum_{j \notin S_n} A_j
\;\le\; A_{(1)}.
\end{aligned}
\end{align}
By \Cref{thm:t1-main:maxweight}, $A_{(1)} \toP 0$
(as $(1-c)^2 > 0$), hence also $p_{S_n} \le k\,A_{(1)} \toP 0$.
Substituting Eq.~\eqref{eq:participation-squeeze} into
Eq.~\eqref{eq:value-identity},
\begin{align*}
\begin{aligned}
\mathrm{MSE}_n
&\;\ge\; d\sigma_V^2\, \frac{(1-p_{S_n})^2}{k}
\\
&\;\ge\; \frac{d\sigma_V^2}{2k}
\\
\text{and}\qquad
\mathrm{MSE}_n
&\;\le\; d\sigma_V^2
\Big[ (1-p_{S_n})^2 + A_{(1)} \Big]
\\
&\;\le\; 2 d \sigma_V^2,
\end{aligned}
\end{align*}
both holding with probability $\to 1$: the left chain uses
$(1-p_{S_n})^2 \toP 1$, the right chain uses $(1-p_{S_n})^2 \le 1$ and
$A_{(1)} \toP 0$.

\textbf{Step 2 (proof of (ii): exponent extraction).} Since $0 < \beta < 1 - c^2$, we have $a_\beta \in
(2c^2, 2c)$ by Eq.~\eqref{eq:abeta-vs-interior}. Fix $\zeta \in (0,
(F - g(a_\beta))/2)$. Each map $u \mapsto G_w[u, v]$, $v \mapsto G_w[u,v]$
is Lipschitz on $[0,2c]$ with the constant $L_w$ of
\Cref{prop:restricted-z}'s proof (as a running maximum of the
$L_w$-Lipschitz $g_w$), so we may fix $\rho \in (0,\,
(a_\beta - 2c^2) \wedge (2c - a_\beta)/2)$ small enough that
\begin{align}\label{eq:rho-choice}
\begin{aligned}
\big| G_2[a_\beta \pm \rho,\, 2c] - h^*_+ \big| \;\le\; \frac{\zeta}{6},
\\
\big| G_2[0,\, a_\beta \pm \rho] - h^*_- \big| \;\le\; \frac{\zeta}{6},
\\
\big| g(a_\beta \pm \rho) - g(a_\beta) \big| \;\le\; \frac{\zeta}{6},
\end{aligned}
\end{align}
where $h^*_+ = G_2[a_\beta, 2c]$ and $h^*_- = G_2[0, a_\beta]$. Work on
the intersection of the rank-sandwich event of \Cref{lem:rank-sandwich}
and the following eight events, each of probability $\to 1$ by
\Cref{prop:restricted-z:upper,prop:restricted-z:lower,prop:restricted-z:full}
with $\gamma = \zeta/6$; a union bound over the nine events keeps the
intersection at probability $\to 1$:
\begin{align*}
\begin{aligned}
Z_2(a_\beta - \rho, \infty)
&\le n^{G_2[a_\beta-\rho,2c] + \zeta/6},\\
Z_2(a_\beta + \rho,\, 2c]
&\ge n^{G_2[a_\beta+\rho,2c] - \zeta/6},\\
Z_2(-\infty, a_\beta + \rho]
&\le n^{G_2[0,a_\beta+\rho] + \zeta/6},\\
Z_2(0, a_\beta - \rho]
&\ge n^{G_2[0,a_\beta-\rho] - \zeta/6},\\
Z_1(a_\beta - \rho, \infty)
&\le n^{g(a_\beta - \rho) + \zeta/6},\\
Z_1(a_\beta + \rho,\, 2c]
&\ge n^{g(a_\beta + \rho) - \zeta/6},\\
n^{F - \zeta/6}
&\;\le\; Z \;\le\; n^{F + \zeta/6},
\end{aligned}
\end{align*}
where we used $G_1[a_\beta \pm \rho, 2c] = g(a_\beta \pm \rho)$ ($g$
decreasing on $[a_\beta - \rho, 2c] \subseteq [2c^2, 2c]$). On this
intersection, Eq.~\eqref{eq:sandwich-kept} and
Eq.~\eqref{eq:sandwich-tail} with $w = 2$, the containments
$Z_2(a_\beta+\rho,\infty) \ge Z_2(a_\beta+\rho, 2c]$ and
$Z_2(-\infty, a_\beta - \rho] \ge Z_2(0, a_\beta - \rho]$, and
Eq.~\eqref{eq:rho-choice} give
\begin{align}\label{eq:qsqt-exponents}
\begin{split}
n^{h^*_+ - 2F - \zeta}
&\;\le\; Q_S\\
&\;=\; \frac{1}{Z^2}\sum_{j \in S_n} e^{2\ell_j}\\
\;\le\; n^{h^*_+ - 2F + \zeta},\\
n^{h^*_- - 2F - \zeta}
&\;\le\; Q_T
\;\le\; n^{h^*_- - 2F + \zeta},
\end{split}
\end{align}
and, with $w = 1$ in Eq.~\eqref{eq:sandwich-kept}, the kept mass is pinned
to its exponent $g(a_\beta) - F < 0$:
\begin{align}\label{eq:ps-exponent}
\begin{split}
n^{g(a_\beta) - F - \zeta}
&\;\le\; \frac{Z_1(a_\beta + \rho,\, 2c]}{Z}\\
\;\le\; p_{S_n}\\
&\;\le\; \frac{Z_1(a_\beta - \rho, \infty)}{Z}
\\
&\;\le\; n^{g(a_\beta) - F + \zeta}
\;\longrightarrow\; 0,
\end{split}
\end{align}
where $g(a_\beta) - F < 0$ holds strictly because $g$ is strictly
decreasing on $[2c^2, 2c]$, $a_\beta > 2c^2$, and $g(2c^2) = F$, and the
limit uses $\zeta < (F - g(a_\beta))/2$. Now combine: by
Eq.~\eqref{eq:value-identity},
\begin{align*}
\mathrm{MSE}_n
\;=\; d\sigma_V^2 \left[
(1-p_{S_n})^2\,\frac{Q_S}{p_{S_n}^2}
+ Q_T \right],
\end{align*}
and on our event $(1 - p_{S_n})^2 \in [1/2,\, 1]$ for $n$ large by
Eq.~\eqref{eq:ps-exponent}. Using $\max(x,y) \le x + y \le 2\max(x,y)$
for $x, y > 0$ together with
Eq.~\eqref{eq:qsqt-exponents} and Eq.~\eqref{eq:ps-exponent},
\begin{align*}
\begin{gathered}
\frac{1}{2}\,
n^{\max\left(
\substack{
h^*_+ - 2g(a_\beta),\\[-2pt]
h^*_- - 2F
}
\right) - 3\zeta}
\\
\;\le\;
\frac{\mathrm{MSE}_n}{d \sigma_V^2}\\
\;\le\;
4\, n^{\max\left(
\substack{
h^*_+ - 2g(a_\beta),\\[-2pt]
h^*_- - 2F
}
\right) + 3\zeta},
\end{gathered}
\end{align*}
where the exponent of $(1-p_{S_n})^2 Q_S/p_{S_n}^2$ lies in
$[\,h^*_+ - 2g(a_\beta) - 3\zeta,\; h^*_+ - 2g(a_\beta) + 3\zeta\,]$ by
Eq.~\eqref{eq:qsqt-exponents}, Eq.~\eqref{eq:ps-exponent}, and the
$(1-p_{S_n})^2$-squeeze. Since
$\max(h^*_+ - 2g(a_\beta), h^*_- - 2F) = -\theta(\beta,c)$ and $\zeta$
was arbitrary in $(0, (F-g(a_\beta))/2)$,
$\log \mathrm{MSE}_n/\log n \toP -\theta(\beta,c)$.

It remains to prove the bounds on $\theta$. For the first term, every
$a \in [a_\beta, 2c]$ satisfies $g(a) \le g(a_\beta)$ ($g$ decreasing
there), so with $h = g + a$ (as $g_2 = g_1 + a$ identically, by
Eq.~\eqref{eq:exponent-functions}),
\begin{align*}
2 g(a_\beta) - h^*_+
&\;=\; \min_{a \in [a_\beta, 2c]}
\big( 2g(a_\beta) - g(a) - a \big)\\
&\;\ge\; \min_{a \in [a_\beta, 2c]}
\big( g(a_\beta) - a \big)\\
&\;=\; g(a_\beta) - 2c
\\
&\;>\; g(2c) - 2c
\;=\; 0,
\end{align*}
where the strict inequality uses $g(a_\beta) > g(2c)$ ($g$ strictly
decreasing on $[2c^2, 2c]$, $2c^2 < a_\beta < 2c$). For the second term,
every $a \in [0, a_\beta]$ satisfies $g(a) \le F$ and $a \le a_\beta$, so
\begin{align*}
2F - h^*_-
&\;=\; \min_{a \in [0, a_\beta]}
\big( 2F - g(a) - a \big)\\
&\;\ge\; F - a_\beta\\
&\;=\; 1 + c^2 - 2c\sqrt{1-\beta}
\\
&\;>\; (1-c)^2
\;>\; 0.
\end{align*}
This proves $\theta \ge \min(g(a_\beta) - 2c, F - a_\beta) > 0$. For the
upper bound, $h^*_+ \ge h(2c) = g(2c) + 2c = 4c$, so
\begin{align*}
\theta
\;\le\; 2g(a_\beta) - h^*_+
\\
\;\le\; 2 g(a_\beta) - 4c
\\
&\;=\; 2\big( g(a_\beta) - 2c \big)\\
&\;\longrightarrow\; 2\big( g(2c) - 2c \big)
\;=\; 0\\
&\qquad \text{as } \beta \downarrow 0,
\end{align*}
since $a_\beta \to 2c$ as $\beta \downarrow 0$ and $g$ is continuous.

\textbf{Step 3 (proof of (iii): bias floor).} By Eq.~\eqref{eq:value-trunc}, dropping the
nonnegative variance term,
\begin{align*}
\E\big[ \norm{O_{S_n} - O}_2^2 \,\big|\, \ell \big]
\;\ge\; (1 - p_{S_n})^2 \norm{\mu_V}_2^2,
\end{align*}
and $p_{S_n} \toP 0$ in both regimes: for fixed $k$ by
$p_{S_n} \le k A_{(1)} \toP 0$ as in Step 1, and for $k_n = \lceil
n^\beta \rceil$, $\beta \in (0, 1-c^2)$, by Eq.~\eqref{eq:ps-exponent}.
This completes the proof.
\end{proof}

\section{Condensed Phase: Poisson--Dirichlet Limit and Coverage Law}\label{sec:app-condensed}

This appendix proves \cref{thm:main-scaled}(ii) of the main text.

Throughout this section we work under $\Ens(c)$ with $c > 1$ fixed and set
$\alpha \coloneqq 1/c \in (0,1)$. We use the classical Gumbel norming
constants (cf.\ \cite{leadbetter1983extremes}; everything we need about
them is proved inline in \Cref{lem:gumbel-calibration}):
\begin{align}\label{eq:gumbel-constants}
\begin{split}
&a_n \;\coloneqq\; \sqrt{2 \log n},\\
&r_n \;\coloneqq\;
\frac{\log \log n + \log 4\pi}{2},\\
&b_n \;\coloneqq\; \sqrt{2\log n}\\
&\hspace{1.8em}
- \frac{\log\log n + \log 4\pi}
{2\sqrt{2 \log n}}\\
&\hspace{1.8em}
\;=\; a_n - \frac{r_n}{a_n},
\end{split}
\end{align}
for $n \ge 3$ (so that $\log\log n > 0$), and the rescaled sorted logits
\begin{align}\label{eq:xjn}
x_j^{(n)} \;\coloneqq\; a_n \big( g_{(j)} - b_n \big),
\qquad j \in [n],
\end{align}
where $g_{(1)} \ge \cdots \ge g_{(n)}$ are the sorted standard Gaussians of
\Cref{def:ensembles} (so $x_1^{(n)} \ge \cdots \ge x_n^{(n)}$). Since
$\ell_{(j)} = \sigma_n g_{(j)}$ and $\sigma_n / a_n = c$, every sorted
weight factorises as
\begin{align}\label{eq:weight-cancellation}
\begin{split}
A_{(j)}
&\;=\; \frac{e^{\sigma_n g_{(j)}}}{\sum_{l=1}^n e^{\sigma_n g_{(l)}}}
\;=\; \frac{e^{\sigma_n b_n}\, e^{c\, x_j^{(n)}}}
{\sum_{l=1}^n e^{\sigma_n b_n}\, e^{c\, x_l^{(n)}}}\\
&\;=\; \frac{e^{c\, x_j^{(n)}}}{D_n},
\qquad
D_n \;\coloneqq\; \sum_{l=1}^n e^{c\, x_l^{(n)}},
\end{split}
\end{align}
so the weight vector is a continuous functional of the rescaled logits.
The programme of this section: calibrate the Gaussian tail at the Gumbel
scale (\Cref{lem:gumbel-calibration}); show that threshold counts of
$\{x_j^{(n)}\}$ converge to independent Poisson slab counts
(\Cref{lem:poissonization}); identify the limit counting process as that
of $\{-\log \Gamma_j\}$ for a unit-rate Poisson process $\Gamma_j = E_1 +
\cdots + E_j$ (\Cref{lem:gamma-counting}), giving convergence of the top
$M$ rescaled logits (\Cref{lem:top-vector}); control the aggregate weight
of all low logits (\Cref{lem:tail-mean} --- the negligibility-of-the-bulk
input); and assemble via a convergence-together lemma
(\Cref{lem:approximation}) into the Poisson--Dirichlet limit
(\Cref{thm:t2-pd}) and its coverage law (\Cref{thm:t2-coverage}).

\subsection{Calibration and Poissonization of the extreme levels}

\begin{lemma}[Gumbel calibration]\label{lem:gumbel-calibration}
With $a_n, b_n, r_n$ as in Eq.~\eqref{eq:gumbel-constants}:
\begin{enumerate}[label=(\roman*), ref=\ref*{lem:gumbel-calibration}(\roman*)]
\item\label{lem:gumbel-calibration:bn} $b_n^2 = 2\log n - 2 r_n +
o(1)$ and $b_n/a_n = 1 - r_n/a_n^2 \to 1$;
\item\label{lem:gumbel-calibration:phi} $\dfrac{n}{a_n}\,\varphi(b_n)
\to 1$;
\item\label{lem:gumbel-calibration:tail} for every fixed $x \in \R$:
$\;n\, \barPhi\big(b_n + x/a_n\big) \to e^{-x}$.
\end{enumerate}
\end{lemma}

\begin{remark}[Slow convergence]\label{rem:gumbel-slow}
The $o(1)$ corrections above decay only at rate $\Theta\big( \log\log n /
\sqrt{\log n} \big)$ (the leading $r_n/a_n = \Theta(\log\log n /
\sqrt{\log n})$ term), so numerical agreement with the Gumbel limit is
slow and one should not expect tight finite-$n$ matching at the scales
plotted in practice.
\end{remark}

\begin{proof}
The strategy is direct expansion of $b_n^2$ and the Gaussian density,
then the two-sided tail bounds of \Cref{fac:gaussian-tail}. For (i),
squaring Eq.~\eqref{eq:gumbel-constants},
\begin{align*}
b_n^2
&\;=\; \Big( a_n - \frac{r_n}{a_n} \Big)^2\\
&\;=\; a_n^2 - 2 r_n + \frac{r_n^2}{a_n^2}\\
&\;=\; 2 \log n - 2 r_n + o(1),
\end{align*}
since $r_n^2/a_n^2 = O\big( (\log\log n)^2 / \log n \big) \to 0$; the
second claim is the definition of $b_n$ divided by $a_n$. For (ii), by
(i) and $e^{r_n} = \sqrt{4\pi \log n}$,
\begin{align*}
\frac{n}{a_n}\,\varphi(b_n)
&\;=\; \frac{n}{a_n}
\cdot \frac{e^{-b_n^2/2}}{\sqrt{2\pi}}\\
&\;=\; \frac{n}{a_n}
\cdot \frac{n^{-1} e^{r_n} e^{o(1)}}{\sqrt{2\pi}}\\
&\;=\; \frac{\sqrt{4\pi \log n}}
{\sqrt{2\pi}\sqrt{2\log n}}\, e^{o(1)}\\
&\;=\; e^{o(1)} \;\to\; 1.
\end{align*}
For (iii), fix $x \in \R$ and set $t_n \coloneqq b_n + x/a_n \to \infty$.
Expanding the square,
\begin{align*}
\varphi(t_n)
&\;=\; \varphi(b_n)\,
\exp\Big(
-\frac{x b_n}{a_n}
- \frac{x^2}{2a_n^2}
\Big)\\
&\;=\; \varphi(b_n)e^{-x}e^{o(1)},
\end{align*}
using $b_n/a_n \to 1$ and $x^2/(2a_n^2) \to 0$ from (i). By
\Cref{fac:gaussian-tail}, for $n$ large enough that $t_n \ge \sqrt 2$,
\begin{gather*}
\Big( \frac{1}{t_n} - \frac{1}{t_n^3} \Big) \varphi(t_n)
\;\le\; \barPhi(t_n)\\
\;\le\; \frac{\varphi(t_n)}{t_n},\\
\text{hence}\\[-0.2em]
n \barPhi(t_n)
\;=\; \frac{n\,\varphi(t_n)}{a_n}
\cdot \frac{a_n}{t_n}\\
\hspace{1.8em}\cdot
\big( 1 + O(t_n^{-2}) \big),
\end{gather*}
and combining the two displays with (ii) and $a_n/t_n \to 1$ gives
$n\barPhi(t_n) \to e^{-x}$. This completes the proof.
\end{proof}

\begin{lemma}[Slab Poissonization]\label{lem:poissonization}
For each $n$ let $Y_1^{(n)}, \dots, Y_n^{(n)}$ be i.i.d.\ real random
variables, fix thresholds $v_1 > v_2 > \cdots > v_M$ (with $v_0 \coloneqq
+\infty$), and define slab probabilities and counts
\begin{align*}
p_i^{(n)}
&\;\coloneqq\; \Pr\big(
Y_1^{(n)} \in (v_i, v_{i-1}]
\big),\\
C_i^{(n)}
&\;\coloneqq\; \#\big\{
j \le n : Y_j^{(n)} \in (v_i,v_{i-1}]
\big\},\\
&\hspace{1.8em} i \in [M].
\end{align*}
If $n\, p_i^{(n)} \to \lambda_i \in [0, \infty)$ for every $i$, then,
with $(C_1, \dots, C_M)$ denoting independent $\Poisson(\lambda_i)$
variables,
\begin{align*}
\sup_{B \subseteq \N_0^M}\;
\Big|
\Pr\big( (C_1^{(n)},\dots,C_M^{(n)})\in B \big)\\
- \Pr\big( (C_1,\dots,C_M)\in B \big)
\Big|
\;\longrightarrow\; 0.
\end{align*}
\end{lemma}

\begin{proof}
The strategy is an exact multinomial pmf computation, a termwise limit,
and Scheff\'e's upgrade to total variation (proved inline). Fix
$k = (k_1, \dots, k_M) \in \N_0^M$ and write $s \coloneqq \sum_i k_i$.
Since each of the $n$ i.i.d.\ points lands in slab $i$ with probability
$p_i^{(n)}$, independently,
\begin{align}\label{eq:multinomial-pmf}
q_n(k)
&\;\coloneqq\; \Pr\big( C^{(n)} = k \big)\\
&\;=\; \frac{n!}{k_1!\cdots k_M!(n-s)!}
\notag\\
&\hspace{1.8em}\cdot
\prod_{i=1}^M \big( p_i^{(n)} \big)^{k_i}
\notag\\
&\hspace{1.8em}\cdot
\Big( 1 - \sum_{i=1}^M p_i^{(n)} \Big)^{n-s},
\notag
\end{align}
valid for $n \ge s$. We take the limit factor by factor:
\begin{align*}
\frac{n!}{(n-s)!\, n^s}
&\;=\; \prod_{r=0}^{s-1}
\Big(1 - \frac{r}{n}\Big)
\;\to\; 1,\\
n^s \prod_{i=1}^M \big( p_i^{(n)} \big)^{k_i}
&\;=\; \prod_{i=1}^M
\big( n p_i^{(n)} \big)^{k_i}\\
&\;\to\; \prod_{i=1}^M \lambda_i^{k_i},
\end{align*}
and, since $\sum_i p_i^{(n)} \to 0$ while $n \sum_i p_i^{(n)} \to
\sum_i \lambda_i$, the elementary expansion $\log(1-u) = -u + O(u^2)$ as
$u \to 0$ gives
\begin{gather*}
\Big( 1 - \sum_i p_i^{(n)} \Big)^{n-s}
\\
\;=\; \exp\Big(
(n-s)\log\Big(1-\sum_i p_i^{(n)}\Big)
\Big)\\
\;=\; \exp\Big(
-n\sum_i p_i^{(n)}(1+o(1))+o(1)
\Big)\\
\;\to\; e^{-\sum_i \lambda_i}.
\end{gather*}
Multiplying the three limits in Eq.~\eqref{eq:multinomial-pmf},
\begin{align}\label{eq:pmf-limit}
q_n(k)
&\;\longrightarrow\; q(k)\\
&\;\coloneqq\;
\prod_{i=1}^M e^{-\lambda_i}
\frac{\lambda_i^{k_i}}{k_i!},
\notag\\
&\text{for every fixed } k \in \N_0^M.
\notag
\end{align}
For the total-variation upgrade, both $q_n$ and $q$ are probability mass
functions on the countable set $\N_0^M$, so $\sum_k (q(k) - q_n(k)) = 0$
and therefore
\begin{gather*}
\sup_{B}
\left|
\begin{aligned}
&\Pr(C^{(n)} \in B)\\[-0.2em]
&\hspace{1em}-\Pr(C \in B)
\end{aligned}
\right|
\;\le\; \sum_{k} \big| q_n(k) - q(k) \big|\\
\;=\; 2 \sum_{k} \big( q(k) - q_n(k) \big)_+\\
\;\longrightarrow\; 0,
\end{gather*}
where the convergence holds by dominated convergence: each term
$(q(k) - q_n(k))_+ \to 0$ by Eq.~\eqref{eq:pmf-limit} and is dominated by
the summable $q(k)$. This completes the proof.
\end{proof}

\subsection{The limiting counting process}

\begin{lemma}[Poisson counting process of $\Gamma$-sums]\label{lem:gamma-counting}
Let $E_1, E_2, \dots$ be i.i.d.\ $\Exp(1)$ and $\Gamma_j \coloneqq E_1 +
\cdots + E_j$. Fix $0 < u_1 < u_2 < \cdots < u_M$ (with $u_0 \coloneqq 0$)
and set $D_i \coloneqq \#\{ j \ge 1 : \Gamma_j \le u_i \}$ (a.s.\ finite).
Then the increments $D_1 - D_0, D_2 - D_1, \dots, D_M - D_{M-1}$ (with
$D_0 \coloneqq 0$) are independent with $D_i - D_{i-1} \sim
\Poisson(u_i - u_{i-1})$.
\end{lemma}

\begin{proof}
The strategy is to integrate the explicit joint density of the partial
sums over the event region. Each $D_i$ is a.s.\ finite since $\Gamma_j \to
\infty$ a.s. Fix integers $m_1, \dots, m_M \ge 0$ and let $s_i \coloneqq
m_1 + \cdots + m_i$, $s \coloneqq s_M$. Since $j \mapsto \Gamma_j$ is
strictly increasing,
\begin{gather}\label{eq:gamma-event}
\big\{ D_i - D_{i-1} = m_i
\ \forall i \in [M] \big\}
\;=\; \big\{ D_i = s_i \ \forall i \big\}\\
\;=\; \big\{
\Gamma_{s_i}\le u_i<\Gamma_{s_i+1}
\ \forall i \big\},
\notag\\
\hspace{1.8em}(\Gamma_0 \coloneqq 0).
\notag
\end{gather}
The map $(E_1, \dots, E_{s+1}) \mapsto (\Gamma_1, \dots, \Gamma_{s+1})$ is
linear with unit Jacobian, so $(\Gamma_1, \dots, \Gamma_{s+1})$ has joint
density
\begin{gather}\label{eq:gamma-density}
\prod_{l=1}^{s+1} e^{-(\gamma_l - \gamma_{l-1})}
\;=\; e^{-\gamma_{s+1}},\\
\hspace{1.8em}
\text{on }\{0<\gamma_1<\cdots<\gamma_{s+1}\},
\quad \gamma_0\coloneqq0.
\notag
\end{gather}
On the event Eq.~\eqref{eq:gamma-event}, the coordinates $\gamma_l$ with
$l \in (s_{i-1}, s_i]$ lie, in increasing order, in the slab
$(u_{i-1}, u_i]$, and $\gamma_{s+1} > u_M$; conversely any such
configuration realises the event (boundary ties are Lebesgue-null). The
density Eq.~\eqref{eq:gamma-density} depends only on $\gamma_{s+1}$, so
the first $s$ coordinates integrate to a product of ordered-block
volumes:
\begin{gather*}
\Pr\big( D_i = s_i \ \forall i \big)
\\
\;=\; \prod_{i=1}^M
\mathrm{Vol}\Big(
\big\{
\substack{
u_{i-1}<\gamma^{(1)}<\cdots\\
<\gamma^{(m_i)}\le u_i
}
\big\}\Big)\\
\cdot \int_{u_M}^{\infty} e^{-\gamma}\, d\gamma \\
\;=\; \prod_{i=1}^M
\frac{(u_i - u_{i-1})^{m_i}}{m_i!}\\
\cdot e^{-u_M},
\end{gather*}
where each ordered block has volume $(u_i - u_{i-1})^{m_i}/m_i!$ (an
ordered simplex is a $1/m_i!$ fraction of the cube). Since
$u_M = \sum_{i=1}^M (u_i - u_{i-1})$,
\begin{align*}
\Pr\big( D_i - D_{i-1} = m_i \ \forall i \big)
&\;=\; \prod_{i=1}^M
e^{-(u_i - u_{i-1})}\\
&\hspace{1.8em}\cdot
\frac{(u_i - u_{i-1})^{m_i}}{m_i!},
\end{align*}
which is exactly the claimed product of Poisson laws. This completes the
proof.
\end{proof}

\begin{lemma}[Law of large numbers for $\Gamma_j$]\label{lem:gamma-lln}
With $\Gamma_j$ as in \Cref{lem:gamma-counting}: for every $\delta \in
(0,1)$ and $j \ge 1$,
\begin{align*}
\Pr\Big( \Big| \frac{\Gamma_j}{j} - 1 \Big| > \delta \Big)
&\;\le\; 2\, e^{-j \delta^2 / 6},\\
&\text{and consequently}\\[-0.2em]
\frac{\Gamma_j}{j}
&\;\toas\; 1 \quad (j \to \infty).
\end{align*}
\end{lemma}

\begin{proof}
The strategy is the Cram\'er--Chernoff method with the exponential
moment $\E e^{\lambda E_1} = (1-\lambda)^{-1}$ ($\lambda < 1$), then a
summable-tail Borel--Cantelli argument, both inline. For the upper tail,
with $\lambda \coloneqq \delta/(1+\delta) \in (0,1)$, by Markov's
inequality and independence,
\begin{align*}
\Pr\big( \Gamma_j \ge (1+\delta) j \big)
&\;\le\; e^{-\lambda(1+\delta)j}
\big( \E e^{\lambda E_1} \big)^j\\
&\;=\; e^{-\delta j}\, (1+\delta)^{j}\\
&\;=\; e^{-j(\delta - \log(1+\delta))}\\
&\;\le\; e^{-j\delta^2/6},
\end{align*}
where the middle equality substitutes $1 - \lambda = 1/(1+\delta)$, and
the last step uses $\log(1+\delta) \le \delta - \delta^2/6$ for $\delta
\in (0,1]$, which follows from $\log(1+\delta) \le \delta - \delta^2/2 +
\delta^3/3$ and $\delta^2/2 - \delta^3/3 \ge \delta^2/6 \Leftrightarrow
\delta^2(1 - \delta) \ge 0$. For the lower tail, with $\lambda \coloneqq
\delta/(1-\delta) > 0$ and $\E e^{-\lambda E_1} = (1+\lambda)^{-1}$,
\begin{gather*}
\Pr\big( \Gamma_j \le (1-\delta) j \big)
\;\le\; e^{\lambda(1-\delta)j}
(1 + \lambda)^{-j}\\
\;=\; e^{\delta j}\, (1-\delta)^{j}\\
\;=\; e^{-j(-\delta - \log(1 - \delta))}
\;\le\; e^{-j \delta^2/2},
\end{gather*}
using $1 + \lambda = 1/(1-\delta)$ and $-\delta - \log(1-\delta) =
\sum_{r \ge 2} \delta^r / r \ge \delta^2/2$. Adding the two tails gives
the claimed bound. For the a.s.\ convergence, for every $\delta \in
(0,1)$ and $J \ge 1$,
\begin{align*}
\Pr\Big( \exists\, j \ge J : \Big|\frac{\Gamma_j}{j} - 1\Big| > \delta
\Big)
&\;\le\; \sum_{j \ge J} 2 e^{-j\delta^2/6}\\
&\;=\; \frac{2 e^{-J \delta^2/6}}
{1 - e^{-\delta^2/6}}\\
&\;\longrightarrow\; 0 \quad (J \to \infty),
\end{align*}
so a.s.\ only finitely many $j$ violate $|\Gamma_j/j - 1| \le \delta$;
intersecting over $\delta \in \{1/2, 1/3, 1/4, \dots\}$ (a countable
family of probability-one events) yields $\Gamma_j / j \to 1$ a.s. This
completes the proof.
\end{proof}

\begin{lemma}[Convergence of the top vector]\label{lem:top-vector}
For every fixed $M \in \N$,
\begin{gather*}
\big( x_1^{(n)}, \dots, x_M^{(n)} \big)
\\[-0.2em]
\;\tod\;\\[-0.2em]
\big( -\log \Gamma_1,\dots,-\log \Gamma_M \big),
\end{gather*}
with $x_j^{(n)}$ as in Eq.~\eqref{eq:xjn} and $\Gamma_j$ as in
\Cref{lem:gamma-counting}.
\end{lemma}

\begin{proof}
The strategy is the cdf criterion \Cref{fac:weak-convergence:cdf}: we
express both cdfs through threshold counts, transfer the count law by
\Cref{lem:poissonization,lem:gamma-counting}, and check continuity of the
limit cdf. Fix $v = (v_1, \dots, v_M) \in \R^M$ and let $\tilde v_j
\coloneqq \min(v_1, \dots, v_j)$, a nonincreasing vector. Since both
$(x_j^{(n)})_j$ and $(-\log \Gamma_j)_j$ are nonincreasing in $j$ (the
$\Gamma_j$ increase), the cdf events reduce monotonically: a nonincreasing
sequence $(y_j)$ satisfies $\{ y_j \le v_j \ \forall j \le M \} = \{ y_j
\le \tilde v_j \ \forall j \le M\}$, because $y_j \le \min_{i \le j} y_i
\le \min_{i\le j} v_i$. For any $v^\ast \in \R$ and the threshold counts
\begin{align*}
N^{(n)}(v^\ast)
&\;\coloneqq\; \#\big\{
l\le n:x_l^{(n)}>v^\ast
\big\},\\
N^{\infty}(v^\ast)
&\;\coloneqq\; \#\big\{
l\ge1:-\log\Gamma_l>v^\ast
\big\},
\end{align*}
the rank--count duality (as in Eq.~\eqref{eq:rank-count-duality}) gives
\begin{align}
F_n(v) &\coloneqq \Pr\big( x_j^{(n)} \le v_j \ \forall j \le M \big)
\label{eq:cdf-as-counts}
\\
&= \Pr\Big(
\bigcap_{j \le M}
\{N^{(n)}(\tilde v_j)\le j-1\}
\Big),
\notag\\
F_\infty(v)
&\coloneqq \Pr\big(
{-\log \Gamma_j}\le v_j
\ \forall j\le M
\big)
\notag\\
&= \Pr\Big(
\bigcap_{j \le M}
\{N^{\infty}(\tilde v_j)\le j-1\}
\Big).
\nonumber
\end{align}
Let $w_1 > w_2 > \cdots > w_{M'}$ be the distinct values among $\{\tilde
v_1, \dots, \tilde v_M\}$. The intersection events in
Eq.~\eqref{eq:cdf-as-counts} are measurable functions of the slab-count
vectors over the thresholds $(w_r)_{r \le M'}$, namely of
\begin{align*}
\big(
N^{(n)}(w_1),\
N^{(n)}(w_2)-N^{(n)}(w_1),\dots
\big),\\
\text{resp.}\\[-0.2em]
\big(
N^{\infty}(w_1),\
N^{\infty}(w_2)-N^{\infty}(w_1),\dots
\big).
\end{align*}
For the prelimit: the unsorted variables $Y_j^{(n)} \coloneqq a_n(g_j -
b_n)$ are i.i.d., and by \Cref{lem:gumbel-calibration:tail} the slab
means converge:
\begin{gather*}
n \Pr\big( Y_1^{(n)} \in (w_r, w_{r-1}] \big)
\;=\; n \barPhi\Big(
b_n + \frac{w_r}{a_n}
\Big)\\
\hspace{1.8em}
- n \barPhi\Big(
b_n + \frac{w_{r-1}}{a_n}
\Big)\\
\;\longrightarrow\;
e^{-w_r} - e^{-w_{r-1}}
\;\eqqcolon\; \lambda_r,
\end{gather*}
with the convention $w_0 = +\infty$, $e^{-w_0} = 0$. Hence
\Cref{lem:poissonization} applies and the prelimit slab counts converge
in total variation to independent $\Poisson(\lambda_r)$. For the limit
object: $N^\infty(w_r) = \#\{l : \Gamma_l < e^{-w_r}\}$, and the
thresholds $u_r \coloneqq e^{-w_r}$ are increasing in $r$, so by
\Cref{lem:gamma-counting} (strict vs.\ weak inequality being a.s.\
immaterial as each $\Gamma_l$ has a continuous law) the limit slab counts
are independent Poisson with means $u_r - u_{r-1} = \lambda_r$ --- the
same law. Therefore, by the total-variation convergence applied to the
count event of Eq.~\eqref{eq:cdf-as-counts},
\begin{align*}
F_n(v) \;\longrightarrow\; F_\infty(v) \qquad \text{for every } v \in
\R^M.
\end{align*}
Finally, $F_\infty$ is continuous: for $v, v' \in \R^M$, a union bound
gives
\begin{gather*}
\big| F_\infty(v) - F_\infty(v') \big|
\;\le\; \sum_{j=1}^M \Pr\Big(
\Gamma_j \text{ lies between}\\
\hspace{4em}
e^{-\tilde v_j}
\text{ and }e^{-\tilde v_j'}
\Big)\\
\;\longrightarrow\; \sum_{j=1}^M
\Pr\big(\Gamma_j=e^{-\tilde v_j}\big)
\;=\; 0
\qquad (v' \to v),
\end{gather*}
since each $\Gamma_j$ has a density and $\tilde v_j' \to \tilde v_j$. By
\Cref{fac:weak-convergence:cdf}, cdf convergence at every (continuity)
point gives the claimed convergence in distribution. This completes the
proof.
\end{proof}

\subsection{Negligibility of the bulk and convergence together}

\begin{lemma}[Tail-mean bound; negligibility of the bulk]\label{lem:tail-mean}
For $x_0 \in \R$ let $T_n(x_0) \coloneqq \sum_{j=1}^n e^{c\, x_j^{(n)}}
\1\{ x_j^{(n)} \le x_0 \}$. Then, for $c > 1$,
\begin{align*}
\limsup_{n \to \infty}\; \E\, T_n(x_0)
\;\le\; \frac{2\, e^{(c-1) x_0}}{c - 1},
\end{align*}
which tends to $0$ as $x_0 \to -\infty$.
\end{lemma}

\begin{proof}
The strategy is to split off the negligible region $g \le 0$ and estimate
the remaining integral after the Gumbel change of variables, using
\Cref{lem:gumbel-calibration}. Since a sum of a fixed function over the
sorted values equals the same sum over the unsorted values (the same
multiset), and the $g_j$ are i.i.d.,
\begin{align*}
\E T_n(x_0)
&\;=\; n\, \E\Big[
e^{c a_n (g - b_n)}\\
&\hspace{3.5em}\cdot
\1\big\{a_n(g-b_n)\le x_0\big\}
\Big],\\
&\hspace{1.8em} g \sim N(0,1).
\end{align*}
\textbf{Region $g \le 0$.} Here $a_n(g - b_n) \le -a_n b_n = -(2\log n -
r_n) \le -\tfrac32 \log n$ for $n$ large, so
\begin{align*}
n\, \E\Big[ e^{c a_n(g-b_n)};\, g \le 0 \Big]
&\;\le\; n\, e^{-\tfrac32 c \log n}\\
&\;=\; n^{1 - 3c/2}\\
&\;\le\; n^{-1/2} \;\longrightarrow\; 0,
\end{align*}
using $c > 1$. \textbf{Region $g > 0$.} Substituting $x = a_n(g - b_n)$
(so $g = b_n + x/a_n$, $dg = dx/a_n$), the contribution is
{\small
\begin{multline*}
n \int_{-a_n b_n}^{x_0} e^{c x}\,
\varphi\Big( b_n + \frac{x}{a_n} \Big)
\frac{dx}{a_n}
\\
\;=\; \frac{n}{a_n} \varphi(b_n)
\cdot
\int_{-a_n b_n}^{x_0} e^{c x}\,
e^{-x b_n / a_n}\, e^{-x^2/(2 a_n^2)}\, dx,
\end{multline*}
}
by the same expansion of $\varphi(b_n + x/a_n)$ as in
\Cref{lem:gumbel-calibration}. Write $b_n/a_n = 1 - \rho_n$ with $\rho_n
\coloneqq r_n / a_n^2 \in (0, 1/8]$ for $n$ large; then, bounding
$e^{-x^2/(2a_n^2)} \le 1$,
\begin{align*}
e^{cx}\, e^{-x b_n/a_n}
&\;=\; e^{(c - 1 + \rho_n) x}
\;=\; e^{(c-1)x}\, e^{\rho_n x}\\
&\;\le\; e^{(c-1)x}
\cdot \max\big( 1,e^{\rho_n x_0} \big)\\
&\hspace{1.8em}
\;\le\; \tfrac{3}{2}\, e^{(c-1)x},
\end{align*}
for all $x \le x_0$ and all $n \ge n_7(x_0)$: indeed $e^{\rho_n x} \le 1$
for $x \le 0$, while for $x \in [0, x_0]$ (a case arising only if $x_0 >
0$) we have $e^{\rho_n x} \le e^{\rho_n x_0} \to 1$. Hence
\begin{multline*}
n\, \E\Big[
e^{c a_n(g - b_n)};\
g > 0,\ a_n(g-b_n) \le x_0
\Big]
\\
\;\le\; \frac{n \varphi(b_n)}{a_n}
\cdot \frac{3}{2}
\cdot
\int_{-\infty}^{x_0} e^{(c-1)x}\, dx
\\
\;=\; \frac{n \varphi(b_n)}{a_n}
\cdot
\frac{3 e^{(c-1)x_0}}{2(c-1)},
\end{multline*}
where the integral converges because $c > 1$. Since $(n/a_n)\varphi(b_n)
\to 1$ by \Cref{lem:gumbel-calibration:phi}, combining the two regions
gives $\limsup_n \E T_n(x_0) \le \tfrac32 e^{(c-1)x_0}/(c-1) + 0 \le 2
e^{(c-1)x_0}/(c-1)$. This completes the proof.
\end{proof}

\begin{lemma}[Convergence together on a compact state space]\label{lem:approximation}
Let $K \in \N$ and let $X_n$, $X_n^{(M)}$ ($n, M \in \N$), $Y^{(M)}$, $Y$
be random vectors with values in $[0,1]^K$ such that:
\begin{enumerate}[label=(\roman*), ref=\ref*{lem:approximation}(\roman*)]
\item\label{lem:approximation:perM} for each fixed $M$:\,
$X_n^{(M)} \tod Y^{(M)}$ as $n \to \infty$;
\item\label{lem:approximation:asM} $Y^{(M)} \to Y$ a.s.\ as $M \to
\infty$;
\item\label{lem:approximation:tight} for every $\varepsilon > 0$,
\[
\lim_{M \to \infty}\limsup_{n \to \infty}
\Pr\bigl(\lVert X_n-X_n^{(M)}\rVert_\infty>\varepsilon\bigr)=0.
\]
\end{enumerate}
Then $X_n \tod Y$.
\end{lemma}

\begin{proof}
The strategy is the three-term split with a modulus of continuity, which
is available because every continuous $f$ is uniformly continuous on the
compact $[0,1]^K$. Fix a bounded continuous $f \colon \R^K \to \R$ and
let $\omega_f(\varepsilon) \coloneqq \sup\{ |f(x) - f(y)| : x, y \in
[0,1]^K, \norm{x - y}_\infty \le \varepsilon \}$, so $\omega_f(\varepsilon)
\to 0$ as $\varepsilon \to 0$. For every $n, M, \varepsilon$,
\begin{gather*}
\big| \E f(X_n) - \E f(Y) \big|
\;\le\;
\underbrace{\big|
\E f(X_n)-\E f(X_n^{(M)})
\big|}_{(\mathrm{I})}\\
\hspace{1.8em}
+ \underbrace{\big|
\E f(X_n^{(M)})-\E f(Y^{(M)})
\big|}_{(\mathrm{II})}\\
\hspace{1.8em}
+ \underbrace{\big|
\E f(Y^{(M)})-\E f(Y)
\big|}_{(\mathrm{III})},
\end{gather*}
and the three pieces obey
\begin{gather*}
(\mathrm{I})
\;\le\; \omega_f(\varepsilon)
+ 2 \norm{f}_\infty\\
\hspace{1.8em}\cdot
\Pr\big( \norm{X_n - X_n^{(M)}}_\infty > \varepsilon \big), \\
\limsup_{n} (\mathrm{II}) \;=\; 0,\\
(\mathrm{III}) \;\longrightarrow\; 0 \ \ (M \to \infty),
\end{gather*}
where the bound on $(\mathrm{I})$ splits the expectation over
$\{\norm{X_n - X_n^{(M)}}_\infty \le \varepsilon\}$ and its complement,
$(\mathrm{II})$ uses hypothesis (i) and the definition of $\tod$ from
\Cref{sec:app-conventions}, and $(\mathrm{III})$ uses hypothesis (ii) with
dominated convergence ($|f(Y^{(M)})| \le \norm{f}_\infty$). Taking
$\limsup_{n \to \infty}$, then $M \to \infty$ using hypothesis (iii),
then $\varepsilon \downarrow 0$,
\begin{gather*}
\limsup_{n \to \infty} \big| \E f(X_n) - \E f(Y) \big|
\;\le\; \omega_f(\varepsilon)
+ 2\norm{f}_\infty\\
\hspace{1.8em}\cdot
\lim_{M\to\infty}\limsup_{n\to\infty}
\Pr\big(
\norm{X_n-X_n^{(M)}}_\infty>\varepsilon
\big)\\
\hspace{1.8em}+ 0\\
\;=\; \omega_f(\varepsilon)
\;\longrightarrow\; 0.
\end{gather*}
Since $f$ was an arbitrary bounded continuous function, $X_n \tod Y$.
This completes the proof.
\end{proof}

\subsection{The Poisson--Dirichlet limit}

\begin{theorem}[Condensation onto a Poisson--Dirichlet weight
profile; formal version of \cref{thm:main-scaled}(ii)]\label{thm:t2-pd}
Let $c > 1$ be fixed, $\alpha = 1/c \in (0,1)$, and let $\Gamma_j = E_1 +
\cdots + E_j$ be as in \Cref{lem:gamma-counting}. Then:
\begin{enumerate}[label=(\roman*), ref=\ref*{thm:t2-pd}(\roman*)]
\item\label{thm:t2-pd:freeenergy} under $\Ens(c)$,
$\dfrac{\log Z}{\log n} \toP 2c$;
\item\label{thm:t2-pd:weights} $W \coloneqq \sum_{j \ge 1}
\Gamma_j^{-1/\alpha} \in (0, \infty)$ a.s., the random sequence
\begin{align*}
P_j \;\coloneqq\; \frac{\Gamma_j^{-1/\alpha}}{W}, \qquad j \ge 1,
\end{align*}
satisfies $P_1 > P_2 > \cdots > 0$ and $\sum_j P_j = 1$ a.s., and for
every fixed $K \in \N$, under $\Ens(c)$,
\begin{align*}
\big( A_{(1)}, \dots, A_{(K)} \big) \;\tod\; \big( P_1, \dots, P_K \big).
\end{align*}
\end{enumerate}
\end{theorem}

\begin{proof}
The strategy: part (i) is the boundary case of the restricted free
energy; for part (ii) we verify the three hypotheses of
\Cref{lem:approximation} with top-$M$ normalised approximants, using
\Cref{lem:top-vector} for (i), monotone convergence for (ii), and
\Cref{lem:tail-mean} plus \Cref{lem:gumbel-calibration} for (iii).

\textbf{Step 1 (proof of (i): free energy).} \Cref{prop:restricted-z:full} with $w = 1$ gives
$\log Z/\log n \toP F_1(c)$, and $F_1(c) = 2c$ for $c \ge 1$ by
Eq.~\eqref{eq:Fw}.

\textbf{Step 2 (proof of (ii): the limit object).} By \Cref{lem:gamma-lln},
$\Gamma_j / j \to 1$ a.s., so a.s.\ $\Gamma_j^{-1/\alpha} \le 2^{1/\alpha}
j^{-1/\alpha}$ for all large $j$; since $1/\alpha = c > 1$, the series
$W = \sum_j \Gamma_j^{-1/\alpha}$ converges a.s., and $W \ge
\Gamma_1^{-1/\alpha} > 0$ a.s. Strict monotonicity $P_1 > P_2 > \cdots >
0$ holds because $\Gamma_j$ is a.s.\ strictly increasing, and $\sum_j P_j
= W/W = 1$.

\textbf{Step 3 (proof of (ii): approximation scheme).} Fix $K$ and let $M \ge K$.
For $n \ge M$ define, recalling Eq.~\eqref{eq:weight-cancellation},
\begin{align*}
X_n
&\;\coloneqq\; \big( A_{(1)},\dots,A_{(K)} \big)\\
&\;=\; \Big(
\frac{e^{c x_j^{(n)}}}{D_n}
\Big)_{j \le K},\\
X_n^{(M)}
&\;\coloneqq\;
\Big(
\frac{e^{c x_j^{(n)}}}{D_n^{(M)}}
\Big)_{j \le K},\\
D_n^{(M)}
&\;\coloneqq\; \sum_{i=1}^M e^{c x_i^{(n)}},
\end{align*}
together with the limit-side vectors
\begin{align*}
Y^{(M)}
&\;\coloneqq\;
\left(
\frac{\Gamma_j^{-1/\alpha}}
{\sum_{i\le M}\Gamma_i^{-1/\alpha}}
\right)_{j \le K},\\
Y &\;\coloneqq\; \big( P_1, \dots, P_K \big).
\end{align*}
All four take values in $[0,1]^K$: each numerator is one of the terms of
the corresponding denominator (using $j \le K \le M$). We check the three
hypotheses of \Cref{lem:approximation}.

\textbf{Step 3a (hypothesis \ref*{lem:approximation}(i): continuity).} The map $\psi \colon \R^M
\to \R^K$, $\psi(y) \coloneqq \big( e^{c y_j} / \sum_{i \le M} e^{c y_i}
\big)_{j \le K}$, is continuous on all of $\R^M$ (its denominator is a
finite sum of positive continuous terms), $X_n^{(M)} = \psi(x_1^{(n)},
\dots, x_M^{(n)})$, and $Y^{(M)} = \psi(-\log \Gamma_1, \dots, -\log
\Gamma_M)$ because $e^{c \cdot (-\log \Gamma_j)} = \Gamma_j^{-c} =
\Gamma_j^{-1/\alpha}$. Hence \Cref{lem:top-vector} and the continuous
mapping theorem \Cref{fac:weak-convergence:cmt} give $X_n^{(M)} \tod
Y^{(M)}$.

\textbf{Step 3b (hypothesis \ref*{lem:approximation}(ii): monotone limit).} As $M \to \infty$ the
denominator $\sum_{i \le M} \Gamma_i^{-1/\alpha} \uparrow W \in (0,
\infty)$ a.s.\ (monotone convergence of a positive series), so $Y^{(M)}
\to Y$ a.s.

\textbf{Step 3c (hypothesis \ref*{lem:approximation}(iii): bulk negligibility).} Fix $\varepsilon > 0$.
For $j \le K$, since $0 < e^{c x_j^{(n)}} \le D_n^{(M)} \le D_n$,
\begin{align}\label{eq:approx-coordinate}
0
&\;\le\; \big( X_n^{(M)} \big)_j
- \big( X_n \big)_j\\
&\;=\; e^{c x_j^{(n)}}
\frac{D_n-D_n^{(M)}}{D_n^{(M)}D_n}
\notag\\
&\;\le\; \frac{D_n-D_n^{(M)}}{D_n^{(M)}},
\notag
\\
&\text{so}
\notag\\[-0.2em]
\norm{X_n - X_n^{(M)}}_\infty
&\;\le\;
\frac{D_n-D_n^{(M)}}{D_n^{(M)}}.
\notag
\end{align}
Fix $x_0 < 0$ and $R > 0$, and define the good event
\begin{gather*}
G_n
\;\coloneqq\; \big\{N^{(n)}(x_0)\le M\big\}
\cap \big\{x_1^{(n)}\ge-R\big\},\\
N^{(n)}(x_0)
\;=\; \#\big\{j:x_j^{(n)}>x_0\big\},\\
\text{as in \Cref{lem:top-vector}}.
\end{gather*}
On $G_n$: every index $l > M \ge N^{(n)}(x_0)$ has $x_l^{(n)} \le x_0$
(the $x^{(n)}$ are sorted), so $D_n - D_n^{(M)} = \sum_{l > M} e^{c
x_l^{(n)}} \le T_n(x_0)$, and $D_n^{(M)} \ge e^{c x_1^{(n)}} \ge
e^{-cR}$. Hence, by Eq.~\eqref{eq:approx-coordinate} and Markov's
inequality,
\begin{gather}\label{eq:approx-three-terms}
\Pr\big( \norm{X_n - X_n^{(M)}}_\infty > \varepsilon \big)
\;\le\;
\frac{e^{cR}\E T_n(x_0)}{\varepsilon}\\
\hspace{1.8em}
+ \Pr\big( N^{(n)}(x_0)>M \big)
\notag\\
\hspace{1.8em}
+ \Pr\big( x_1^{(n)}<-R \big).
\notag
\end{gather}
We bound the three limsups. First, by \Cref{lem:tail-mean},
\begin{align*}
\limsup_{n} \frac{e^{cR} \E T_n(x_0)}{\varepsilon}
&\;\le\;
\frac{2 e^{cR} e^{(c-1)x_0}}
{(c-1)\varepsilon}.
\end{align*}
Second, $N^{(n)}(x_0) \sim \Bin\big(n, \barPhi(b_n + x_0/a_n)\big)$, so by
Markov's inequality and \Cref{lem:gumbel-calibration:tail},
\begin{gather*}
\limsup_n \Pr\big( N^{(n)}(x_0) > M \big)
\\
\;\le\; \limsup_n
\frac{n \barPhi(b_n + x_0/a_n)}{M + 1}\\
\;=\; \frac{e^{-x_0}}{M+1}.
\end{gather*}
Third, with $q_n \coloneqq \barPhi(b_n - R/a_n)$ we have $n q_n \to
e^{R}$ and $q_n \to 0$ by \Cref{lem:gumbel-calibration:tail}, so
\begin{gather*}
\Pr\big( x_1^{(n)} < -R \big)
\;=\; (1 - q_n)^n
\\
\;=\; \exp\big( n \log(1 - q_n) \big)\\
\;=\; \exp\big(
-n q_n (1 + O(q_n))
\big)
\;\longrightarrow\; e^{-e^{R}}.
\end{gather*}
Now fix $\zeta > 0$ and choose, in order: $R = R(\zeta)$ with $e^{-e^R}
\le \zeta/3$; then $x_0 = x_0(\zeta, \varepsilon, R, c) < 0$ with
$2 e^{cR} e^{(c-1)x_0} / ((c-1)\varepsilon) \le \zeta/3$ (possible since
$c > 1$); then $M_0 = M_0(\zeta, x_0)$ with $e^{-x_0}/(M_0+1) \le
\zeta/3$. Substituting into Eq.~\eqref{eq:approx-three-terms}, for every
$M \ge M_0$,
\begin{align*}
\limsup_{n \to \infty}
\Pr\big(
\norm{X_n-X_n^{(M)}}_\infty>\varepsilon
\big)
\;\le\; \zeta,
\end{align*}
and since $\zeta > 0$ was arbitrary, $\lim_{M} \limsup_n \Pr(\norm{X_n -
X_n^{(M)}}_\infty > \varepsilon) = 0$, which is hypothesis
\ref*{lem:approximation}(iii).

\textbf{Step 4 (conclusion).} \Cref{lem:approximation} yields $X_n =
(A_{(1)}, \dots, A_{(K)}) \tod Y = (P_1, \dots, P_K)$. This completes
the proof.
\end{proof}

\begin{remark}[Naming: the two-parameter Poisson--Dirichlet law
$\PD(\alpha, 0)$]\label{rem:pd-naming}
The limit law in \Cref{thm:t2-pd:weights} is a standard object. By
\Cref{lem:gamma-counting}, for every $t > 0$,
\begin{align*}
\#\big\{ j : \Gamma_j^{-1/\alpha} > t \big\}
&\;=\; \#\big\{j:\Gamma_j<t^{-\alpha}\big\}\\
&\;\sim\; \Poisson\big( t^{-\alpha} \big),
\end{align*}
with independent counts over disjoint slabs, i.e.\ the points
$\{\Gamma_j^{-1/\alpha}\}_{j \ge 1}$ are the ranked atoms of a Poisson
point process on $(0,\infty)$ with intensity $\alpha t^{-\alpha - 1} dt$
--- the jump intensity of an $\alpha$-stable subordinator up to a
constant time-change. The constant is immaterial after normalisation:
replacing the intensity by $C \alpha t^{-\alpha-1} dt$ ($C>0$) realises
the ranked atoms as $\{ (\Gamma_j / C)^{-1/\alpha} \}_j = C^{1/\alpha}
\{\Gamma_j^{-1/\alpha}\}_j$ (again by \Cref{lem:gamma-counting}, since
$\#\{j : (\Gamma_j/C)^{-1/\alpha} > t\} = \#\{j : \Gamma_j < C t^{-\alpha}\}
\sim \Poisson(C t^{-\alpha})$), and the common factor $C^{1/\alpha}$
cancels from $P_j = \Gamma_j^{-1/\alpha}/W$. Hence $(P_j)_{j \ge 1}$ is
the sequence of ranked, sum-normalised jumps of an $\alpha$-stable
subordinator, whose law is the two-parameter Poisson--Dirichlet
distribution $\PD(\alpha, 0)$ of \cite{pitman1997two} --- we use the
result above self-contained and the citation only to name the limit. In
the language of disordered systems, \Cref{thm:t2-pd} is the analogue for
softmax rows of the Gibbs-weight condensation of the Random Energy Model
\cite{derrida1981random}: with $n$ states and inverse-temperature ratio
$c$, the REM's low-temperature Gibbs weights converge to
$\PD(\beta_c/\beta, 0)$ \cite{bovier2006statistical}, matching
$\alpha = 1/c$ here; the free-energy kink of
\Cref{prop:restricted-z:full} at $c = 1$ ($1 + c^2$ vs.\ $2c$) is the
freezing transition.
\end{remark}

\subsection{The coverage law}

\begin{lemma}[Atomlessness of partial sums of the limit
weights]\label{lem:atomless}
In the setting of \Cref{thm:t2-pd:weights}, for every fixed $m \ge 1$
and every $s \in (0,1)$,
\begin{align*}
\Pr\Big( \sum_{j \le m} P_j = s \Big) \;=\; 0.
\end{align*}
\end{lemma}

\begin{proof}
The strategy is to freeze $(E_2, E_3, \dots)$ and show the partial sum is
a strictly decreasing continuous function of $E_1$ alone, whence its
conditional law is atomless. Write $c_1 \coloneqq 0$ and $c_j \coloneqq
E_2 + \cdots + E_j$ for $j \ge 2$, so that $\Gamma_j = E_1 + c_j$, and
work on the probability-one event
\begin{align*}
\Omega_0
&\;\coloneqq\;
\Big\{0=c_1<c_2<c_3<\cdots\Big\}\\
&\hspace{1.8em}\cap
\Big\{\frac{c_j}{j}\to1\Big\},
\end{align*}
where the second part holds by \Cref{lem:gamma-lln} applied to the
shifted i.i.d.\ sequence $(E_2, E_3, \dots)$. For $e > 0$ define
\begin{align*}
N(e)
&\;\coloneqq\; \sum_{j \le m}
(e + c_j)^{-1/\alpha},\\
R(e)
&\;\coloneqq\; \sum_{j > m}
(e + c_j)^{-1/\alpha},\\
\phi(e) \;\coloneqq\; \frac{N(e)}{N(e) + R(e)},
\end{align*}
so that $\sum_{j \le m} P_j = \phi(E_1)$ identically. On $\Omega_0$ and
on any compact $[e_0, e_1] \subset (0, \infty)$: the series $R$ and its
termwise derivative series $\sum_{j>m} (-1/\alpha)(e + c_j)^{-1/\alpha -
1}$ converge uniformly, since their terms are dominated by
$(e_0 + c_j)^{-1/\alpha}$ resp.\ $(1/\alpha)(e_0 + c_j)^{-1/\alpha-1}$,
both summable because $c_j / j \to 1$ and $1/\alpha > 1$; by the standard
theorem on termwise differentiation of uniformly convergent series of
$C^1$ functions, $R \in C^1(0,\infty)$ with the termwise derivative, and
$N \in C^1$ as a finite sum. Differentiating the ratio,
\begin{gather*}
\operatorname{sign} \phi'(e)
\;=\; \operatorname{sign}\Big(
N'(e)\big(N(e)+R(e)\big)\\
\hspace{5.5em}
-N(e)\big(N'(e)+R'(e)\big)
\Big)\\
\;=\; \operatorname{sign}\Big(
N'(e)R(e)-N(e)R'(e)
\Big),
\end{gather*}
and, expanding both products into an absolutely convergent double series
and pairing the $(j, l)$ terms,
\begin{gather*}
N'(e) R(e) - N(e) R'(e)
\\
\;=\; -\frac{1}{\alpha}
\sum_{j \le m} \sum_{l > m}
\Big[
\begin{aligned}
&(e+c_j)^{-\frac1\alpha - 1}
(e+c_l)^{-\frac1\alpha}\\
&- (e+c_j)^{-\frac1\alpha}
(e+c_l)^{-\frac1\alpha - 1}
\end{aligned}
\Big]\\
\;=\; -\frac{1}{\alpha}
\sum_{j \le m}\sum_{l > m}\\
\hspace{1.8em}\cdot
(e+c_j)^{-\frac1\alpha - 1}
(e+c_l)^{-\frac1\alpha - 1}\\
\hspace{1.8em}\cdot(c_l-c_j)
\;<\; 0,
\end{gather*}
strictly, because every pair $j \le m < l$ has $c_l > c_j$ on
$\Omega_0$. Hence $\phi$ is continuous and strictly decreasing on
$(0,\infty)$, so for each fixed $s$ the set $\{ e > 0 : \phi(e) = s \}$
is empty or a single point $e^\ast = e^\ast(E_2, E_3, \dots)$.
Conditioning on $\mathcal{G} \coloneqq \sigma(E_2, E_3, \dots)$ and using
the independence of $E_1$ from $\mathcal{G}$ together with the
absolute continuity of $\Exp(1)$,
\begin{align*}
\Pr\Big( \sum_{j \le m} P_j = s \Big)
&\;=\; \E\Big[
\Pr\big( \phi(E_1)=s \mid \mathcal{G} \big)
\Big]\\
&\;\le\; \E\Big[
\Pr\big(E_1=e^\ast\mid\mathcal{G}\big);\\
&\hspace{5.5em}
e^\ast\text{ exists}
\Big]\\
&\;=\; 0.
\end{align*}
This completes the proof.
\end{proof}

\begin{theorem}[Coverage law; formal version of \cref{thm:main-scaled}(ii)]\label{thm:t2-coverage}
Let $c > 1$, $\alpha = 1/c$, and let $(P_j)_{j\ge1}$, $W$ be as in
\Cref{thm:t2-pd}. Define the limit coverage budget
\begin{align*}
K_\eta
&\;\coloneqq\; \min\Big\{
k\ge1:\sum_{j\le k}P_j\ge1-\eta
\Big\},\\
&\hspace{1.8em}\eta\in(0,1).
\end{align*}
Then:
\begin{enumerate}[label=(\roman*), ref=\ref*{thm:t2-coverage}(\roman*)]
\item\label{thm:t2-coverage:keta} for every fixed $\eta \in (0,1)$:
$K_\eta$ is finite a.s., and under $\Ens(c)$, as $n \to \infty$,
\begin{align*}
k_\eta &\;\tod\; K_\eta,\\
&\text{equivalently}\\[-0.2em]
\Pr(k_\eta\le m)
&\;\to\; \Pr(K_\eta\le m),
\quad m\in\N,
\end{align*}
and in particular $k_\eta = O_P(1)$: condensed rows have bounded
effective support;
\item\label{thm:t2-coverage:eta} on the limit object, almost surely,
\begin{align*}
\eta^{\frac{\alpha}{1-\alpha}}\, K_\eta
&\;\longrightarrow\;
\Big( \frac{\alpha}{(1-\alpha)\, W} \Big)^{\frac{\alpha}{1-\alpha}}
\\
&\;\in\; (0, \infty)
\quad \text{as } \eta \downarrow 0.
\end{align*}
\end{enumerate}
\end{theorem}

\begin{proof}
The strategy: (i) follows from \Cref{thm:t2-pd} by the continuous mapping
theorem and the atomlessness of \Cref{lem:atomless} at the level $1 -
\eta$; (ii) follows from the strong law for $\Gamma_j$ by a termwise
squeeze of the tail sum and inversion. Note the quantifier order: in (i),
$\eta$ is fixed and $n \to \infty$; in (ii), the softmax row no longer
appears and $\eta \downarrow 0$ on the limit object alone.

\textbf{Step 1 (proof of (i): cdf transfer).} Since $\sum_{j \le m} P_j \uparrow \sum_j P_j = 1$
a.s.\ as $m \to \infty$, the threshold $1 - \eta < 1$ is a.s.\ reached at
some finite $m$, so $K_\eta < \infty$ a.s. Fix $m \in \N$. By
\Cref{def:coverage} and the definition of $K_\eta$,
\begin{align*}
\{ k_\eta \le m \}
&\;=\; \Big\{
\sum_{j \le m} A_{(j)} \ge 1-\eta
\Big\},\\
\{ K_\eta \le m \}
&\;=\; \Big\{
\sum_{j \le m} P_j \ge 1-\eta
\Big\}.
\end{align*}
By \Cref{thm:t2-pd:weights} with $K = m$ and the continuous mapping
theorem \Cref{fac:weak-convergence:cmt} applied to $h(y) = \sum_{j\le m}
y_j$,
\begin{align*}
S_m^{(n)}
&\;\coloneqq\; \sum_{j \le m} A_{(j)}\\
&\;\tod\; S_m
\;\coloneqq\; \sum_{j \le m} P_j .
\end{align*}
The law of $S_m$ has no atom at $s \coloneqq 1 - \eta \in (0,1)$ by
\Cref{lem:atomless}; in fact it has no atom anywhere in $(0,1)$, so its
cdf $F_{S_m}$ is continuous on $(0,1)$. Therefore, for every small
$\delta > 0$ with $s - \delta \in (0,1)$, the cdf criterion
\Cref{fac:weak-convergence:cdf} gives
\begin{align*}
F_{S_m}(s - \delta)
&\;=\; \lim_n F_{S_m^{(n)}}(s-\delta)\\
&\;\le\; \liminf_n \Pr\big(S_m^{(n)}<s\big)\\
&\;\le\; \limsup_n \Pr\big(S_m^{(n)}<s\big)\\
&\;\le\; \lim_n F_{S_m^{(n)}}(s)\\
&\;=\; F_{S_m}(s),
\end{align*}
and letting $\delta \downarrow 0$ (using continuity of $F_{S_m}$ at $s$
and atomlessness, $F_{S_m}(s-\delta) \to F_{S_m}(s^-) = F_{S_m}(s)$)
shows $\Pr(S_m^{(n)} < s) \to F_{S_m}(s) = \Pr(S_m < s)$, hence
\begin{align*}
\Pr( k_\eta \le m )
&\;=\; 1 - \Pr\big(S_m^{(n)}<1-\eta\big)\\
&\;\longrightarrow\;
1 - \Pr\big(S_m<1-\eta\big)\\
&\;=\; \Pr( K_\eta \le m ).
\end{align*}
Since $k_\eta$ and $K_\eta$ are positive-integer valued, their cdfs are
flat between integers: for every $t \in \R$,
\begin{align*}
\Pr( k_\eta \le t )
&\;=\; \Pr\big(k_\eta\le\lfloor t\rfloor\big),\\
&\text{and}\\[-0.2em]
\Pr( K_\eta \le t )
&\;=\; \Pr\big(K_\eta\le\lfloor t\rfloor\big),
\end{align*}
so the convergence just proved at every positive integer (trivial at
integers $m \le 0$, where both cdfs vanish) extends to every real $t$,
and $k_\eta \tod K_\eta$ by \Cref{fac:weak-convergence:cdf}. Tightness
($k_\eta = O_P(1)$)
follows: given $\delta > 0$, pick $m$ with $\Pr(K_\eta \le m) \ge 1 -
\delta/2$; then $\Pr(k_\eta \le m) \ge 1 - \delta$ for all large $n$.

\textbf{Step 2 (proof of (ii): small-$\eta$ law).} All statements in this part are almost-sure
statements about $(P_j)_j$; abbreviate the tail and its normalised limit
\begin{align*}
T_k &\;\coloneqq\; \sum_{j > k} P_j,\\
B \;\coloneqq\; \frac{\alpha}{(1-\alpha) W}.
\end{align*}
\textbf{Step 2a (tail asymptotics: $k^{(1-\alpha)/\alpha}\, T_k \to B$ a.s.).}
By \Cref{lem:gamma-lln} and continuity of $t \mapsto t^{-1/\alpha}$ at
$t = 1$,
\begin{align*}
j^{1/\alpha}\, \Gamma_j^{-1/\alpha}
\;=\; \Big( \frac{\Gamma_j}{j} \Big)^{-1/\alpha}
\;\toas\; 1,
\end{align*}
so for every $\varepsilon \in (0,1)$ there is a.s.\ a finite
$J(\varepsilon)$ with $(1-\varepsilon) j^{-1/\alpha} \le
\Gamma_j^{-1/\alpha} \le (1+\varepsilon) j^{-1/\alpha}$ for all $j >
J(\varepsilon)$. Summing over $j > k \ge J(\varepsilon)$ and comparing
with the integral of the decreasing function $x \mapsto x^{-1/\alpha}$,
\begin{align*}
\int_{k+1}^\infty x^{-1/\alpha}\, dx
&\;\le\; \sum_{j > k} j^{-1/\alpha}\\
&\;\le\; \int_{k}^\infty x^{-1/\alpha}\, dx,\\
\int_{k}^{\infty} x^{-1/\alpha} dx
&\;=\; \frac{\alpha}{1 - \alpha}
k^{-\frac{1-\alpha}{\alpha}},
\end{align*}
(the integral converges since $1/\alpha > 1$), whence, multiplying by
$k^{(1-\alpha)/\alpha}$ and using $((k+1)/k)^{-(1-\alpha)/\alpha} \to 1$,
\begin{align*}
(1 - \varepsilon)\, \frac{\alpha}{1-\alpha}
&\;\le\; \liminf_{k}
k^{\frac{1-\alpha}{\alpha}}W T_k\\
&\;\le\; \limsup_{k}
k^{\frac{1-\alpha}{\alpha}}W T_k\\
&\;\le\; (1+\varepsilon)
\frac{\alpha}{1-\alpha}
\quad\text{a.s.},
\end{align*}
where $W T_k = \sum_{j>k} \Gamma_j^{-1/\alpha}$. Letting $\varepsilon
\downarrow 0$ along a countable sequence gives
$k^{(1-\alpha)/\alpha} T_k \to B$ a.s.

\textbf{Step 2b (inversion).} Since every $P_j > 0$, we have $T_k > 0$ for
all $k$, so $K_\eta \to \infty$ as $\eta \downarrow 0$ a.s.\ (for each
fixed $k$, $\eta < T_k$ forces $K_\eta > k$). By minimality of $K_\eta$,
\begin{align*}
T_{K_\eta} \;\le\; \eta \;<\; T_{K_\eta - 1}
\qquad \text{for all } \eta \in (0, 1),
\end{align*}
with the convention $T_0 = 1$. Write $\psi(k) \coloneqq
k^{(1-\alpha)/\alpha} T_k$, so $\psi(k) \to B$ a.s.\ by Step 2a. The two
sandwich inequalities invert to
\begin{align*}
\eta^{\frac{\alpha}{1-\alpha}} K_\eta
&\;\ge\; \big( T_{K_\eta} \big)^{
\frac{\alpha}{1-\alpha}} K_\eta\\
&\;=\; \psi(K_\eta)^{
\frac{\alpha}{1-\alpha}},\\
\eta^{\frac{\alpha}{1-\alpha}} ( K_\eta - 1 )
&\;\le\; \big( T_{K_\eta - 1} \big)^{
\frac{\alpha}{1-\alpha}}
(K_\eta - 1)\\
&\;=\; \psi(K_\eta - 1)^{
\frac{\alpha}{1-\alpha}},
\end{align*}
using that $t \mapsto t^{\alpha/(1-\alpha)}$ is increasing on $(0,\infty)$
and, in the first inequality, $T_{K_\eta} \le \eta$; in the second,
$\eta < T_{K_\eta-1}$ (valid once $K_\eta \ge 2$, which holds for all
small $\eta$ a.s.). As $\eta \downarrow 0$ we have $K_\eta \to \infty$
a.s., so both $\psi(K_\eta)$ and $\psi(K_\eta - 1)$ converge to $B$ a.s.,
and
\begin{align*}
B^{\frac{\alpha}{1-\alpha}}
&\;\le\; \liminf_{\eta \downarrow 0}
\eta^{\frac{\alpha}{1-\alpha}} K_\eta\\
&\;\le\; \limsup_{\eta \downarrow 0}
\eta^{\frac{\alpha}{1-\alpha}} K_\eta\\
&\;\le\; \lim_{\eta \downarrow 0}
\Big(
\psi(K_\eta - 1)^{
\frac{\alpha}{1-\alpha}}\\
&\hspace{5.5em}
+ \eta^{\frac{\alpha}{1-\alpha}}
\Big)\\
&\;=\; B^{\frac{\alpha}{1-\alpha}},
\end{align*}
which proves the claim, with the limit $B^{\alpha/(1-\alpha)} \in
(0,\infty)$ a.s.\ because $W \in (0,\infty)$ a.s. This completes the
proof.
\end{proof}

\begin{remark}[Budget blow-up as $c \downarrow 1$]\label{rem:c-blowup}
The exponent in \Cref{thm:t2-coverage:eta} is
\begin{align*}
\frac{\alpha}{1 - \alpha}
&\;=\; \frac{1/c}{1 - 1/c}\\
&\;=\; \frac{1}{c - 1},
\end{align*}
so on the limit object $K_\eta \asymp \eta^{-1/(c-1)}$ as $\eta
\downarrow 0$: deep in the condensed phase ($c$ large) a handful of
entries cover any fixed fraction of the mass, but as $c \downarrow 1$
the required budget blows up at rate $\eta^{-1/(c-1)}$ --- the condensed
phase degenerates continuously into the dispersed phase
(\cref{sec:app-dispersed}), where
\Cref{thm:t1-main:keta} puts $k_\eta$ at a positive power of $n$.
\end{remark}

\section{Spiked Rows: Detection, Dominance, Denoising}\label{sec:app-spike}

This appendix proves \cref{thm:main-spike,thm:main-denoise} of the main text: \cref{thm:main-spike} is \cref{thm:t3-rank,prop:t3-dominance}, and \cref{thm:main-denoise} is \cref{thm:t4}.

\subsection{Spike detectability and dominance}\label{sec:app-spike-detect}

Throughout this subsection we work under $\Espike(\mu;\sigma)$ of
\Cref{def:ensembles} with $\sigma > 0$ fixed: the spike index $j^\star$
carries the deterministic logit $\ell_{j^\star} = \mu_n \ge 0$, and the
$n - 1$ \emph{bulk} logits are i.i.d.\ $N(0,\sigma^2)$. Following the
natural scale of the bulk maximum, write
\begin{align}\label{eq:m-scale}
m_n \;\coloneqq\; \frac{\mu_n}{\sigma \sqrt{2 \log n}},
\end{align}
and quantify all statements over sequences $(\mu_n)_n$ for which $m_n \to
m \in [0, \infty)$. The two relevant statistics are the bulk exceedance
count and the bulk partition function,
\begin{align}\label{eq:bulk-stats}
\begin{aligned}
N_b(v)
&\;\coloneqq\; \#\big\{ j \ne j^\star : \ell_j > v \big\}
\quad (v \in \R),\\
Z_b
&\;\coloneqq\; \sum_{j \ne j^\star} e^{\ell_j},
\end{aligned}
\end{align}
and the \emph{rank} of the spike is $\mathrm{rank}(j^\star) \coloneqq
\min\{ k : j^\star \in S_k \}$, where $S_k$ is the set of indices of the
$k$ largest weights (ties broken by index). Since the bulk logits are
atomless and $\mu_n$ is deterministic, almost surely no bulk logit equals
$\mu_n$, and (the softmax being monotone)
\begin{align}\label{eq:rank-identity}
\mathrm{rank}(j^\star) \;=\; 1 + N_b(\mu_n) \quad \text{almost surely}.
\end{align}
The \emph{minimal detecting budget} is $k^\ast(\mu_n) \coloneqq
\mathrm{rank}(j^\star)$: the smallest top-$k$ budget whose kept set
contains the spike.

\begin{lemma}[Bulk exceedance counts and bulk maximum]\label{lem:spike-count}
Let $\sigma > 0$ be fixed, let $m_n \to m \in [0,\infty)$, and set
$v_n \coloneqq m_n\, \sigma \sqrt{2\log n}$. (Here $(m_n)_n$ may be any
real sequence with $\limsup_n |m_n| < \infty$ converging to such an $m$;
in particular individual $m_n$ may be negative, and this quantification is
independent of the $\mu_n \ge 0$ hypothesis used in \Cref{thm:t3-rank}.) Write $M_b \coloneqq
\max_{j \ne j^\star} \ell_j$. Then:
\begin{enumerate}[label=(\roman*), ref=\ref*{lem:spike-count}(\roman*)]
\item\label{lem:spike-count:twosided} if $m < 1$: for every fixed
$\varepsilon \in (0, 1-m^2)$ there is $n_8(\sigma, (m_n)_n,
\varepsilon)$ such that for all $n \ge n_8$,
\begin{align*}
&\Pr\Big(
\begin{aligned}
n^{1 - m^2 - \varepsilon}
&\le N_b(v_n)\\
&\le n^{1-m^2+\varepsilon}
\end{aligned}
\Big)\\
&\;\ge\; 1 - 2\, e^{-n^{(1-m^2)/2}/8}.
\end{align*}
\item\label{lem:spike-count:zero} if $m > 1$:
$\Pr\big( N_b(v_n) \ge 1 \big) \le n^{1 - m_n^2} \to 0$;
\item\label{lem:spike-count:max} $M_b / (\sigma\sqrt{2 \log n}) \toP 1$;
\item\label{lem:spike-count:sharp} $\Pr\big( M_b > \sigma \sqrt{2\log n}
\big) \le (4 \pi \log n)^{-1/2} \to 0$.
\end{enumerate}
\end{lemma}

\begin{proof}
The strategy is the same level bookkeeping as in \Cref{sec:app-level},
specialised to a single threshold: compute the mean of the binomial count
from the Gaussian tail and apply \Cref{lem:chernoff}. Since the bulk
logits are $\sigma g$ with $g \sim N(0,1)$, $N_b(v_n) \sim \Bin(n-1,
\barPhi(x_n))$ with $x_n \coloneqq v_n/\sigma = m_n \sqrt{2 \log n}$.

\textbf{Step 1 (mean bounds).} For the upper bound, whenever $m_n \ge 0$ (which
holds eventually if $m > 0$; for $m = 0$ and $m_n < 0$ we use the
trivial bound $\E N_b \le n$ instead, which suffices below),
\Cref{fac:gaussian-tail} gives
\begin{align}\label{eq:spike-mean-upper}
\E N_b(v_n) \;\le\; n\, \barPhi(x_n)
\;\le\; n\, e^{-x_n^2/2} \;=\; n^{1 - m_n^2}.
\end{align}
For the lower bound assume $m < 1$ and fix $\varepsilon \in (0, 1-m^2)$.
If $x_n \le \sqrt 2$ then $\barPhi(x_n) \ge \barPhi(\sqrt2) \ge 1/20$,
so $\E N_b \ge (n-1)/20 \ge n^{1 - \varepsilon/2}$ for $n$ large. If
$x_n \ge \sqrt 2$ then \Cref{fac:gaussian-tail} gives $\barPhi(x_n) \ge
\varphi(x_n)/(2 x_n)$, so, with $\bar m \coloneqq \sup_n |m_n| <
\infty$,
\begin{align}\label{eq:spike-mean-lower}
\begin{split}
\E N_b(v_n)
&\;\ge\; (n-1)\, \frac{e^{-x_n^2/2}}{2\sqrt{2\pi}\, x_n}
\\
&\;\ge\;
\frac{n}{2} \cdot
\frac{n^{-m_n^2}}{2\sqrt{2\pi}\, \bar m \sqrt{2\log n}}\\
&\;\ge\; n^{1 - m^2 - \varepsilon/2},
\end{split}
\end{align}
for all $n \ge n_8$, using $n - 1 \ge n/2$, $\sqrt{\log n} = n^{o(1)}$,
and $|m_n^2 - m^2| \le \varepsilon/8$ eventually. In both cases $\E
N_b(v_n) \ge n^{1 - m^2 - \varepsilon/2}$ for $n \ge n_8$.

\textbf{Step 2 (proof of (i): two-sided count).} By Eq.~\eqref{eq:spike-mean-upper} (or the trivial
bound $n$ when $m = 0 > m_n$), $\E N_b \le n^{1-m_n^2} \le
n^{1-m^2+\varepsilon/2}$ eventually, so $t \coloneqq
n^{1-m^2+\varepsilon}/2 \ge \E N_b$ for $n$ large and
\Cref{lem:chernoff:upper} gives
\begin{align*}
\Pr\Big( N_b(v_n) \ge n^{1-m^2+\varepsilon} \Big)
\;\le\; e^{-n^{1-m^2+\varepsilon}/6},
\end{align*}
while \Cref{lem:chernoff:lower} together with $n^{1-m^2-\varepsilon} \le
\tfrac12 \E N_b$ (valid for $n$ large by the mean lower bound) gives
\begin{align*}
&\Pr\Big(
N_b(v_n) \le n^{1-m^2-\varepsilon} \Big)\\
&\;\le\; \Pr\Big(
N_b(v_n) \le \tfrac12 \E N_b(v_n) \Big)\\
&\;\le\; e^{-\E N_b(v_n)/8}\\
&\;\le\; e^{-n^{1-m^2-\varepsilon/2}/8}.
\end{align*}
By a union bound over the two tails, and since both exponents
$1-m^2+\varepsilon$ and $1 - m^2 - \varepsilon/2$ exceed $(1-m^2)/2$ (as
$\varepsilon < 1 - m^2$), the failure probability is at most
$2e^{-n^{(1-m^2)/2}/8}$ for $n \ge n_8$.

\textbf{Step 3 (proof of (ii): no exceedances).} Markov's inequality and
Eq.~\eqref{eq:spike-mean-upper}: $\Pr(N_b \ge 1) \le \E N_b \le
n^{1-m_n^2}$, and $1 - m_n^2 \to 1 - m^2 < 0$.

\textbf{Step 4 (proof of (iii): bulk maximum).} Fix $\delta \in (0,1)$. For the upper deviation,
applying Eq.~\eqref{eq:spike-mean-upper} with the constant sequence
$m_n' = 1 + \delta$ and Markov,
\begin{align*}
&\Pr\big(
M_b > (1+\delta)\sigma\sqrt{2\log n} \big)
\\
&\;=\; \Pr\big(
N_b\big((1+\delta)\sigma\sqrt{2\log n}\big)
\ge 1 \big)\\
&\;\le\; n^{1 - (1+\delta)^2}\\
&\;=\; n^{-2\delta - \delta^2}\\
&\;\to\; 0.
\end{align*}
For the lower deviation, part (i) with the constant sequence $m_n' = 1 -
\delta$ (limit $m' = 1-\delta < 1$) and any fixed $\varepsilon \in (0,
1-(1-\delta)^2)$ gives $N_b((1-\delta)\sigma\sqrt{2\log n}) \ge 1$ with
probability $\to 1$, i.e.\ $\Pr(M_b \le (1-\delta)\sigma\sqrt{2\log n})
\to 0$.

\textbf{Step 5 (proof of (iv): sharp no-slack bound).} By a union bound over the $n - 1$ bulk entries and
\Cref{fac:gaussian-tail} ($\barPhi(x) \le \varphi(x)/x$ for $x > 0$),
\begin{align*}
\Pr\big( M_b > \sigma\sqrt{2\log n} \big)
&\;\le\; n\, \barPhi\big(
\sqrt{2 \log n} \big)\\
&\;\le\; n \cdot
\frac{\varphi(\sqrt{2\log n})}{\sqrt{2\log n}}\\
&\;=\; n \cdot
\frac{n^{-1}}{\sqrt{2\pi}\,\sqrt{2 \log n}}\\
&\;=\; \frac{1}{\sqrt{4\pi \log n}},
\end{align*}
using $\varphi(\sqrt{2\log n}) = (2\pi)^{-1/2} e^{-\log n} =
n^{-1}/\sqrt{2\pi}$. This completes the proof.
\end{proof}

\begin{theorem}[Minimal detecting budget; formal version of \cref{thm:main-spike}]\label{thm:t3-rank}
Let $\sigma > 0$ be fixed and let $\mu_n \ge 0$ satisfy $m_n \to m \in
[0, \infty)$. Under $\Espike(\mu_n; \sigma)$:
\begin{enumerate}[label=(\roman*), ref=\ref*{thm:t3-rank}(\roman*)]
\item\label{thm:t3-rank:exponent}
$\dfrac{\log k^\ast(\mu_n)}{\log n} \;\toP\; (1 - m^2)_+ \,$;
\item\label{thm:t3-rank:rankone} if $m > 1$, then
$\Pr\big( k^\ast(\mu_n) = 1 \big) \to 1$: the spike is whp the
top-ranked entry;
\item\label{thm:t3-rank:band} for every fixed $\beta \in (0,1)$, with
$k_n = \lceil n^\beta \rceil$:
\begin{align*}
\Pr\big( j^\star \in S_{k_n} \big)
&\to 1
\quad \text{ if } m^2 > 1 - \beta,\\
\Pr\big( j^\star \in S_{k_n} \big)
&\to 0
\quad \text{ if } m^2 < 1 - \beta.
\end{align*}
\end{enumerate}
\end{theorem}

\begin{proof}
The strategy is to translate everything to the bulk count through the
rank identity Eq.~\eqref{eq:rank-identity} and apply
\Cref{lem:spike-count} with $v_n = \mu_n$ (i.e.\ with the given sequence
$(m_n)_n$).

\textbf{Step 1 (proof of (i), case $m < 1$: count exponent).} Fix $\varepsilon \in (0, (1-m^2)/2)$.
On the event of \Cref{lem:spike-count:twosided}, whose probability tends
to one,
\begin{align*}
n^{1-m^2-\varepsilon}
&\;\le\; N_b(\mu_n)
\\
&\;\le\; k^\ast(\mu_n)
\;=\; 1 + N_b(\mu_n)\\
&\;\le\; 2\, n^{1-m^2+\varepsilon}
\;\le\; n^{1-m^2+2\varepsilon},
\end{align*}
the last step for $n$ large, whence $\big| \log k^\ast(\mu_n)/\log n - (1-m^2)
\big| \le 2\varepsilon$ with probability $\to 1$; as $\varepsilon$ was
arbitrary, part (i) follows in this case (note $(1-m^2)_+ = 1-m^2$).

\textbf{Step 2 (proof of (i), case $m \ge 1$, and of (ii)).} Always $k^\ast \ge 1$,
so $\log k^\ast / \log n \ge 0 = (1-m^2)_+$. For the upper bound fix
$\varepsilon > 0$; by Markov's inequality, Eq.~\eqref{eq:rank-identity},
and Eq.~\eqref{eq:spike-mean-upper},
\begin{align*}
\Pr\big( k^\ast(\mu_n) > n^{\varepsilon} \big)
&\;\le\; \Pr\big(
N_b(\mu_n) \ge n^{\varepsilon}/2 \big)\\
&\;\le\; \frac{2\,\E N_b(\mu_n)}{n^{\varepsilon}}\\
&\;\le\; 2\, n^{1 - m_n^2 - \varepsilon}
\;\longrightarrow\; 0,
\end{align*}
since $1 - m_n^2 - \varepsilon \to 1 - m^2 - \varepsilon \le
-\varepsilon < 0$. Hence $\log k^\ast/\log n \toP 0 = (1-m^2)_+$. If
moreover $m > 1$, \Cref{lem:spike-count:zero} gives $N_b(\mu_n) = 0$ with
probability $\to 1$, i.e.\ $k^\ast = 1$, proving (ii).

\textbf{Step 3 (proof of (iii): budget comparison).} The events coincide: $j^\star \in S_{k_n}
\Leftrightarrow k^\ast(\mu_n) \le k_n$. If $m^2 > 1 - \beta$, part (i)
with $\varepsilon \coloneqq (\beta - (1-m^2)_+)/2 > 0$ gives, with
probability $\to 1$,
\begin{align*}
\log k^\ast(\mu_n)
&\;\le\; \big(
(1-m^2)_+ + \varepsilon \big) \log n\\
&\;<\; \beta \log n\\
&\;\le\; \log k_n,
\end{align*}
so $\Pr(j^\star \in S_{k_n}) \to 1$. If $m^2 < 1 - \beta$ then $m < 1$
and $1 - m^2 > \beta$; part (i) with $\varepsilon \coloneqq (1 - m^2 -
\beta)/2$ gives, with probability $\to 1$,
\begin{align*}
\log k^\ast(\mu_n)
&\;\ge\; \big(
1 - m^2 - \varepsilon \big) \log n\\
&\;=\; \Big(
\beta + \tfrac{1-m^2-\beta}{2} \Big) \log n\\
&\;>\; \log\big( 2 n^{\beta} \big)\\
&\;\ge\; \log k_n,
\end{align*}
the second-to-last step for $n$ large, so $\Pr(j^\star \in S_{k_n}) \to
0$. This completes the proof.
\end{proof}

\begin{proposition}[Dominance threshold; formal version of \cref{thm:main-spike}]\label{prop:t3-dominance}
Let $\sigma > 0$ be fixed and $\mu_n \ge 0$. Under
$\Espike(\mu_n;\sigma)$:
\begin{enumerate}[label=(\roman*), ref=\ref*{prop:t3-dominance}(\roman*)]
\item\label{prop:t3-dominance:lln} $\dfrac{Z_b}{n-1} \toP
e^{\sigma^2/2}$;
\item\label{prop:t3-dominance:theta} if $\mu_n \ge \log n - C_0$ for a
fixed $C_0 \ge 0$ and all $n$, then with probability $\to 1$,
$\;A_{j^\star} \ge \big( 1 + 2 e^{\sigma^2/2 + C_0} \big)^{-1}$;
\item\label{prop:t3-dominance:small} if $\mu_n \le (1-\varepsilon) \log
n$ for a fixed $\varepsilon \in (0,1)$ and all $n$, then with
probability $\to 1$, $\;A_{j^\star} \le 2\, e^{-\sigma^2/2}\,
n^{-\varepsilon}$;
\item\label{prop:t3-dominance:vanish} if $\mu_n - \log n \to -\infty$,
then $A_{j^\star} \toP 0$.
\end{enumerate}
\end{proposition}

\begin{proof}
The strategy is a Chebyshev weak law for $Z_b$ (the summands are i.i.d.\
log-normals with finite variance because $\sigma$ is fixed), then direct
substitution into $A_{j^\star} = e^{\mu_n}/(e^{\mu_n} + Z_b)$.

\textbf{Step 1 (proof of (i): Chebyshev LLN).} For $\ell \sim N(0,\sigma^2)$, completing the square,
\begin{align*}
\E e^{\ell}
&\;=\; \int e^{\sigma x} \varphi(x)\, dx
\;=\; e^{\sigma^2/2},\\
\E e^{2\ell}
&\;=\; e^{2\sigma^2},\\
\operatorname{Var}\big( e^{\ell} \big) &\;=\; e^{2\sigma^2} -
e^{\sigma^2} \;<\; \infty.
\end{align*}
By independence and Chebyshev's inequality, for every fixed $\tau > 0$,
\begin{align}\label{eq:zb-lln}
\Pr\Big(
\Big| \frac{Z_b}{n-1} - e^{\sigma^2/2} \Big|
> \tau \Big)
&\;\le\;
\frac{e^{2\sigma^2} - e^{\sigma^2}}{(n-1)\,\tau^2}\\
&\;\longrightarrow\; 0.
\notag
\end{align}
In particular, applying Eq.~\eqref{eq:zb-lln} with $\tau =
e^{\sigma^2/2}/3$ and using $n - 1 \ge \tfrac34 n$ for $n \ge 4$: with
probability $\to 1$,
\begin{align}\label{eq:zb-sandwich}
\begin{split}
\tfrac{1}{2}\, n\, e^{\sigma^2/2}
&\;\le\; \tfrac{2}{3}\,(n-1)\, e^{\sigma^2/2}
\;\le\; Z_b\\
&\;\le\; \tfrac{4}{3}\,(n-1)\, e^{\sigma^2/2}
\;\le\; 2\, n\, e^{\sigma^2/2}.
\end{split}
\end{align}

\textbf{Step 2 (proof of (ii): dominance).} Since $x \mapsto 1/(1+x)$ is decreasing and
$A_{j^\star} = \big(1 + e^{-\mu_n} Z_b\big)^{-1}$, on the event
Eq.~\eqref{eq:zb-sandwich}, and using $-\mu_n \le C_0 - \log n$ by the
hypothesis $\mu_n \ge \log n - C_0$,
\begin{align*}
e^{-\mu_n} Z_b
&\;\le\; e^{C_0 - \log n}
\cdot 2 n e^{\sigma^2/2}\\
&\;=\; 2 e^{\sigma^2/2 + C_0},\\
\text{hence}\qquad
A_{j^\star}
&\;\ge\; \big(
1 + 2 e^{\sigma^2/2 + C_0} \big)^{-1}.
\end{align*}

\textbf{Step 3 (proof of (iii): polynomially small weight).} On the event Eq.~\eqref{eq:zb-sandwich}, dropping
$e^{\mu_n}$ from the denominator,
\begin{align}\label{eq:dominance-upper}
A_{j^\star} \;\le\; \frac{e^{\mu_n}}{Z_b}
\;\le\; \frac{n^{1-\varepsilon}}{\tfrac12\, n\, e^{\sigma^2/2}}
\;=\; 2\, e^{-\sigma^2/2}\, n^{-\varepsilon}.
\end{align}

\textbf{Step 4 (proof of (iv): vanishing weight).} As in Eq.~\eqref{eq:dominance-upper},
$A_{j^\star} \le 2 e^{-\sigma^2/2} e^{\mu_n - \log n} \to 0$ on the
event Eq.~\eqref{eq:zb-sandwich}, and the event has probability $\to 1$,
so $A_{j^\star} \toP 0$. This completes the proof.
\end{proof}

\begin{remark}[Interpretation: detect early, dominate late]\label{rem:t3-gap}
\Cref{thm:t3-rank,prop:t3-dominance} separate two scales. The spike
enters the top-$n^\beta$ as soon as $\mu_n$ passes
$\sigma\sqrt{2(1-\beta)\log n} \asymp \sqrt{\log n}$, but its
\emph{softmax weight} stays polynomially small until $\mu_n$ reaches
$\log n - O(1)$ --- a multiplicative gap of order $\sqrt{\log n}/\sigma$
in the logit. In the whole intermediate band the spike is selectable by
a sparse scheme yet invisible to the dense softmax average; this is the
window \Cref{sec:app-spike-denoise} exploits.
\end{remark}

\begin{remark}[One-sided sub-Gaussian extension]\label{rem:subgaussian}
Suppose the bulk variables are merely $\sigma^2$-sub-Gaussian, i.e.\
$\Pr(\ell_j > t) \le e^{-t^2/(2\sigma^2)}$ for all $t \ge 0$. Then the
mean bound Eq.~\eqref{eq:spike-mean-upper} and hence all
\emph{upper-bound} conclusions persist verbatim: $\E N_b(v_n) \le
n^{1-m_n^2}$, so $k^\ast(\mu_n) \le n^{(1-m^2)_+ + \varepsilon}$ whp
(detectability with budget $n^{(1-m^2)_+ + \varepsilon}$ suffices), and
$m > 1$ still forces $\mathrm{rank}(j^\star) = 1$ whp. The
\emph{matching lower bounds} are distribution-specific and fail under
bare sub-Gaussianity: the deterministic bulk $\ell_j \equiv 0$ is
$\sigma^2$-sub-Gaussian and has no exceedances of any positive level, so
$k^\ast = 1$ regardless of $m$. Likewise the dominance threshold of
\Cref{prop:t3-dominance} uses the exact log-normal mean $e^{\sigma^2/2}$
and does not transfer. We claim nothing two-sided beyond the Gaussian
bulk.
\end{remark}
\subsection{Denoising: sparse attention amplifies a planted token}\label{sec:app-spike-denoise}

This subsection proves the headline of the spiked model: in an explicit band
of spike strengths, renormalised top-$k$ attention gives the planted
token constant weight while exact softmax provably gives it vanishing
weight. We work under $\Espike(\mu_n; \sigma)$ with $\sigma > 0$ fixed
and use the notation of \Cref{sec:app-spike-detect} ($N_b$, $Z_b$, $M_b$,
$\mathrm{rank}(j^\star)$, $S_k$). The budget is a sequence $k_n$ with
\begin{align}\label{eq:budget-condition}
\begin{aligned}
1 &\;\le\; k_n,\\
\log k_n &\;=\; o\big( \sqrt{\log n} \big)\\
&\qquad (\text{e.g.\ } k_n \text{ constant}\\
&\qquad\hspace{4.25em}
\text{or polylogarithmic}),
\end{aligned}
\end{align}
the kept set is $S_n \coloneqq S_{k_n}$ (top-$k_n$ by weight), the
\emph{kept-bulk sum} and the \emph{renormalised spike weight} are
\begin{align}\label{eq:bs-def}
\begin{aligned}
B_S
&\;\coloneqq\;
\sum_{j \in S_n \setminus \{j^\star\}} e^{\ell_j},\\
\wt A_{j^\star}
&\;\coloneqq\; \frac{A_{j^\star}}{p_{S_n}}\\
&\;=\; \frac{e^{\mu_n}}
{\sum_{j \in S_n} e^{\ell_j}}\\
&\;=\; \frac{1}{1 + e^{-\mu_n} B_S}\\
&\qquad \text{on } \{ j^\star \in S_n \},
\end{aligned}
\end{align}
consistently with the renormalised scheme of \Cref{def:schemes} (the
weight that $\wt O_{S_n}$ places on $V_{j^\star}$).

\begin{theorem}[The denoising band; formal version of \cref{thm:main-denoise}]\label{thm:t4}
Fix $\sigma > 0$, $C > 0$, and $\varepsilon \in (0,1)$, let $k_n$
satisfy Eq.~\eqref{eq:budget-condition}, and let $\mu_n$ satisfy the
\emph{band condition}
\begin{align}\label{eq:app-band}
\begin{aligned}
\sigma \sqrt{2 \log n} + \log k_n + C
&\;\le\; \mu_n\\
&\;\le\; (1 - \varepsilon) \log n\\
&\qquad \text{for all } n \ge n_{10},
\end{aligned}
\end{align}
which is satisfiable for all $n$ large since $(1-\varepsilon)\log n -
\sigma\sqrt{2\log n} - \log k_n - C \to \infty$ under
Eq.~\eqref{eq:budget-condition}. Then, under $\Espike(\mu_n;\sigma)$,
with probability $\to 1$:
\begin{enumerate}[label=(\roman*), ref=\ref*{thm:t4}(\roman*)]
\item\label{thm:t4:rank} $\mathrm{rank}(j^\star) = 1$; in particular
$j^\star \in S_n$;
\item\label{thm:t4:bulk} the kept-bulk sum is controlled two-sided:
\begin{align*}
B_S
&\;\le\; k_n\, e^{\sigma\sqrt{2\log n}},\\
&\qquad\text{and, if } k_n \ge 2
\text{, for every fixed}\\
&\qquad \delta \in (0,1): \quad\\
&\qquad B_S \;\ge\;
e^{(1-\delta)\,\sigma \sqrt{2\log n}},
\end{align*}
the lower bound holding on its own probability-$\to 1$ event;
\item\label{thm:t4:weights} sparse versus dense weight:
\begin{align*}
\wt A_{j^\star}
&\;\ge\; \frac{1}{1 + e^{-C}},\\
\text{while}\qquad
A_{j^\star}
&\;\le\; 2\, e^{-\sigma^2/2}\, n^{-\varepsilon},
\end{align*}
so the sparse weight is constant and the dense weight polynomially
small;
\item\label{thm:t4:amplification} log-amplification: for every fixed
$\delta > 0$,
\begin{align*}
\log \frac{\wt A_{j^\star}}{A_{j^\star}}
\;\ge\; (\varepsilon - \delta) \log n,
\end{align*}
and moreover the proof shows the asymptotic expansion
$\log\big( \wt A_{j^\star}/A_{j^\star} \big) = \log n - \mu_n +
\sigma^2/2 + \xi_n - \zeta_n$ with $\xi_n \toP 0$ and $\zeta_n \in
[0, e^{-C}]$ on the same event.
\end{enumerate}
\end{theorem}

\begin{proof}
The strategy: the band's lower edge places the spike strictly above the
bulk maximum plus a $\log k_n + C$ margin, which simultaneously forces
rank one and caps the kept-bulk sum after renormalisation at $e^{-C}$;
the band's upper edge keeps the spike negligible for the dense softmax
via the LLN of \Cref{prop:t3-dominance}. Throughout write $r_n \coloneqq
\log k_n + C \ge C$.

\textbf{Step 1 (proof of (i): rank one).} Since $\mu_n \ge \sigma\sqrt{2\log n} + r_n \ge 0$,
\Cref{fac:gaussian-tail} and the expansion of the square give
\begin{align*}
\frac{\mu_n^2}{2\sigma^2}
&\;\ge\; \frac{\big(
\sigma\sqrt{2\log n} + r_n \big)^2}{2\sigma^2}\\
&\;=\; \log n
+ \frac{r_n \sqrt{2 \log n}}{\sigma}
+ \frac{r_n^2}{2\sigma^2}\\
&\;\ge\; \log n
+ \frac{r_n\sqrt{2\log n}}{\sigma},
\end{align*}
so that, by Markov's inequality applied to the binomial count
$N_b(\mu_n)$ of Eq.~\eqref{eq:bulk-stats},
\begin{align}\label{eq:t4-rank-one}
\Pr\big( \mathrm{rank}(j^\star) \ne 1 \big)
\begin{aligned}
&\;=\; \Pr\big( N_b(\mu_n) \ge 1 \big)\\
&\;\le\; n\, \barPhi\Big( \frac{\mu_n}{\sigma} \Big)\\
&\;\le\; n\, e^{-\mu_n^2/(2\sigma^2)}\\
&\;\le\; e^{-r_n \sqrt{2\log n}/\sigma}\\
&\;\le\; e^{-C\sqrt{2\log n}/\sigma}\\
&\;\to\; 0,
\end{aligned}
\end{align}
using Eq.~\eqref{eq:rank-identity} in the first equality. On
$\{\mathrm{rank}(j^\star) = 1\}$ we have $j^\star \in S_n$ because
$k_n \ge 1$.

\textbf{Step 2 (proof of (ii): kept-bulk control).} Work on the intersection of
$\{\mathrm{rank}(j^\star) = 1\}$, $\{M_b \le \sigma\sqrt{2\log n}\}$,
and (for the lower bound) $\{M_b > (1-\delta)\sigma\sqrt{2\log n}\}$; by
Eq.~\eqref{eq:t4-rank-one}, \Cref{lem:spike-count:sharp}, and
\Cref{lem:spike-count:max}, together with a union bound, this
intersection has probability $\to 1$. On $\{\mathrm{rank}(j^\star) =
1\}$ the kept set is $S_n = \{j^\star\} \cup (\text{the } k_n - 1
\text{ top bulk indices})$, so
\begin{align*}
B_S
&\;\le\; (k_n - 1)\, e^{M_b}\\
&\;\le\; k_n\, e^{\sigma\sqrt{2\log n}},\\
\text{and, if } k_n \ge 2: \quad
B_S
&\;\ge\; e^{M_b}\\
&\;\ge\; e^{(1-\delta)\sigma\sqrt{2\log n}},
\end{align*}
where the lower chain uses that for $k_n \ge 2$ the top bulk entry
belongs to $S_n$.

\textbf{Step 3 (proof of (iii): sparse versus dense weight).} On the event of Step 2, by the band's lower edge
Eq.~\eqref{eq:app-band},
\begin{align}\label{eq:t4-bs-margin}
\begin{split}
e^{-\mu_n} B_S
&\;\le\; k_n\, e^{\sigma\sqrt{2\log n} - \mu_n}
\\
&\;\le\; k_n\, e^{-\log k_n - C}
\;=\; e^{-C},
\\
\text{hence}\qquad
\wt A_{j^\star}
&\;=\; \frac{1}{1 + e^{-\mu_n} B_S}\\
&\;\ge\; \frac{1}{1 + e^{-C}},
\end{split}
\end{align}
by Eq.~\eqref{eq:bs-def}. The dense bound $A_{j^\star} \le
2e^{-\sigma^2/2} n^{-\varepsilon}$ is
\Cref{prop:t3-dominance:small} applied to $\mu_n' \coloneqq
\min\big( \mu_n,\, (1-\varepsilon)\log n \big) \vee 0$, which satisfies
$0 \le \mu_n' \le (1-\varepsilon)\log n$ for all $n$ and, by the band's
upper edge Eq.~\eqref{eq:app-band}, equals $\mu_n$ for $n \ge n_{10}$, so
the probability-$\to 1$ conclusion transfers to
$\Espike(\mu_n;\sigma)$; intersect with its event (again a union bound,
probability $\to 1$).

\textbf{Step 4 (proof of (iv): log-amplification).} On the intersection of all previous events with the
event of Eq.~\eqref{eq:zb-sandwich}, write the exact decomposition
\begin{align}\label{eq:t4-log-ratio}
\begin{aligned}
\log \frac{\wt A_{j^\star}}{A_{j^\star}}
&\;=\; \log \frac{Z}
{\sum_{j \in S_n} e^{\ell_j}}\\
&\;=\; \log Z - \mu_n - \zeta_n,\\
\zeta_n
&\;\coloneqq\; \log\big(
1 + e^{-\mu_n} B_S \big)\\
&\;\in\; \big[ 0,\, e^{-C} \big],
\end{aligned}
\end{align}
where the range of $\zeta_n$ uses Eq.~\eqref{eq:t4-bs-margin} and
$\log(1+x) \le x$. For the partition function, $Z = e^{\mu_n} + Z_b$,
so
\begin{align*}
\log Z
&\;=\; \log Z_b
+ \log\Big( 1 + \frac{e^{\mu_n}}{Z_b} \Big)\\
&\;=\; \log n + \frac{\sigma^2}{2} + \xi_n,\\
\xi_n
&\;\coloneqq\; \log \frac{Z_b}{n\, e^{\sigma^2/2}}\\
&\hphantom{\;\coloneqq\;}
+ \log\Big( 1 + \frac{e^{\mu_n}}{Z_b} \Big),
\end{align*}
and $\xi_n \toP 0$: the first term because $Z_b/(ne^{\sigma^2/2}) =
\big(Z_b/((n-1)e^{\sigma^2/2})\big) \cdot (n-1)/n \toP 1$ by
\Cref{prop:t3-dominance:lln} and the continuous mapping theorem
\Cref{fac:weak-convergence:cmt}, the second because $e^{\mu_n}/Z_b \le
2e^{-\sigma^2/2} n^{-\varepsilon} \toP 0$ as in
Eq.~\eqref{eq:dominance-upper}. Substituting into
Eq.~\eqref{eq:t4-log-ratio} and using the band's upper edge $\mu_n \le
(1-\varepsilon)\log n$,
\begin{align*}
\log \frac{\wt A_{j^\star}}{A_{j^\star}}
&\;=\; \log n - \mu_n
+ \frac{\sigma^2}{2} + \xi_n - \zeta_n\\
&\;\ge\; \varepsilon \log n
+ \frac{\sigma^2}{2} - e^{-C} + \xi_n\\
&\;\ge\; (\varepsilon - \delta) \log n,
\end{align*}
with probability $\to 1$ for every fixed $\delta > 0$, since
$\sigma^2/2 - e^{-C} + \xi_n \ge -\delta \log n$ with probability $\to
1$. This completes the proof.
\end{proof}

\begin{corollary}[Task loss inside the denoising band]\label{cor:t4-task-loss}
Assume \Cref{thm:t4}.  Let \(V_{j^\star}\) be fixed, and let the bulk values
\((V_j)_{j\ne j^\star}\) be i.i.d., independent of the logits, with mean
\(\bar V_b\) and
\(\E\|V_j-\bar V_b\|_2^2\le\sigma_V^2\).  Write
\(w_e=A_{j^\star}\), \(w_s=\widetilde A_{j^\star}\), and
\(D=\|V_{j^\star}-\bar V_b\|_2>0\).  In addition, suppose the normalized
kept-bulk weights
\[
 b_j\coloneqq
 \frac{\widetilde A_j}{1-w_s},
 \qquad j\in S_n\setminus\{j^\star\},
\]
satisfy
\[
 \sum_{j\in S_n\setminus\{j^\star\}}b_j^2=O_\Pr(k_n^{-1}).
\]
Then for
\(y_\alpha=\alpha V_{j^\star}+(1-\alpha)\bar V_b\),
\[
\begin{aligned}
\big|\|O-y_\alpha\|_2-|\alpha-w_e|D\big|
 &=O_\Pr(\sigma_Vn^{-1/2}),\\
\big|\|\widetilde O_{S_n}-y_\alpha\|_2-|\alpha-w_s|D\big|
 &=O_\Pr(\sigma_Vk_n^{-1/2}).
\end{aligned}
\]
Write
\begin{align*}
\alpha_n^\star&=\frac{w_e+w_s}{2},\\
\Delta_n(\alpha)&=\bigl||\alpha-w_e|-|\alpha-w_s|\bigr|.
\end{align*}
Consequently, for any deterministic or logit-measurable
\(\alpha_n\in[0,1]\) satisfying
\[
D\,\Delta_n(\alpha_n)
\gg_\Pr
\sigma_V\!\left(n^{-1/2}+k_n^{-1/2}\right),
\]
where \(\gg_\Pr\) means that the ratio of the left- and right-hand sides
diverges in probability,
exact attention has smaller loss when
\(\alpha_n<\alpha_n^\star\), and top-\(k_n\) attention has smaller loss when
\(\alpha_n>\alpha_n^\star\), with probability tending to one.  For the value-aware
oracle \(S^\star\in\arg\min_{|S|=k_n}\|\widetilde O_S-y_\alpha\|_2\),
\(\|\widetilde O_{S^\star}-y_\alpha\|_2
\le\|\widetilde O_{S_n}-y_\alpha\|_2\) deterministically.
\end{corollary}

\begin{proof}
Let \(Z_b=\sum_{j\ne j^\star}e^{\ell_j}\) and
\(c_j=e^{\ell_j}/Z_b\).  The exact output decomposes as
\[
 O=w_eV_{j^\star}+(1-w_e)\sum_{j\ne j^\star}c_jV_j.
\]
Because the lognormal bulk has finite first and second moments, the law of
large numbers gives
\[
 \sum_{j\ne j^\star}c_j^2
 =\frac{\sum_{j\ne j^\star}e^{2\ell_j}}{Z_b^2}
 =O_\Pr(n^{-1}).
\]
Conditional on the logits, independence and centering eliminate the cross
terms, so
\[
 \E\!\left[
 \left\|\sum_{j\ne j^\star}c_j(V_j-\bar V_b)\right\|_2^2
 \,\middle|\,\ell\right]
 \le \sigma_V^2\sum_{j\ne j^\star}c_j^2.
\]
Markov's inequality therefore makes the exact bulk deviation
\(O_\Pr(\sigma_V/\sqrt n)\).  On the event \(j^\star\in S_n\), the sparse
output similarly decomposes as
\[
 \widetilde O_{S_n}
 =w_sV_{j^\star}
 +(1-w_s)\sum_{j\in S_n\setminus\{j^\star\}}b_jV_j.
\]
The stated diffuseness condition and the same conditional second-moment
calculation make its bulk deviation
\(O_\Pr(\sigma_V/\sqrt{k_n})\).

Subtract \(y_\alpha\) from the two decompositions and use
\(\big|\|x+r\|_2-\|x\|_2\big|\le\|r\|_2\) with
\(x=(w-\alpha)(V_{j^\star}-\bar V_b)\).  This proves both loss expansions.
\Cref{thm:t4:weights} gives \(w_s>w_e\) with probability tending to one.
The noiseless distances to \(w_e\) and \(w_s\) cross only at their midpoint,
and their exact gap is
\[
D\,\Delta_n(\alpha)
=D\min\!\left\{w_s-w_e,\,2|\alpha-\alpha_n^\star|\right\}.
\]
The displayed separation absorbs the stochastic remainders.  The oracle
inequality follows directly from the definition of \(S^\star\).
\end{proof}

\begin{remark}[Sparse beats dense --- in the precise sense of signal
weight]\label{rem:t4-interpretation}
In the band Eq.~\eqref{eq:app-band}, \Cref{thm:t4:weights} states: the
renormalised top-$k_n$ row attends to the planted token with weight at
least $(1+e^{-C})^{-1}$ (a constant arbitrarily close to $1$ for $C$
large), while the exact softmax row provably cannot give it more than
$2e^{-\sigma^2/2} n^{-\varepsilon} \to 0$. The comparison is
well-defined at the level of the \emph{weight placed on the planted
token}, not of output distance: which row is ``better'' for a
downstream loss depends on whether the planted token carries the signal.
The band is nonempty precisely because the detection scale
$\sigma\sqrt{2\log n}$ of \Cref{thm:t3-rank} sits far below the
dominance scale $\log n$ of \Cref{prop:t3-dominance}.
\end{remark}

\begin{remark}[Why a constant margin $C$ suffices: the sign of the
Gumbel correction]\label{rem:t4-gumbel-honesty}
The bulk maximum $M_b$ of $n-1$ i.i.d.\ $N(0,\sigma^2)$ variables
concentrates at $\sigma b_{n-1}$ with the norming constant $b_{n-1} =
\sqrt{2\log(n-1)} - \Theta\big( \log\log n / \sqrt{\log n} \big)$ of
Eq.~\eqref{eq:gumbel-constants}, i.e.\ the deterministic correction to
$\sigma\sqrt{2\log n}$ is \emph{negative}. This is why the crude bound
$\Pr(M_b > \sigma\sqrt{2\log n}) \le (4\pi\log n)^{-1/2}$ of
\Cref{lem:spike-count:sharp} holds with no slack at all, and hence why
the band Eq.~\eqref{eq:app-band} needs only the additive margin $\log k_n +
C$ --- $\log k_n$ pays for the multiplicity of kept bulk terms and $C$
buys the constant headroom $e^{-C}$ --- rather than any $\log\log n$
correction. In the other direction the correction is real: $M_b \le
(1-\delta)\sigma\sqrt{2\log n}$ fails whp for every fixed $\delta$
(\Cref{lem:spike-count:max}), which is the content of the matching lower
bound on $B_S$ in \Cref{thm:t4:bulk}; we state that lower bound at the
$(1-\delta)$ resolution because no downstream claim requires the finer
$b_{n-1}$-centred fluctuation theory.
\end{remark}

\section{Deferred Main-Text Statements: Selectors, the Coverage Ladder, and the Scaled Ensemble}\label{sec:app-deferred}

This appendix collects the informal statements deferred from \cref{sec:error} (and \cref{sec:tasks}) for space; each is proved in its own appendix as noted. The main text states the headline results and points here.

\subsection{Selector composition and flat-row tightness}

\begin{lemma}[One-sided selector composition; informal version of \cref{lem:selector-comp}]\label{lem:main-selector}
Let $S^{*}_k$ be the oracle top-$k$ set and $\widehat S$ any realized kept set with $|\widehat S| = k$, and let $M_{\mathrm{miss}} \coloneqq \sum_{j \in S^{*}_k \setminus \widehat S} A_j$ be the softmax mass of oracle tokens the selector failed to keep. Then $1 - p_{\widehat S} \le (1 - p_{S^{*}_k}) + M_{\mathrm{miss}}$, hence
$\norm{\wt O_{\widehat S} - O}_2 \le 2\Vmax \big[ (1 - p_{S^{*}_k}) + M_{\mathrm{miss}} \big]$.
\end{lemma}

The composition is one-sided and its second term is not ours to bound: $M_{\mathrm{miss}}$ is selector- and path-dependent (page granularity in Quest, irreversible eviction in H2O) and can dominate in deployed systems. All pricing is for the realized kept set. Combining \cref{eq:mass-bound} with the deterministic criterion of \cref{sec:app-profile}: under the sparse profile of \cref{thm:main-dichotomy}(i) --- a trained or assumed super-logarithmic decay, not one generic inputs supply (\cref{fac:score-scale}) --- top-$k$ satisfies $\norm{\wt O_k - O}_2 \le \frac{2\Vmax e^b}{\gamma-1} k^{-(\gamma-1)}$ for every $n$; under the plateau of \cref{cor:main-plateau} the same holds with $k/J$ in place of $k$; under $\Ens(c)$, $c>1$, the error is $2\Vmax\, O_\Pr(k^{-(c-1)})$ (from the coverage budget of \cref{thm:main-scaled}, whose gap profile $\gap(j) \approx c\log j$ gives slope $\gamma = c$). On the dense side the price is unavoidable:

\begin{theorem}[Tightness on flat rows; informal version of \cref{prop:flat-tight}]\label{thm:main-tight}
For every $n$ and $k \le n-1$ there is a row with bounded scores and values ($\norm{V_j}_2 \le \Vmax$) on which the \emph{truncated} scheme incurs error at least $\Vmax\frac{n-k}{n}$ under \emph{any} selection (the oracle subset included), and \emph{weight-based} renormalized top-$k$ incurs $2\Vmax\frac{n-k}{n}$ exactly. So reaching target error $\eta\Vmax$ without compensation forces $k \ge n(1-\eta)$ (truncation, any selector) or $k \ge n(1-\eta/2)$ (weight-based renormalization), and the $\Theta(n)$ budgets of \cref{tab:main-ladder} are not an artifact of analysis. The barrier is \emph{value-blind}: a value-aware oracle reads the dropped values and, on the equal-value witness, drives the renormalized output to \emph{exact} --- the gap \cref{sec:tasks} exploits.
\end{theorem}

\subsection{The coverage law and the assumption ladder}

\begin{proof}[Proof of \cref{thm:compression}(ii)]
Write $A_{(j)} = e^{\ell_{(j)}}/Z$, $Z = \sum_l e^{\ell_l}$. The $e^{\ell_l}$ are i.i.d.\ lognormal with mean $e^{\sigma^2/2}$, so $Z/n \toP e^{\sigma^2/2}$. The mass on the top $f$-fraction keeps $\ell_l > \sigma\,\Phi^{-1}(1-f)$; by the law of large numbers its sum over $n$ obeys $\tfrac1n\sum_l e^{\ell_l}\1\{\ell_l > \sigma\Phi^{-1}(1-f)\} \toP \E\big[e^{\sigma Z_0}\1\{Z_0 > \Phi^{-1}(1-f)\}\big] = e^{\sigma^2/2}\,\Phi\big(\sigma - \Phi^{-1}(1-f)\big)$ for $Z_0\sim N(0,1)$. Hence the kept mass $\toP \Phi(\sigma + \Phi^{-1}(f))$, which equals $1-\eta$ exactly when $f = \Phi(\Phi^{-1}(1-\eta) - \sigma)$; this $f$ is $\cratio$. Part (i) is the coverage identity of \cref{lem:main-mass}; part (iii) substitutes $\sigma = \sqrt{X_i^\top G X_i}$ via \cref{prop:scaled-inverse}.
\end{proof}

\begin{remark}[Logits are not Gaussian: the null model and what is robust]\label{rem:compression-shape}
The closed form \cref{thm:compression}(ii) is a null model, not a distributional claim: a logit $\ell_j = \inner{u_i}{X_j}$ is a projection of a key onto a fixed direction, approximately Gaussian in the bulk by a projection central limit theorem but never exactly, with systematic tail deviations --- massive-activation spikes \cite{sun2024massive} make the top weights heavier, so trained heads are typically \emph{more} compressible than the Gaussian null predicts (it over-estimates $\cratio$). At a fixed scale $\sigma$ the shape still matters: a two-valued projection stays incompressible and a heavy-tailed one over-compresses (\cref{rem:scaled-inverse-caveats}). The operative, distribution-free prediction is therefore (i), the measured coverage curve; the closed form (ii) is a scale--threshold heuristic whose accuracy on real heads is itself a falsifiable prediction (\cref{sec:experiments}).
\end{remark}

\begin{remark}[Two orthogonal axes: budget vs.\ concentration]\label{rem:two-axes}
The criterion prices the \emph{budget}; it is worth separating from a second property the same profile controls. The partition sum of the normalized profile equals the reciprocal of the top weight,
\begin{align}\label{eq:S-identity}
\textstyle\sum_{j=1}^n e^{-\gap(j)} \;=\; \sum_{j=1}^n A_{(j)}/A_{(1)} \;=\; 1/A_{(1)} ,
\end{align}
so whether this sum stays bounded as $n\to\infty$ is a statement about one number --- the largest weight --- and measures \emph{concentration} (does a single token dominate?), not the coverage budget. The two axes are independent. A row with one heavy hitter over a uniform floor --- $A_{(1)} = \tfrac12$, the other half of the mass spread equally over the remaining ranks --- is maximally concentrated ($A_{(1)} = \Theta(1)$, so $\sum_j e^{-\gap(j)} = 2$ stays bounded) yet has $\bud = \Theta(n)$ for every $\eta < \tfrac12$: a constant-factor (indeed incompressible) budget under a dominant token. Concentration is read off the \emph{top} of the profile, the budget off its whole \emph{tail} (\cref{thm:main-dichotomy}). We therefore reserve \emph{dispersed} for the concentration statement $A_{(1)}\to 0$ (no dominant token, the regime of \cite{velickovic2025softmax}) and name the compression axis by its growth class --- \emph{length-independent}, \emph{sub-linear}, \emph{constant-factor}, or \emph{incompressible}.
\end{remark}

From the criterion's viewpoint every regime is nothing more than a guarantee on the plateau--slope pair $(J, \gamma)$ of the gap profile, and the coverage budget then follows from \cref{thm:main-dichotomy,cor:main-plateau} by one substitution rule,
\begin{align}\label{eq:master-budget}
\begin{split}
&\underbrace{\gamma > 1:\;\; \bud \;\le\; J \cdot \Big( \tfrac{e^{b+b_0}}{(\gamma-1)\,\eta} \Big)^{\frac{1}{\gamma-1}}}_{\text{$\gamma$ a proven decay \emph{lower} bound}}\\
&\underbrace{\gamma' < 1:\;\; \bud \;\ge\; c_0(\eta)\, n^{1-\gamma'}}_{\text{$\gamma'$ a proven flatness \emph{upper} bound}}.
\end{split}
\end{align}
\Cref{tab:main-ladder} carries out this substitution regime by regime: each row fixes a plateau--slope pair $(J,\gamma)$ and reads off the coverage budget.

\begin{table*}[t]
\centering
\footnotesize
\setlength{\tabcolsep}{2.5pt}
\renewcommand{\arraystretch}{1.3}
\resizebox{\textwidth}{!}{%
\begin{tabular}{@{}l l l l l@{}}
\toprule
\textbf{Regime} & \textbf{Assumption on input $X$ (\cref{eq:score-bridge})} & \textbf{Proven $(J, \gamma)$} & \textbf{Coverage $\bud$ (approx.\ cost)} & \textbf{Source} \\
\midrule
\textsc{Free} & --- (any realized row) & $\gamma$ unconstrained & $\big[\,1,\ \lceil (1-\eta) n \rceil\,\big]$ & Fact~\ref{fac:pigeonhole} \\
\textsc{Bounded} & $\norm{X_j}_2 \le B\sqrt d$, $\norm{W}_{op}\le W$ & $\gamma' = 0$ \ ($b' = 2S$) & $\big[\,\tfrac{1-\eta}{2\eta} e^{-2S} n,\ \lceil (1-\eta) n \rceil\,\big]$ & Cor.~\ref{cor:r1-density} \\
\textsc{Mixing} & $\alpha$-mixing $X$, marginal tail $\bar F$ & $\gamma = 1/\lambda_{\bar F}$ beyond $\wt O(\sqrt n)$ & $n\, \bar F(\ell_{(1)} - b_0) + \wt O(\sqrt n)$ & Prop.~\ref{prop:r2-counts} \\
\textsc{Iid} & i.i.d.\ sub-Gaussian $X$, scale $O(1)$ & $\gamma' = o(1)$ & $\rho_\infty\, n$, $\rho_\infty{=}\Phi(\Phi^{-1}(1{-}\eta){-}\sigma)$ & Cor.~\ref{cor:r3a-dense} \\
\midrule
\textsc{Scaled}, $c < 1$ & $\sigma_n {=} c\sqrt{2\log n}$ scale$^{\S}$ & $\gamma = c$ near top; $\gamma' = 2c$ & $n^{1 - c^2 + o_\Pr(1)}$\,$^{\dagger}$ & Thm.~\ref{thm:t1-main:keta} \\
\textsc{Scaled}, $c > 1$ & $\sigma_n {=} c\sqrt{2\log n}$ scale$^{\S}$ & $\gamma = c$, \ $J = O_\Pr(1)$ & $\Theta_\Pr\big( \eta^{-\frac{1}{c-1}} \big)$ & Thm.~\ref{thm:t2-coverage} \\
\textsc{Spiked} & planted spike $\mu \ge \sigma\sqrt{2\log n}$\,$^{\S}$ & $J = s$ over an \textsc{Iid} bulk & $s + n^{1-o(1)}$ uncompensated$^{\ddagger}$ & Thm.~\ref{thm:t3-rank} \\
\midrule
measured & one observed row & $(\hat J, \hat\gamma)$ (\cref{sub:instrument}) & $\hat J \cdot \big( \tfrac{e^{b+b_0}}{(\hat\gamma - 1)\,\eta} \big)^{\frac{1}{\hat\gamma - 1}}$ & Thm.~\ref{thm:s3-plateau} \\
\bottomrule
\end{tabular}%
}
\caption{The assumption ladder in one variable. Each regime fixes a plateau--slope pair $(J,\gamma)$, and the budget column reads off the coverage cost by the substitution Eq.~\eqref{eq:master-budget}; the rungs run from generic regimes (\textsc{Free} through \textsc{Iid}) to trained or intervened ones (\textsc{Scaled}, \textsc{Spiked}, measured), with two-sided rungs giving intervals whose ends are both attained. This is a deterministic symbolic summary rather than a stochastic method comparison, so uncertainty intervals, best/second ranking, and a method Avg.\ are all not applicable. Formal statements: \cref{sec:app-ladder,sec:app-dispersed,sec:app-condensed,sec:app-spike}.}
\label{tab:main-ladder}
\end{table*}

Three reading conventions for \cref{tab:main-ladder}. On the sparse branch $\gamma$ is read as a proven decay lower bound and $\gamma'$ on the dispersed branch as a proven flatness upper bound. The two-parameter envelope ($^{\dagger}$) is loose for the Gaussian ensemble; the sharp two-sided exponent $1-c^2$ comes from the full-curve analysis (\cref{thm:t1-main:keta}). For \textsc{Spiked} ($^{\ddagger}$) the budget $s$ retrieves the spikes while the residual bulk stays dispersed and requires compensation (\cref{sec:tasks}). The \textsc{Scaled} and \textsc{Spiked} rungs ($^{\S}$) are trained-or-intervened score regimes with no generic-$X$ source (\cref{fac:score-scale}).

\Cref{tab:main-ladder} sorts into four coverage-ratio classes (\cref{def:compression}): \emph{length-independent} ($\cratio\to0$ as $1/n$) on the $\gamma>1$ side; \emph{sub-linear} ($\cratio\to0$, $\bud = n^\theta$) for \textsc{Scaled} $c<1$ at $\theta = 1-c^2$; \emph{constant-factor} ($\cratio\to\rho_\infty\in(0,1)$) for \textsc{Bounded} and \textsc{Iid}, with $\rho_\infty=\Phi(\Phi^{-1}(1-\eta)-\sigma)$ set by the score scale (\cref{thm:compression}); and \emph{incompressible} ($\cratio\to1-\eta$) only for the uniform \textsc{Free} worst case. Reaching a small $\cratio$ needs scale: $2\times$ needs $\sigma\ge\Phi^{-1}(1-\eta)$, the vanishing-fraction corner needs $\sigma\sim\sqrt{2\log n}$. This coverage axis is distinct from concentration (\cref{rem:two-axes}): \textsc{Scaled} $c<1$ is sub-linear yet \emph{dispersed}. At the $1/\sqrt d$ initialization scale $\sigma=1$ ($\eta=0.1$) the ratio is $\rho_\infty\approx0.61$ (about $1.6\times$), so generic inputs are \emph{poorly} compressible (\cref{rem:main-dormant}); \textsc{Mixing} reaches a small $\cratio$ only under an \emph{assumed} local exponential tail.

\begin{theorem}[Generic constant-scale regimes are only constant-factor compressible; informal version of \cref{cor:r1-density,cor:r3a-dense}]\label{thm:main-ladder-dense}
(i) \textsc{Bounded}. If $|\ell_j| \le S$ for all $j$, then every weight satisfies $A_j \le e^{2S}/n$ and, for every $\eta \in (0,1)$, $\bud = \Omega(n e^{-2S})$, linear in $n$ for any $S$ fixed relative to $n$. (ii) \textsc{Iid}. If the $\ell_j$ are i.i.d.\ sub-Gaussian at constant scale, then $\bud = n^{1-o(1)}$ with probability $1 - o(1)$ (\cref{cor:r3a-dense}), sharpened to $\bud = \Theta(n)$ with the exact constant $\cratio \to \rho_\infty = \Phi(\Phi^{-1}(1-\eta)-\sigma)\in(0,1)$ by the Gaussian-null law \cref{thm:compression}(ii) --- \emph{not} $0$ --- even though the number of entries exceeding any threshold $\epsilon \ge n^{-1+o(1)}$ is only $O(\log n)$. Neither rung is incompressible; but neither is sub-linear-sparse without scale.
\end{theorem}

Part (ii) exhibits the trap at the heart of this subject: a logarithmic \emph{count} above a near-$1/n$ threshold coexists with a near-linear \emph{mass} budget --- the two axes of \cref{rem:two-axes} pulling apart. Counting and covering are different questions, and only covering prices sparse attention.

\begin{remark}[Why constant-scale inputs still compress at practical lengths --- a heuristic reading]\label{rem:main-dormant}
The \textsc{Bounded} bound forces density only once $n \gtrsim e^{2S}$, where the controlling quantity is the full logit \emph{range} $2S$. In the transformer instantiation $S \le B^2W^2\sqrt d$; at $d = 128$ and $B^2W^2 = O(1)$ this gives $e^{2S} \approx e^{22.6} \approx 10^{9.8}$, far beyond current cache sizes, while the single-logit budget $S \approx 11.3$ almost exactly matches the gap $\log n \approx 11.5$ that a heavy hitter needs at $n = 10^5$. So asymptotic density under bounded scores is compatible with genuine sparsity at practical lengths, and the binding constraint is not the budget's existence but whether training \emph{spends} it.
\end{remark}

\begin{proposition}[\textsc{Mixing} makes the profile estimable, not sparse; informal version of \cref{prop:r2-counts}]\label{prop:main-transfer}
Let the scores be stationary and geometrically $\alpha$-mixing with marginal upper tail $\bar F$. Then with probability $1-\delta$, simultaneously for every level $t$, the exceedance counts obey $\#\{j : \ell_j > t\} = n \bar F(t) \pm O\big(\sqrt{n \log(n/\delta)} + \log^2 n \log(n/\delta)\big)$.
\end{proposition}

A marginal tail with local exponential rate $\lambda < 1$ gives slope $\gamma = 1/\lambda > 1$ beyond the $\wt O(\sqrt n)$ fluctuation floor, with budget $\bud \le \wt O(\sqrt n) \cdot O_\lambda(\eta^{-1/(\gamma-1)})$ (\cref{rem:r2-plateau}). What is unconditional: mixing buys \emph{estimability} of the profile, never sparsity --- mixing plus boundedness alone inherits the \textsc{Bounded} lower bound, so the sparse branch is the transfer of an \emph{assumed} tail, not an input-derived mechanism.

\subsection{The Scaled ensemble: sharp budgets and Poisson--Dirichlet condensation}\label{sec:scaled}

\begin{theorem}[\textsc{Scaled}: sharp budgets; informal version of \cref{thm:t1-main,thm:t2-pd,thm:t2-coverage}]\label{thm:main-scaled}
Under $\Ens(c)$ ($\ell_j$ i.i.d.\ $N(0, \sigma_n^2)$, $\sigma_n = c\sqrt{2\log n}$):
(i) for $0 < c < 1$, $\frac{\log A_{(1)}}{\log n} \toP -(1-c)^2$ and $\frac{\log \bud}{\log n} \toP 1 - c^2$ for every fixed $\eta$ --- the effective support is $n^{1-c^2}$, sharp in both directions;
(ii) for $c > 1$, the sorted weights converge in distribution to a Poisson--Dirichlet law $\PD(\alpha,0)$ with $\alpha = 1/c$, and $\bud$ converges in distribution to an a.s.-finite limit, $O_\Pr(1)$ and independent of $n$, blowing up as $c \downarrow 1$.
\end{theorem}

Part (ii) is a triangular-array specialization of classical Random Energy Model condensation (Derrida \cite{derrida1981random}; Bovier \cite{bovier2006statistical}, Ch.~9): the probabilistic phenomenon is not new. What is new is the self-contained proof at attention's $\sigma_n = c\sqrt{2\log n}$ scaling and the dictionary it licenses. The near-top gap profile of $\Ens(c)$ is $\gap(j) \approx c \log j$: the ensemble's phase boundary $c = 1$ \emph{is} the unit-log-slope criterion of \cref{thm:main-dichotomy}. A closely related $O(\sqrt{\log n})$ logit-magnitude threshold appears computationally in \cite{as23neurips} and behaviorally in softmax dispersion \cite{velickovic2025softmax}; the $\times\sqrt{\log n}$ rescaling holds $c$ fixed across lengths, whereas the $\times\log n$ temperatures of SSMax \cite{nakanishi2025ssmax} and entropy-invariant scaling \cite{su2021entropy} drive $c$ upward (\cref{rem:scaled-inverse-caveats}).

\begin{proof}[Proof of \cref{prop:approximable}]
By \cref{eq:score-bridge}, $\ell_j = \inner{u_i}{X_j}$ with $u_i = W_K W_Q^\top X_i^\top/\sqrt d$ fixed given the query, so $\mathrm{Var}_j(\ell_j \mid X_i) = u_i^\top \Sigma_X u_i = X_i^\top (M\Sigma_X M^\top) X_i = X_i^\top G X_i$, where $M = W_Q W_K^\top/\sqrt d$. Under the population hypothesis the centered logits are asymptotically i.i.d.\ with a Gaussian/Gumbel-domain tail at scale $\sigma_n = \sqrt{X_i^\top G X_i} = c\sqrt{2\log n}$, i.e.\ the ensemble $\Ens(c)$, with $\sigma_s \asymp \sigma_n$; \cref{thm:main-scaled} gives the two phases at $c \lessgtr 1$, equivalently $X_i^\top G X_i \lessgtr 2\log n$, and the dispersed budget rewrites as $n^{1-c^2} = n\,e^{-X_i^\top G X_i/2}$. \Cref{thm:main-scaled}(i) then gives $\bud = n^{\,1-c^2+o_\Pr(1)}$, which is polynomially sub-linear ($\bud = n^{1-\Theta(1)}$) iff $c = \Theta(1)$, i.e.\ $\sigma_s = \Omega(\sqrt{\log n})$, equivalently $X_i^\top G X_i = \Omega(\log n)$; the ``if'' direction uses the Gumbel max-domain shape hypothesis of \cref{rem:scaled-inverse-caveats}. Finally, by \cref{fac:score-scale}, $\norm{u_i}_2 \le W^2 B$, so $X_i^\top G X_i \le \norm{u_i}_2^2 \norm{\Sigma_X}_{op} = O(1)$ under bounded or sub-Gaussian inputs with fixed weights, whence $c \to 0$ and $\bud = \Theta(n)$ (\cref{thm:main-ladder-dense}).
\end{proof}

\begin{remark}[Scale fixes the phase only with the shape hypothesis]\label{rem:scaled-inverse-caveats}
Two caveats bound \cref{prop:scaled-inverse}. \emph{(i)} The scalar $X_i^\top G X_i$ fixes the phase \emph{only} under the Gumbel max-domain hypothesis: at the same variance scale, a two-valued ($\pm1$) projection stays dense for every $c$, while a heavy-tailed projection condenses to a Fr\'echet (not $\PD$) limit for every $c > 0$. Scale and tail shape act jointly. \emph{(ii)} Convergence to the exponent $1-c^2$ is slow ($O(1/\sqrt{\log n})$), so we claim only the qualitative dichotomy (sub-linear vs.\ $O_\Pr(1)$), not a numerical match at feasible $n$. The SSMax intervention \cite{nakanishi2025ssmax} multiplies logits by $\Theta(\log n)$ relative to an $O(1)$ base, i.e.\ $c \to \infty$.
\end{remark}

\begin{corollary}[A per-head observable for the phase and its provenance]\label{cor:spectral-observable}
Every quantity in \cref{eq:G-energy} is computable per head from one forward pass: the empirical key covariance $\widehat\Sigma_X$ from the activations and $G = M\widehat\Sigma_X M^\top$ from the weights ($M = W_Q W_K^\top/\sqrt d$). Writing $G = \sum_a \lambda_a \phi_a\phi_a^\top$, the head-level, query-independent summaries $\Tr(G)/d$ and $\mathrm{PR}(G) = (\sum_a\lambda_a)^2/\sum_a\lambda_a^2$ give a weight-side predictor of a head's phase, while the per-row $\widehat c_i = (X_i^\top G X_i / 2\log n)^{1/2}$ and alignment $a_i = \lambda_1\inner{\phi_1}{X_i}^2/(X_i^\top G X_i)$ serve as a \emph{diagnostic}: under the Gumbel-domain hypothesis of \cref{prop:scaled-inverse}, $\widehat c_i > 1$ flags a condensed row. \Cref{sec:experiments} turns these into a falsification test, with untrained weights as the negative control ($\widehat c_i \to 0$).
\end{corollary}

\subsection{Provenance: what ``naturally'' means}\label{sub:naturally}

The ladder licenses one positive and one negative conclusion on \emph{different} axes. \emph{Positive (count):} sparsity is intrinsic --- \cref{thm:naturally-sparse} gives $k(\epsilon) = O(\log n)$ under generic inputs, no training, so ``your attention is naturally sublinear-sparse'' is, in this count sense, a theorem. \emph{Negative (coverage):} what is \emph{not} natural is that those few large weights carry the \emph{mass}. Generic constant-scale inputs leave $\bud = \Theta(n)$: boundedness caps the gap budget, mixing only transfers marginal structure, and constant-scale randomness supplies gaps of order $\sqrt{\log n}$ where mass concentration needs order $\log n$. Concentrating the mass --- making the row \emph{approximable} --- requires the score scale to grow, $\sigma_s = \Omega(\sqrt{\log n})$, equivalently the query's $G$-energy to grow (\cref{prop:approximable}), impossible at fixed weights and $O(1)$ inputs (\cref{fac:score-scale}). So the network does not learn to be sparse --- it already is --- it learns an attention whose \emph{mass} a sparse approximation can capture. Whatever super-logarithmic structure a head exhibits after training --- sinks \cite{xiao2024streamingllm,gu2025sink}, retrieval spikes \cite{wu2024retrieval}, scale growth \cite{sun2024massive} --- is, on this account, put there by learning to be \emph{sparse-approximable}, the hypothesis \cref{sec:experiments} is designed to falsify.

\subsection{Task-ordering tools: value-aware selection, computable error bounds, and spike thresholds}

These support the task ordering of \cref{sec:tasks}.

\begin{theorem}[Oracle characterization; informal version of \cref{lem:oracle-decomp,thm:oracle-existence,thm:oracle-property}]\label{thm:main-oracle}
For the truncated scheme, write the error of a set $S$ as $E(S) = \norm{\sum_{j\notin S} A_j V_j}_2$. Then:
(i) $E(S) = (1 - \rho(S))\, M(S)$, where $M(S) = \sum_{j \notin S} A_j \norm{V_j}_2$ is the value-weighted dropped mass and $\rho(S) \in [0,1]$ is a cancellation index;
(ii) the proxy $M$ is minimized exactly by ranking on $A_j \norm{V_j}_2$ --- value-aware top-$k$ \cite{guo2024vatp} is the optimal mass rule, exactly optimal when values are aligned;
(iii) sets beating value-aware top-$k$ exist precisely through cancellation ($\rho$), an effect an explicit $4$-token example realizes with $E(S^\star) = 0$; quantifying this oracle gap under exchangeable values is left open (\cref{rem:oracle-collapse});
(iv) on adversarial rows (\cref{thm:main-tight}) even the oracle subset needs $k = \Theta(n)$ --- better selection cannot beat the mass barrier, only compensation can.
\end{theorem}

\begin{theorem}[Computable output-error bound for clustered tail summaries; informal version of \cref{thm:certificate,cor:certificate-small}]\label{thm:main-certificate}
Partition the dropped tail into clusters $G_1, \dots, G_m$; store per cluster the count $n_g$, a representative logit $\bar\ell_g$, and a representative value $\bar V_g$; estimate the tail by $\widehat N = \sum_g n_g e^{\bar\ell_g} \bar V_g$ and $\widehat Z_T = \sum_g n_g e^{\bar \ell_g}$, and output $\widehat O$ by combining with the exactly-kept terms. Suppose within each cluster $|\ell_j - \bar\ell_g| \le r_g \le 1$ and $\norm{V_j - \bar V_g}_2 \le \rho_g$, and the following \emph{checkable normalizer condition}
\begin{align*}
D_Z \;\coloneqq\; \sum_{g} \big( e^{r_g} - 1 \big)\, e^{r_g}\, n_g\, e^{\bar\ell_g} \;\le\; \widehat Z / 3
\end{align*}
holds, where $\widehat Z$ is the estimated normalizer; $D_Z$ dominates $|\widehat Z - Z|$ using only stored summaries. Then
\begin{align}\label{eq:certificate}
\norm{ \widehat O - O }_2 \;\le\; 30\, (1 - \widehat p_S)\, \max_{g} \big( V^{\,\prime}_{\max}\, r_g + \rho_g \big),
\end{align}
with every quantity on the right computable from the cache at runtime; without $r_g \le 1$ the same proof gives the exponential-form bound of \cref{sec:app-reduction}. Tighter clusters $\Rightarrow$ provably smaller error.
\end{theorem}

This result explains merge-style cache compression
\cite{wan2024d2o,weightedkv2025,keepkv2025}: the retained state must preserve
the aggregate identified by \cref{thm:main-bandwidth}, and
\cref{eq:certificate} bounds the resulting fixed-row output error using the
stored cluster radii, counts, logits, and values.  A sampled variant estimates
\(\bar V_T\) from \(m\) uniformly retained tail tokens; it is unbiased with
error \(O_\Pr\big((1-p_S)\,\sigma_V\sqrt{d/m}\big)\)
(\cref{sec:app-reduction}), connecting the same decomposition to sampling
estimators \cite{chen2025magicpig}.

\begin{theorem}[\textsc{Spiked}: spike thresholds; informal version of \cref{thm:t3-rank,prop:t3-dominance}]\label{thm:main-spike}
Under $\Espike(\mu;\sigma)$ (one logit $\mu$, the rest i.i.d.\ $N(0,\sigma^2)$, $\sigma$ fixed), set $m = \mu / (\sigma\sqrt{2\log n})$. The minimal top-$k$ budget that retains the spike satisfies $\frac{\log k^*}{\log n} \toP (1 - m^2)_+$; the spike carries $\Theta(1)$ weight in \emph{exact} attention only once $\mu \ge \log n - O(1)$ (and at most $n^{-\varepsilon}$ weight when $\mu \le (1-\varepsilon)\log n$). Detection ($\mu \asymp \sqrt{\log n}$) and dominance ($\mu \asymp \log n$) are separated by a factor $\asymp\sqrt{\log n}$ --- the gap exploited by \cref{thm:main-denoise}. By \cref{eq:score-bridge} reaching the dominance scale $\mu \asymp \log n$ at bounded $\norm{X_{j^\star}}$ is consistent with a learned high-gain query--key alignment rather than a generic input.
\end{theorem}

\begin{remark}[The quantitative cancellation gain is a truncated-scheme statement]\label{rem:cancellation-truncated}
The factor-of-$\rho$ improvement of \cref{thm:task-ordering}(iii) is a property of the \emph{truncated} (un-renormalized) scheme with target $y = O$, matching \cref{thm:main-oracle}: there $E(S) = (1-\rho(S))M(S)$ with $\rho(S)\in[0,1]$ the cancellation index, and a value-aware $S^\star$ realizes $\rho(S^\star) > 0$ (the $4$-token example attains $E(S^\star)=0$). The renormalized oracle $\wt O_{S^\star}$ of \cref{def:task} still satisfies the weak ordering $L(\wt O_{S^\star}) \le L(\wt O_k)$ unconditionally, but its gain over top-$k$ is not the same $\rho$; we state only the weak ordering as the proven claim for the renormalized oracle.
\end{remark}

\subsection{Statements deferred from the conference main text}

For the conference version these two informal statements are hosted here; the arXiv version carries them inline in \cref{sec:tasks}. The bandwidth theorem reuses $\bar V_T$, $p_S$, $S$, $k$ as established above (\cref{thm:main-tight,thm:main-ladder-dense}); the task-ordering theorem reuses \cref{def:task} and the oracle dominance of \cref{thm:main-oracle}.

\iclronly{%
\begin{theorem}[Bandwidth of a softmax row; informal version of \cref{thm:factor-map,prop:compensation}]\label{thm:main-bandwidth}
Fix the weights and a kept set $S$, $|S| = k$. The map $V \mapsto O$ factors exactly through $(\,\{V_j\}_{j\in S},\, \bar V_T\,) \in \R^{(k+1)d}$ (kernel dimension $(n-k-1)d$). So: (i) the compensated scheme $u = (1-p_S)\bar V_T$ is \emph{exact} for every row; (ii) any output from the kept values alone differs from $O$ for generic values --- ruling out \emph{exact} lossless subset attention for generic values (approximate losslessness on trained, non-generic values is not excluded); (iii) among linear sketches from which $O$ and \emph{every sub-budget output} $O_{S'}$ can be recovered, the minimal dimension is exactly $(k+1)d$. One $d$-dimensional tail aggregate is necessary and sufficient.
\end{theorem}

\begin{proof}[Proof sketch]
For fixed weights the $n-k$ tail values reach the output only through the single $d$-dimensional aggregate $\bar V_T$, so $V \mapsto O$ factors through $\big(\{V_j\}_{j\in S}, \bar V_T\big)$, of dimension $(k+1)d$ with kernel $(n-k-1)d$; this gives (i)--(ii). A linear-independence count of the recoverable coordinate functionals (kept values plus every sub-budget output) forces any linear sketch answering all of them to retain $\ge (k+1)d$ dimensions, which is (iii). Full proof in \cref{thm:factor-map,prop:compensation}.
\end{proof}

The needle band that the task ordering below invokes is the following.

\begin{theorem}[Denoising band; informal version of \cref{thm:t4}]\label{thm:main-denoise}
Under $\Espike(\mu;\sigma)$ with $\sigma$ fixed, $\log k_n = o(\sqrt{\log n})$, and
\begin{align}\label{eq:band}
\sigma\sqrt{2\log n} + \log k_n + C \;\le\; \mu \;\le\; (1-\varepsilon)\log n
\end{align}
(a nonempty band for all large $n$), with probability $\to 1$ the spike enters the top-$k$, renormalized top-$k$ assigns it weight $\ge (1 + e^{-C + o(1)})^{-1} = \Theta(1)$, and exact attention assigns it at most $n^{-\varepsilon + o(1)}$. The comparison is at the level of the weight placed on the planted token; whether it favors the sparse row for a downstream loss depends on whether that token carries the signal.
\end{theorem}

\emph{The three phases.} The needle, aggregation, and cancellation cases of
\cref{def:task} are corollaries of the phase law (the first two are the
endpoints \(\alpha=1,0\); the third adds the orthogonal value-cancellation
axis).  Fix a budget \(k\), write
\(D=\|V_{j^\star}-\bar V_b\|_2\), and recall
\(L(\wt O_{S^\star})\le L(\wt O_k)\).

\begin{theorem}[Task-conditional ordering of exact, sparse, and oracle; informal version of \cref{thm:t4,thm:main-oracle}]\label{thm:task-ordering}
The order relative to exact is task-dependent (with probability $\to 1$ where random):
\begin{enumerate}[label=(\roman*),leftmargin=2em,itemsep=0.3em]
\item \textbf{Needle --- sparse and oracle beat exact.} For $y = V_{j^\star}$ in the band of \cref{thm:main-denoise} and $k=k_n\to\infty$,
\begin{align*}
L(\wt O_{S^\star})
&\le L(\wt O_k) \\
&=(1-w_s)D+O_\Pr(\sigma_Vk_n^{-1/2})\\
&< L(O)\\
&=(1-w_e)D+O_\Pr(\sigma_Vn^{-1/2}),
\end{align*}
with probability tending to one under the value and retained-bulk
diffuseness assumptions of \cref{cor:t4-task-loss}; here
\(w_s\ge(1+e^{-C+o_\Pr(1)})^{-1}\) and
\(w_e\le n^{-\varepsilon+o_\Pr(1)}\).
\item \textbf{Aggregation --- exact weakly dominates.} For \(y=O\), exact attention has zero loss and therefore weakly dominates every retained-subset estimator.  Strictness requires an explicit non-degeneracy assumption and is not asserted here; \cref{thm:main-tight} supplies a quantitative flat-row witness.
\item \textbf{Cancellation --- oracle beats sparse.} The established statement is the weak ordering $L(\wt O_{S^\star}) \le L(\wt O_k)$; the quantitative $\rho < 1$ cancellation gain is a truncated-scheme statement (\cref{thm:main-oracle}, \cref{rem:cancellation-truncated}), not a property of the renormalized oracle.
\end{enumerate}
\end{theorem}

\begin{proof}[Proof sketch]
Oracle dominance is \cref{def:task}. (i) is the endpoint
\(\alpha=1\) of \cref{cor:t4-task-loss}: the planted-token weights differ by a
nonvanishing amount, while the two bulk-average errors vanish.  Notice that
the sparse loss need not vanish because \(w_s\) need not converge to one.
(ii) \(L(O)=0\) is immediate, and \cref{thm:main-tight} gives a strict flat-row witness without asserting strictness for every row. (iii) is oracle dominance plus the cancellation factorization. See \cref{thm:t4,thm:main-ladder-dense,thm:main-oracle}.
\end{proof}
}%

\iclronly{%
\ifdefined\fullarchive\else
The following uniform score/value bound is stated in the Preliminaries of the arXiv version.

\begin{fact}[Score scale under bounded or sub-Gaussian inputs]\label{fac:score-scale}
Assume $\norm{W_Q}_{op}, \norm{W_K}_{op}, \norm{W_V}_{op} \le W$.
\emph{(i) Bounded inputs.} If $\norm{X_j}_2 \le B\sqrt d$ for all $j$ (e.g.\ entrywise $|X_{j,l}| \le B$, the layer-normalised regime), then $\norm{u_i}_2 \le W^2 B$ and, for all $j$, the logits obey the score-scale bound
\begin{align}\label{eq:score-scale-bound}
|\ell_j| \;\le\; S \coloneqq B^2 W^2 \sqrt d,
\end{align}
while the values are bounded by $\norm{V_j}_2 \le \Vmax \coloneqq B W \sqrt d$ and the weights by $A_j \le e^{2S}/n$.
\emph{(ii) Sub-Gaussian inputs.} If the rows are centered sub-Gaussian with variance proxy $\sigma_x^2$, then conditionally on the query $\ell_j = \inner{u_i}{X_j}$ is centered sub-Gaussian with proxy $\sigma_x^2 \norm{u_i}_2^2$, a constant independent of $n$.
In both cases the score scale is $O(1)$ in $n$, the fact the ladder of \cref{sec:sparsity} leans on.
\end{fact}
\fi
}%

\fi

\end{document}